\documentclass{article} % For LaTeX2e
\usepackage{iclr2025_conference,times}

\usepackage{amsmath,amsfonts,bm}

\def\eqref#1{equation~\ref{#1}}
\def\1{\bm{1}}

\DeclareMathAlphabet{\mathsfit}{\encodingdefault}{\sfdefault}{m}{sl}
\SetMathAlphabet{\mathsfit}{bold}{\encodingdefault}{\sfdefault}{bx}{n}

\DeclareMathOperator*{\argmax}{arg\,max}

\usepackage{hyperref}
\usepackage{url}
\usepackage{amsmath}
\usepackage{amssymb}
\usepackage{mathtools}
\usepackage{amsthm}

\theoremstyle{plain}

\theoremstyle{definition}

\theoremstyle{remark}

\usepackage{algorithm}
\usepackage{algorithmic}
\usepackage[algo2e]{algorithm2e}
\usepackage[utf8]{inputenc}
\usepackage[cjk]{kotex}
\usepackage{subfigure}
\usepackage{multirow}
\usepackage{pbox}
\usepackage{graphicx}
\usepackage{wrapfig}
\usepackage{bm}
\usepackage{siunitx}
\usepackage{booktabs}
\usepackage{lipsum}
\usepackage{xfrac}
\usepackage{bbding}
\usepackage{colortbl}
\usepackage{enumitem}
\usepackage[noabbrev,capitalize,nameinlink]{cleveref}
\hypersetup{colorlinks={true},linkcolor={darkred},citecolor=darkblue}
\definecolor{bluegray}{rgb}{0.4, 0.6, 0.8}
\definecolor{electriclime}{rgb}{0.8, 1.0, 0.0}
\definecolor{malachite}{rgb}{0.04, 0.85, 0.32}
\definecolor{darkred}{rgb}{0.55, 0.0, 0.0}
\definecolor{darkblue}{rgb}{0.0, 0.0, 0.55}
\definecolor{darkgreen}{rgb}{0.0, 0.2, 0.13}
\definecolor{darkorchid}{rgb}{0.6, 0.2, 0.8}

\Crefname{figure}{Fig.}{Figs.}% {<type>}{<singular>}{<plural>}
\usepackage[textsize=tiny]{todonotes}
\usepackage{marginnote}
 
\newcommand{\std}{\scriptsize{\;±\;}}

\definecolor{lightRed}{HTML}{F8CECC}
\definecolor{redBorder}{HTML}{CC0000}
\definecolor{greenBorder}{HTML}{82B366}
\definecolor{lightGreen}{HTML}{D5E8D4}
\definecolor{blueBorder}{HTML}{6C8EBF}
\definecolor{lightBlue}{HTML}{DAE8FC}
\definecolor{blackBorder}{HTML}{000000}

\DeclareRobustCommand\myredcircle{\raisebox{-2pt} {\tikz[]{\node[shape=circle,draw=redBorder,fill=white,text=white, inner sep=1pt, line width=1pt]{\scriptsize{A}};}}}

\DeclareRobustCommand\myblackcircle{\raisebox{-2pt} {\tikz[]{\node[shape=circle,draw=blackBorder,fill=white,text=white,inner sep=1pt, line width=1pt]{\scriptsize{A}};}}}
\DeclareRobustCommand\myblackbox{\raisebox{-1pt} {\tikz[]{\node[shape=rectangle,draw=blackBorder,fill=white,text=white,inner sep=1pt, line width=1pt]{\scriptsize{A}};}}}

\usepackage[toc,page,header]{appendix}
\usepackage{minitoc}

\title{PIORF: Physics-Informed Ollivier--Ricci Flow for Long-Range Interactions in Mesh Graph Neural Networks}

\author{Youn-Yeol Yu\textsuperscript{\rm 1}\thanks{Equal contribution.}\;\;\textsuperscript{\href{mailto:yyyou@yonsei.ac.kr}{\Envelope}}, 
\;Jeongwhan Choi\textsuperscript{\rm 1}\footnotemark[1]\;\;\textsuperscript{\href{mailto:jeongwhan.choi@yonsei.ac.kr}{\Envelope}},
\;Jaehyeon Park\textsuperscript{\rm 2 \href{mailto:jaehyeon.park@kaist.ac.kr}{\Envelope}} ,
\;Kookjin Lee\textsuperscript{\rm 3 \href{mailto:kookjin.lee@asu.edu}{\Envelope}} ,
\;Noseong Park\textsuperscript{\rm 2}\thanks{Corresponding author.}\;\;\textsuperscript{\href{mailto:noseong@kaist.ac.kr}{\Envelope}} \\
\textsuperscript{1}Yonsei University \quad \textsuperscript{2}KAIST \quad \textsuperscript{3}Arizona State University
}

\iclrfinalcopy % Uncomment for camera-ready version, but NOT for submission.
\begin{document}
\doparttoc % Tell to minitoc to generate a toc for the parts
\faketableofcontents % Run a fake tableofcontents command for the partocs

\maketitle

\begin{abstract}
Recently, data-driven simulators based on graph neural networks have gained attention in modeling physical systems on unstructured meshes.
However, they struggle with long-range dependencies in fluid flows, particularly in refined mesh regions. This challenge, known as the `over-squashing' problem, hinders information propagation. While existing graph rewiring methods address this issue to some extent, they only consider graph topology, overlooking the underlying physical phenomena. 
We propose Physics-Informed Ollivier--Ricci Flow (PIORF), a novel rewiring method that combines physical correlations with graph topology. PIORF uses Ollivier--Ricci curvature (ORC) to identify bottleneck regions and connects these areas with nodes in high-velocity gradient nodes, enabling long-range interactions and mitigating over-squashing. Our approach is computationally efficient in rewiring edges and can scale to larger simulations. 
Experimental results on 3 fluid dynamics benchmark datasets show that PIORF consistently outperforms baseline models and existing rewiring methods, achieving up to 26.2\% improvement.
\end{abstract}

\begin{figure}[h]
    \centering
    \begin{minipage}[b]{0.53\textwidth}
        \centering
        \subfigure[ORC distribution]{\includegraphics[width=0.8\textwidth]{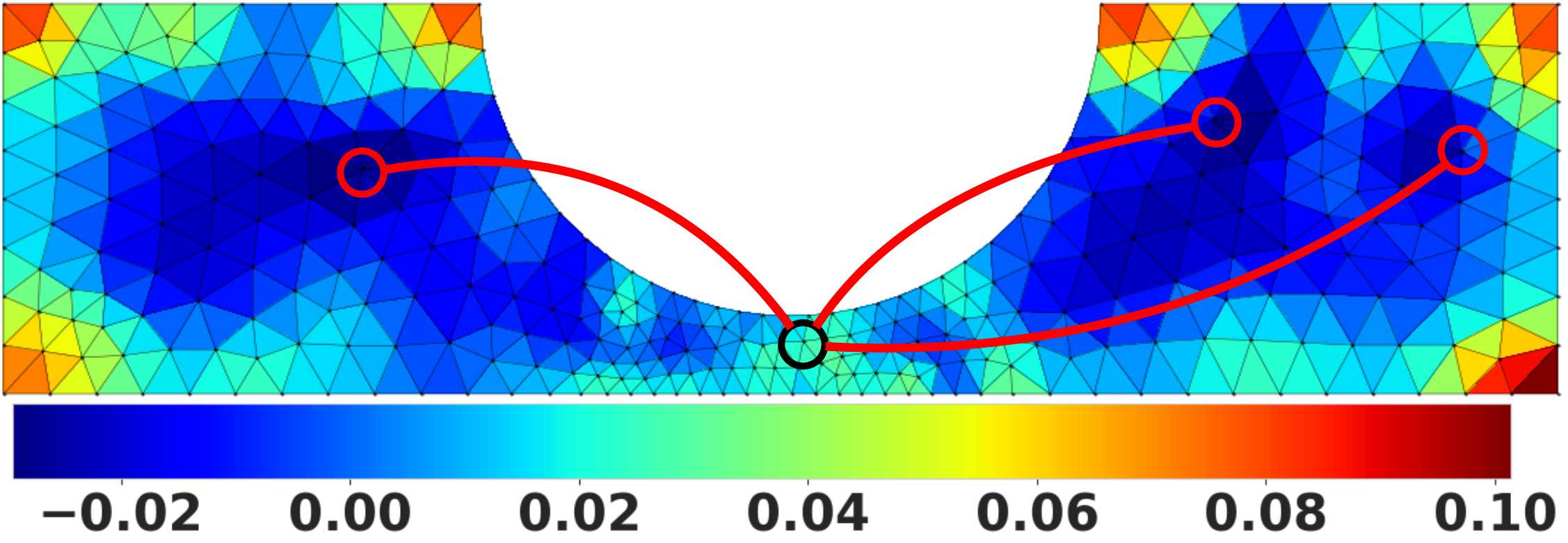}\label{fig:ansys_orc_a}}
        \vspace{-0.2cm}
        \subfigure[Velocity contour]{\includegraphics[width=0.8\textwidth]{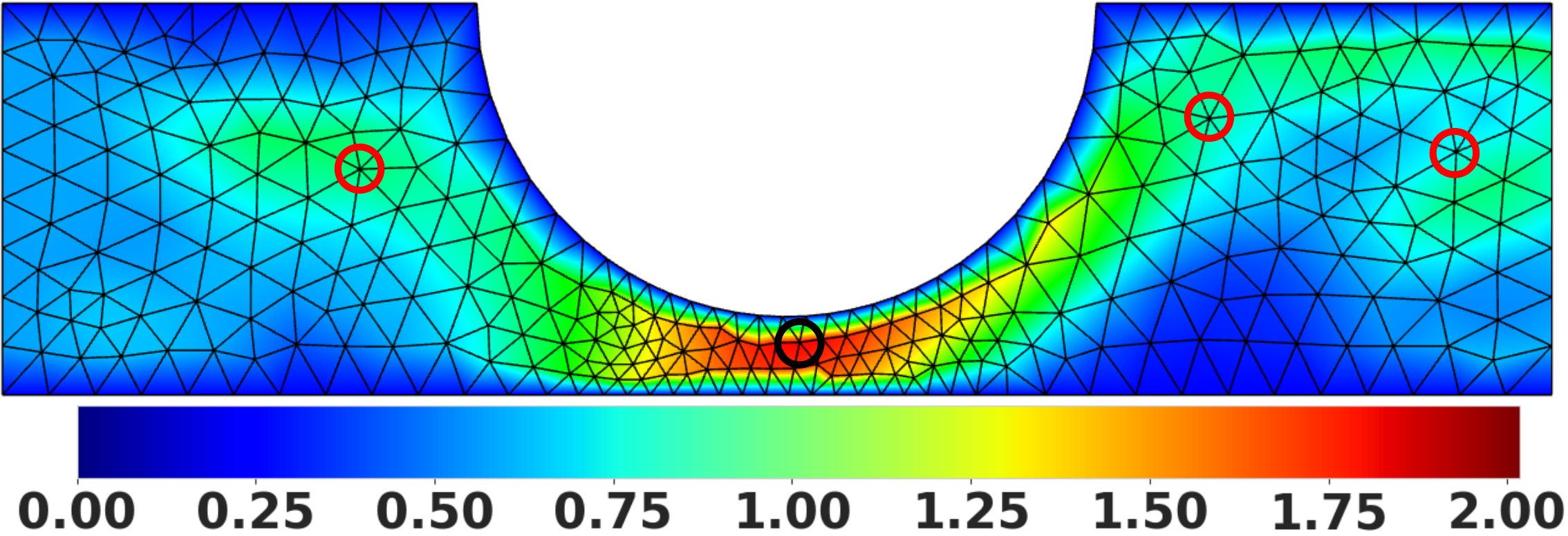}\label{fig:ansys_orc_b}}
        \caption{Visualization of PIORF rewiring in \textsc{CylinderFlow-Tiny}. (a) Blue areas indicate potential bottlenecks. Red circles (\myredcircle) denote critical bottleneck nodes. (b) The black circle (\myblackcircle) denotes the highest velocity node. PIORF connects bottleneck nodes (\myredcircle) with high-velocity nodes (\myblackcircle).}
        \label{fig:ansys_orc}
    \end{minipage}
    \hfill
    \begin{minipage}[b]{0.44\textwidth}
        \centering
        \includegraphics[width=\textwidth]{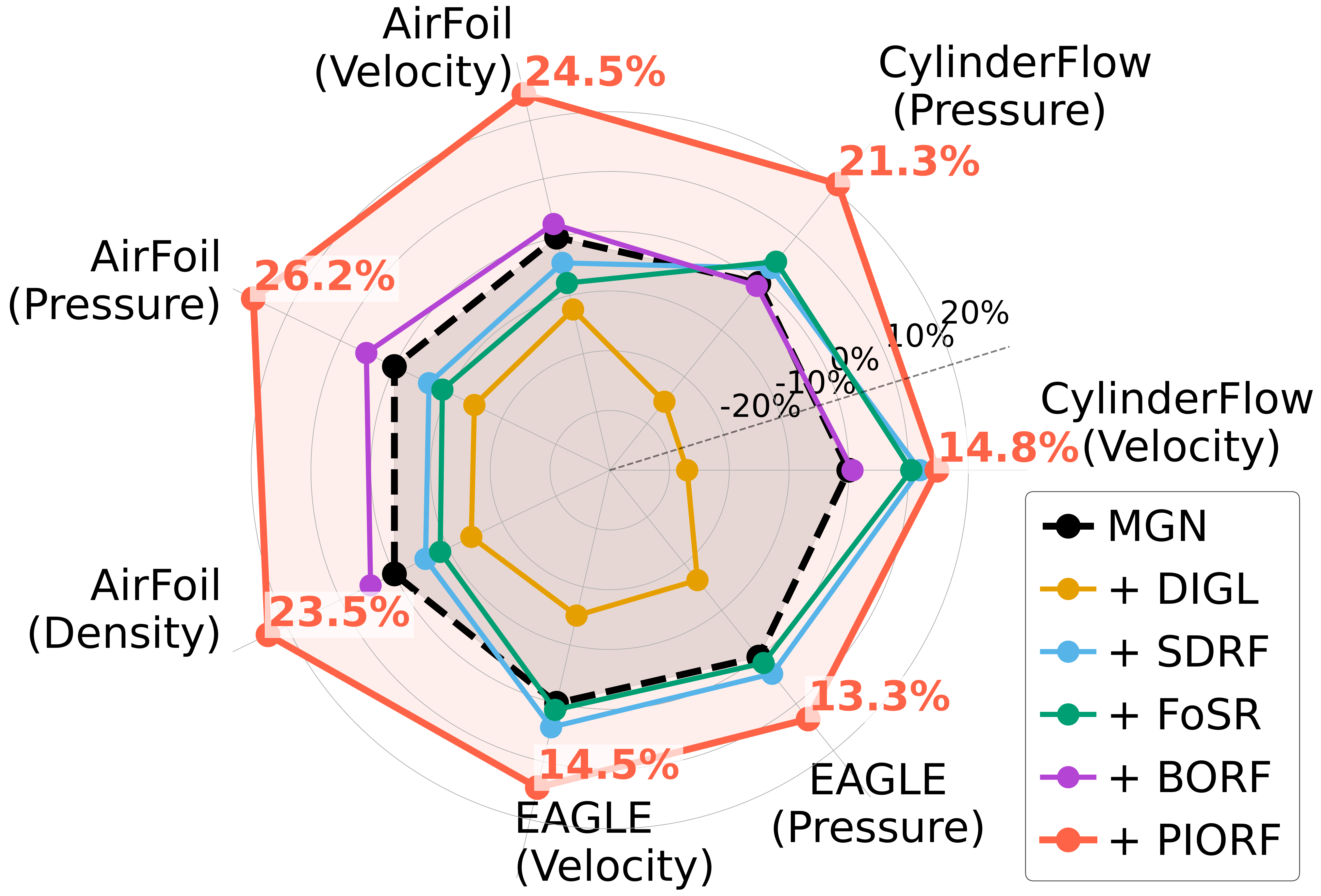}
        \vspace{-0.2cm}
        % \caption{Performance comparison of PIORF against other rewiring methods on MGN. The plot shows the percentage improvement over MGN for each method.}
        \caption{The radar plot shows the percentage improvement over MGN for each method on 3 datasets. The radial distance indicates the magnitude of improvement. PIORF consistently outperforms other methods with substantial gains particularly in \textsc{AirFoil} (24.5\% for Velocity) and \textsc{CylinderFlow} (21.3\% for Pressure).}
        \label{fig:radar}
    \end{minipage}
\end{figure}

\section{Introduction}
Solving the Navier--Stokes equations that govern fluid dynamics remains an open problem. In the absence of an analytical solution, most studies use numerical methods, representatively, finite element methods (FEMs)~\citep{madenci2015finite, stolarski2018engineering, abaqus2011abaqus, dhatt2012finite} to discretize differential equations spatially and temporally to account for complex physics. 
To optimize computational resources while maintaining accuracy in simulations involving unstructured surfaces, mesh refinement techniques are commonly used. These methods allocate higher resolution to regions of interest that require more detailed analysis, such as areas with steep gradients or complex geometries. 
While this approach balances computational cost with simulation accuracy, it results in a complex and irregular mesh structure~\citep{lohner1995mesh,liu2022sph}.

The high computational cost of traditional numerical solutions has sparked interest in data-driven simulators based on graph neural networks (GNNs). Graph machine learning approaches, particularly MeshGraphNets (MGNs)~\citep{pfaff2020mgn}, have shown promising results in modeling physical systems on unstructured meshes. So far, studies using MGNs have shown accurate predictions for various physical systems~\citep{sanchez2020learning,fortunato2022msmgn,yu2024hcmt}. However, these methods face the challenge of capturing the long-range dependence of fluid flows, which is essential for accurately simulating complex phenomena such as turbulence~\citep{benzi2023lectures}.

\paragraph{Mesh refinement and over-squashing problem.}
The core problem in using GNNs for fluid dynamics simulations lies in balancing mesh refinement and information propagation. 
To achieve accurate simulations, it is essential to use finer meshes, especially in regions with significant velocity gradients, such as in boundary conditions (e.g. walls, holes, inlets, and outlets)~\citep{katz2011mesh, baker2005relationship}. 
However, this refinement introduces two critical issues:
i) as information propagates through the graph, it is repeatedly compressed, leading to an `over-squashing' problem~\citep{alon2021oversquashing,topping2021understanding}. The over-squashing occurs in areas of local mesh refinement~\citep{imai2006higher} and near boundary conditions where the mesh is non-uniform, resulting in some nodes having few neighbors.
ii) As the mesh becomes finer, MGNs need to perform more message-passing steps to propagate information over the same physical distance. This leads to `under-reaching' problems~\citep{fortunato2022msmgn}, where the model struggles to capture interactions beyond a certain range.
These issues are particularly pronounced in fluid dynamics simulations. As the mesh becomes finer, the challenges increase, creating a trade-off between the demand for high-resolution simulations and the capacity of GNNs to efficiently process the graphs.

\paragraph{Limitations of existing solutions.}
While several graph rewiring methods have been proposed to address over-squashing~\citep{topping2021understanding,karhadkar2022fosr,nguyen2023borf,black2023understanding,arnaiz2022diffwire}, they typically consider only the graph topology. This approach is insufficient for fluid dynamics simulations, where the underlying physical phenomena play a crucial role in determining important long-range interactions.

\paragraph{Main idea.}
To address these challenges, we propose Physics-Informed Ollivier--Ricci Flow (PIORF)\footnote{Our code is available here: \url{https://github.com/yuyudeep/piorf}}, a novel method that incorporates physical quantities such as flow velocity into graph rewiring. PIORF uses graph topology and physical phenomena to reduce over-squashing and enhance information flow. We use the Ollivier--Ricci curvature (ORC)~\citep{ollivier2009ricci} to identify bottleneck regions in the graph structure.
\Cref{fig:ansys_orc} depicts the key idea behind our PIORF using a \textsc{CylinderFlow-Tiny} simulation. The ORC distribution (\Cref{fig:ansys_orc_a}) reveals potential bottleneck areas (blue regions), with red circles (\myredcircle) marking nodes of minimum curvature. The velocity magnitude contour (\Cref{fig:ansys_orc_b}) shows areas of rapid fluid velocity changes, with the black circle (\myblackcircle) indicating the highest velocity node. Our approach connects these bottleneck nodes with nodes in high-velocity gradient regions, enabling long-range interactions and mitigating over-squashing.

\paragraph{Contributions.} Our contributions are summarized as follows:
\begin{itemize}[leftmargin=10pt]
    \item To the best of our knowledge, we are the first to introduce a rewiring method that considers both graph topology and physical phenomena for fluid dynamics simulations.
    \item Our PIORF method shows excellent computational efficiency by adding multiple edges with a single calculation compared to existing rewiring methods.
    \item We extend PIORF to handle temporal mesh graphs and apply it to dynamic simulation environments such as the \textsc{EAGLE} dataset, demonstrating the scalability of PIORF to larger mesh graphs.
    \item As shown in \Cref{fig:radar}, PIORF consistently outperforms MGN model and other rewiring methods across 3 benchmark datasets, achieving up to 26.2\% improvement.
\end{itemize}

\section{Related Work}
\subsection{Mesh-based Simulation Models} 
Using GNNs to predict the results of complex physical systems is a popular area of scientific machine learning (SciML)~\citep{li2020fourier, michalowska2023neural,belbute2020combining,mrowca2018flexible,li2019propagation,li2018learning, pfaff2020mgn}. Among them, MGN performs local message passing by re-expressing it as a graph from a mesh. The strength of MGN lies in its ability to use mesh-based representations commonly used in many commercial simulation tools to numerically solve partial differential equations (PDEs). Instead of solving the PDEs directly, MGN learns the underlying dynamics from data and can be applied to a variety of systems while incorporating boundary conditions. However, in order to obtain a more accurate solution approximate, MGN often requires finer meshes. 
A larger number of nodes causes the GNN's under-reaching problem and requires more layers for effective long-range interactions, which reduces learning efficiency. To address this, recent studies have investigated methods to enable long-range interaction by forming a hierarchical structure~\citep{fortunato2022msmgn, cao2023bistride} or using a Transformer~\citep{janny2023eagle, yu2024hcmt}. \citet{fortunato2022msmgn} introduce a dual-layer structure designed to propagate messages at two different resolutions. \citet{janny2023eagle} proposes a clustering-based pooling method and performs global self-attention. \citet{cao2023bistride} reviews the shortcomings of current pooling methods and proposes Bi-Stride Multi-Scale (BSMS), a hierarchical GNN using bi-stride pooling. \citet{yu2024hcmt} use hierarchical mesh graphs and has an ability to capture long-range interactions between spatially distant locations within an object.

\subsection{Over-squashing and Graph Rewiring Methods}
The issue of over-squashing was initially identified by \citet{alon2021oversquashing} and has since emerged as a significant challenge in GNNs when dealing with long-range dependencies. This phenomenon occurs when the information aggregated from a large number of neighbors is compressed into a fixed-sized node feature vector, resulting in a considerable loss of information~\citep{alon2021oversquashing}.
Several approaches have been studied to address the over-squashing problem in GNNs~\citep{finkelshtein2023cooperative,shi2023exposition,errica2023adaptive,choi2024panda,fesser2024mitigating,choi2025fractalinspired}. 
While alternative message-passing strategies, such as expanded width-aware message passing~\citep{choi2024panda}, have gained attention, graph rewiring -- adding or removing edges -- has been the most actively proposed~\citep{gasteiger2019diffusion,topping2021understanding,nguyen2023borf,arnaiz2022diffwire,karhadkar2022fosr,black2023understanding,banerjee2022oversquashing,attali2024delaunay}.
\citet{gasteiger2019diffusion} propose DIGL rewiring method that computes kernel evaluation and sparsification of the adjacency matrix. DIGL smooths the adjacency of the graph, which makes it tend to connect nodes at short distances~\citep{coifman2006diffusion}.
However, this makes it not suitable for tasks that require longer diffusion distances. \citet{topping2021understanding} propose a curvature-based graph rewiring strategy. This method identifies edges with minimal negative curvature and adds new edges around them. First-order spectral rewiring (FoSR) proposed by \citet{karhadkar2022fosr} calculates the change in spectral gap due to edge addition and selects the edge that maximizes the gap. \citet{nguyen2023borf} propose batch Ollivier--Ricci flow (BORF) using ORC to simultaneously solve the over-smoothing and over-squashing problems. BORF works in batches and calculates the curvature with a minimum and maximum in each batch. Then, connections are added to the set with the minimum edge value to uniformly weaken the graph bottleneck. BORF does not recalculate the graph curvature within each batch, but rather reuses the already computed optimal transfer plan between sets to determine which edges should be added. Recently, \citet{attali2024delaunay} alleviate over-squashing by using Delaunay triangulation, but this is not appropriate because mesh-based simulations are already constructed by the triangulation.

Despite interest in over-squashing in GNNs, over-squashing in mesh-based GNNs such as MGN remains unexplored (See \Cref{tab:comp}). Since mesh structures have different characteristics from graph structures used in existing research, and existing rewiring methods define bottlenecks for graph topologies from a geometric perspective, it is necessary to verify that existing rewiring methods are suitable for mesh graphs with a certain number of distributed edges.

\section{Preliminaries}
\subsection{MeshGraphNets (MGN)}
MGNs~\citep{pfaff2020mgn} are a class of GNNs designed for mesh-based simulation, using an Encoder-Processor-Decoder framework.
The encoder encodes as multigraph, the nodes of the mesh are converted to graph nodes, and the mesh edges become bidirectional mesh-edges. 
The processor updates all node and edge embeddings by performing multiple message passing along the mesh edges through multiple GraphNet blocks~\citep{sanchez2020learning}. 
% Each block consists of modules that sequentially update the nodes and edges from the output of the previous block.
Finally, the decoder predicts the subsequent state by using the updated latent node representations.

\paragraph{Encoder.}
The mesh $\mathcal{M}^t$ at time $t$ is transformed into a graph $\mathcal{G}=(\mathcal{V},\mathcal{E})$, where the mesh nodes become graph nodes $v_i\in\mathcal{V}$, and the mesh edges become bidirectional edges $(i,j)\in \mathcal{E}$.
For each edge, we define the mesh edge feature $\mathbf{m}_{ij}$, which encodes connectivity information.
The edge features are derived from the relative displacement vector $\mathbf{x}_{ij}=\mathbf{x}_i-\mathbf{x}_j$ and its norm $|\mathbf{x}_{ij}|$.
Node features include the velocity $\mathbf{w}_{i}$ and the node type $\mathbf{n}_{i}$, which indicates the boundary conditions.
The input and output characteristics for each dataset are detailed in \Cref{app:models}.

\paragraph{Processor.}
The processor consists of several GraphNet blocks. Each block sequentially updates node and edge embeddings through message passing operations.
$\mathbf{v}_i^l$ and $\mathbf{e}_{ij}^l$ denote the node and edge embeddings at layer $l$, respectively. The update equations are:
\begin{align}
\mathbf{e}_{ij}^{l+1} = f_E(\mathbf{e}_{ij}^l, \mathbf{v}_i^l, \mathbf{v}_j^l), \quad
\mathbf{v}_i^{l+1} = f_V \Bigl(\mathbf{v}_i^l, \sum_{j \in \mathcal{N}_{i}} \mathbf{e}_{ij}^{l+1} \Bigr),
\end{align}
where $f_E$ and $f_V$ are learnable functions parameterized as multi-layer perceptrons (MLPs), and $\mathcal{N}_{i}$ denotes the set of neighbors of node $i$.

\paragraph{Decoder and updater.}
To predict the next time state from the current time, an MLP decoder is used to predict one or more output features $\mathbf{o}_i$, such as the velocity gradient $\hat{\dot{\mathbf{w}_{i}}}$, density gradient $\hat{\dot{\rho_i}}$ and the next pressure $\hat{p_i}$. The velocity gradient is used to calculate the next velocity $\hat{\mathbf{w}}^{t+1}_i$ through an updater, which performs a first-order integration ($\hat{\mathbf{w}}^{t+1}_i=\hat{\dot{\mathbf{w}}}^t_i+\mathbf{w}^t_i$).

\paragraph{Training loss.}
Following the MGN approach, the training loss uses the mean squared error (MSE):
\begin{align}
    \mathcal{L} = \frac{1}{|\mathcal{V}|}\sum_{i=1}^{|\mathcal{V}|} (\mathbf{w}^{t+1}_i- \hat{\mathbf{w}}^{t+1}_i)^2 + \frac{1}{|\mathcal{V}|}\sum_{i=1}^{|\mathcal{V}|} (p^{t+1}_i- \hat{p}^{t+1}_i)^2,
\end{align}
where $|\mathcal{V}|$ is the number of nodes.
% [Jaehyeon] |mathcal{V}|로 통일해도 되는지 확인하고 수정하겠습니다.
% [KL] 네 좋습니다. 지금 거의 page limit에 딱 맞춰져 있는거 같아서 최대한 길이를 늘리지 않는 방향으로 진행해주세요.
% 넵 알겠습니다.

\subsection{Ollivier--Ricci Curvature on Graphs}
Ricci curvature, a fundamental concept in differential geometry, describes the average dispersion of geodesics in the local region of a Riemannian manifold. 
In the context of graphs, ORC~\citep{ollivier2009ricci} extends these concepts to graphs and considers random walks between nearby points using Wasserstein distances between Markov chains. 

Given a graph $G=(\mathcal{V},\mathcal{E})$ and a pair of nodes $i,j \in \mathcal{V}$, ORC $\kappa(i,j)$ of edge $ (i,j) \in \mathcal{E}$ is defined as:
\begin{align}\label{eq:orc}
    \kappa(i,j) = 1 - \frac{W_1(\mathbf{m}_i ,\mathbf{m}_j)}{d(i,j)},
\end{align}
where $d(i,j)$ is the shortest-path distance between nodes $i$ and $j$, $\mathbf{m}_i$ is probability distribution of 1-step random walk from node $i$, and $W_1$ is the Wasserstein distance of order 1. For a node $p \in \mathcal{V}$, $\mathbf{m}_i(p)$ represents the probability that a random walker starting at $i$ will reach $p$ in one step.
The Wasserstein distance $W_1(\mathbf{m}_i, \mathbf{m}_j)$ between probability distributions $\mathbf{m}_i$ and $\mathbf{m}_j$ is defined as:
\begin{align}\label{eq:wasserstein}
    W_1(\mathbf{m}_i, \mathbf{m}_j) = \inf_{\pi \in \Pi(\mathbf{m}_i, \mathbf{m}_j)} \left(\sum_{(p,q) \in \mathcal{V}^2} \pi(p,q)d(p,q)\right),
\end{align}
where $\Pi(\mathbf{m}_i, \mathbf{m}_j)$ is the set of joint probability distributions with marginals $\mathbf{m}_i$ and $\mathbf{m}_j$.
% where $\pi\in\Pi(\mu_i,\mu_j)$ is the family of joint probability distributions of $\mathbf{m}_i$ and $\mathbf{m}_j$.

ORC quantifies the variance of a geodesic and has positive, negative, and zero values. When it is 0 ($\kappa(i,j)=0$), the geodesics tend to remain parallel, when it is a negative value ($\kappa(i,j)<0$), they diverge, and when it is a positive value ($\kappa(i,j)>0$), they converge. ORC on edges with high negative values is known to cause over-squashing~\citep{topping2021understanding}. \Cref{eq:wasserstein} requires defining the probability distribution for function neighbor nodes. Since the radius of the neighbor nodes in the graph is 1, a given one-step random walk $\mathbf{m}$ from node $i$ to node $p$ is defined as:
\begin{align}
    \mathbf{m}_i(p) = \begin{cases}
    \frac{1}{\text{deg} (i)} & \text{if}\; p \in \mathcal{N}_{i}, \\
    0 & \text{otherwise},
    \end{cases}
\end{align}
where $\text{deg}(i)$ is the degree of node $i$, which means the number of element in $\mathcal{N}_{i}$.

\section{PIORF: Physics-Informed Ollivier--Ricci Flow}\label{sec:method}
In this section, we introduce PIORF, a novel rewiring method to improve long-range dependencies.

\paragraph{Design Goals.} Our proposed method is designed with the following 3 goals:
\begin{itemize}[leftmargin=10pt]
    \item (\emph{Physical Context}) The method should incorporate physical quantities (e.g., velocity) with topology (e.g., ORC) to improve long-range interactions.
    \item (\emph{Efficiency}) The computational cost of adding new edges should be lower than that of existing rewiring methods.
    \item (\emph{Accuracy}) The prediction error should be lower than that of other rewiring methods.
\end{itemize}

\begin{algorithm}[b!]
   \caption{Physics-Informed Ollivier--Ricci Flow (PIORF)}
   \label{alg:algorithm}
\begin{algorithmic}[1]
    \STATE {\bfseries Input:} A graph $\mathcal{G}=(\mathcal{V},\mathcal{E})$, pooling ratio $\delta \in (0,1)$, velocity $w_i$ of node $i$
    \STATE {\bfseries Output:} Rewired graph $\mathcal{G}'=(\mathcal{V},\mathcal{E}')$ 
    \STATE Calculate the ORC, $\gamma_i$, for all nodes $i$ in $\mathcal{V}$ using \Cref{eq:orc_node}
    \STATE Selects $\lfloor \delta |\mathcal{V}| \rfloor$ nodes with the lowest $\gamma_i$ in order to form a set $S$, where $|S| = \lfloor \delta |\mathcal{V}| \rfloor$. %Sort nodes by $\gamma_i$ in ascending order and 
    \STATE For each node $s \in S$, calculate the Euclidean distance $d(w_s, w_i)$ between velocities of $s$ and all other nodes $i \in \mathcal{V} \setminus {s}$.
    
    \STATE Find the node $r = \argmax_{i \in \mathcal{V} \setminus {s}}{d(w_s, w_i)}$ with the largest velocity differences.
    \STATE Add bidirectional edges $(s,r)$ and $(r,s)$ to $\mathcal{E}'$.
    \RETURN $\mathcal{G}' = (\mathcal{V}, \mathcal{E}')$
\end{algorithmic}
\end{algorithm}

\paragraph{Rewiring with PIORF.}
To achieve these goals, PIORF selects nodes based on their topological properties and physical quantities.  We extend the ORC to node-level curvature, denoted as $\gamma_i$ for a node $i$. This node curvature $\gamma_i$ is computed as:
\begin{align}\label{eq:orc_node}
    \gamma_{i}=\frac{1}{|\mathcal{N}_{i}|}\sum_{j \in \mathcal{N}_{i}}\kappa(i,j).
\end{align}
The rewiring proceeds as \Cref{alg:algorithm}, of which the key steps are described as follows:
\begin{enumerate}[label=\roman*),leftmargin=15pt]
    \item PIORF selects $\lfloor \delta |\mathcal{V}| \rfloor$ nodes with the lowest $\gamma_i$ from $\mathcal{V}$ in order to form a set $S$, so that $|S| = \lfloor \delta |\mathcal{V}| \rfloor$ where $\delta \in (0,1)$ is the pooling ratio. $\delta$ is our \emph{sole hyperparameter}.
    \item For each $s \in S$, PIORF computes the Euclidean distance $d(w_s, w_i)$ between velocities $w_s$ and $w_i$ for all nodes $i \in \mathcal{V} \setminus {s}$.
    \item For each $s \in S$, PIORF identifies nodes $r = \argmax_{i \in \mathcal{V} \setminus {s}}{d(w_s, w_i)}$  with the largest velocity differences and defines their set as $R_s$.
    \item PIORF adds bidirectional edges $(s,r)$ and $(r,s)$ to the graph $\mathcal{G}$ for all $s \in S$ and $r \in R_s$.
\end{enumerate}
For the sake of convenience in explanation, the physical quantity used in PIORF is described based on the use of velocity from the node features. By integrating both physical and topological properties, PIORF enhances long-range interactions and mitigates over-squashing. The detailed description of the notations used in the formulas written so far is summarized in \Cref{app:notations}.

\paragraph{Physical interpretation.}
In fluid dynamics, the distinction between laminar~\citep{schubauer1947laminar} and turbulent~\citep{mathieu2000introduction} flows, as quantified by velocity and the Reynolds numbers~\citep{lissaman1983low}, is important for understanding system behavior. The relationship between velocity and pressure is described through the rate of change and is explained by the Navier-Stokes equations~\citep{temam2001navier}. The velocity refers to the speed at which a fluid moves at a specific point in space. The pressure is the force exerted by a fluid per unit area on the surfaces. PIORF integrates this physical context by adding edges between nodes with significant velocity differences. This allows the model to help with long-range interactions to better simulate real-world phenomena such as fluid turbulence. Physically, connecting nodes with large velocity differences indicates regions of instability. 
Unlike existing rewiring methods, PIORF ensures that the rewiring process takes on the actual physical context of the system, leading to physically meaningful signal propagation. With this physical insight, PIORF can improve long-range interactions and prediction performance in physics-based simulations.
\paragraph{Computational efficiency.}
Unlike existing rewiring methods, which rely on greedy algorithms to iteratively add edges based on their objective functions~\citep{karhadkar2022fosr,black2023understanding}, PIORF introduces a more efficient approach. PIORF identifies nodes with significant differences in physical quantities and adds new edges in a single pass. 
This avoids the high computational cost of iterative edge addition and, thus, improves scalability.
We show that PIORF is more efficient than other rewiring methods in rewiring new edges in \Cref{sec:complex}.

\section{Discussion}\label{sec:discussion}
In this section, we analyze mesh graphs and distinguish our method from graph pooling techniques.

\begin{figure}[h!]
    \centering
    \subfigure[]
    {\includegraphics[width=0.33\textwidth]{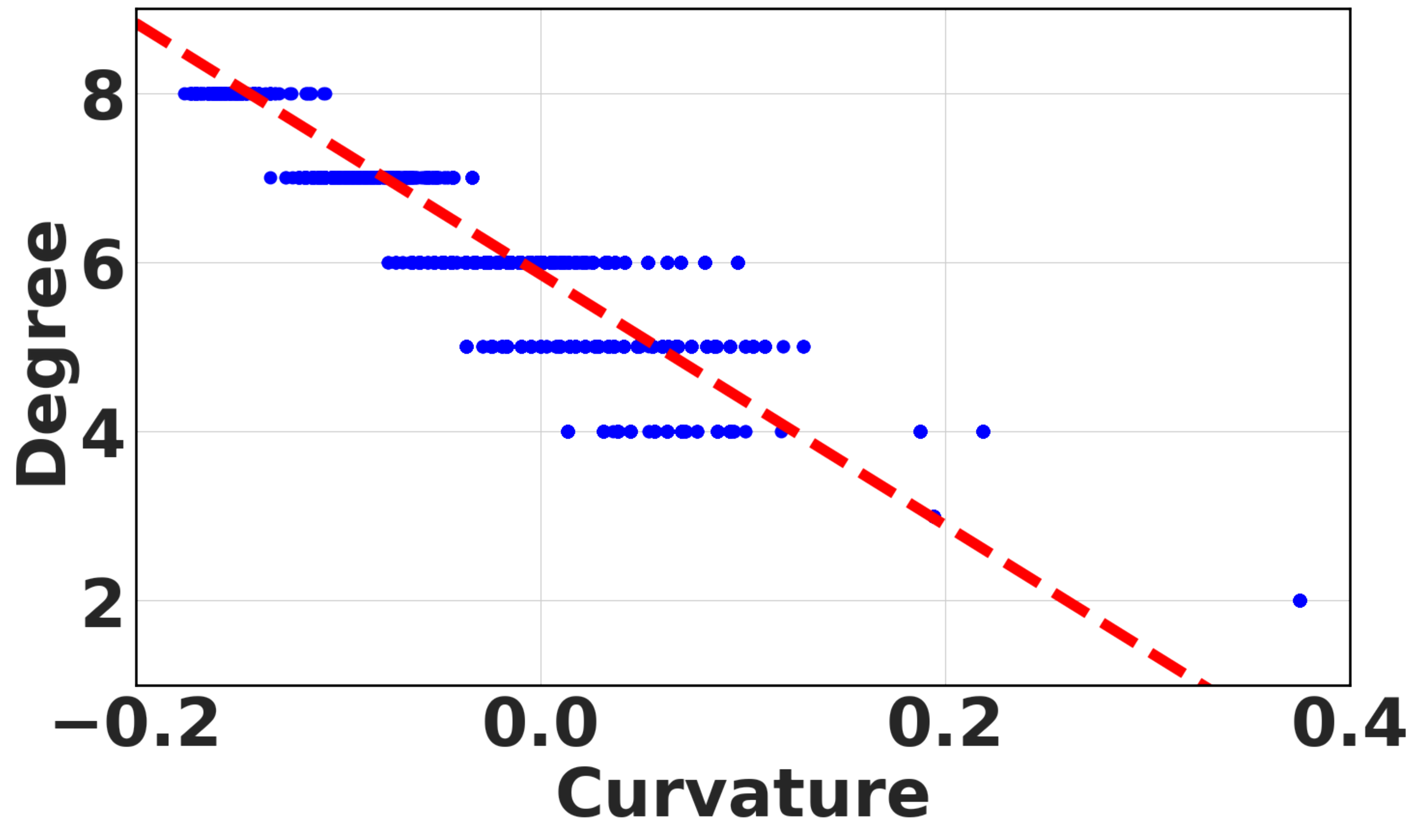}\label{fig:degrees_corr}}
    \hspace*{\baselineskip}
    \subfigure[]
    {\includegraphics[width=0.33\textwidth]{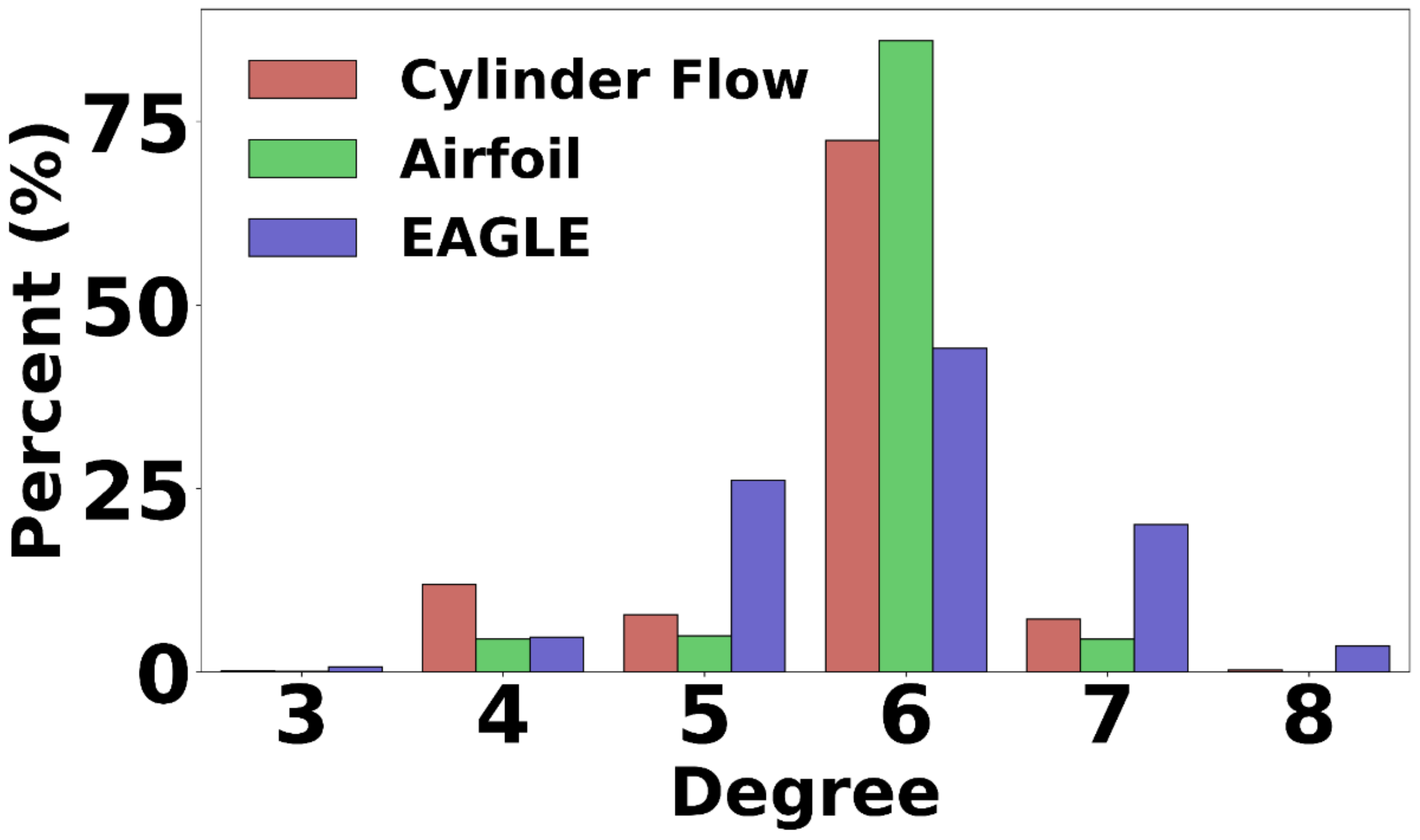}\label{fig:degrees_dist}}
    \hspace*{\baselineskip}
    \subfigure[]
    {\includegraphics[width=0.25\textwidth]{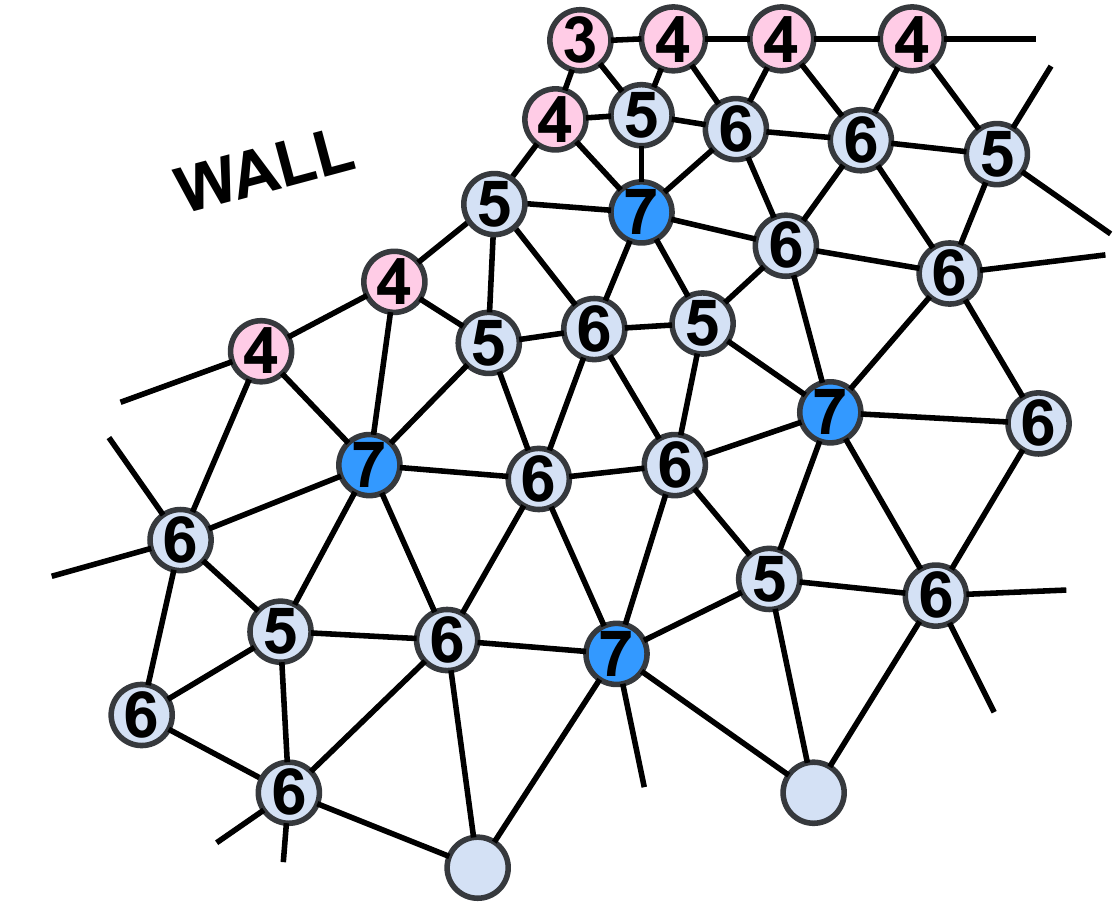}\label{fig:degrees_bound}}
    \vspace{-1em}
    \caption{Structural analyses of mesh graphs: (a) Correlation between ORC and node degree in training dataset of \textsc{CylinderFlow}, revealing potential information bottlenecks. (b) Node degree distribution across datasets, showing the prevalence of degree-6 nodes in uniform regions. (c) Non-uniform mesh refinement near boundary conditions.}
    \label{fig:degrees}
\end{figure}

\paragraph{Analysis of mesh graphs.}
We use ORC to analyze the topology of mesh graphs of fluid dynamics benchmark datasets. This analysis reveals several key insights:
\begin{itemize}[leftmargin=10pt]
    \item \Cref{fig:degrees_corr} shows a strong negative correlation between ORC and node degree. This relationship identifies potential information bottlenecks in the mesh graph, particularly in high-degree nodes.
    \item \Cref{fig:degrees_dist} indicates a prevalence of degree-6 nodes in uniform regions, typical of Delaunay triangulation~\citep{weatherill1992delaunay}. However, boundary condition nodes (e.g.,  holes, walls, inlets, and outlets) show lower degrees due to their sparse distribution, as shown in \Cref{fig:degrees_bound}.
    \item In computational fluid dynamics (CFD)~\citep{anderson1995computational}, local mesh refinement is often applied to enhance accuracy in specific areas. This process leads to a gradual transition from fine meshes near boundaries to coarser meshes, resulting in non-uniform structures (\Cref{fig:degrees_bound}).
\end{itemize}
These findings emphasize the relationship between the mesh configuration, boundary conditions, and the risk of information bottlenecks in GNNs used for fluid dynamics simulations. \Cref{fig:app_rollout_orc} in \Cref{app:datasets} shows the ORC distribution obtained through this information for each dataset.

\paragraph{Pooling and rewiring methods in mesh graphs.}
Due to mesh graphs with more than thousands of nodes, node pooling techniques \citep{fortunato2022msmgn,cao2023bistride,yu2024hcmt} are widely used to reduce computational complexity and enhance the capture of long-range interactions. We extend the application of our PIORF beyond MGN to hierarchical models such as BSMS~\citep{cao2023bistride} and HMT~\citep{yu2024hcmt}. While these models already incorporate pooling to effectively reduce the number of nodes, we hypothesize that applying PIORF to the pooled structures could further optimize edge connections. This integration of pooling and rewiring aims to refine capacity of the model to represent complex physical relationships across different scales. In \Cref{sec:exp}, we explore whether this combination can yield additional improvements in fluid dynamics benchmarks. 

\section{Experiments}\label{sec:exp}
\subsection{Experiments on Fluid Dynamics Benchmark Datasets}
\paragraph{Datasets.}
We evaluate our method on two publicly available datasets: \textsc{CylinderFlow} and \textsc{AirFoil}. Both datasets follow the Navies--Stokes equations~\citep{temam2001navier}, but differ in their flow characteristics. \textsc{CylinderFlow} shows laminar flow behavior. In contrast, \textsc{AirFoil} represents a turbulent flow model with high velocity, where fluid particles move irregularly in time and space.
(See \Cref{app:datasets} for a detailed description of datasets.)

\paragraph{Setting.} \label{par:setting1}
We compare our PIORF against rewiring methods: DIGL~\citep{gasteiger2019diffusion}, FoSR~\citep{karhadkar2022fosr}, SDRF~\citep{topping2021understanding}, and BORF~\citep{nguyen2023borf}
These are apllied to 4 different models architectures: MGN~\citep{pfaff2020mgn}, BSMS~\citep{cao2023bistride}, Graph Transformer (GT)~\citep{dwivedi2020generalization}  and HMT~\citep{yu2024hcmt}. 
BSMS is a hierarchical GNN and HMT is a hierarchical Transformer. 
For MGN, we use 15 blocks. For optimal performance, BSMS is set to level 7 for \textsc{CylinderFlow} and level 9 for \textsc{AirFoil}.
Detailed hyperparameters for all baselines are provided in \Cref{app:base}. 
Each experiment is repeated 5 times with different random seeds.
All experiments are performed on NVIDIA 3090 and Intel Core-i9 CPUs.

\begin{table}[t]
    \small
    \setlength{\tabcolsep}{10pt}
    \centering
    \caption{RMSE (rollout-all, $\times10^{3}$) for our PIORF and other rewiring methods.}
    \begin{tabular}{l cc ccc}\toprule
        \multirow{2}{*}{Method} & \multicolumn{2}{c}{\textsc{CylinderFlow}} & \multicolumn{3}{c}{\textsc{AirFoil}} \\ \cmidrule(lr){2-3}\cmidrule(lr){4-6}
         & Velocity & Pressure & Velocity & Pressure & Density \\\midrule
        MGN      & 48.8\std{5.6} & 36.7\std{2.4} & 10,261\std{832} & 3,043,186\std{282,514} & 29.4\std{2.7}\\
        \;+ DIGL & 62.0\std{1.7} & 46.0\std{0.4} & 11,534\std{623} & 3,495,260\std{252,832} & 33.6\std{2.2}\\
        \;+ SDRF & 43.0\std{3.0} & 35.5\std{1.0} & 10,714\std{669} & 3,238,730\std{183,094} & 31.1\std{1.9}\\
        \;+ FoSR & 43.7\std{3.2} & 35.0\std{1.2} & 11,068\std{377} & 3,314,506\std{164,026} & 31.9\std{1.5}\\
        \;+ BORF & 48.5\std{7.8} & 36.9\std{2.2} & 10,029\std{410} & 2,884,555\std{186,003} & 28.1\std{1.8}\\
        \cmidrule(lr){1-6} \rowcolor{gray! 20}
        \;+ PIORF & \textbf{41.6\std{3.9}} & \textbf{28.9\std{1.5}} & \textbf{7,743\std{584}} & \textbf{2,245,858\std{142,452}} & \textbf{22.5\std{1.4}}\\ 
        \midrule
        BSMS       & 78.7\std{2.8} & 50.7\std{2.2} & 10,883\std{460} & 2,640,398\std{158,480} & 26.5\std{2.1}\\
        \;+ DIGL   & 237.8\std{7.3} & 163.6\std{8.5} & 40,312\std{3,936} & 8,218,660\std{1,281,200} & 81.3\std{11.5}\\
        \;+ SDRF   & 78.0\std{4.1} & 50.7\std{1.9} & 36,539\std{3,980} & 7,426,023\std{642,555} & 74.2\std{8.1}\\
        \;+ FoSR   & 82.2\std{3.8} & 52.3\std{3.0} & 41,831\std{2,011} & 8,490,283\std{352,622} & 84.0\std{2.4}\\
        \;+ BORF   & 84.9\std{2.3} & 54.2\std{1.6} & 10,750\std{430} & 2,632,487\std{126,177} & 25.8\std{1.2}\\
        \cmidrule(lr){1-6} \rowcolor{gray! 20}
        \;+ PIORF   & \textbf{76.9\std{3.8}} & \textbf{50.6\std{2.4}} & \textbf{10,482\std{500}} & \textbf{2,584,690\std{163,680}} & \textbf{25.4\std{1.6}}\\ \midrule
        GT         & 54.3\std{7.3} & 40.0\std{2.0} & 10,002\std{218} & 2,979,573\std{99,293}  & 29.0\std{0.9}\\
        \;+ DIGL   & 68.4\std{3.6} & 49.5\std{1.0} & 11,004\std{511} & 3,331,160\std{192,098} & 32.7\std{1.9}\\
        \;+ SDRF   & 52.1\std{9.1} & 39.0\std{1.3} & 10,354\std{610} & 3,120,743\std{179,587} & 30.1\std{1.9}\\
        \;+ FoSR   & 50.7\std{8.7} & 39.3\std{1.6} & 11,211\std{868} & 3,415,094\std{312,517} & 33.6\std{3.1}\\
        \;+ BORF   & 58.9\std{9.7} & 40.6\std{2.4} &  9,830\std{416} & 2,883,648\std{136,064} & 28.6\std{1.3}\\
        \cmidrule(lr){1-6} \rowcolor{gray! 20}
        \;+ PIORF   & \textbf{48.5\std{4.5}} & \textbf{31.3\std{2.3}} &  \textbf{7,429\std{778}} & \textbf{2,124,920\std{130,279}} & \textbf{21.4\std{1.2}}\\  \midrule
        HMT        & 71.0\std{1.2} & 51.1\std{1.5} & 5,303\std{414} & 1,251,955\std{79,764} & 12.8\std{0.8}\\
        \;+ DIGL   & 76.3\std{1.7} & 54.0\std{0.7} & 5,176\std{409} & 1,232,486\std{79,250} & 12.5\std{0.8}\\
        \;+ SDRF   & 71.0\std{1.0} & 51.3\std{0.6} & 32,695\std{1,013} & 7,579,699\std{247,004} & 74.4\std{2.3}\\
        \;+ FoSR   & 72.1\std{1.5} & 52.3\std{1.1} & 35,474\std{1,011} & 8,137,115\std{231,038} & 79.4\std{2.0}\\
        \;+ BORF   & 74.2\std{3.1} & 53.6\std{1.2} & 5,591\std{416} & 1,306,555\std{82,509} & 13.3\std{0.8}\\
        \cmidrule(lr){1-6} \rowcolor{gray! 20}
        \;+ PIORF   & \textbf{70.9\std{1.6}} & \textbf{50.9\std{0.8}} & \textbf{4,961\std{378}} & \textbf{1,182,495\std{67,499}} & \textbf{12.1\std{0.7}}\\ 
        \bottomrule
    \end{tabular}
    \label{tab:maintable}
\end{table}

\begin{figure}[t!]
    \centering
    \subfigure[Ground Truth]{\includegraphics[width=0.235\textwidth]{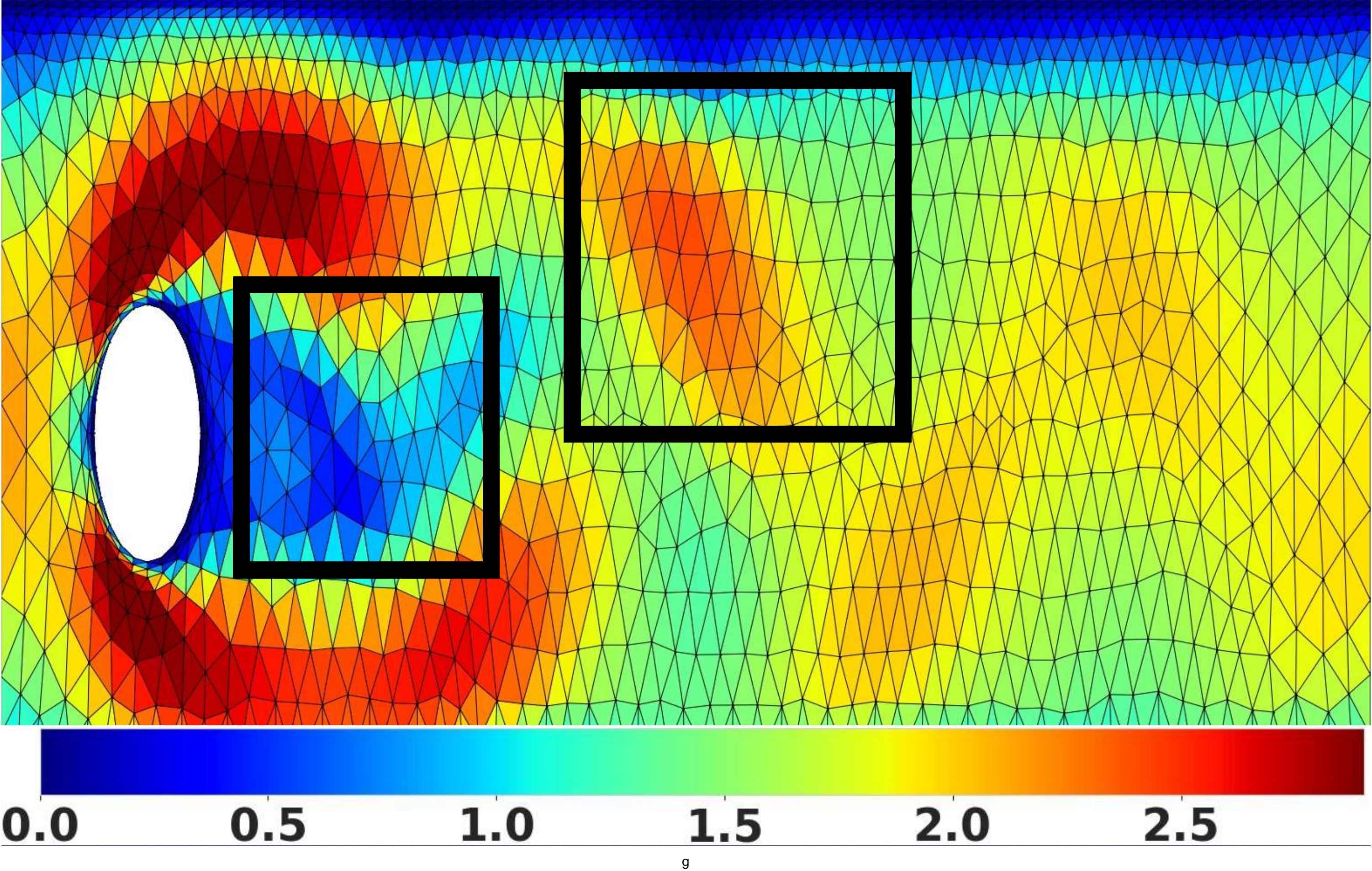}}
    \subfigure[MGN + PIORF]{\includegraphics[width=0.235\textwidth]{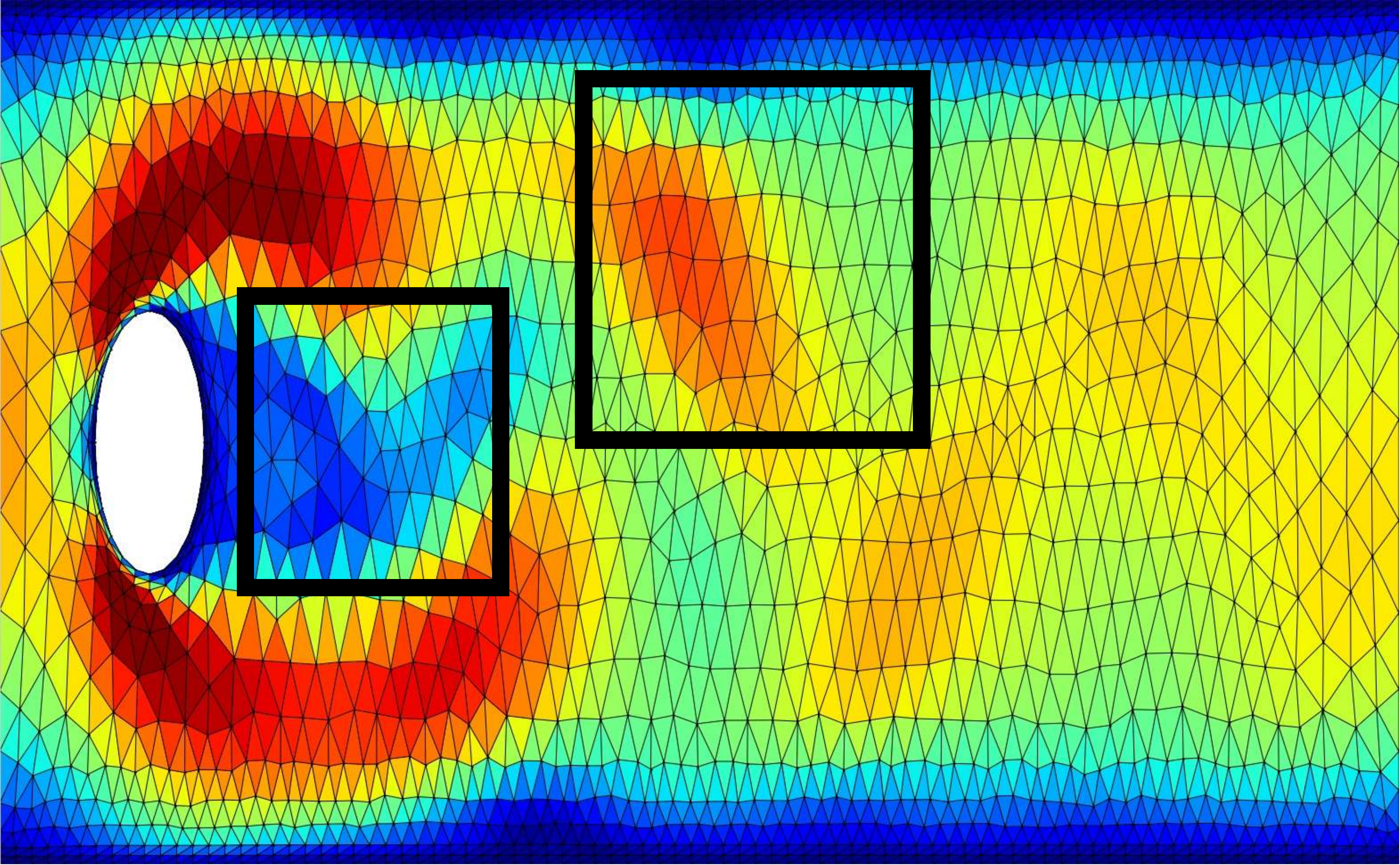}}
    \subfigure[MGN + BORF]{\includegraphics[width=0.235\textwidth]{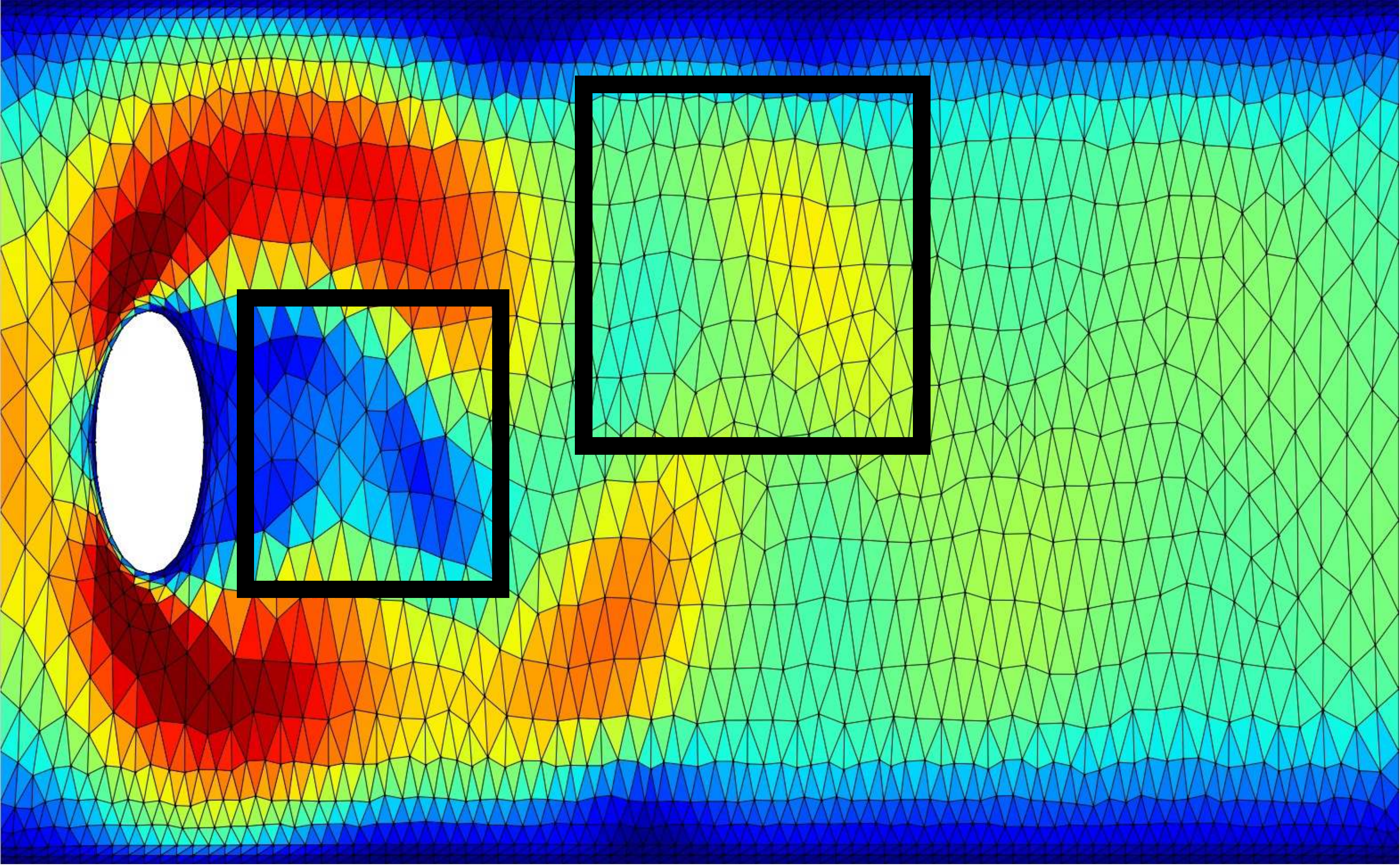}}
    \subfigure[MGN]{\includegraphics[width=0.235\textwidth]{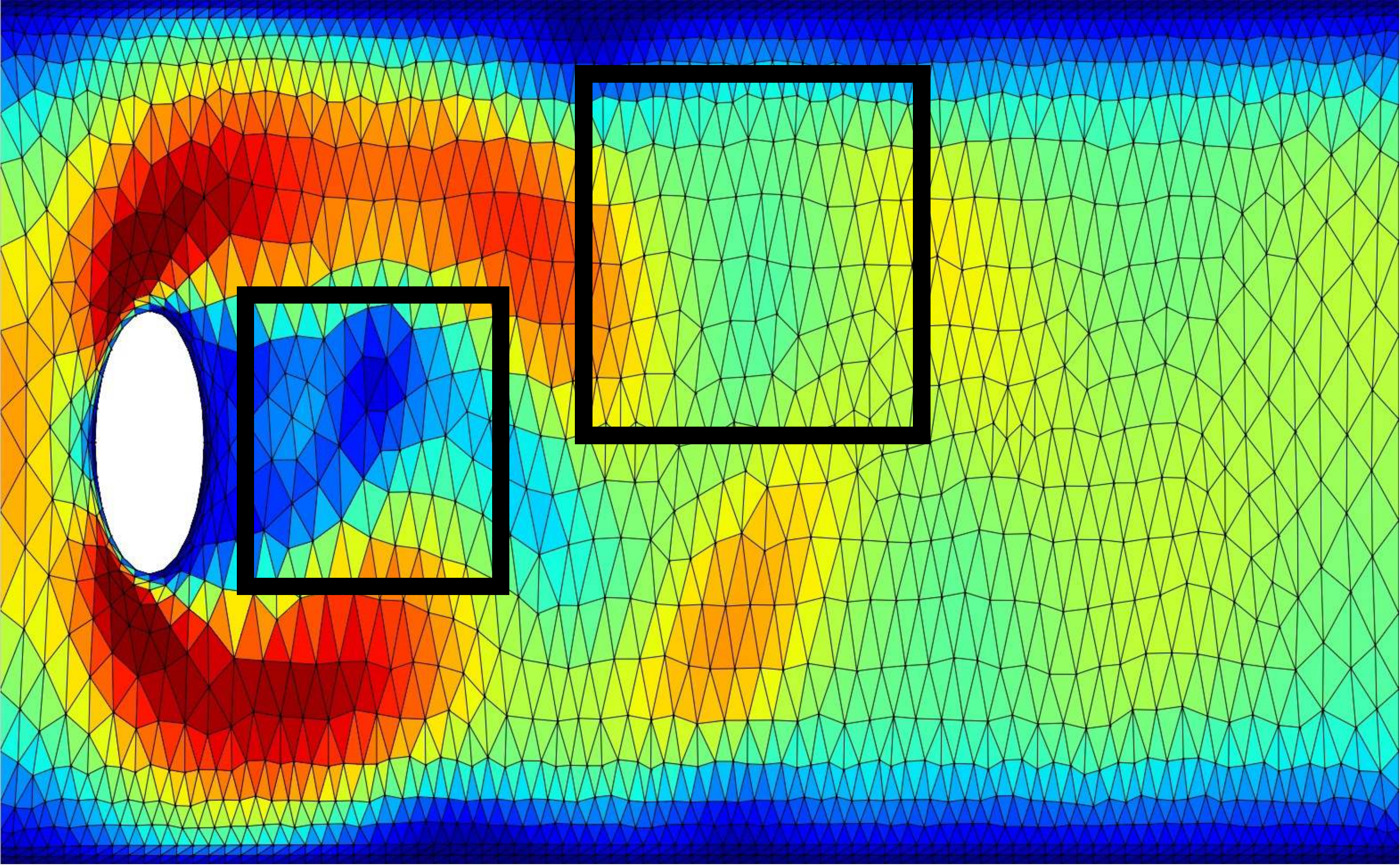}}
    \subfigure[Ground Truth]{\includegraphics[width=0.235\textwidth]{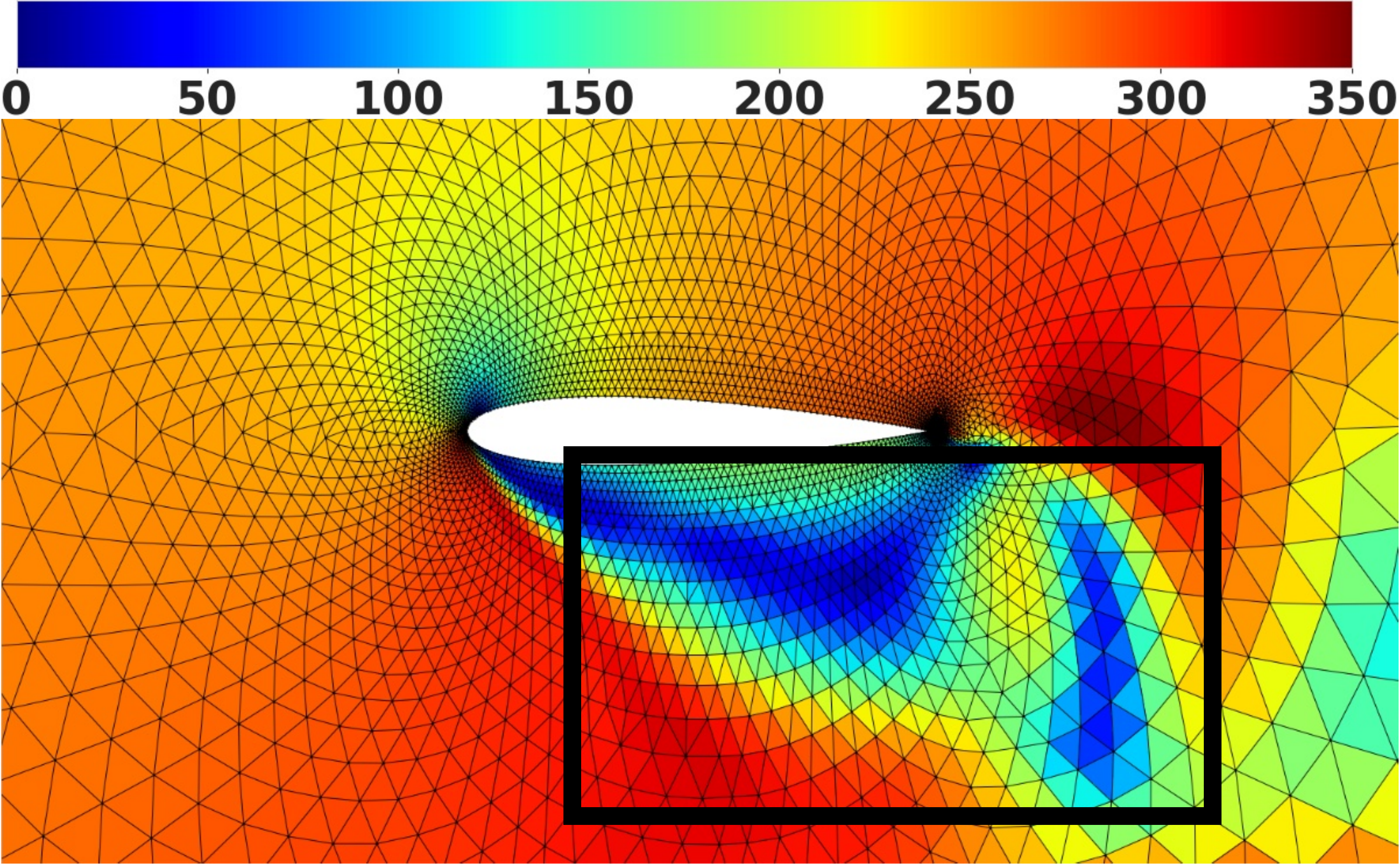}}
    \subfigure[MGN + PIORF]{\includegraphics[width=0.235\textwidth]{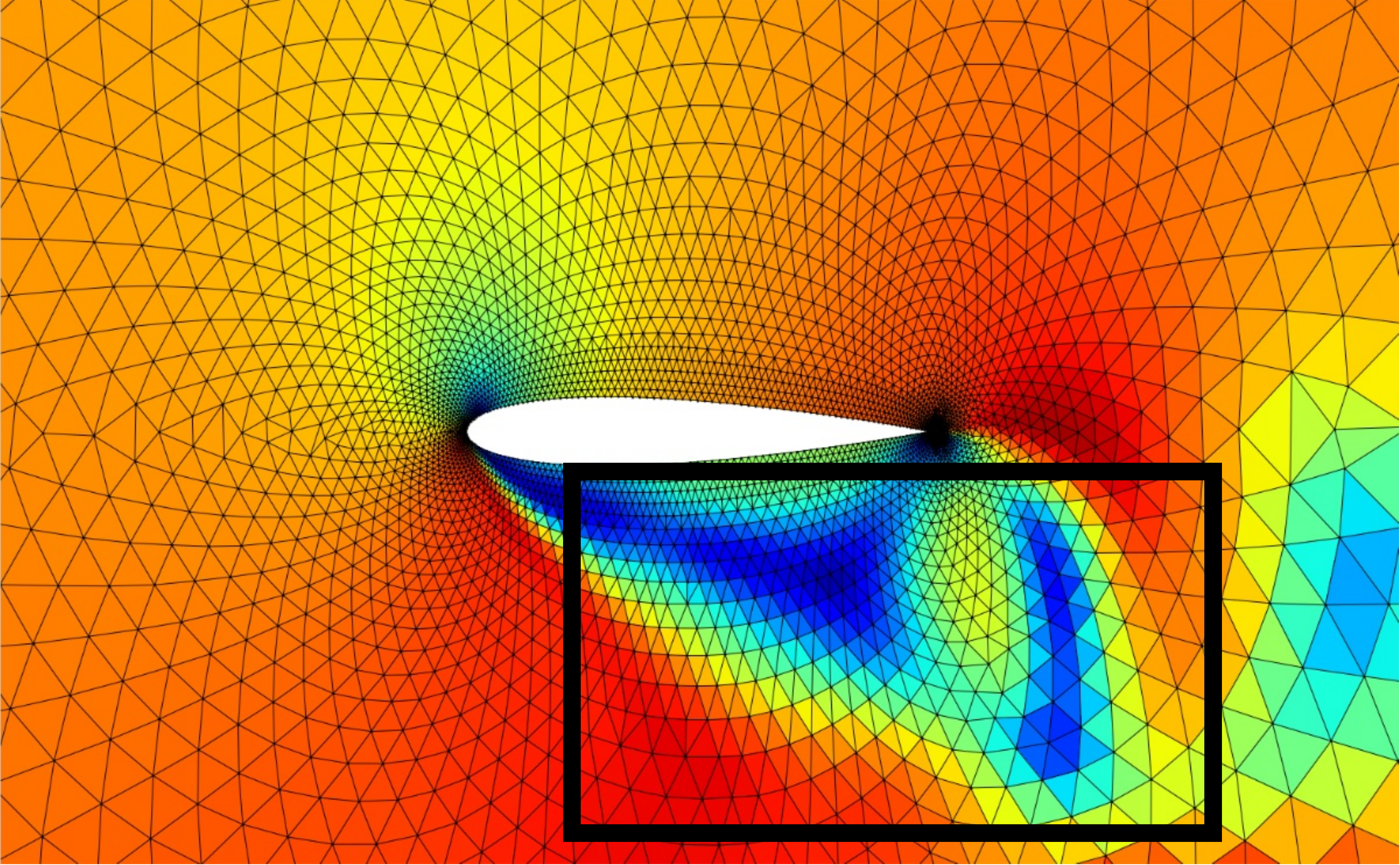}}
    \subfigure[MGN + BORF]{\includegraphics[width=0.235\textwidth]{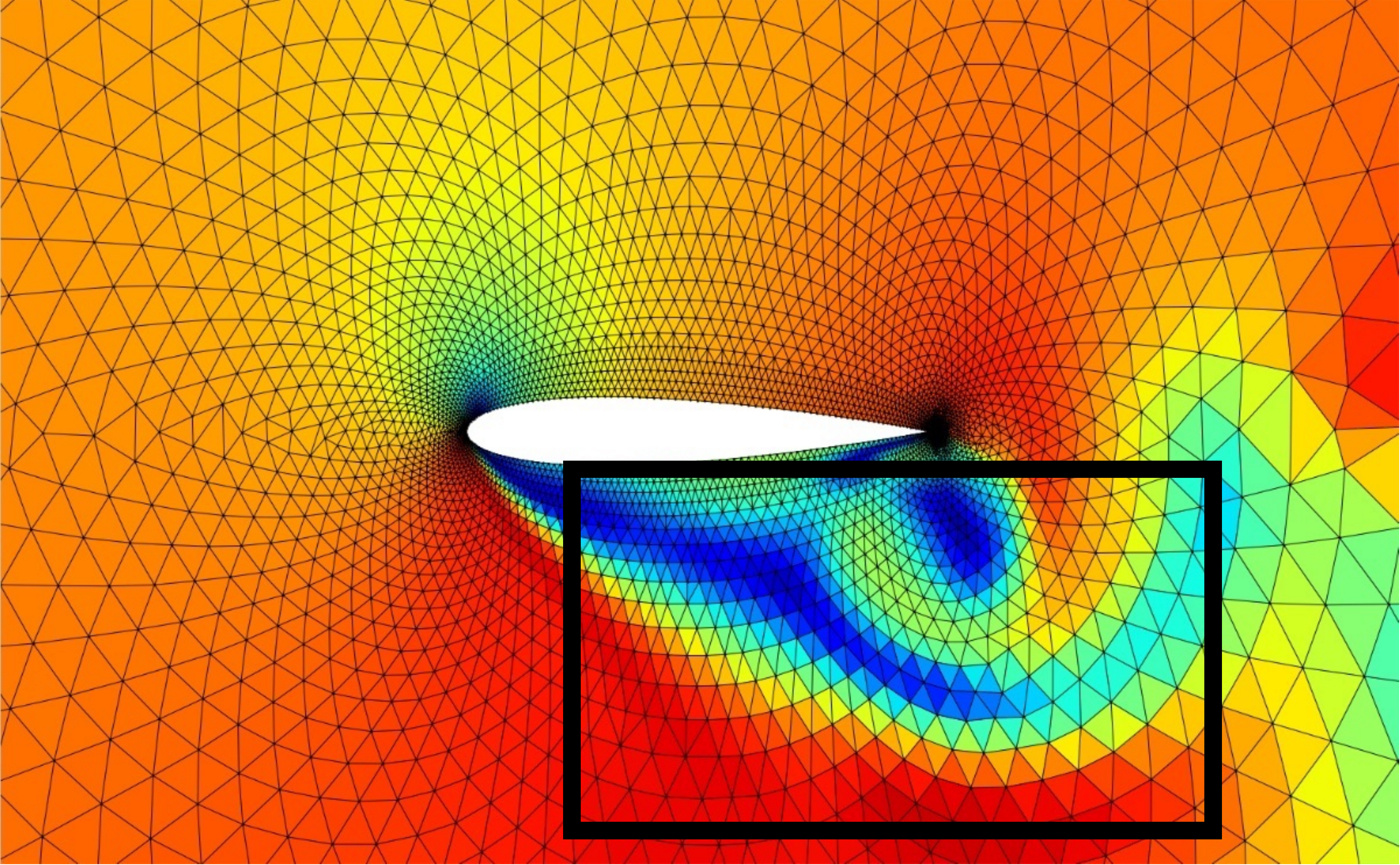}}
    \subfigure[MGN]{\includegraphics[width=0.235\textwidth]{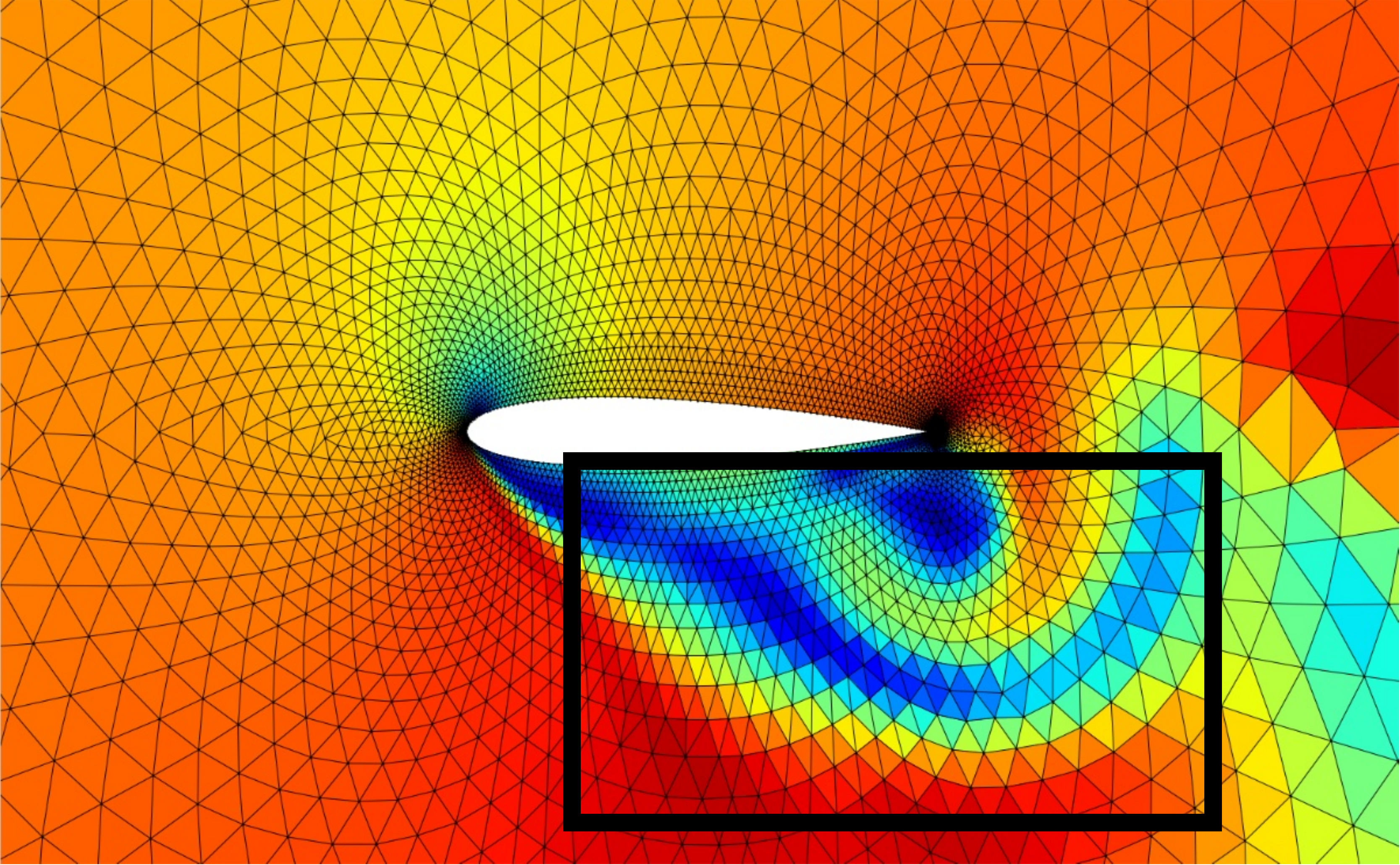}}
    \caption{Comparison of 2D cross-sectional velocity magnitude contours for \textsc{CylinderFlow} (a)-(d) and \textsc{AirFoil} (e)-(h) at the last time step with the largest cumulative error. It is most similar to ground truth when PIORF is applied. The closer the color is to red, the faster the velocity. The black boxes (\myblackbox) highlight regions where PIORF shows particular accuracy in predicting complex flow structures. PIORF consistently achieves the closest match to ground truth on both datasets. More rollout images can be found in \Cref{app:contour}.}
    \label{fig:rollout}
\end{figure}

\paragraph{Results.}
\Cref{tab:maintable} shows a comprehensive performance comparison of different rewiring methods across the 4 model architectures.
PIORF consistently outperforms other rewiring baselines when applied to MGN, BSMS, GT, and HMT models.
For \textsc{CylinderFlow}, PIORF achieves the lowest RMSE in both velocity and pressure when applied to MGN. This improvement is especially significant compared to MGN and other rewiring methods. For \textsc{AirFoil}, PIORF achieves the best performance in all cases.
\Cref{fig:rollout} shows the superiority of PIORF by showing velocity magnitude contours at the final timestep. Our PIORF results closely align with the ground truth, especially in regions marked by black boxes.

\begin{figure}[t]
    \centering
    \subfigure[\textsc{CylinderFlow}]{\includegraphics[width=0.245\textwidth]{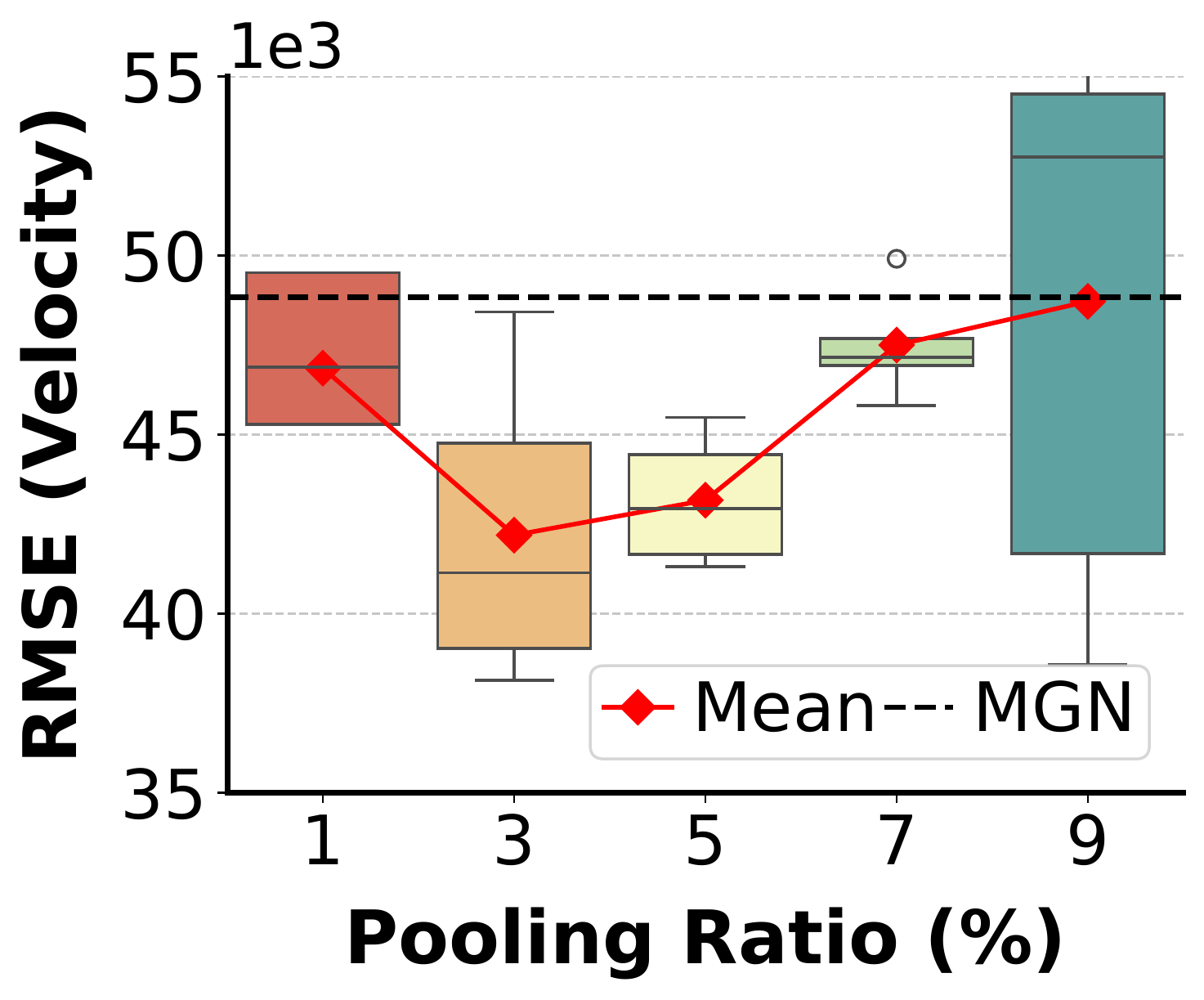}}
    \subfigure[\textsc{CylinderFlow}]{\includegraphics[width=0.245\textwidth]{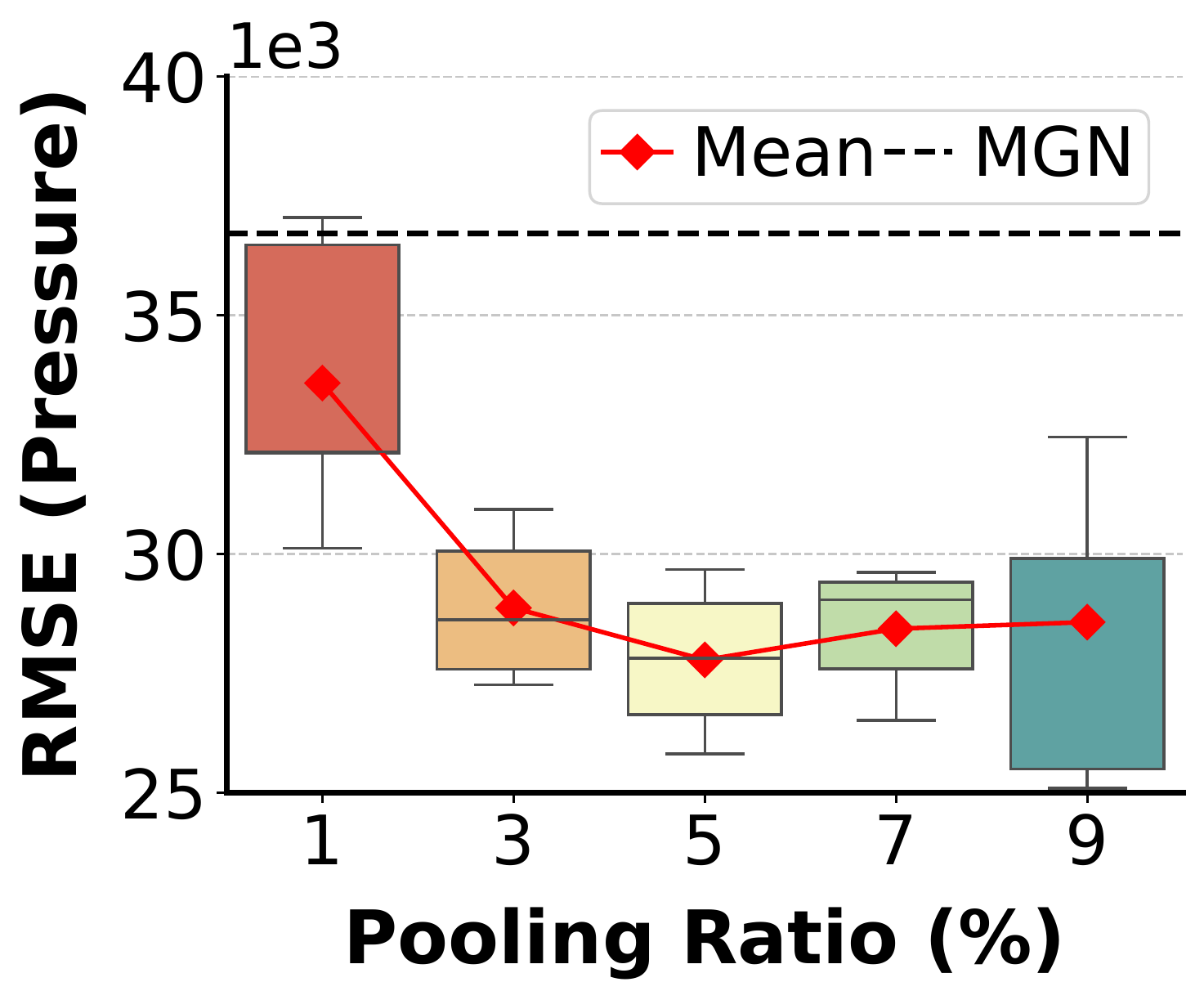}}
    \subfigure[\textsc{AirFoil}]{\includegraphics[width=0.245\textwidth]{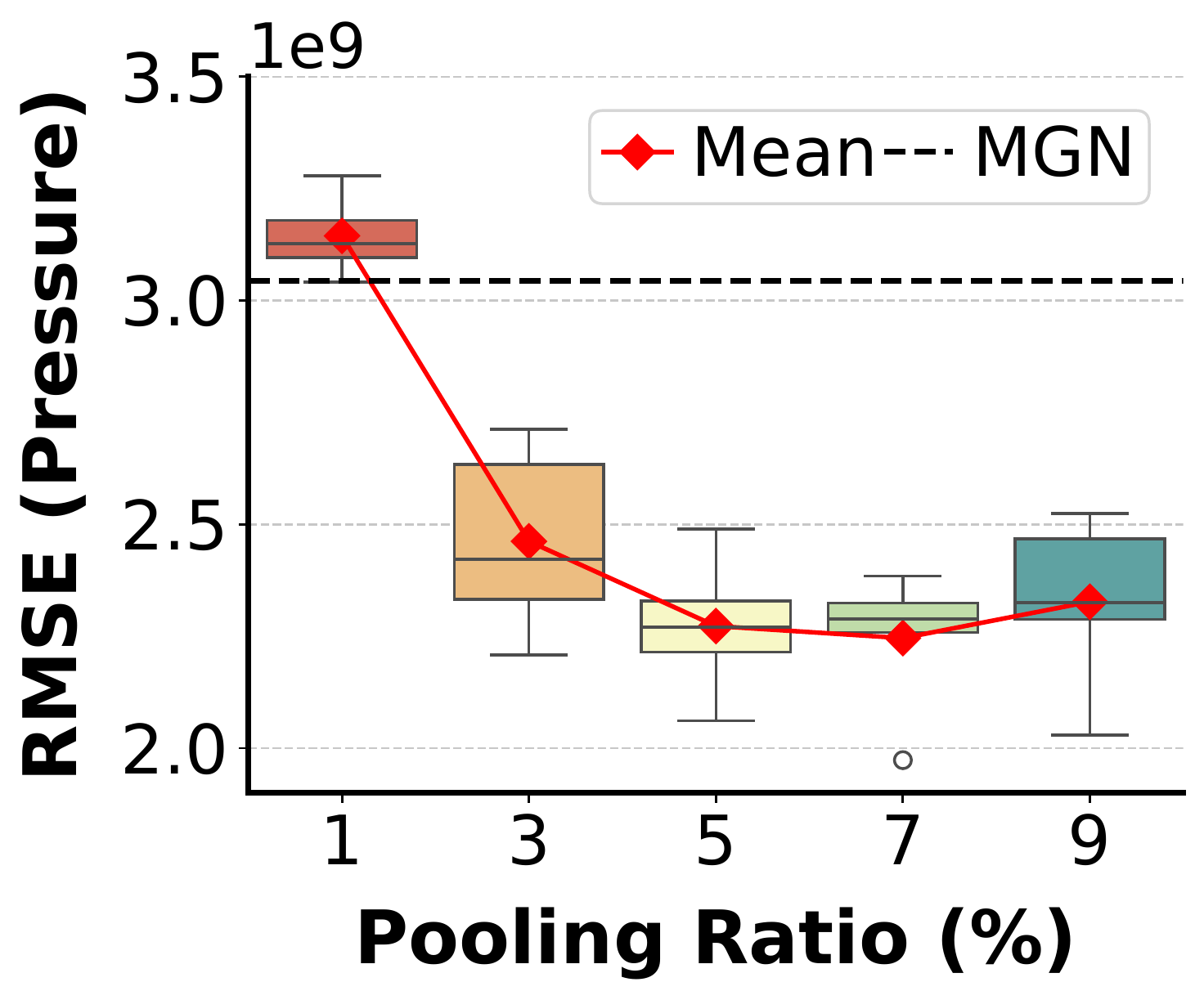}}
    \subfigure[\textsc{AirFoil}]{\includegraphics[width=0.245\textwidth]{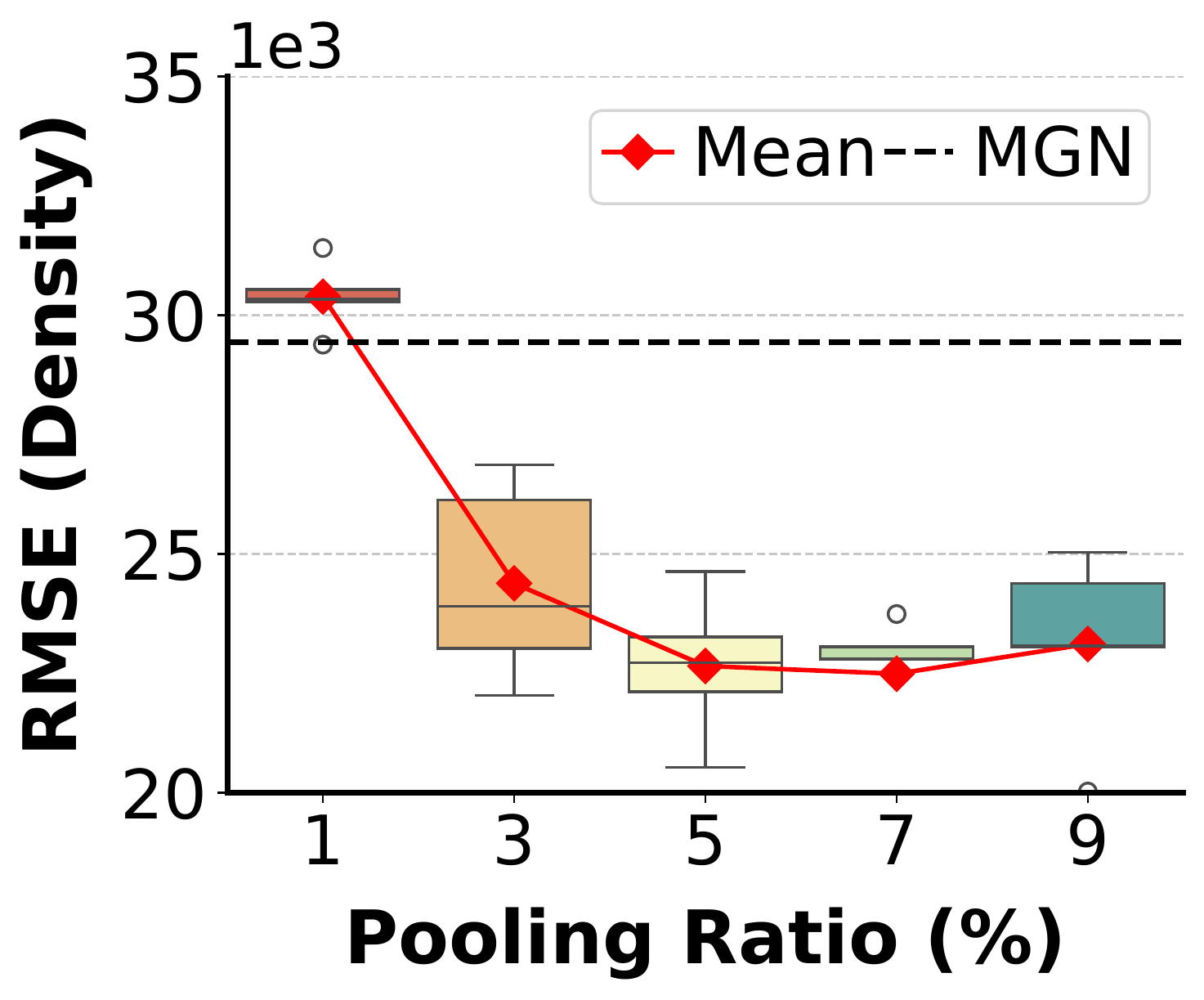}}
    \caption{Sensitivity to pooling ratio $\delta$. 
    The dashed lines represent RMSE of MGN without rewiring.}
    \label{fig:sensitivity}
\end{figure}

\paragraph{Sensitivity to pooling ratio.}
We analyze sensitivity to pooling ratio $\delta$, which is \emph{our sole hyperparameter} and determines the number of new edge connections. 
\Cref{fig:sensitivity} shows how rollout RMSE varies with $\delta$ for both datasets.
Velocity RMSE of \textsc{CylinderFlow} is optimal at 3\%, while pressure RMSE generally improves with higher ratios. For \textsc{Airfoil}, velocity RMSE is best at 7\%, and pressure RMSE at 7\%. Across all cases, 1\% pooling ratio often performs worse than MGN, while 9\% increases standard deviations. The results show the need to tune $\delta$ specfically for each dataset.
\Cref{fig:sensitivity} (a) shows the Velocity RMSE of \textsc{CylinderFlow}, and it can be seen that the average and standard deviation of RMSE increase at 9\% where a large number of edges are connected.

\subsection{Scaling to Larger Fluid Dynamics}

\paragraph{Datasets.}\label{app:eagle}
\begin{wraptable}{r}{0.53\textwidth}
\vspace{-3em}
    \small
    \setlength{\tabcolsep}{1.2pt}
    \centering
    \caption{Comparison of fluid dynamics datasets}
    \begin{tabular}{l ccccc}\toprule
        Dataset       & Size & Dynamic Scene & Dynamic Mesh \\ \midrule
        \textsc{CylinderFlow} & 15GB & \XSolidBrush & \XSolidBrush \\
        \textsc{AirFoil} & 56GB & \XSolidBrush & \XSolidBrush \\
        \textsc{EAGLE} & 270GB & \Checkmark & \Checkmark \\\bottomrule
    \end{tabular}
    \label{tab:eagle_dataset}
\vspace{-1em}
\end{wraptable}
To evaluate scalability and efficiency of PIORF, we use \textsc{EAGLE}~\citep{janny2023eagle}, which simulates turbulent flows created by drones in various scenes.
As shown in \Cref{tab:eagle_dataset}, \textsc{EAGLE} significantly surpasses \textsc{CylinderFlow} and \textsc{AirFoil} in scale and complexity.
\textsc{EAGLE} has dynamic meshes~\citep{malcevic2002dynamic, jasak2009dynamic}, where the mesh positions and boundary conditions change at each time step. This dynamic nature requires temporal graph rewiring, presenting a more challenging and realistic scenario compared to the static meshes of \textsc{CylinderFlow} and \textsc{AirFoil}.

\paragraph{Setting.} \label{par:setting2}
We use MGN with 15 layers and maintain the same baseline rewiring methods, adjusting only dataset-specific hyperparameters. 
We set the velocity noise standard deviation to 0.02 in all methods.
DIGL is set with \texttt{alpha} to 0.01 and \texttt{eps} at 0.4.
For SDRF, we set a maximum of 10 iterations and no edge removal, and for FoSR, we use an initial power of 5 and a maximum of 20 iterations.
To ensure statistical significance, we repeat each experiment 5 times with different seeds.

\paragraph{Results.}
\begin{wraptable}{r}{0.35\textwidth}
    \vspace{-3em}
    \footnotesize
    \setlength{\tabcolsep}{2pt}
    \centering
    \caption{Rollout-all RMSE ($\times10^{3}$)}
    \begin{tabular}{c cc}\toprule
        \multirow{2}{*}{Model} & \multicolumn{2}{c}{EAGLE}  \\ \cmidrule(lr){2-3} & Velocity & Pressure  \\\midrule
        MGN           & 2,280\std{135} & 10,893\std{632}  \\
        \;+ DIGL  & 2,623\std{114} & 12,688\std{698}  \\
        \;+ SDRF  & 2,186\std{70}  & 10,504\std{297}  \\
        \;+ FoSR  & 2,254\std{63}  & 10,755\std{246}  \\
        \;+ BORF  & Time-out  & Time-out  \\
        \cmidrule(lr){1-3} \rowcolor{gray! 20}
        \;+ PIORF  & 1,950\std{28}  & 9,449\std{167}  \\
        \bottomrule
    \end{tabular}
    \label{tab:eagle_result}
    \vspace{-3.5em}
\end{wraptable}
The timeout of BORF highlights the computational challenges in applying rewiring to the large-scale dataset.
As shown in \Cref{tab:eagle_result}, our PIORF outperforms all baselines, achieving a 14.5\% improvement in velocity RMSE over MGN. 
While other rewiring methods such as SDRF and FoSR show some improvements, they are significantly smaller compared to PIORF.
\Cref{fig:roll_eagle} in \Cref{app:contour} shows the result of the last step with different rewiring methods applied.

\subsection{Computational Efficiency}\label{sec:complex}
\begin{figure}[h!]
    \centering
    \subfigure[\textsc{CylinderFlow}]
    {\includegraphics[width=0.27\textwidth]{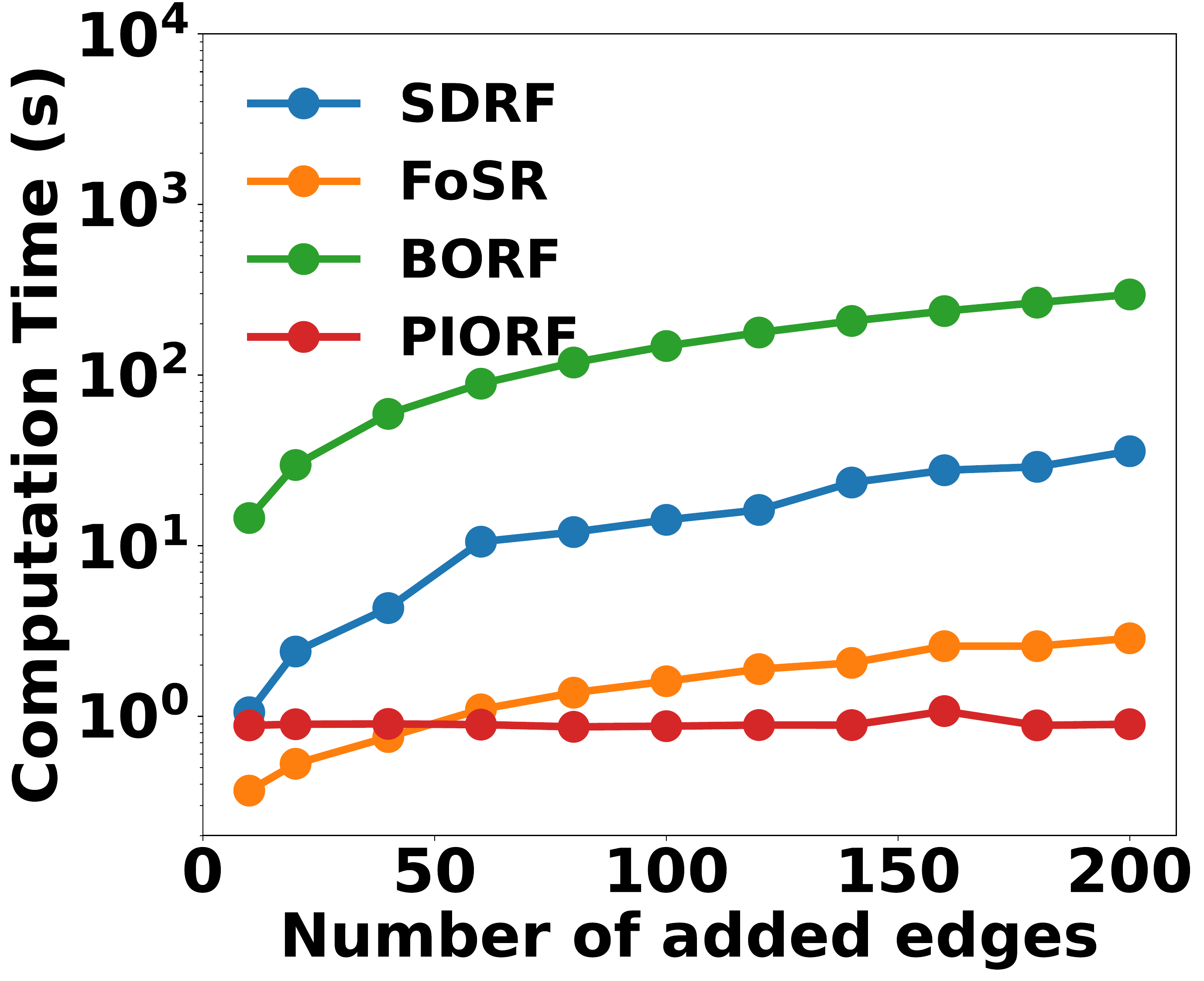}}
    \hfill
    \subfigure[\textsc{AirFoil}]{\includegraphics[width=0.27\textwidth]{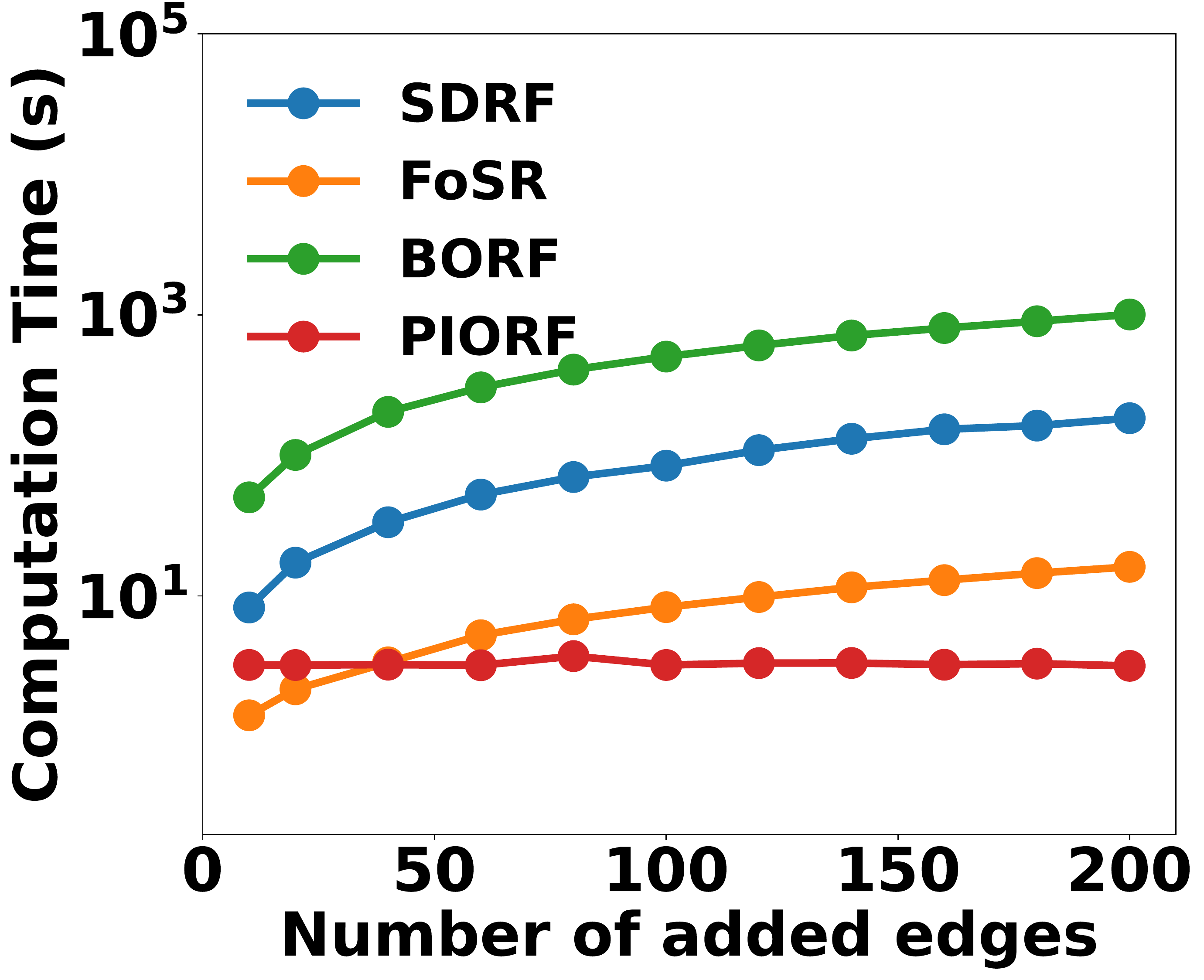}}
    \hfill
    \subfigure[\textsc{EAGLE}]
    {\includegraphics[width=0.27\textwidth]{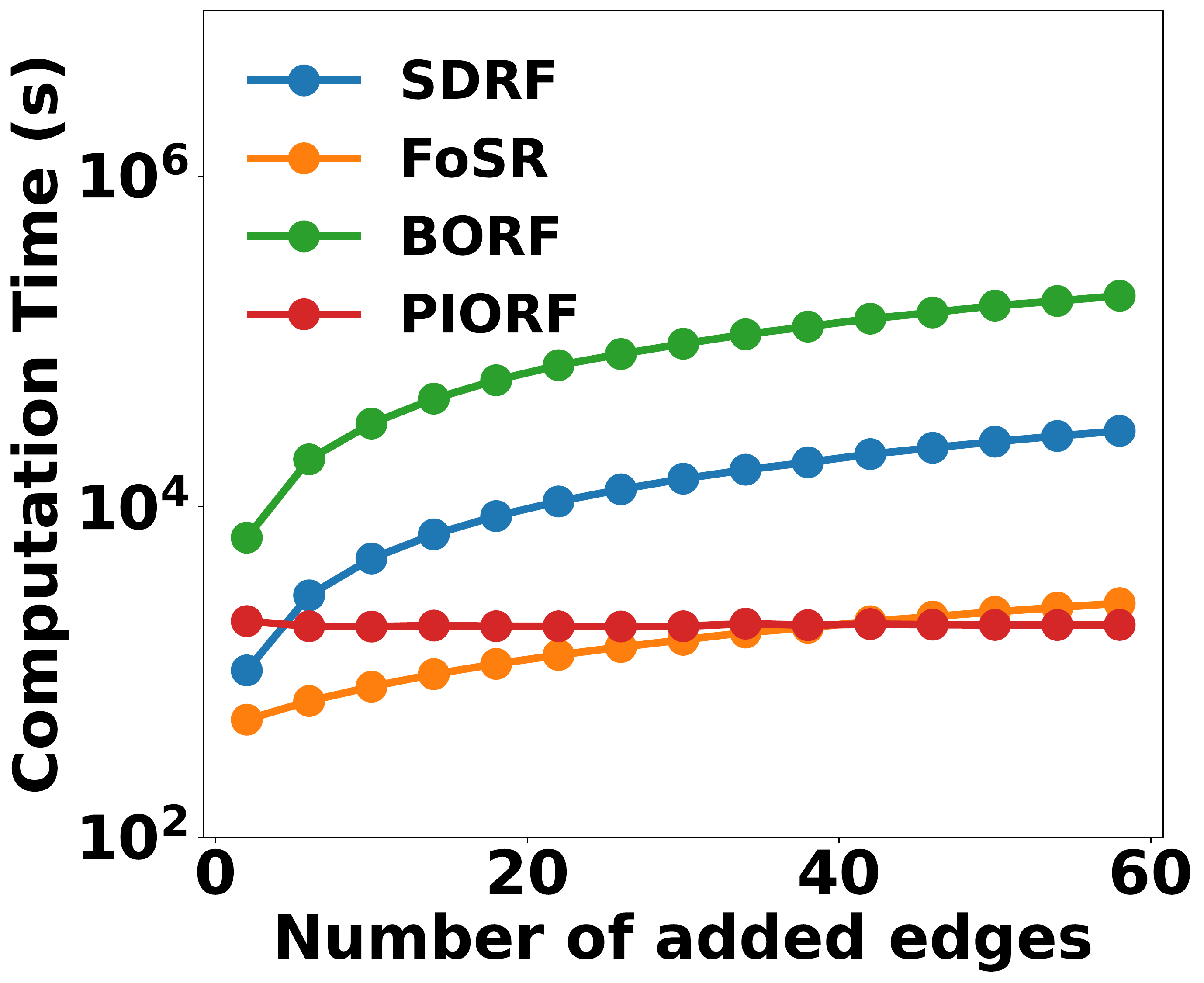}}
    \caption{Comparison of computation time as the number of edges added increases.}
    \label{fig:complex}
    % \vspace
\end{figure}

Given the large scale of mesh graphs, with thousands of nodes and tens of thousands of edges (see \Cref{tab:data} in \Cref{app:datasets}), we need to add a large number of edges to alleviate over-squashing. However, existing rewiring methods require multiple iterations to add or delete edges, leading to increased computational overhead. \Cref{fig:complex} shows the computation time required to add varying numbers of edges when rewiring one trajectory in the \textsc{CylinderFlow}, \textsc{AirFoil}, and \textsc{EAGLE} datasets. PIORF maintains the lowest computation time in all datasets and edge counts. 
This is due to the ability of PIORF to compute all the necessary rewiring in a single pass, avoiding an iterative process.
In contrast, BORF shows a steep increase in computation time as the number of added edges grows, particularly evident in \textsc{EAGLE}. 
Although SDRF and FoSR are more efficient than BORF, they still show a trend of increasing computational time, emphasizing the scalability advantage of PIORF in handling large-scale fluid dynamics simulations.

\subsection{Ablation Studies}
We conduct ablation studies to evaluate components of PIORF and \Cref{tab:ablation} summarizes our findings.

\begin{table}[t]
    \footnotesize
    \setlength{\tabcolsep}{4pt}
    \centering
    \caption{Rollout-all RMSE ($\times10^{3}$) for PIORF and the ablations.}
    \begin{tabular}{l cc cc ccc}\toprule
        \multirow{2}{*}{Ablation Model} & \multicolumn{2}{c}{Actions} & \multicolumn{2}{c}{\textsc{CylinderFlow}} & \multicolumn{3}{c}{\textsc{AirFoil}} \\ \cmidrule(lr){2-3}\cmidrule(lr){4-5}\cmidrule(lr){6-8}
        & Add & Remove & Velocity & Pressure & Velocity & Pressure & Density \\\midrule
        MGN          & & & 48.8\std{5.6} & 36.7\std{2.4} & 10,261\std{832} & 3,043,186\std{282,514} & 29.4\std{2.7}\\
        \cmidrule(lr){1-8}
        Velocity     &\Checkmark & & \textbf{41.6\std{3.9}} & \textbf{28.9\std{1.5}} & \textbf{7,743\std{584}} & \textbf{2,245,858\std{142,452}} & \textbf{22.5\std{1.4}}\\
        Pressure     &\Checkmark &           & 80.9\std{13.4} & 75.8\std{15.3} & 7,768\std{288} & 2,293,481\std{108,098} & 23.1\std{1.2}\\
        \cmidrule(lr){1-8}
        Random       &\Checkmark &           & 43.4\std{2.4} & 32.3\std{1.2} & 10,317\std{771} & 3,115,406\std{230,796} & 30.2\std{2.2}\\
        \cmidrule(lr){1-8}
        Only Removal  & &\Checkmark          & 42.0\std{2.3} & 36.8\std{0.5} & 10,890\std{438} & 3,289,964\std{94,568} & 31.7\std{0.8}\\
        Both         &\Checkmark &\Checkmark & 49.0\std{7.5} & 31.1\std{2.9} & 7,813\std{551} & 2,334,583\std{182,600} & 23.4\std{1.9}\\    
        \cmidrule(lr){1-8}
        Weighted Edges &\Checkmark &           & 53.2\std{8.6} & 44.1\std{4.5} & 11,214\std{563} & 3,486,655\std{203,277} & 32.9\std{1.6}\\
        \cmidrule(lr){1-8}
        To Senders &\Checkmark & & 53.4\std{7.2} & 35.8\std{1.0} & 10,358\std{866} & 3,099,548\std{320,553} & 29.7\std{2.8}\\ 
        To Receivers &\Checkmark & & 47.9\std{4.9} & 35.0\std{4.9} & 10,421\std{704} & 3,132,703\std{119,582} & 30.3\std{1.2}\\ 
        \bottomrule
    \end{tabular}
    \label{tab:ablation}
    \vspace{-1em}
\end{table}

\paragraph{Choice of physical value for rewiring.}
We analyze the impact of using velocity or pressure to identify nodes for edge rewiring in PIORF. 
For \textsc{CylinderFlow}, an incompressible flow~\citep{panton2024incompressible}, velocity-based rewiring significantly outperforms pressure-based rewiring. 
This aligns with Bernoulli's principle for incompressible flows, where velocity changes more indicate key flow dynamics. 
For \textsc{AirFoil}, a compressible~\citep{saad1985compressible} and turbulent flow~\citep{mathieu2000introduction}, pressure-based and velocity-based rewiring performs well and outperforms other rewiring methods.

\paragraph{Effect of physical-informed node selection.}
PIORF selects the nodes based on ORC-identified bottlenecks and nodes with high physical changes. To assess the impact of using physical values in this selection process, we compare our approach (``Velocity'') with a modified version (``Random'') where nodes are chosen based on ORC bottlenecks but the second node is selected randomly, ignoring physical values. Results show that physics-informed selection outperforms random selection.

\paragraph{Effect of edge addition/removal.}
We analyze the effects of edge addition (``Velocity''), removal (``Only Removal''), and both (``Both''). Removal (``Only Removal'') removes the edge with the highest positive curvature. Interestingly, edge addition alone yields the best performance for all datasets, suggesting that adding new edges is beneficial than removing existing ones.

\paragraph{Weighted edges.}
We explore the impact of weighted edges by the L2 distance of velocity when calculating ORC in \Cref{eq:orc} and \Cref{eq:wasserstein}. The ``Weighted Edges'' results indicate that this approach does not improve performance. It means that binary edge existence might be sufficient for capturing relevant physical relationships.

\paragraph{Directionality in rewiring.}
We dissect the effect of directional rewiring by adding one-way edge sets. `To Senders'' is when aggregation is performed from receivers to senders, while ``To Receivers'' is the opposite. The results show that bidirectional rewiring outperforms unidirectional approaches.

\section{Conclusions}
We introduce PIORF as a new rewiring method that simultaneously considers the topology and physical correlation of the mesh graph and experimentally demonstrate best performance in the field of physics mesh simulation. Moreover, we show for the first time that applying our rewiring method to hierarchical GNNs and Transformer also improves model performance in mesh graph. 
\paragraph{Limitations and future work.}
One limitation of PIORF is its dependence on the choice of physical values for rewiring. Future research could focus on developing adaptive mechanisms for selecting the most relevant physical quantities automatically. Another important direction is to extend PIORF to handle dynamic adaptive mesh refinement~\citep{bangerth2003adaptive,cerveny2019nonconforming}, which could include integrating PIORF with error estimation techniques that enable more targeted refinement in areas with large solution errors. Additionally, extending our PIORF to applications in multi-body dynamics~\citep{choi2013general}, equivariant graphs~\citep{satorras2021n}, and particle-based simulations~\citep{li2018learning} is an important area of future work.

\clearpage

\section*{Ethics Statement}
Our proposed PIORF is a rewiring method designed for modeling physical systems on unstructured meshes, and thus it poses no clear negative societal or ethical implications. However, potential misuse or application of the algorithm in unintended areas could result in unintended consequences. 

Additionally, this paper may have implications regarding the carbon footprint and accessibility of learning algorithms. Recently, as the computational demands in machine learning research have grown, they have led to an increasing carbon footprint. Our proposed method contributes to reducing this carbon footprint by not only improving performance but also enhancing computational efficiency in such contexts.

\section*{Reproducibility}
We provide the source code for our experimental environments and the proposed method. In the future, we intend to make this source code available for the benefit of the community.
PIORF source code can be found in the following: \url{https://github.com/yuyudeep/piorf}

PIORF has a single hyperparameter, the pooling ratio $\delta$. The best hyperparameter option for reproduction in each dataset is described in Section 5, along with sensitivity analysis. Additionally, the experimental settings for the proposed method and baseline can be found in \Cref{par:setting1}, \Cref{par:setting2}, and \Cref{app:base}.

\section*{Acknowledgements}
This work was supported by the LG Display and an IITP grant funded by the Korean government (MSIT) (No. RS-2020-II201361, Artificial Intelligence Graduate School Program (Yonsei University)).
K. Lee acknowledges support from the U.S. National Science Foundation under grant IIS 2338909.

\bibliography{reference}
\bibliographystyle{iclr2025_conference}
%%%%%%%
\clearpage
\appendix

\addcontentsline{toc}{section}{Appendix} % Add the appendix text to the document TOC
% {\huge Appendix} % Start the appendix part
\part{\Large{Supplementary Materials for ``PIORF''}} % Start the appendix part
\parttoc % Insert the appendix TOC
%%%%%%%
\section{Comparison of Rewiring Methods and Complexcity}\label{app:comp}
We further compare existing graph rewiring methods with our proposed method. As shown in \Cref{tab:comp}, our method, PIORF, takes the physical context into account, which other rewiring methods do not. This is a key characteristic of our approach, which aims to overcome the limitations of existing methods for learning fluid dynamics simulations that are not designed for this purpose. 

The complexity of PIORF is $\mathcal{O}(|\mathcal{E}|d^3_{\mathrm{max}})$, where $|\mathcal{E}|$ is the number of edges and $d_\mathrm{max}$ is the maximal degree. In particular, simulation datasets in fluid dynamics use thousands to tens of thousands of nodes and edges to ensure solution accuracy, so the PIORF method is advantageous in applying more edges to these datasets. One of the biggest differences is the computational cost. Existing methods such as DIGL, SDRF, FoSR, and BORF incur significant computational cost in the process of selecting which edges to rewiring to optimize their own defined objective function (See \Cref{fig:complex}). Our method, on the other hand, performs the rewiring without any objective function optimization, which is beneficial in terms of computational cost.
Another important difference is the number of hyperparameters. Existing rewiring methods typically require two or more hyperparameters, while our PIORF uses \emph{only one} hyperparameter, the pooling ratio. This has the advantage of reducing the hyperparameter search space.

\begin{table}[h]
    \small
    \centering
    \caption{Comparison of different rewiring methods and our PIORF.}
    \label{tab:comp}
    \begin{tabular}{ll ccc}\toprule
        Methods & Indicator & Complexity & Geometry & Physics \\\midrule
        SDRF & Balanced Forman curvature & $\mathcal{O}(|\mathcal{E}|d^2_{\mathrm{max}})$ & \Checkmark & \XSolidBrush\\
        FoSR & Spectral gap              & $\mathcal{O}(\mathcal{V}^2)$ & \Checkmark & \XSolidBrush\\
        BORF & Ollivier--Ricci curvature & $\mathcal{O}(|\mathcal{E}|d^3_{\mathrm{max}})$ & \Checkmark & \XSolidBrush\\ \cmidrule(lr){1-5}
        PIORF & Ollivier--Ricci curvature with physical context& $\mathcal{O}(|\mathcal{E}|d^3_{\mathrm{max}})$  & \Checkmark & \Checkmark \\
        \bottomrule
    \end{tabular}
\end{table}

\section{Datasets Details}\label{app:datasets}
\begin{figure}[h]
    \centering
    \subfigure[\textsc{CylinderFlow}]{\includegraphics[width=0.32\textwidth]{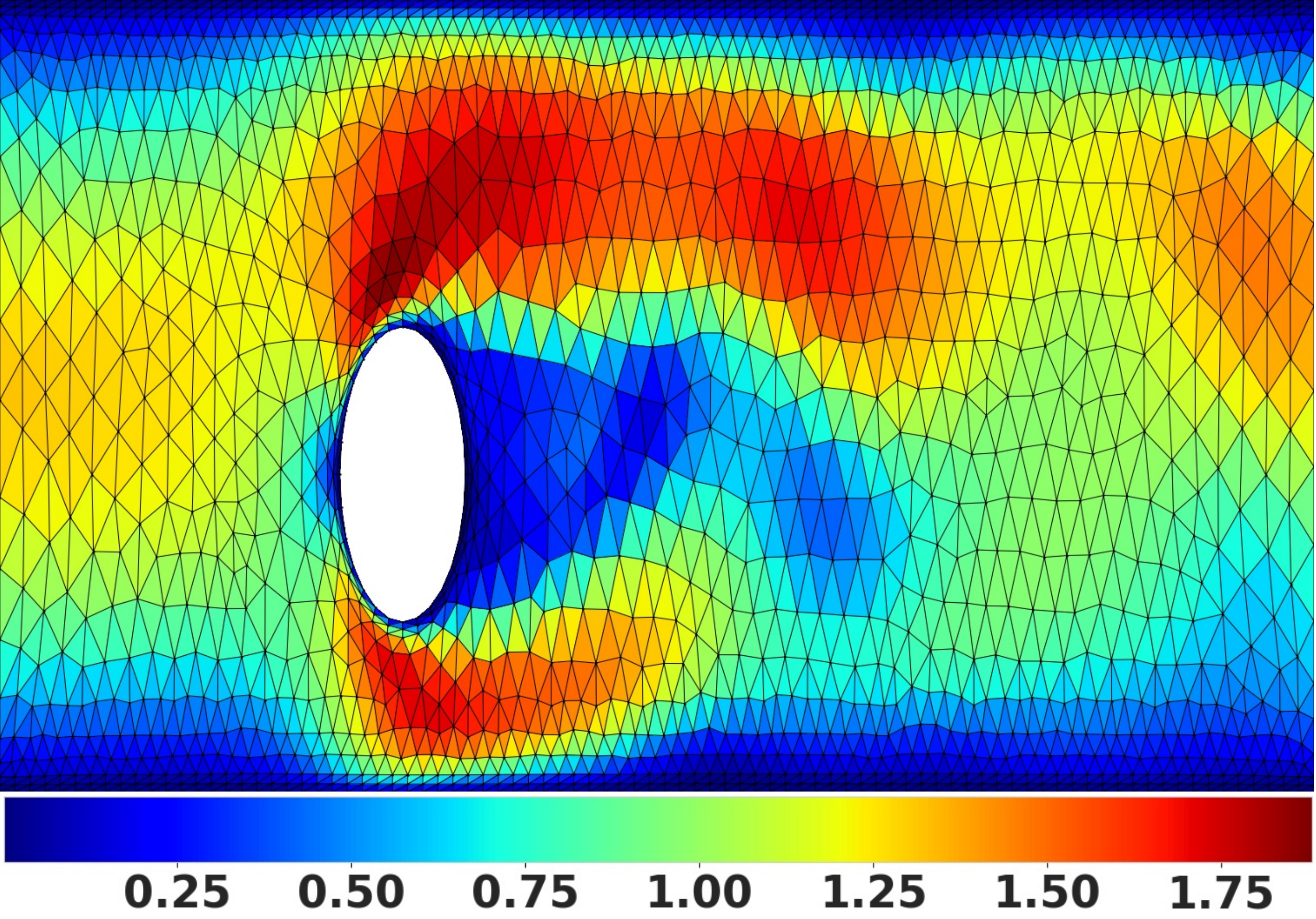}}
    \subfigure[\textsc{AirFoil}]{\includegraphics[width=0.32\textwidth]{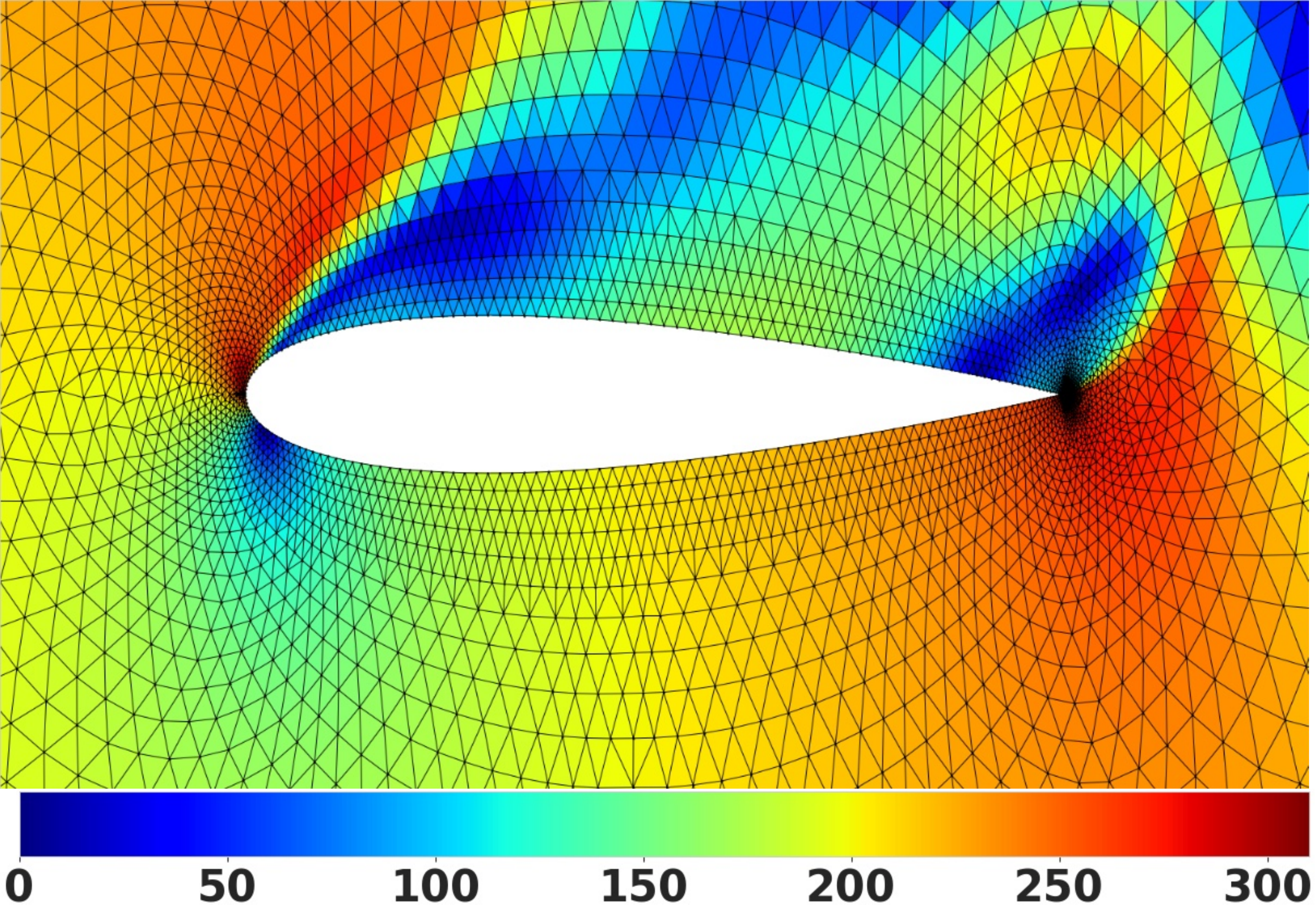}}
    \subfigure[\textsc{EAGLE}]{\includegraphics[width=0.32\textwidth]{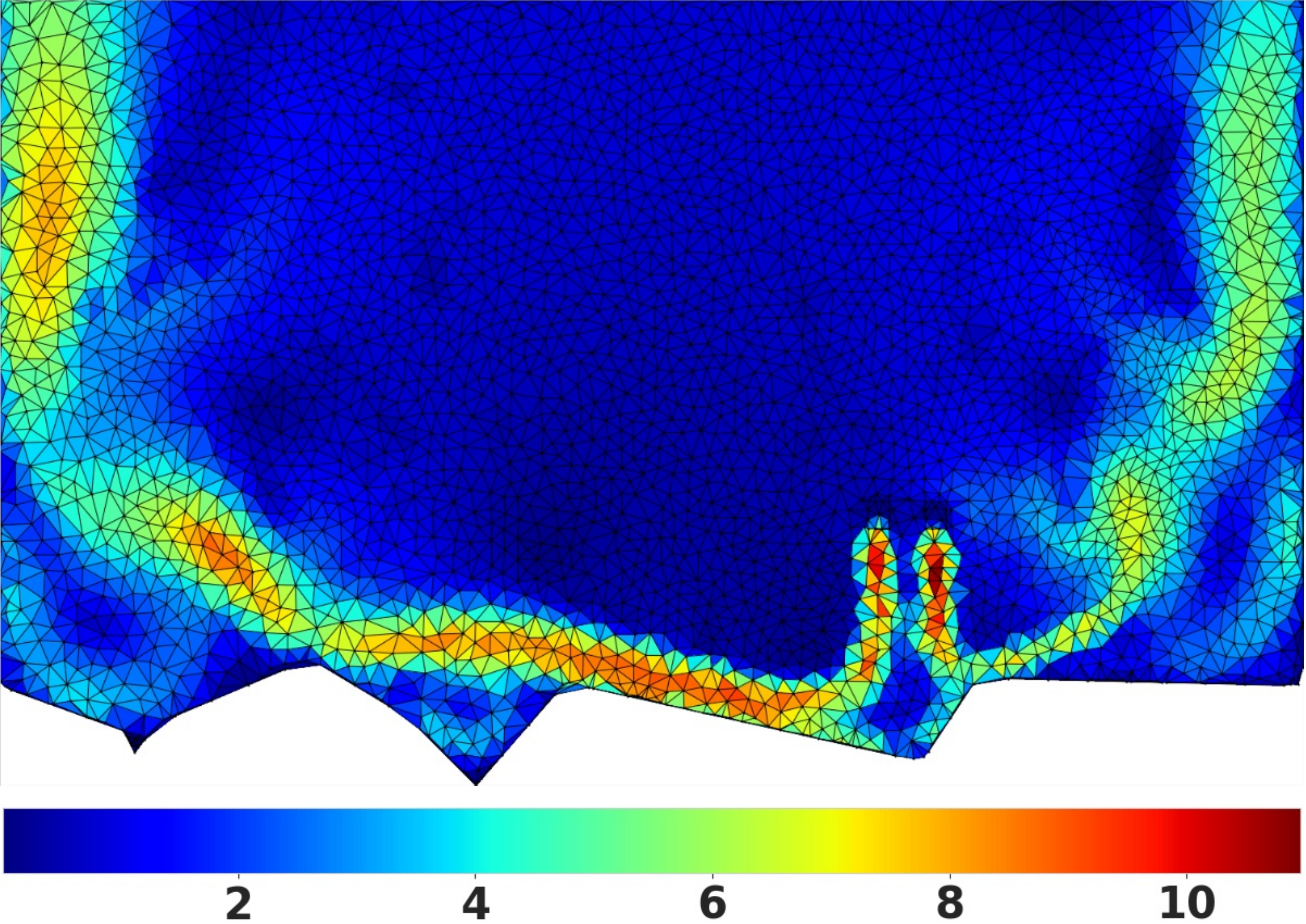}}
    \caption{Velocity magnitude contour image for each dataset. In all cases, changes in velocity occur in walls where fluid cannot flow.}
    \label{fig:app_rollout}
\end{figure}

\begin{figure}[h]
    \centering
    \subfigure[\textsc{CylinderFlow}]
    {\includegraphics[width=0.32\textwidth]{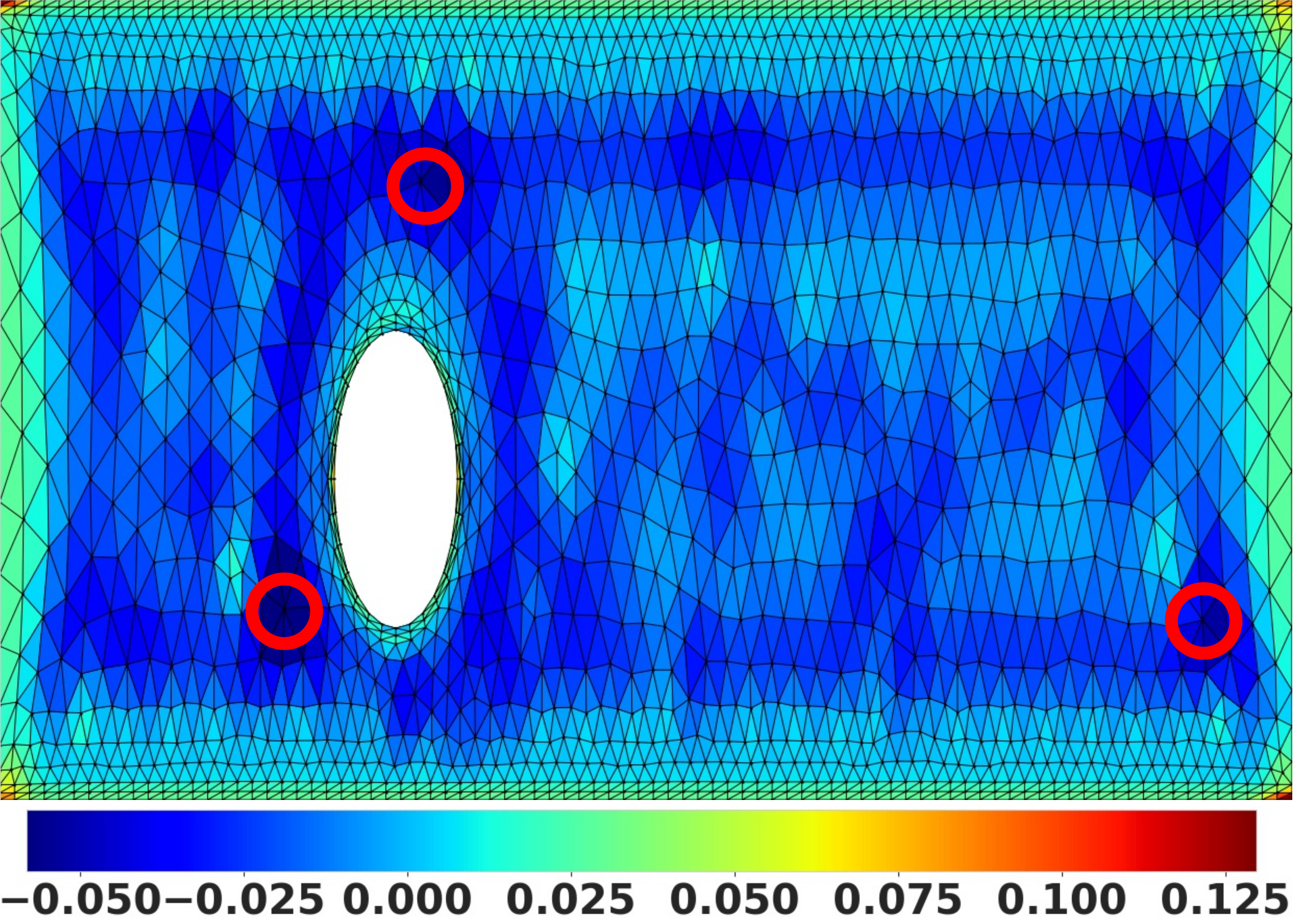}}
    \subfigure[\textsc{AirFoil}]{\includegraphics[width=0.32\textwidth]{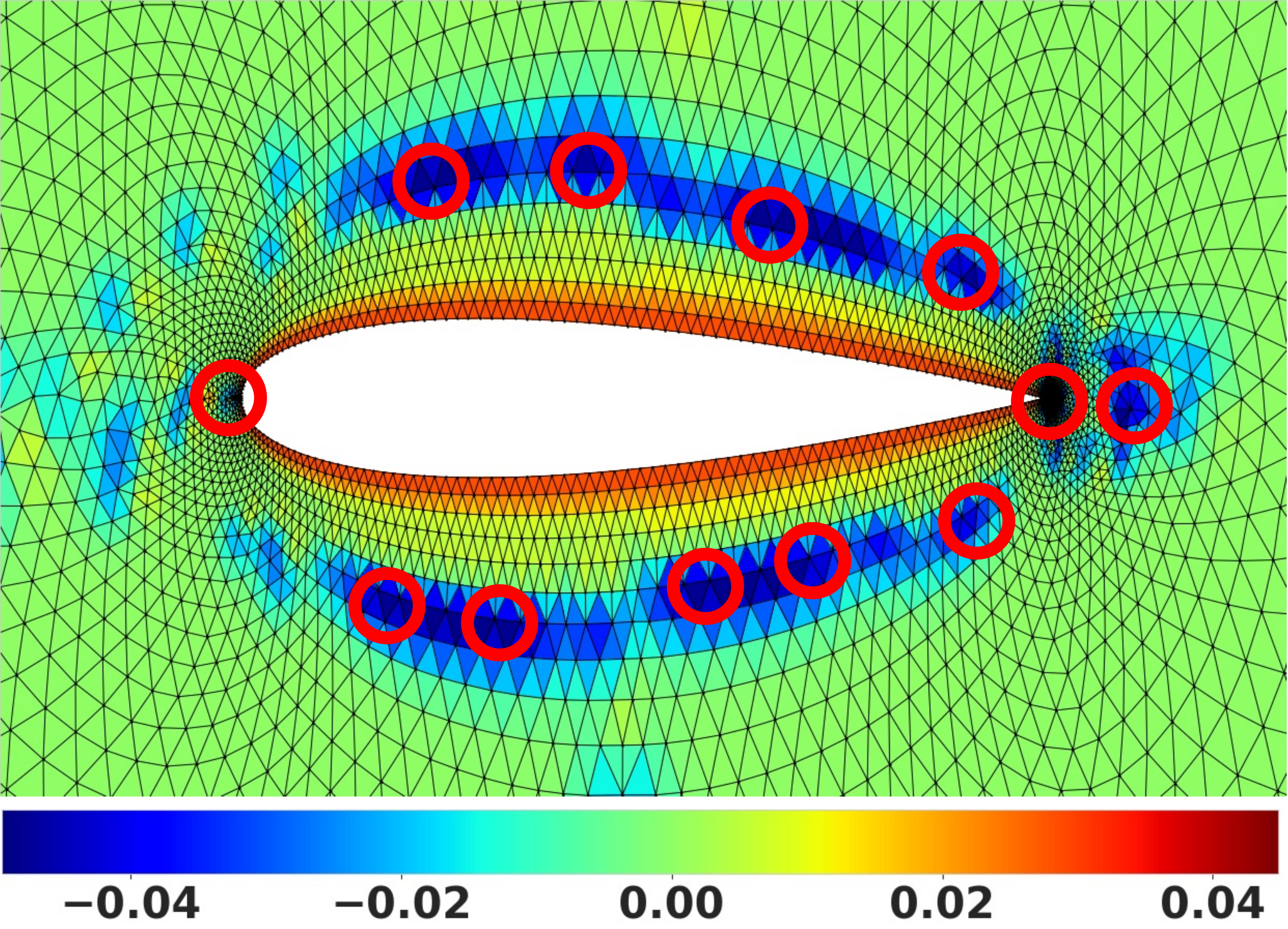}}
    \subfigure[EAGLE]
    {\includegraphics[width=0.32\textwidth]{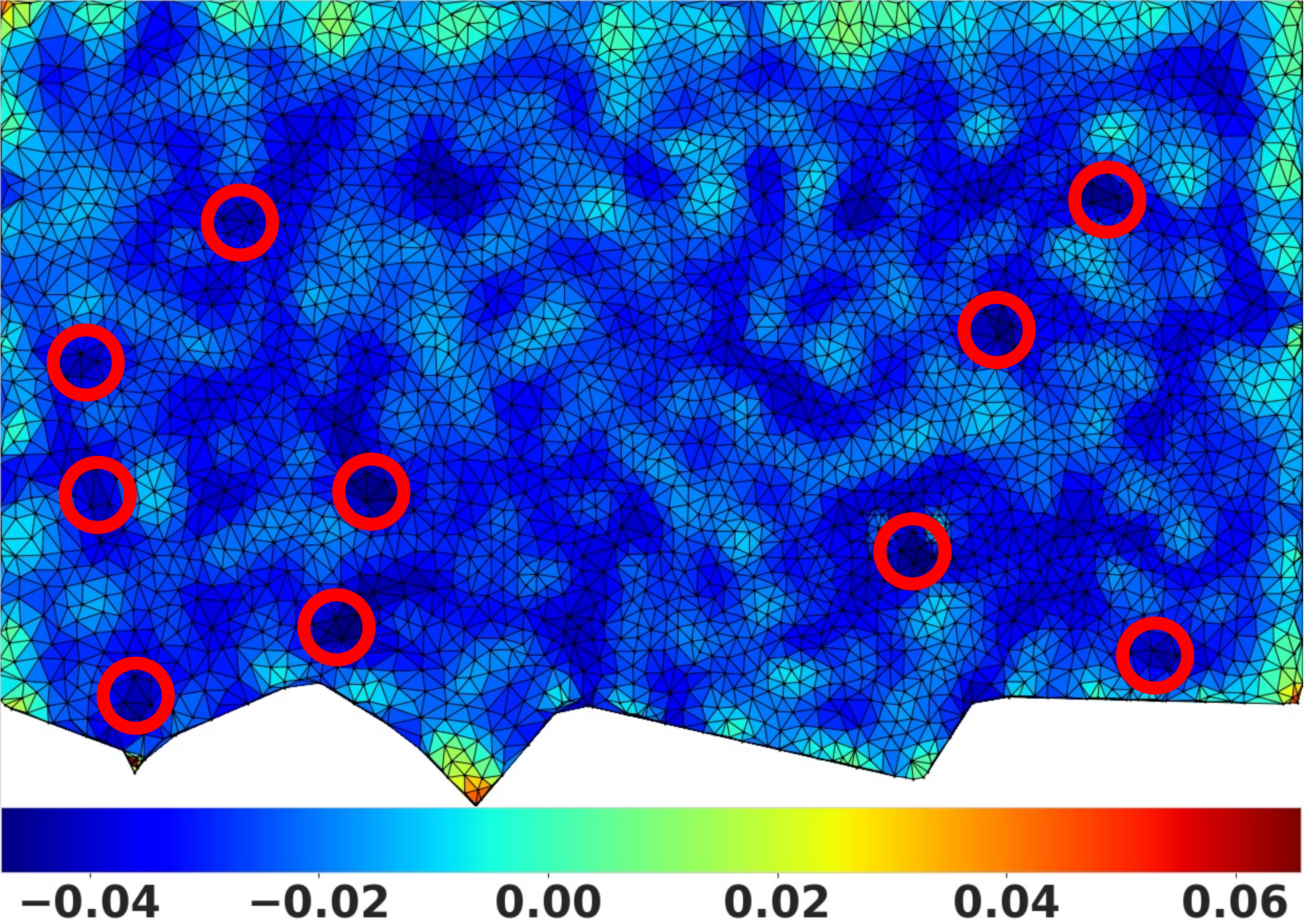}}
    \caption{ORC distribution image for each dataset. Red circles (\myredcircle) are the nodes where the degree is high and a bottleneck occurs.}
    \label{fig:app_rollout_orc}
\end{figure}

\textsc{CylinderFlow-Tiny} dataset used in \Cref{fig:ansys_orc} is used to illustrate the concept of PIORF and is not desinged for evaluation.
We use three public datasets for evaluation, and \Cref{tab:data} shows information such as the number of cases, number of steps, number of nodes and number of edges for each dataset. \textsc{AirFoil} and \textsc{EAGLE} datasets are turbulent flow models, and \textsc{CylinderFlow} is a laminar flow model. \Cref{fig:app_rollout} shows the velocity magnitude contour image of each datasets. In all datasets, the velocity is high in areas near boundary conditions such as walls. \Cref{fig:app_rollout_orc} shows the distribution of ORC by each dataset. When creating a mesh, nodes with high degrees occur due to local fine mesh and boundary conditions. Red circles are nodes where the degree is 7 or higher and bottlenecks occur.

\begin{table}[h]
    \setlength{\tabcolsep}{1.6pt}
    \centering
    \caption{Dataset description: Fluid dynamics behavior, number of trajectories for each data set, time step, and average number of nodes, edges, and cells in the training data set. A cell refers to an element and is a small unit that makes up a mesh. In the case of a triangular mesh, one cell consists of three nodes.}
    \begin{tabular}{l ccccccccc}\toprule
      \multicolumn{1}{l}{Datasets} &
      \multicolumn{1}{c}{Behavior} &
      \multicolumn{1}{c}{Cases (Train)} &
      \multicolumn{1}{c}{Cases (Test)} &
      \multicolumn{1}{c}{Steps} & 
      \multicolumn{1}{c}{Nodes (avg)} &
      \multicolumn{1}{c}{Edges (avg)} &
      \multicolumn{1}{c}{Cells (avg)} \\ \midrule
    \textsc{CylinderFlow}   &Laminar   &1,000 &100 &600  &1,886 &10,848 &3,538 \\ 
    \textsc{AirFoil}         &Turbulent &1,000 &100 &600  &5,233 &30,898 &10,216 \\
    \textsc{EAGLE}           &Turbulent &947   &118 &990  &3,389 &20,023 &6,623 \\ 
    \bottomrule
    \end{tabular}
    \label{tab:data}
\end{table}

\paragraph{\textsc{CylinderFlow-Tiny}.}
\textsc{CylinderFlow-Tiny} is the dataset used to illustrate the concept of our PIORF for understanding the flow of fluid in narrow passages around a cylinder.
To create \textsc{CylinderFlow-Tiny} dataset, we consider performing simulation modelling in an environment similar to that of \textsc{CylinderFlow}. We use the Ansys Fluent\textsuperscript{\textregistered} solver~\citep{stolarski2018engineering} to generate the dataset. The number of nodes is approximately 300 and the fluid input is air.

\paragraph{\textsc{CylinderFlow}.}
\textsc{CylinderFlow} is important in many industrial applications, such as the cooling of cylindrical pipes, by analyzing the flow of fluid around a cylinder. The flow can exhibit laminar or turbulent flow behavior depending on factors such as flow rate, fluid density, and cylinder size. \textsc{CylinderFlow} dataset~\citep{pfaff2020mgn} consists of 1,000 analysis results, with each case containing 600 time steps. The dataset contains a single cylinder, but includes a variety of Reynolds numbers, sizes, and positions.

\paragraph{\textsc{AirFoil}.}
\textsc{AirFoil} is an application of aerodynamics and the most basic CFD modeling. \textsc{AirFoil}, also known as wings, is utilized in the design of airfoils and various other aerodynamic applications such as aircraft, helicopters, and spacecraft. \textsc{AirFoil} plays a central role in designing an airplane's wings to generate lift, control flight, and move through airflow. Moreover, it is very important to design an aerodynamic design that is effective in a specific range of flow conditions. \textsc{AirFoil} dataset~\citep{pfaff2020mgn} consists of 1,000 analysis results, with each case containing 600 time steps. The data set contains one \textsc{AirFoil} and various input conditions, such as velocity and pressure, with the fluid density changing at every step.

\paragraph{\textsc{EAGLE}.}
\textsc{EAGLE} is a large-scale dataset for learning non-steady fluid dynamics. This is a simulation of the airflow generated by a drone moving in a 2D environment with various boundary shapes. It is much more difficult than other datasets such as \textsc{CylinderFlow} or \textsc{AirFoil} as it models the complex ground effect turbulence created by the drone's airflow according to its control laws. Different scene geometries produce completely different results, resulting in highly turbulent and non-periodic eddies and high flow diversity. In the field of learned simulators, \textsc{EAGLE} is the first to apply a dynamic mesh effect in which the shape and position of the mesh change at every time step. It accurately simulates fluid behavior by changing the drone's position over time.

\paragraph{Representative physical quantity.}
The velocity refers to the speed at which a fluid moves at a specific point in space. The pressure is the force exerted by a fluid per unit area on the surfaces. The density $\rho$ is a measure of how much mass is contained within a given volume of the substance. It is defined as mass per unit volume. The density of a fluid depends on temperature and pressure. These three physical quantities are related by Bernoulli's equation. When density is constant, increasing velocity causes pressure to decrease.

\section{Baseline Details}\label{app:base}
We compare four competitive rewiring methodologies and four models. For models, MGN~\citep{pfaff2020mgn}, BSMS, GT, and HMT are used, along with rewiring methods such as DIGL~\citep{gasteiger2019diffusion}, SDRF~\citep{topping2021understanding}, FoSR~\citep{karhadkar2022fosr}, and BORF~\citep{nguyen2023borf}. For all datasets, training steps are set to 10,000,000. Velocity noise standard deviation is 0.02 and 10 for \textsc{CylinderFlow} and \textsc{AirFoil} datasets, respectively.

\subsection{Rewiring Methods}
For DIGL, we set \texttt{alpha} to 0.01 and use 0.4 for \texttt{eps}. For SDRF, max number of iterations is 10. Edge removal is not used. For FoSR, initial power and max number of iterations are set to 5 and 20, respectively. In the case of BORF, the max number of iterations is set to 10, and edge addition and deletion for each batch are set to 4 and 2, respectively.
We use the official implementation released by the authors on GitHub for all rewiring baselines: 
\begin{itemize}
    \item DIGL: \url{https://github.com/gasteigerjo/gdc.git}
    \item SDRF: \url{https://github.com/jctops/understanding-oversquashing}
    \item FoSR: \url{https://github.com/kedar2/FoSR}
    \item BORF: \url{https://github.com/hieubkvn123/revisiting-gnn-curvature}
\end{itemize}

\subsection{Models}\label{app:models}

\paragraph{MGN.}
To align with the MGN methodology, we apply 15 iterations of message passing in all datasets. All MLPs have a hidden vector size of 128. \Cref{tab:feature} indicates the input, edge, and output features used for each dataset.

\begin{table}[h]
    \centering
    \caption{Details of features for each dataset. $\rho_i$ is the fluid density and $\mathbf{w}_i$ is the velocity of the fluid. $\dot{\mathbf{w}}_i$ is the gradient of velocity, $\mathbf{n}_{i}$ is the node type, and $\mathbf{x}_{i}$ is the position of the node.}
    \begin{tabular}{l cccc}\toprule
    \multicolumn{1}{l}{Datasets} &
      \multicolumn{1}{c}{Inputs $\mathbf{m}_{ij}$} &
      \multicolumn{1}{c}{Inputs $\mathbf{v}_i$} &
      \multicolumn{1}{c}{Outputs $\mathbf{o}_i$}  \\ \midrule
    \textsc{CylinderFlow}     & $\mathbf{x}_{ij},|\mathbf{x}_{ij}|$ & $\mathbf{n}_{i},\mathbf{w}_{i}$ & $\dot{\mathbf{w}}_i,p_i$\\ 
    \textsc{AirFoil}           & $\mathbf{x}_{ij},|\mathbf{x}_{ij}|$ & $\mathbf{n}_{i},\mathbf{w}_{i},\rho_i$ & $\dot{\mathbf{w}}_i,p_i,\dot{\rho_i}$\\ 
    \textsc{EAGLE}             & $\mathbf{x}_{ij},|\mathbf{x}_{ij}|$ & $\mathbf{n}_{i},\mathbf{w}_{i}$ & $\dot{\mathbf{w}}_i,p_i$\\ 
    \bottomrule
    \end{tabular}
    \label{tab:feature}
\end{table}

\paragraph{BSMS.}
We implement the BSMS model with \texttt{Tensorflow}. And according to the best hyperparameters of BSMS, levels 7 and 9 are used for \textsc{CylinderFlow} and \textsc{AirFoil}, respectively. Noise standard deviation is set the same as MGN. All MLPs have a hidden vector size of 128. 
The Encoder/decoder are set to those in MGN.

\paragraph{GT.}
The hidden dimension size inside its Transformer is set to 128. FFN used three linear layers and two ReLU activations. To ensure numerical stability, the results obtained with the exponential term within the softmax function are constrained to fall in the range of $[-2, 2]$. We use the FFN without using positional encoding. There are 15 transformer blocks with 4 heads. The encoder/decoder are set to those in MGN.

\paragraph{HMT.}
Because contact edges are not used in fluid datasets, we only use HMT among the modules of the HCMT model. The hidden dimension size of HMT is set to 128, and both FFN and numerical stability are set to the same as GT. There are 15 transformer blocks with 4 heads. The encoder/decoder are set to those in MGN.

We use the official implementation released by the authors on GitHub for all baselines models: 
\begin{itemize}
    \item MGN: \url{https://github.com/google-deepmind/deepmind-research/tree/master/meshgraphnets}
    \item BSMS: \url{https://github.com/Eydcao/BSMS-GNN}
    \item GT: \url{https://github.com/graphdeeplearning/graphtransformer}
    \item HMT: \url{https://github.com/yuyudeep/hcmt}
\end{itemize}

\clearpage
\section{Other Variable Contour and Rollout Figures}\label{app:contour}
\begin{figure}[h]
    \centering
    \setlength{\tabcolsep}{1pt}
    \begin{tabular} {l ccc}\toprule
    Models & Traj. 1 & Traj. 2 & Traj. 3 \\ \midrule
    Ground Truth & \includegraphics[width=0.28\textwidth]{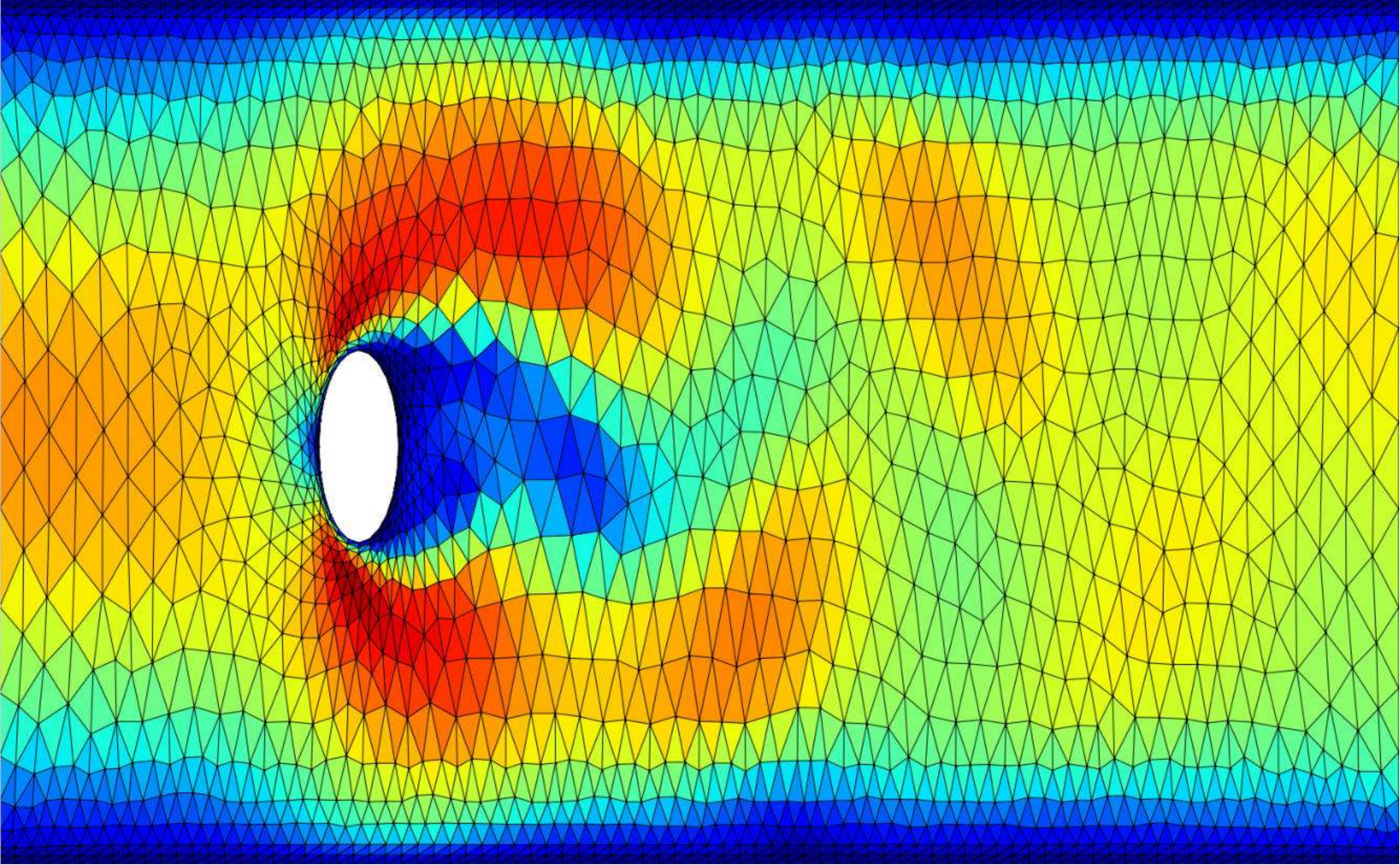} & \includegraphics[width=0.28\textwidth]{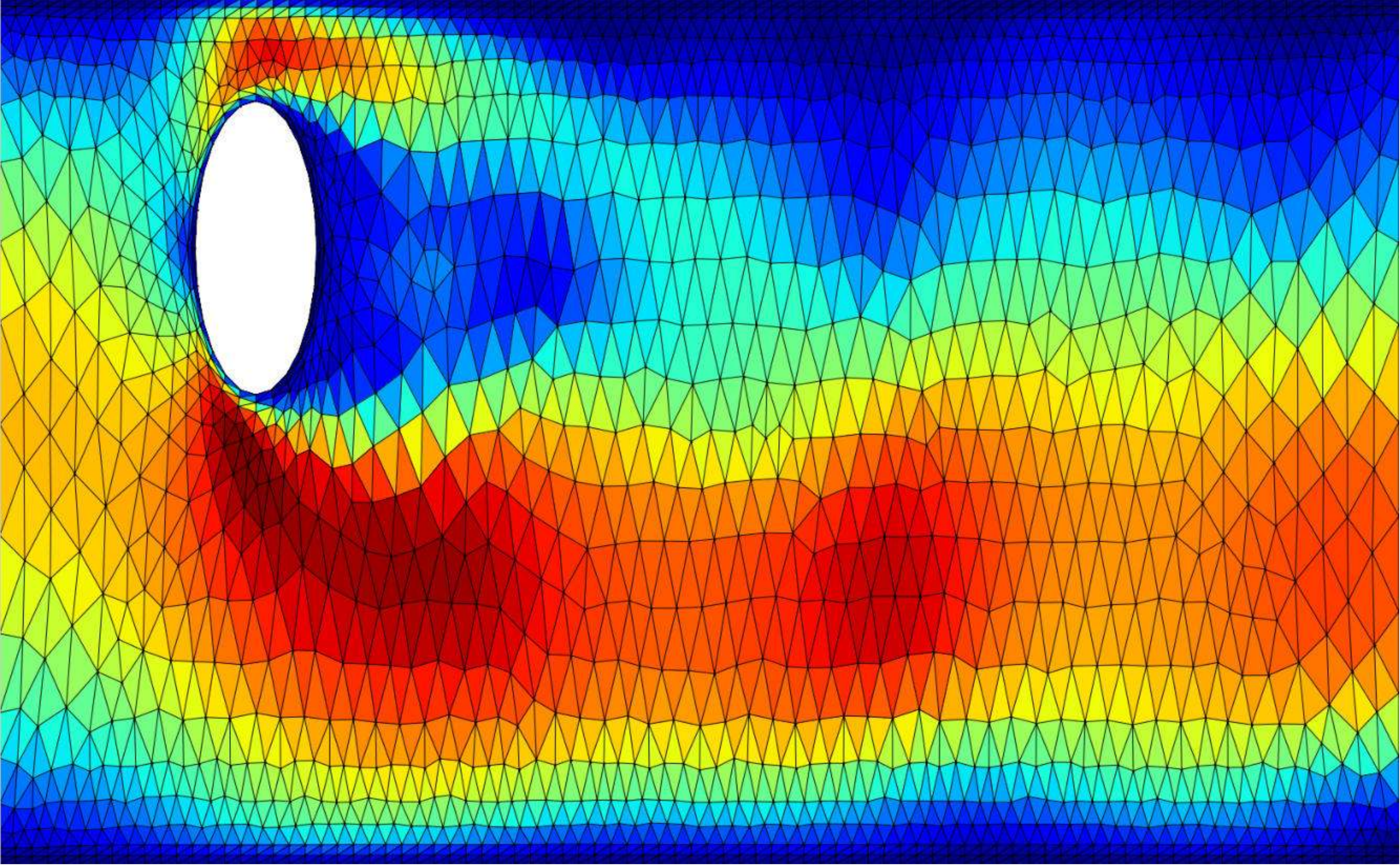} & \includegraphics[width=0.28\textwidth]{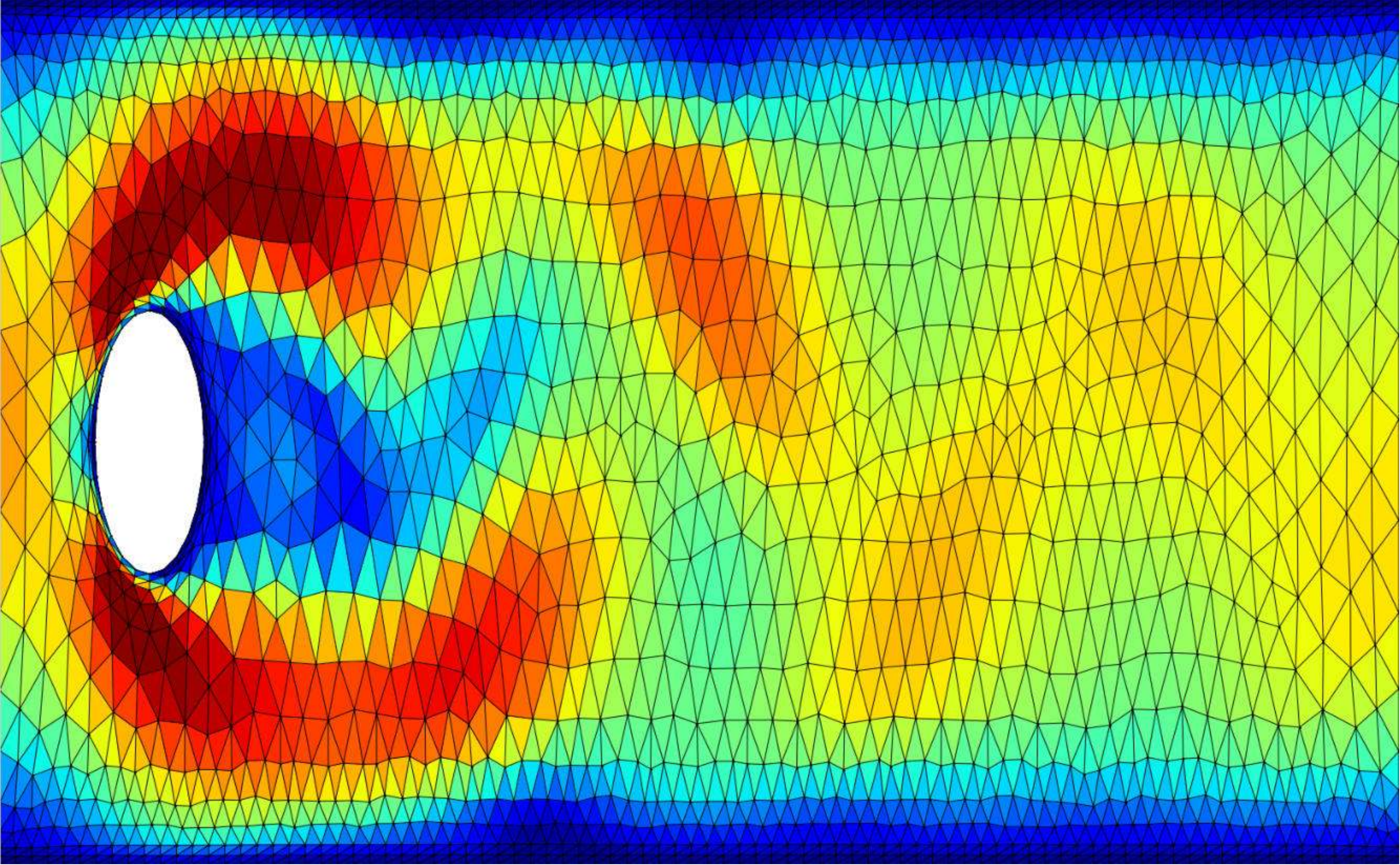}   \\
    MGN & \includegraphics[width=0.28\textwidth]{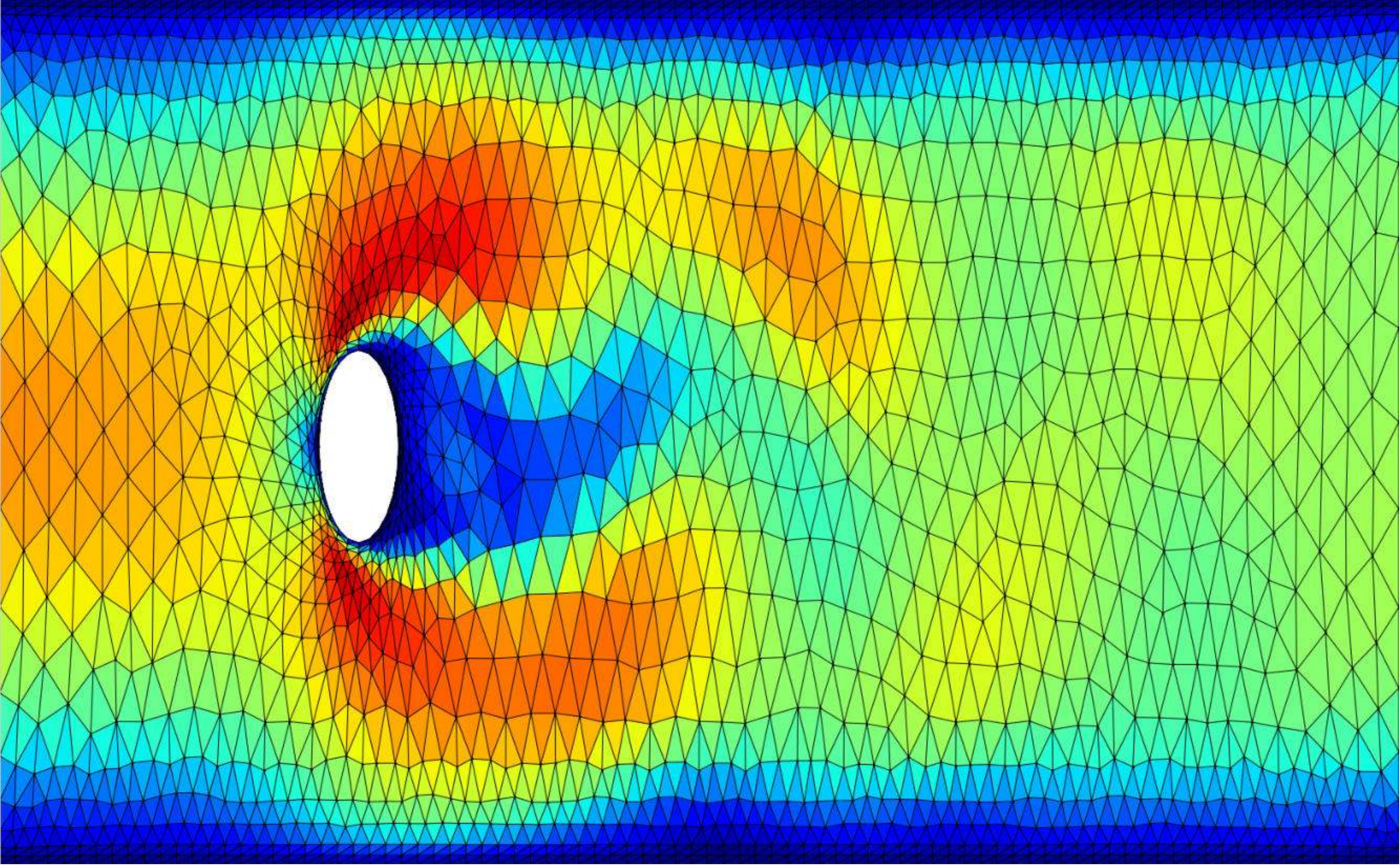} & \includegraphics[width=0.28\textwidth]{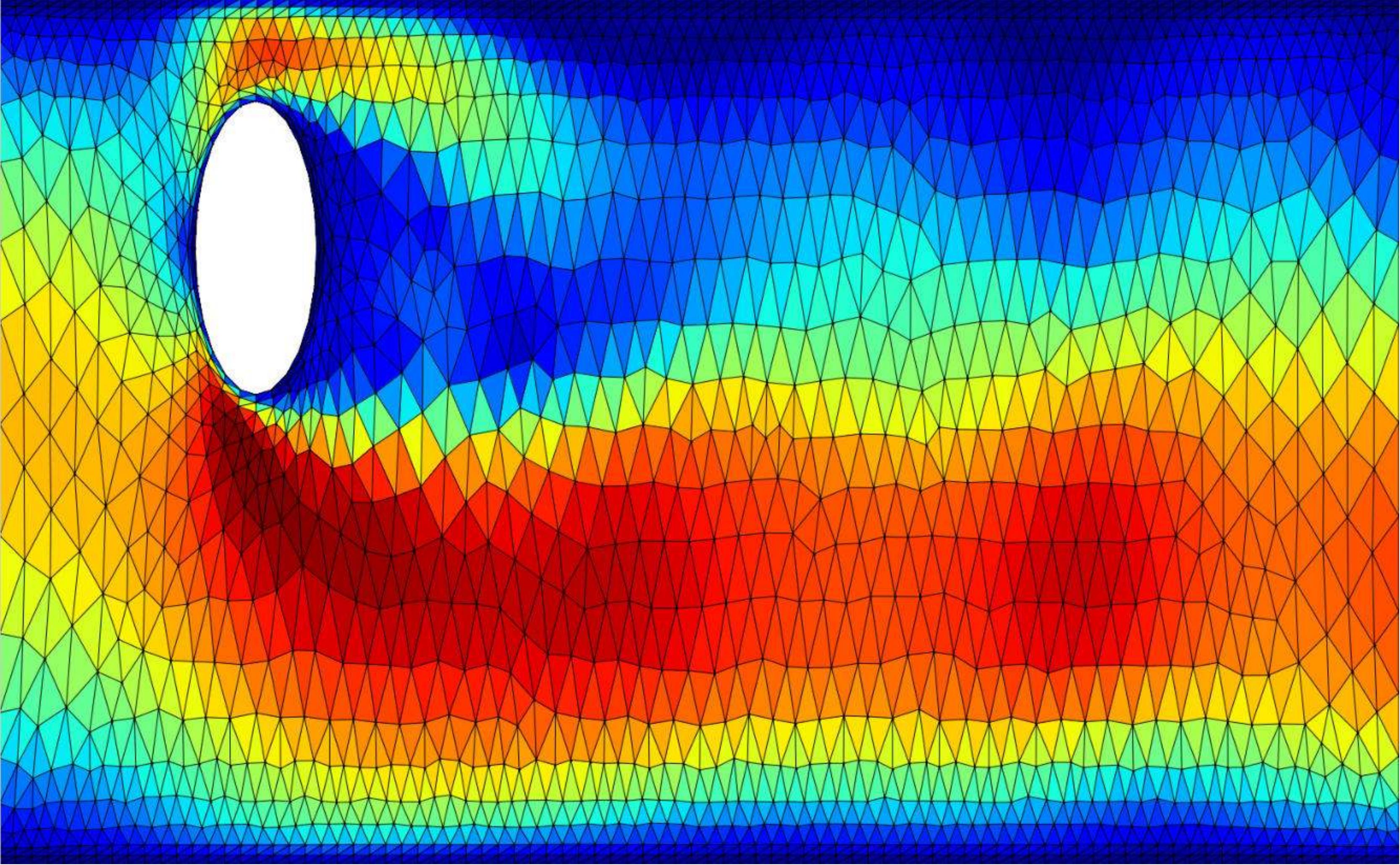} & \includegraphics[width=0.28\textwidth]{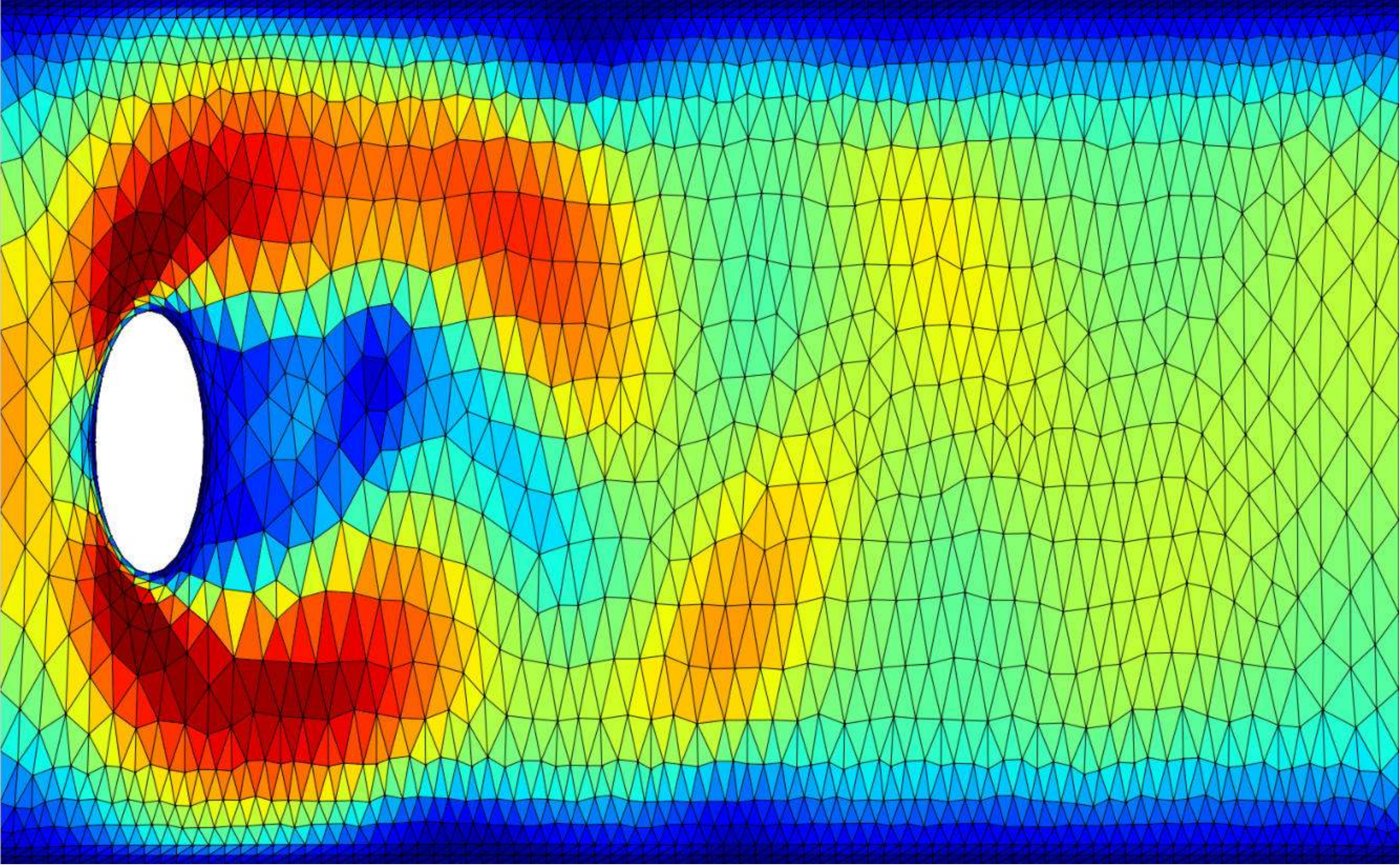} \\
    \;+ DIGL & \includegraphics[width=0.28\textwidth]{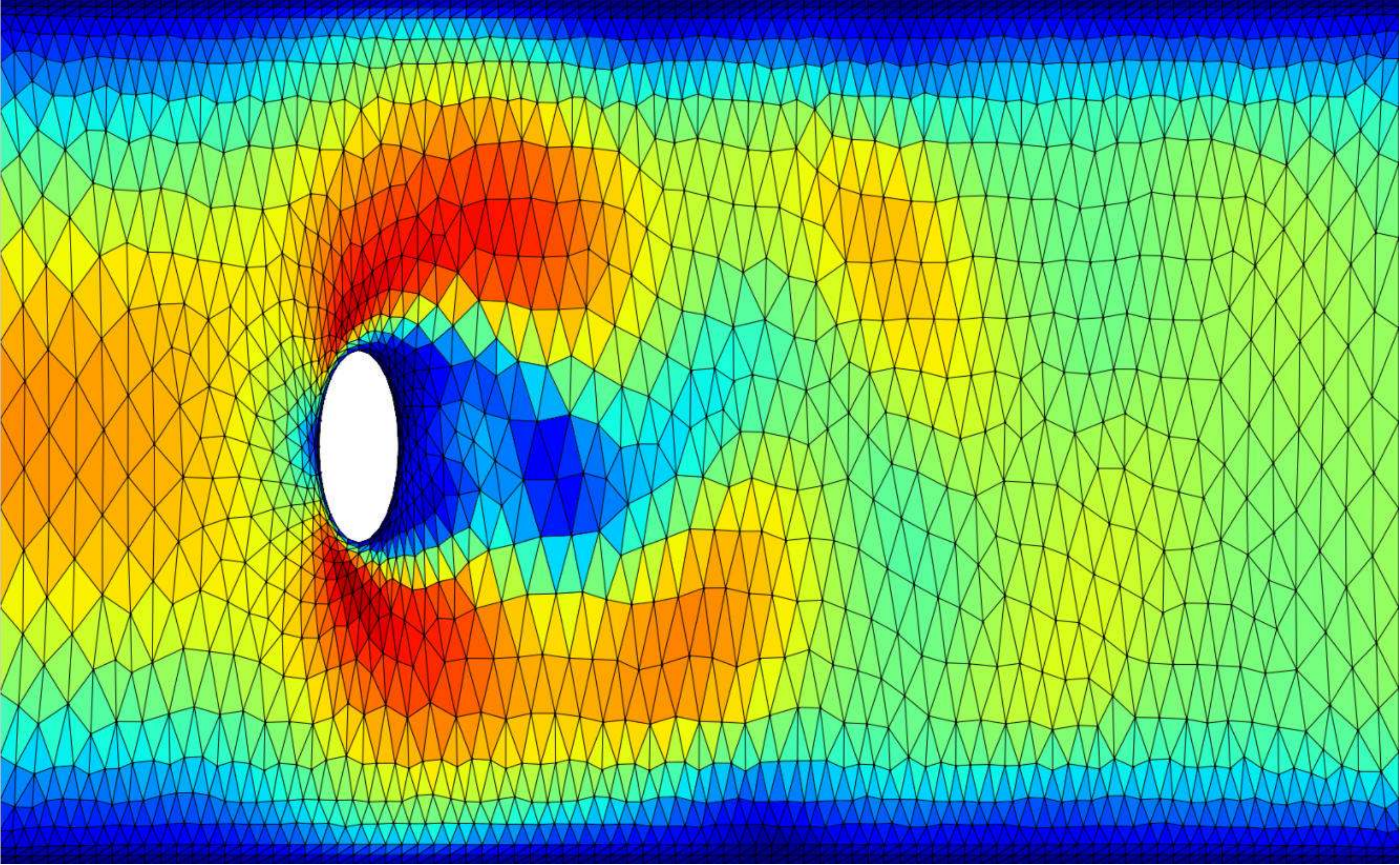} & \includegraphics[width=0.28\textwidth]{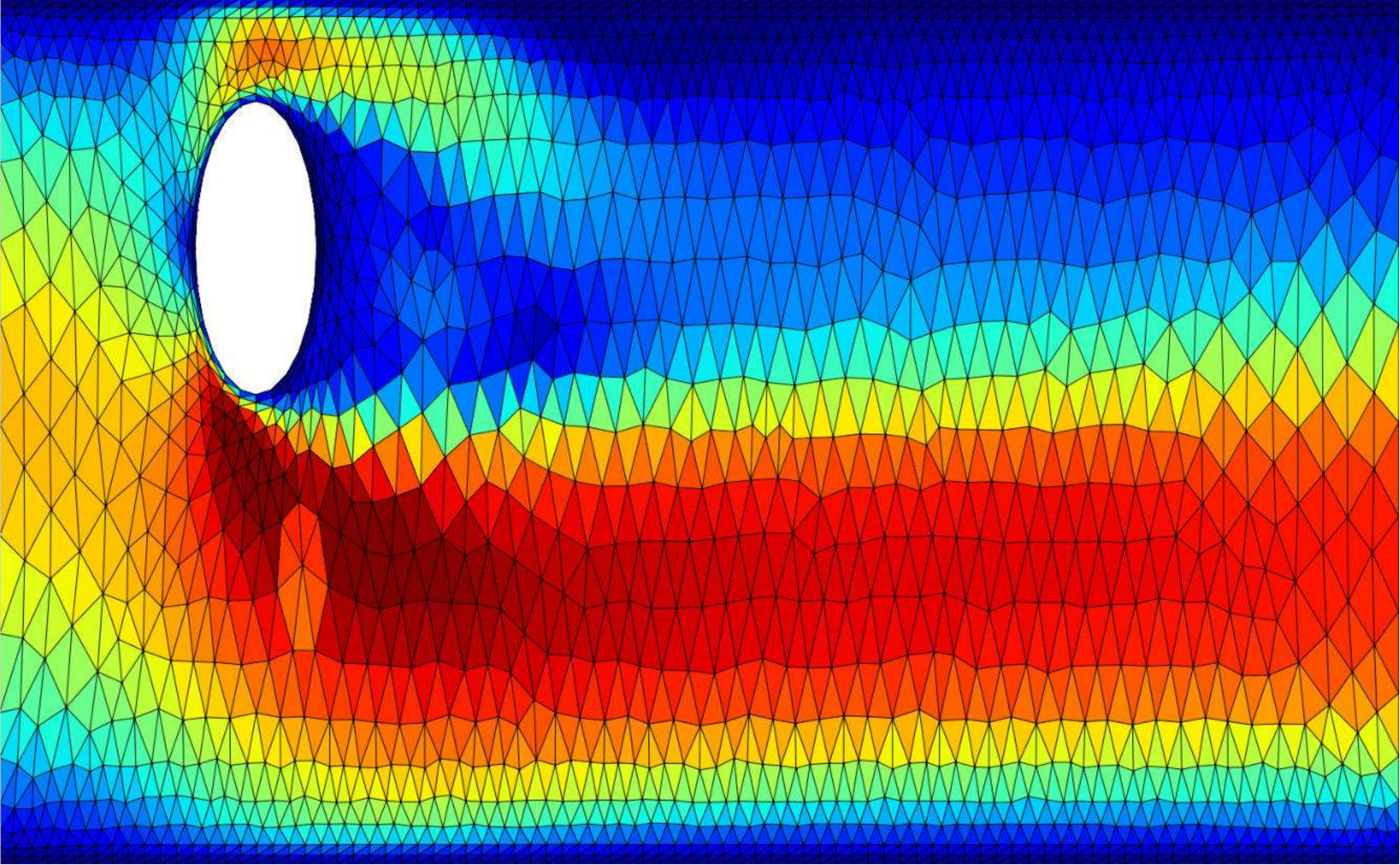} & \includegraphics[width=0.28\textwidth]{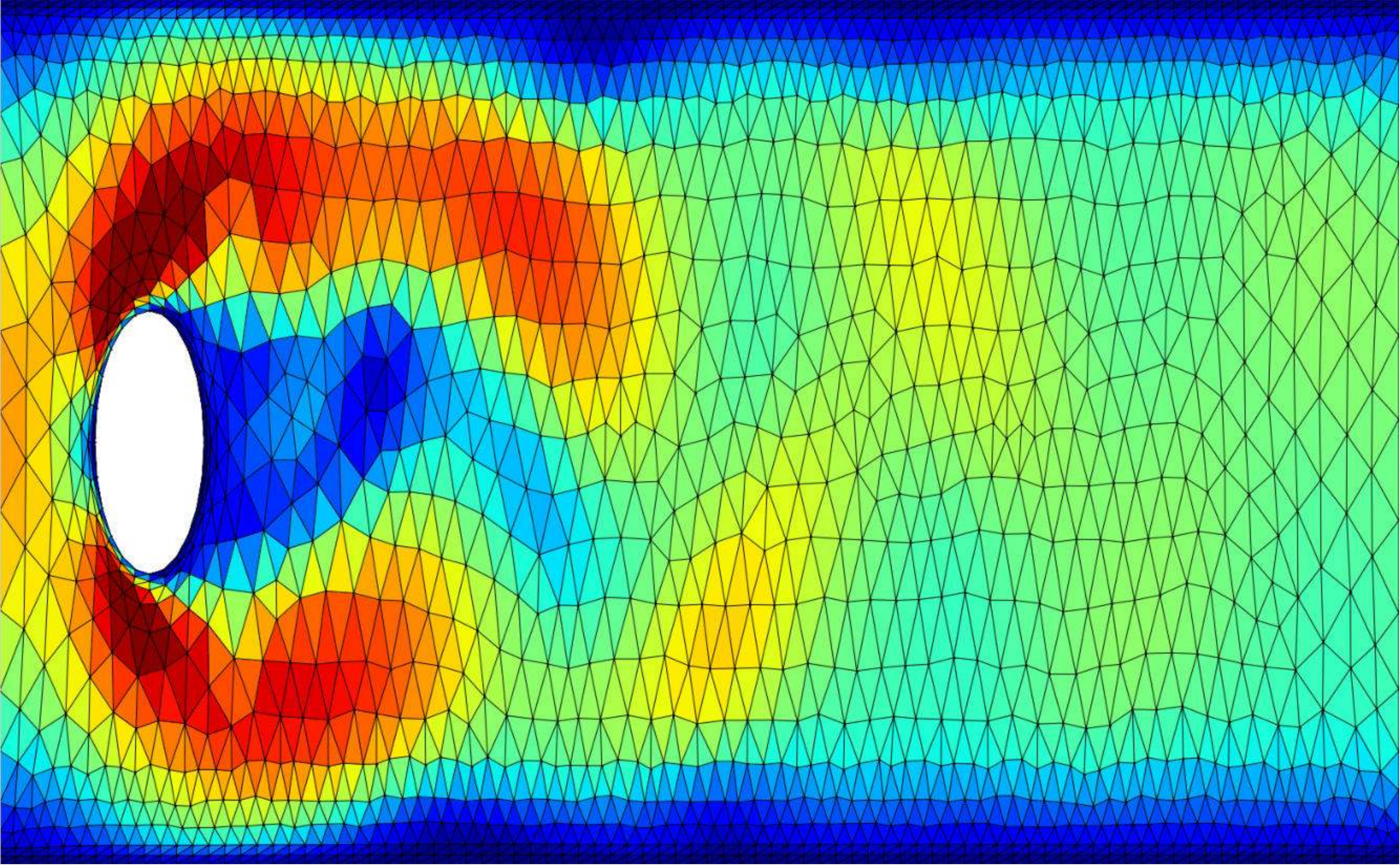}  \\       
    \;+ FoSR & \includegraphics[width=0.28\textwidth]{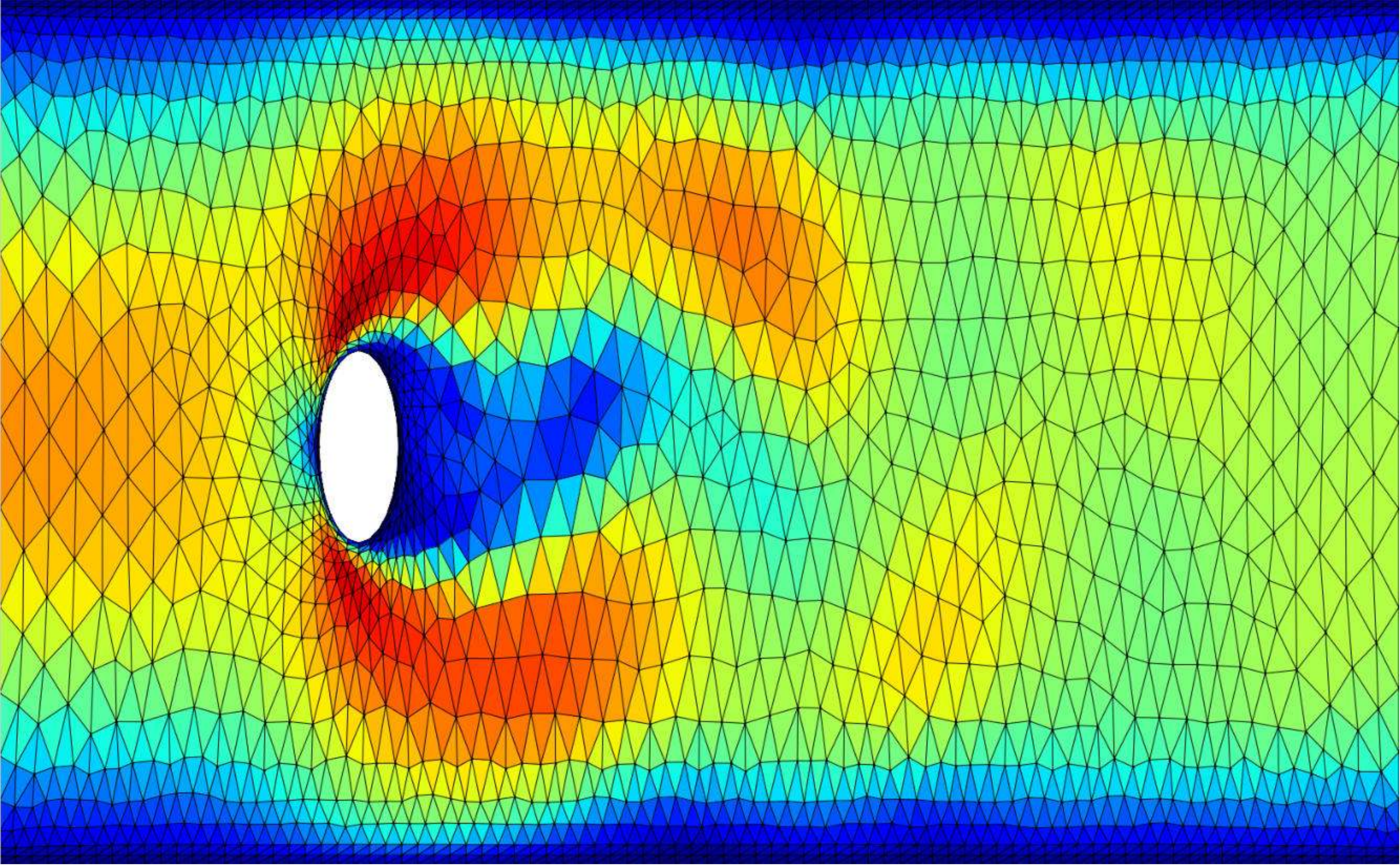} & \includegraphics[width=0.28\textwidth]{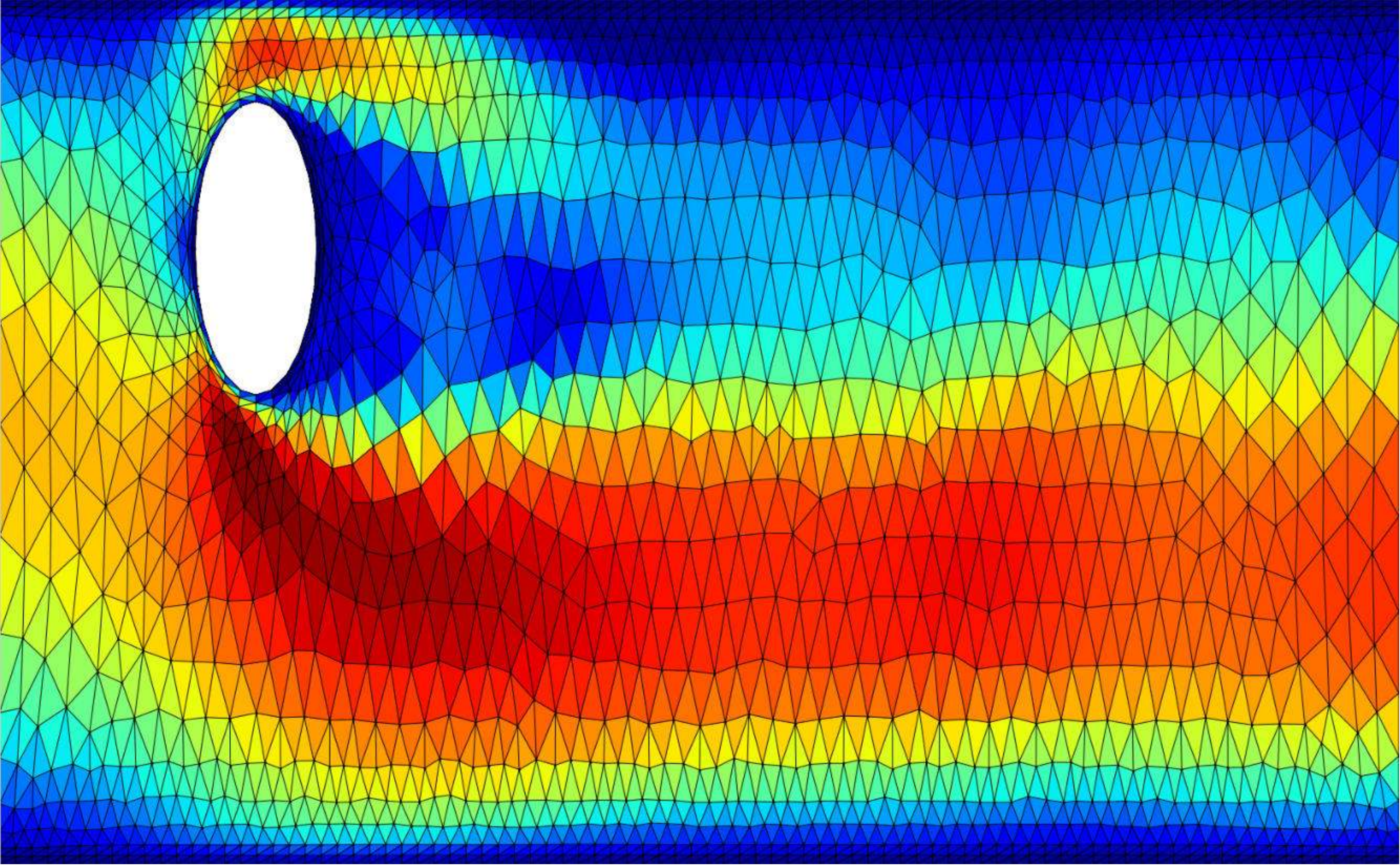} & \includegraphics[width=0.28\textwidth]{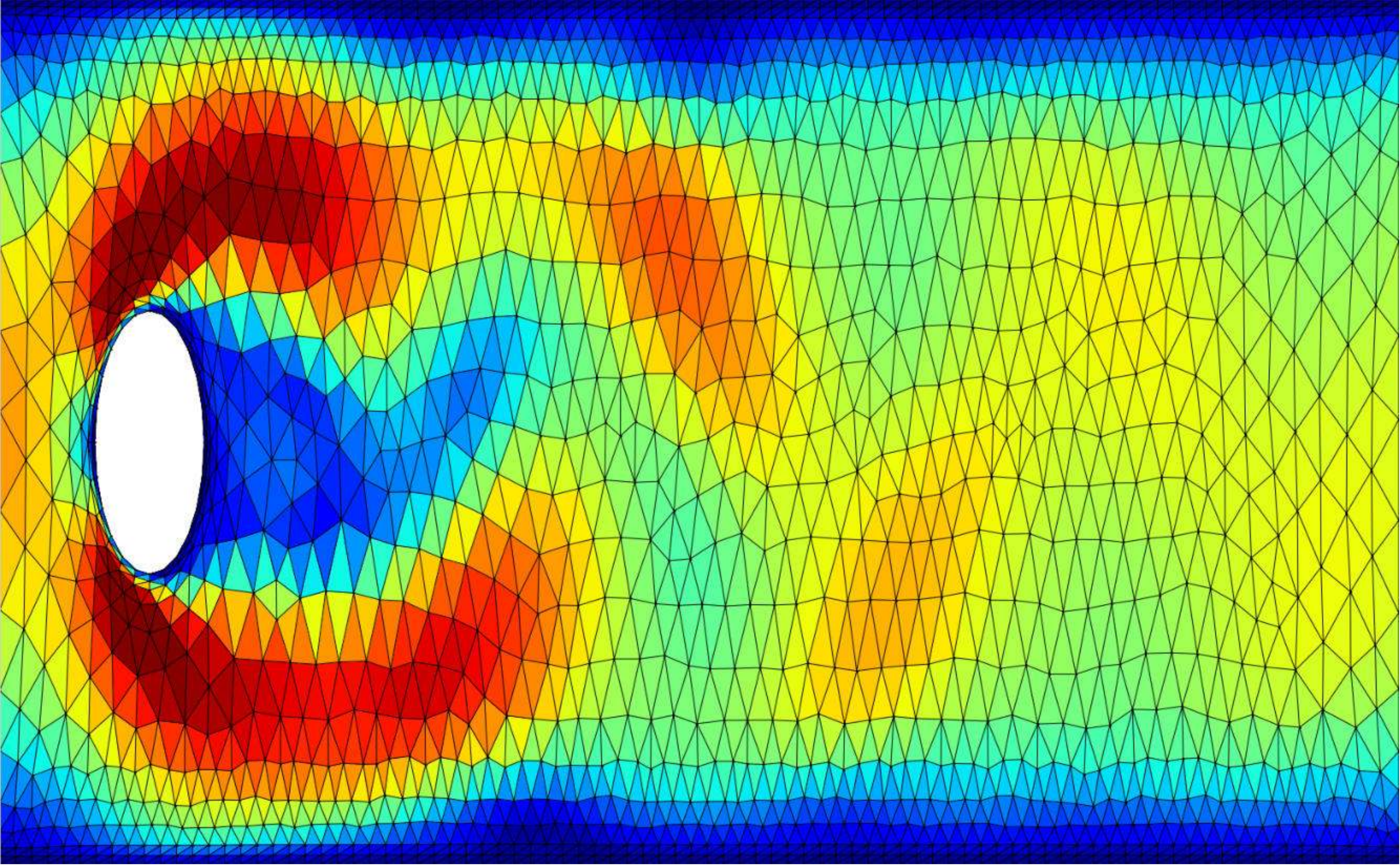}  \\    
    \;+ SDRF & \includegraphics[width=0.28\textwidth]{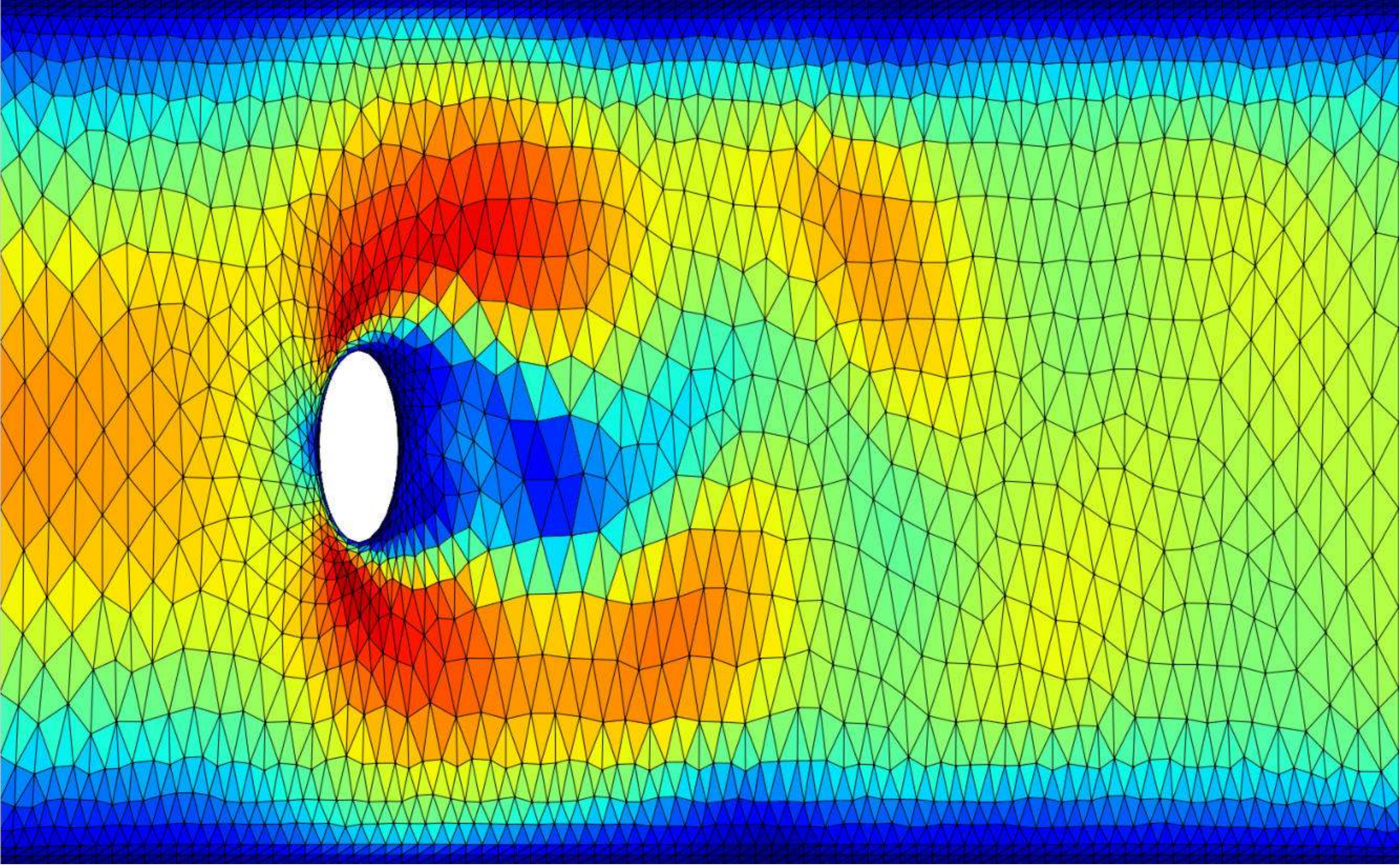} & \includegraphics[width=0.28\textwidth]{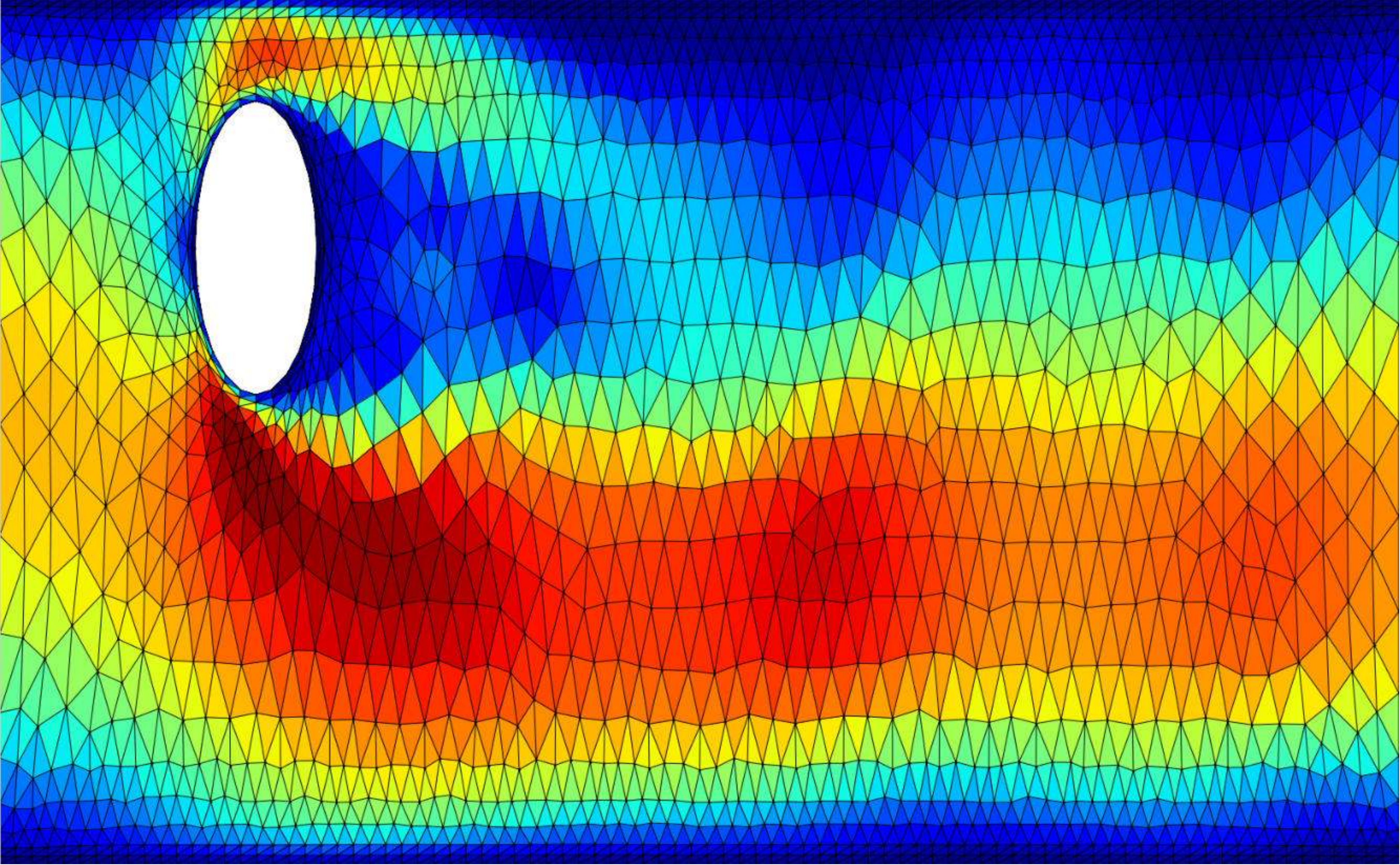} & \includegraphics[width=0.28\textwidth]{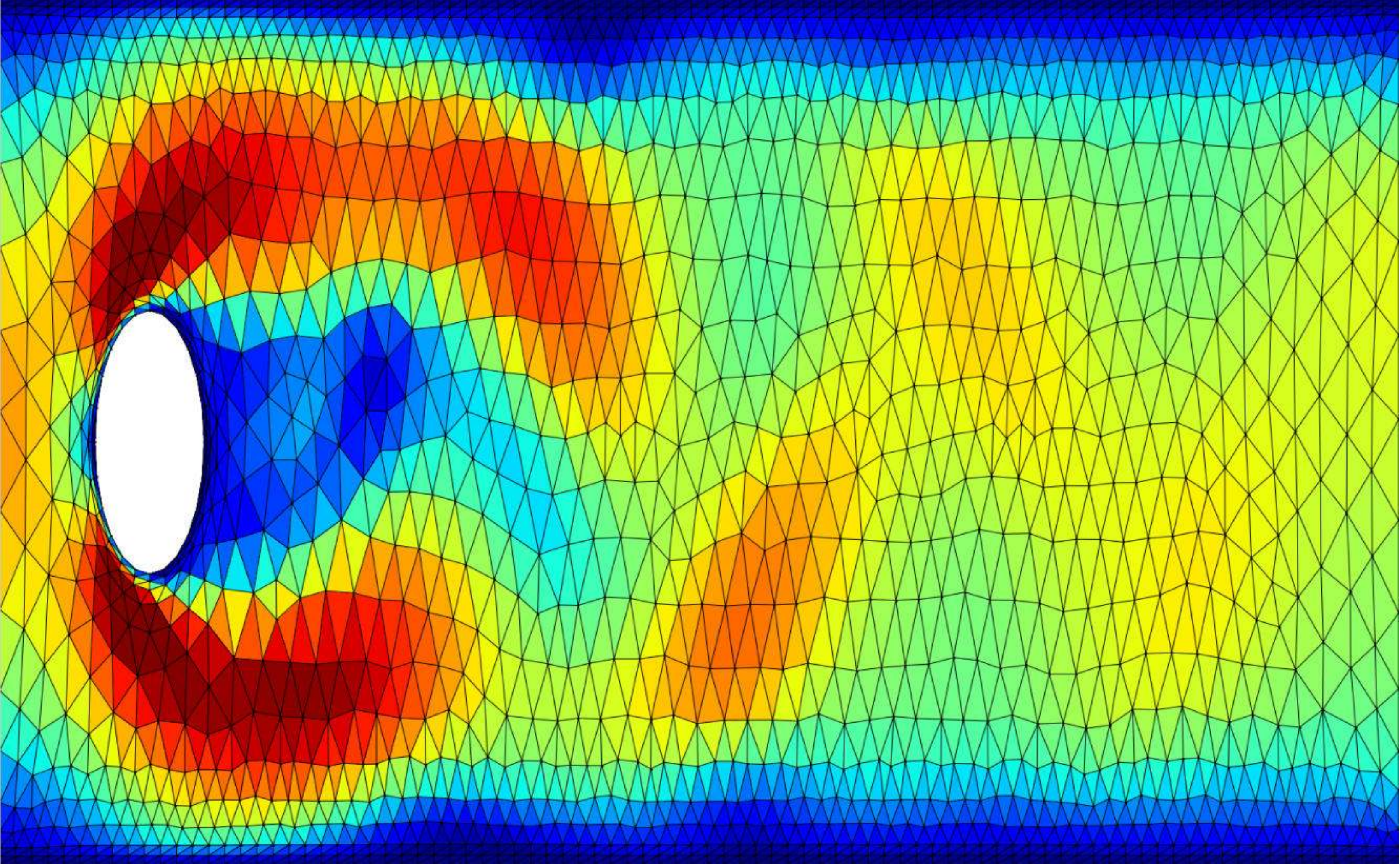}  \\   
    \;+ BORF & \includegraphics[width=0.28\textwidth]{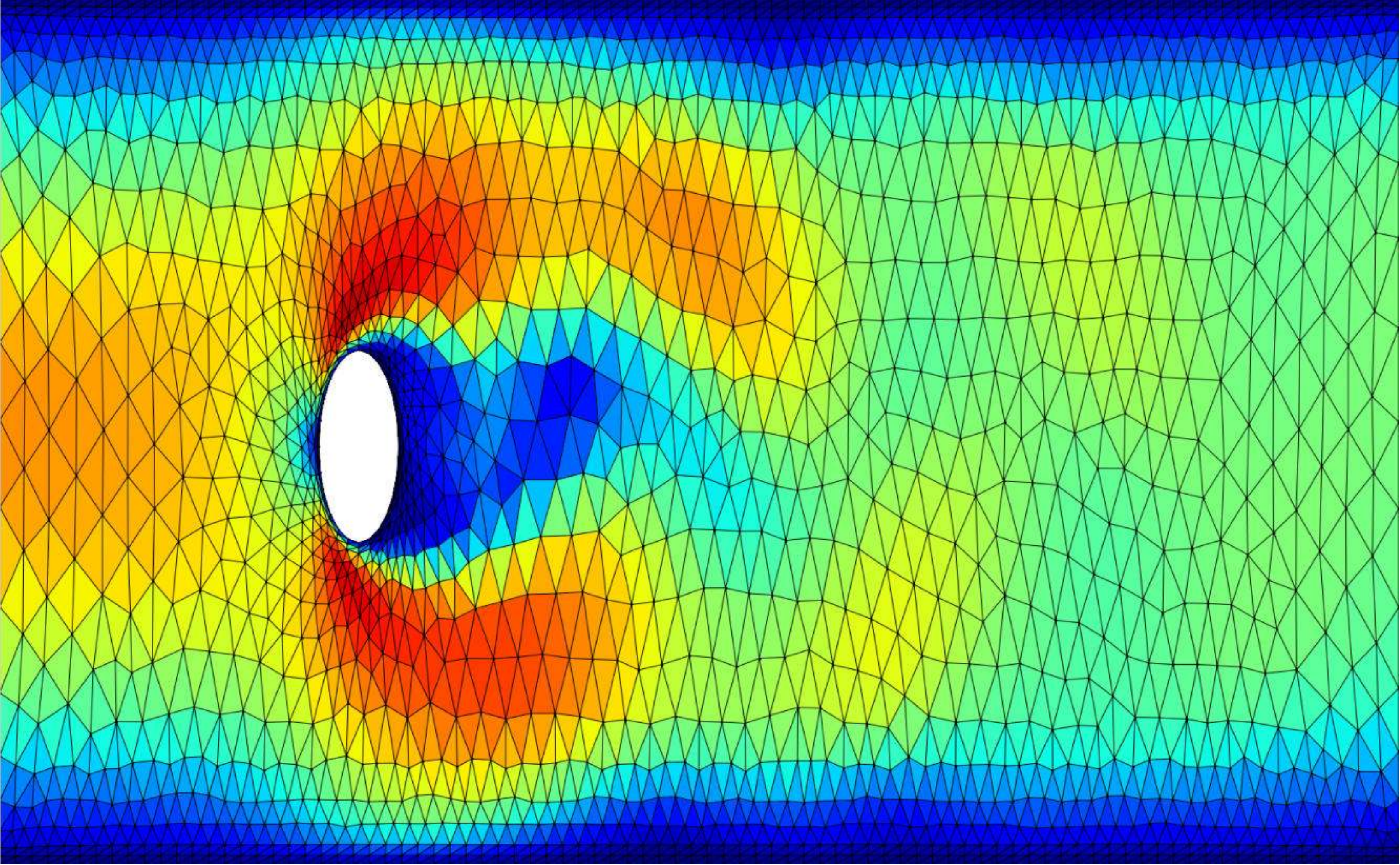} & \includegraphics[width=0.28\textwidth]{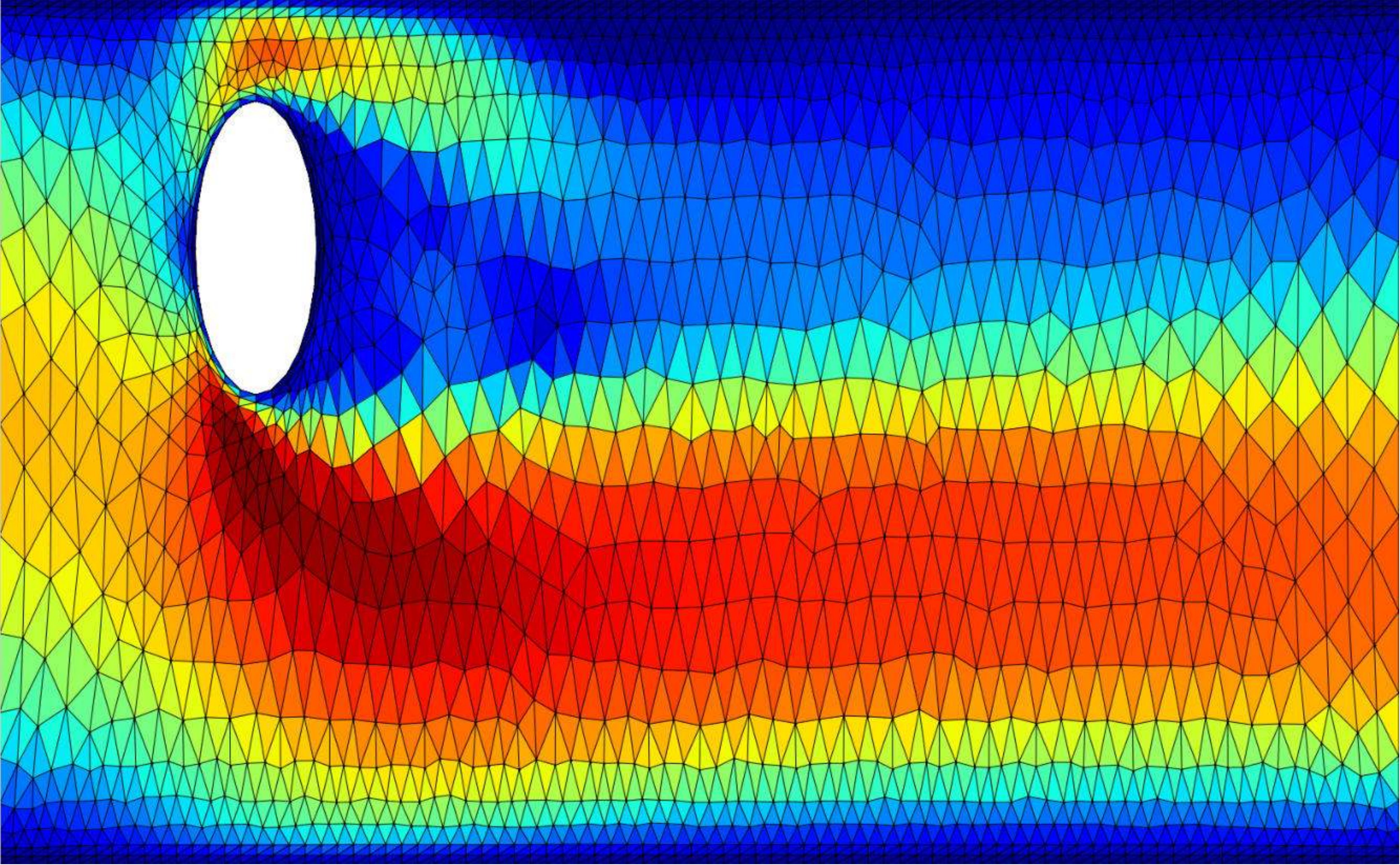} & \includegraphics[width=0.28\textwidth]{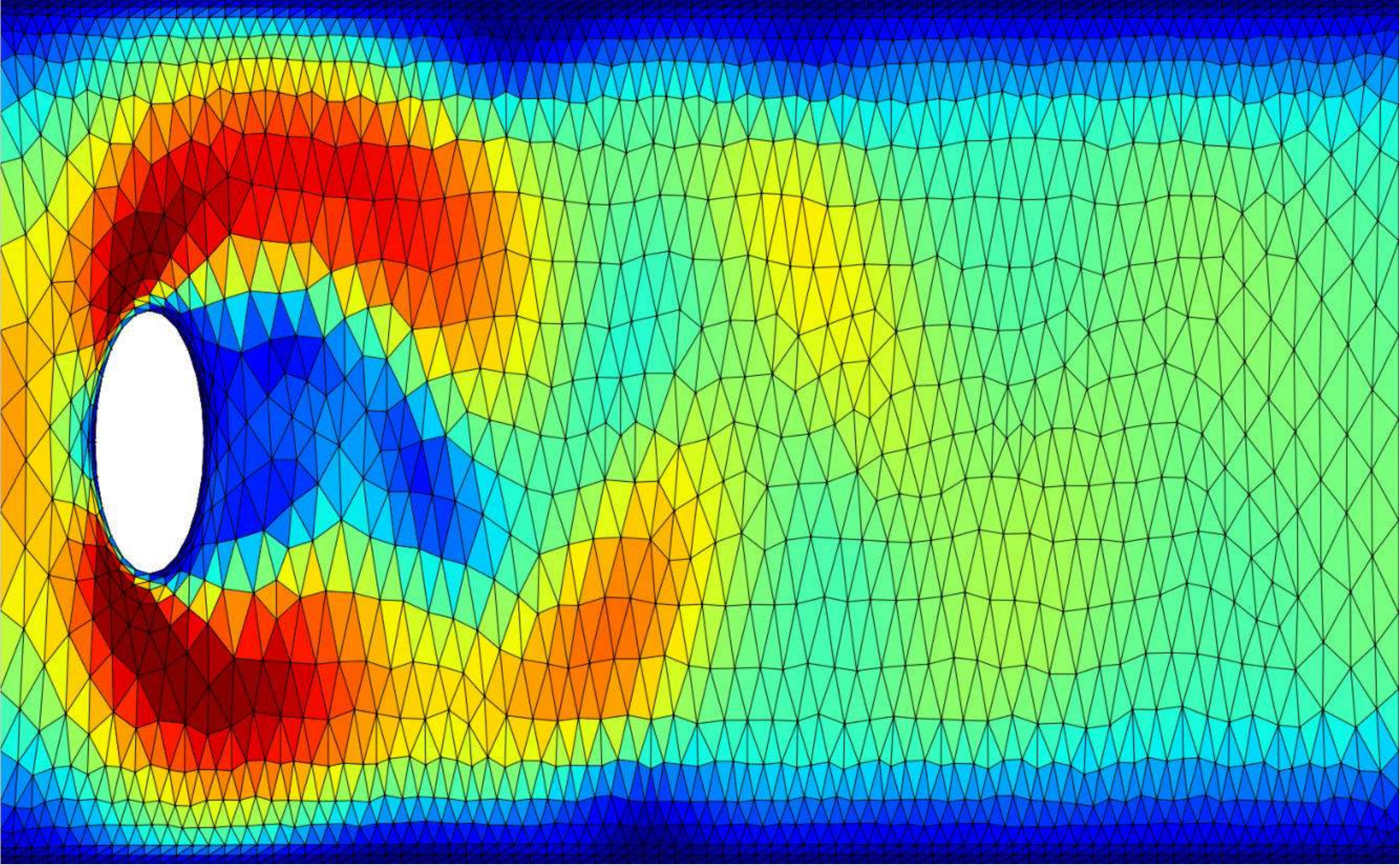}  \\       
    \;+ PIORF & \includegraphics[width=0.28\textwidth]{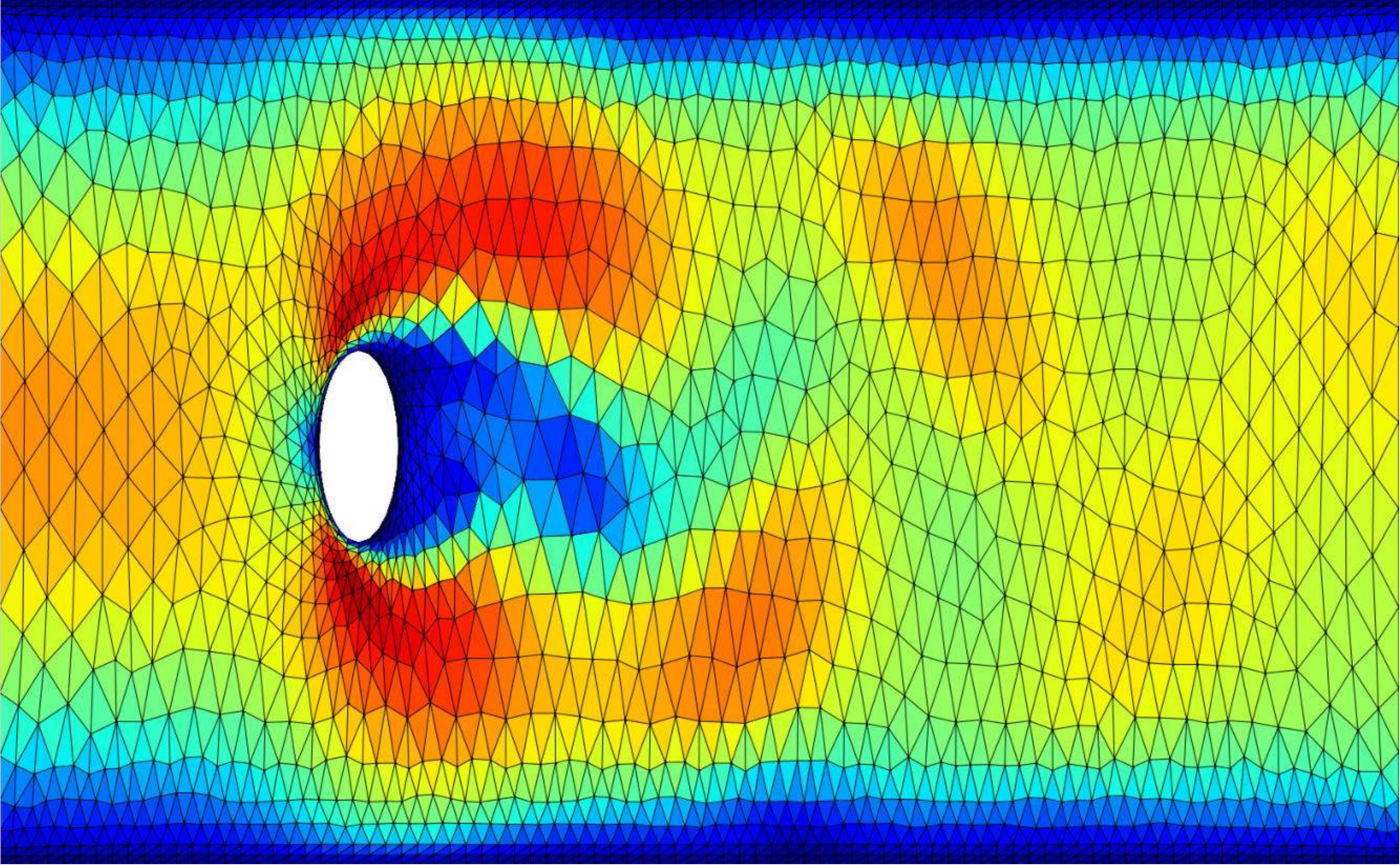} & \includegraphics[width=0.28\textwidth]{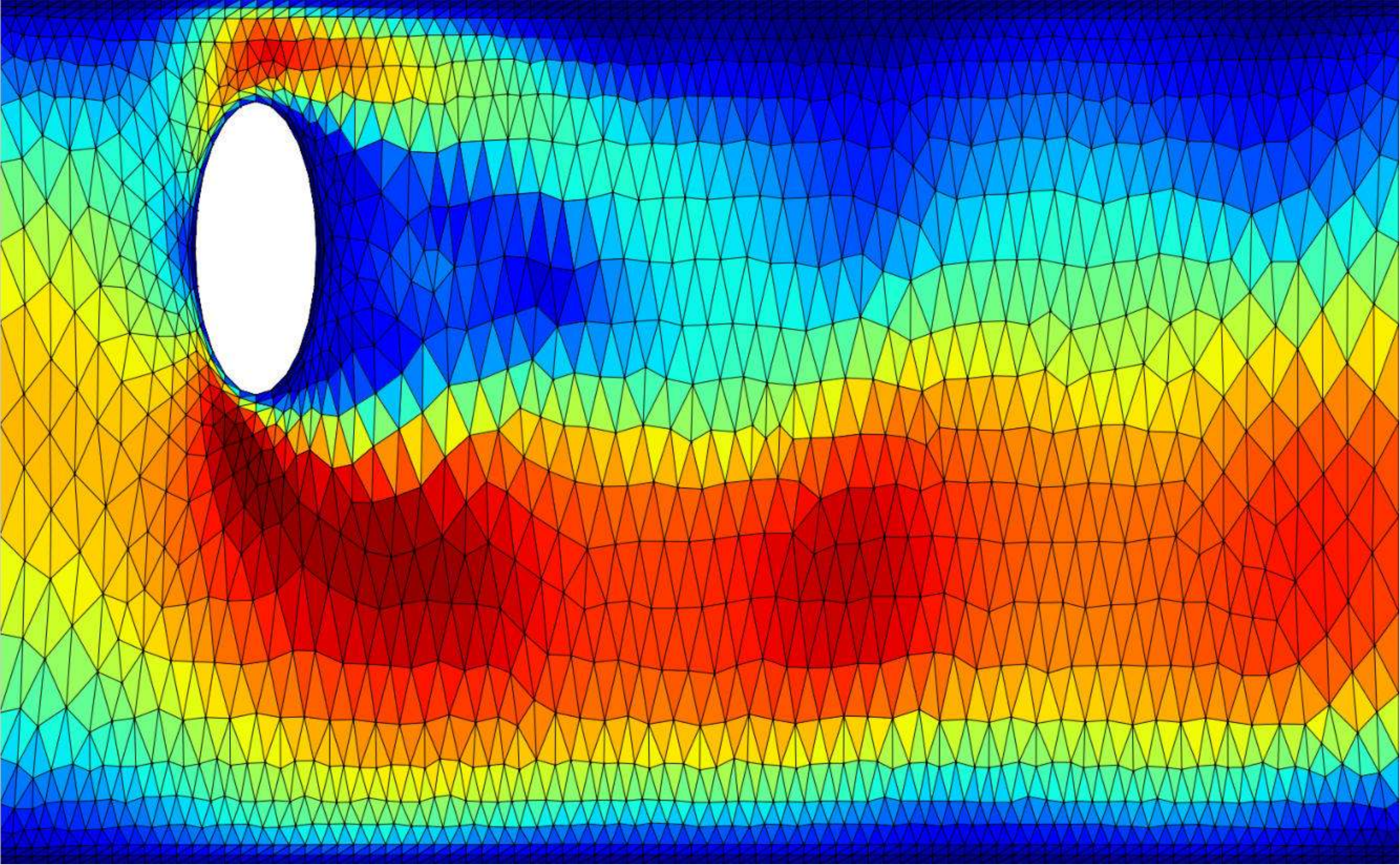} & \includegraphics[width=0.28\textwidth]{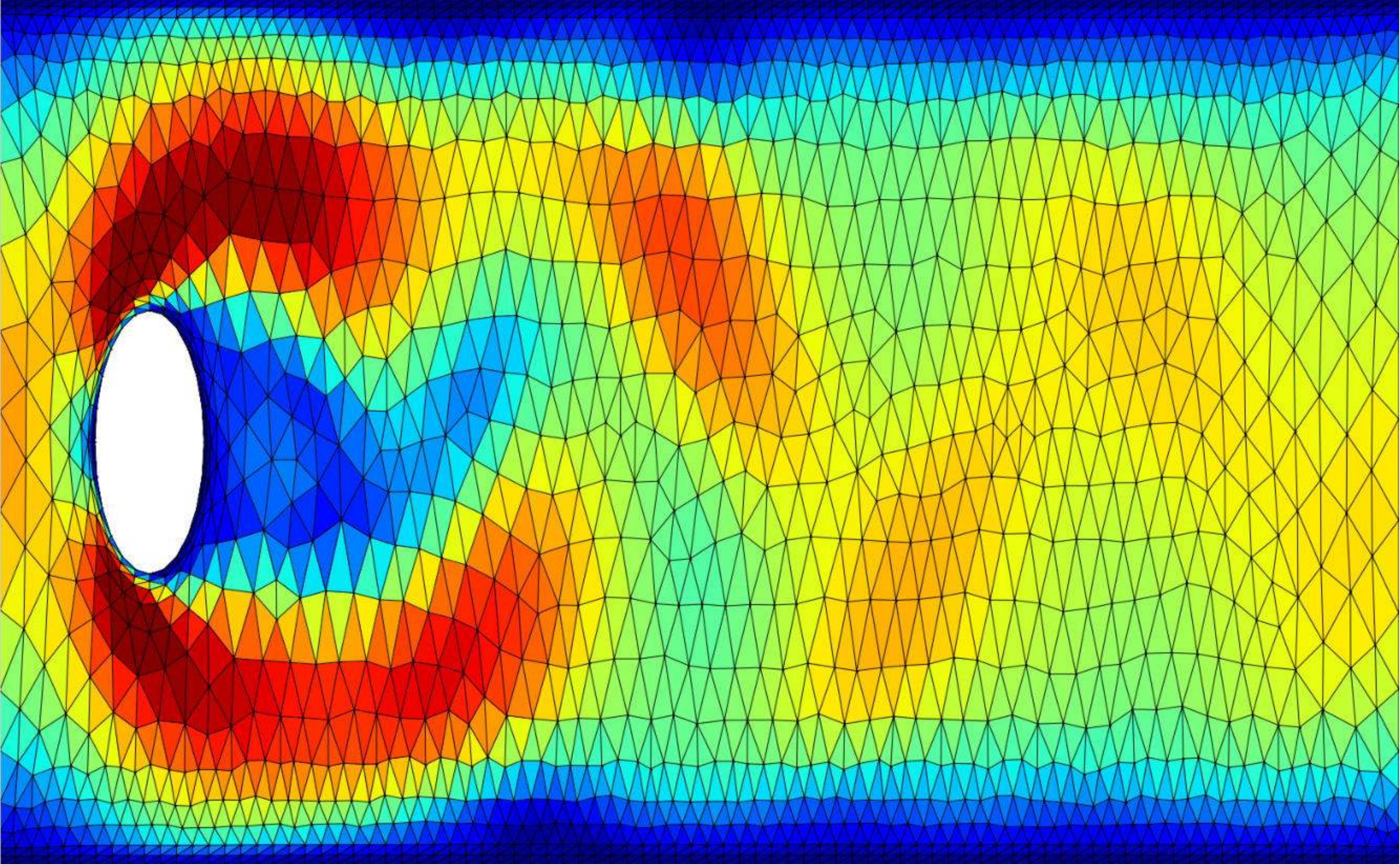}  \\      
    \bottomrule
    \end{tabular}
    \includegraphics[width=0.5\textwidth]{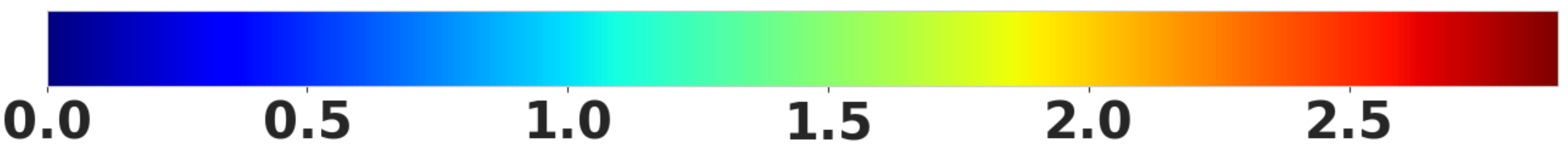}
    \caption{The velocity magnitude contours of various rewiring methods compared to the ground truth at \textsc{CylinderFlow}}
    \label{fig:roll_cylinder}
\end{figure}

\begin{figure}[h]
    \centering
    \setlength{\tabcolsep}{1pt}
    \begin{tabular} {l ccc}\toprule
    Models & Traj. 1 & Traj. 2 & Traj. 3 \\ \midrule
        Ground Truth & \includegraphics[width=0.28\textwidth]{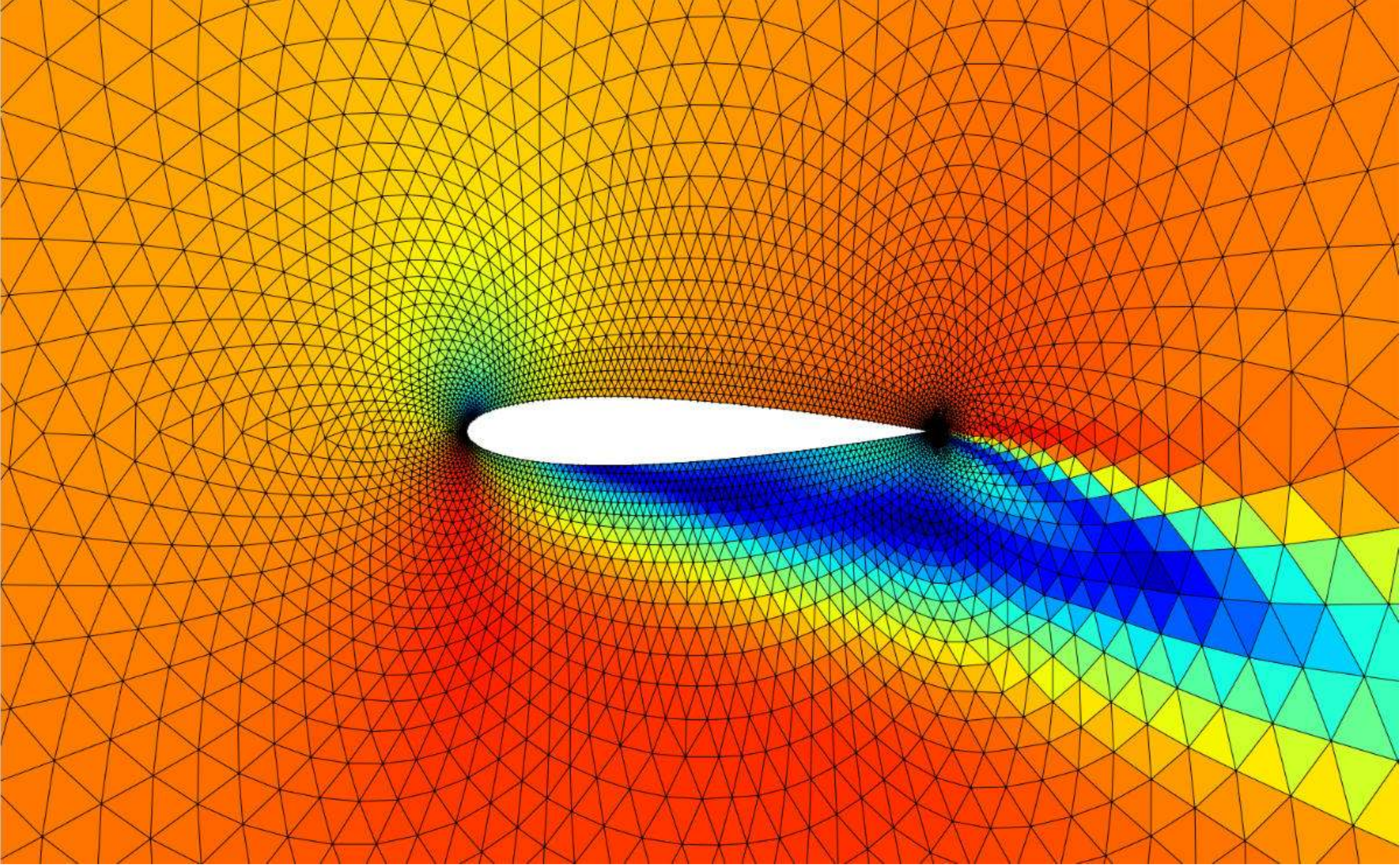} & \includegraphics[width=0.28\textwidth]{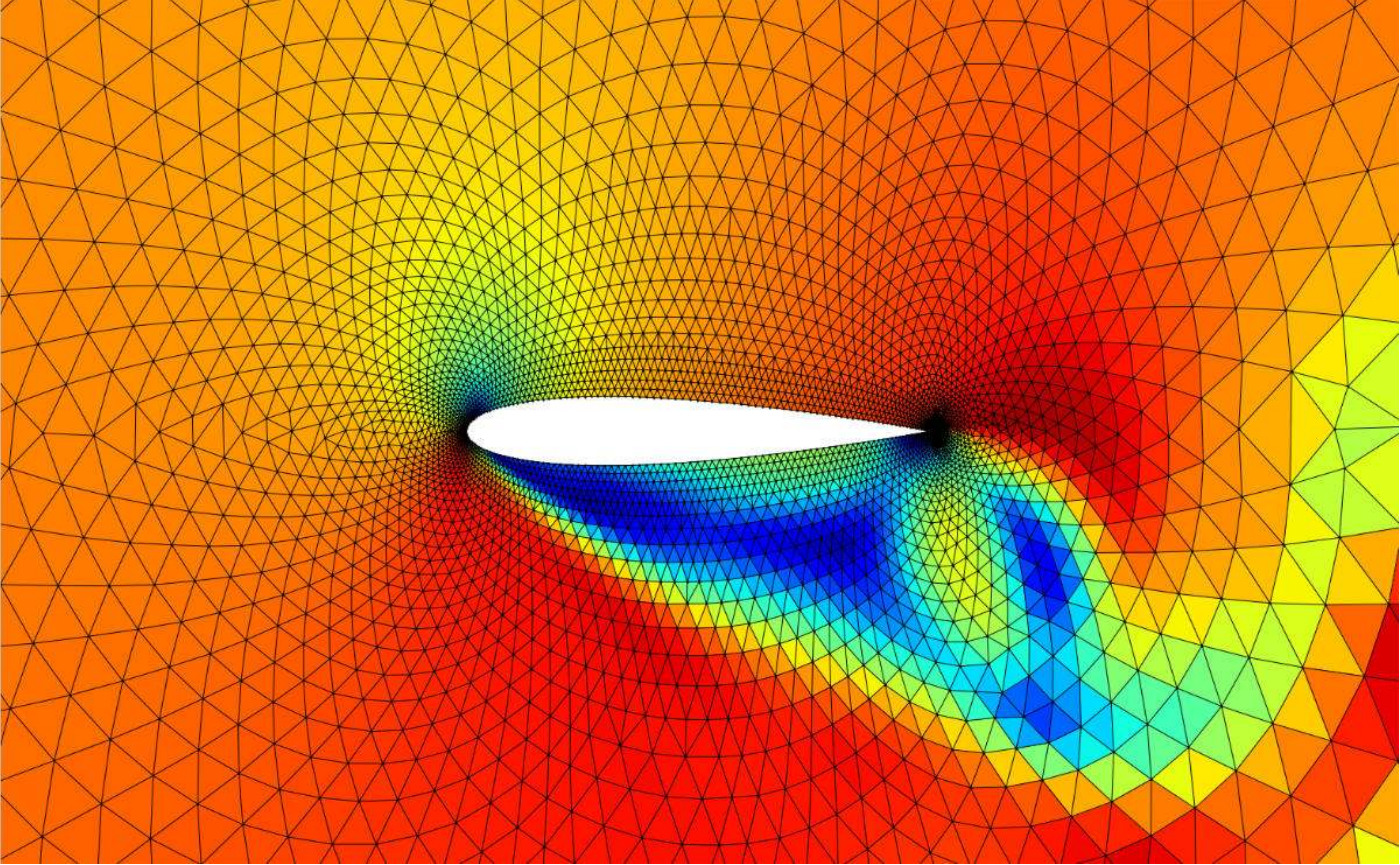} & \includegraphics[width=0.28\textwidth]{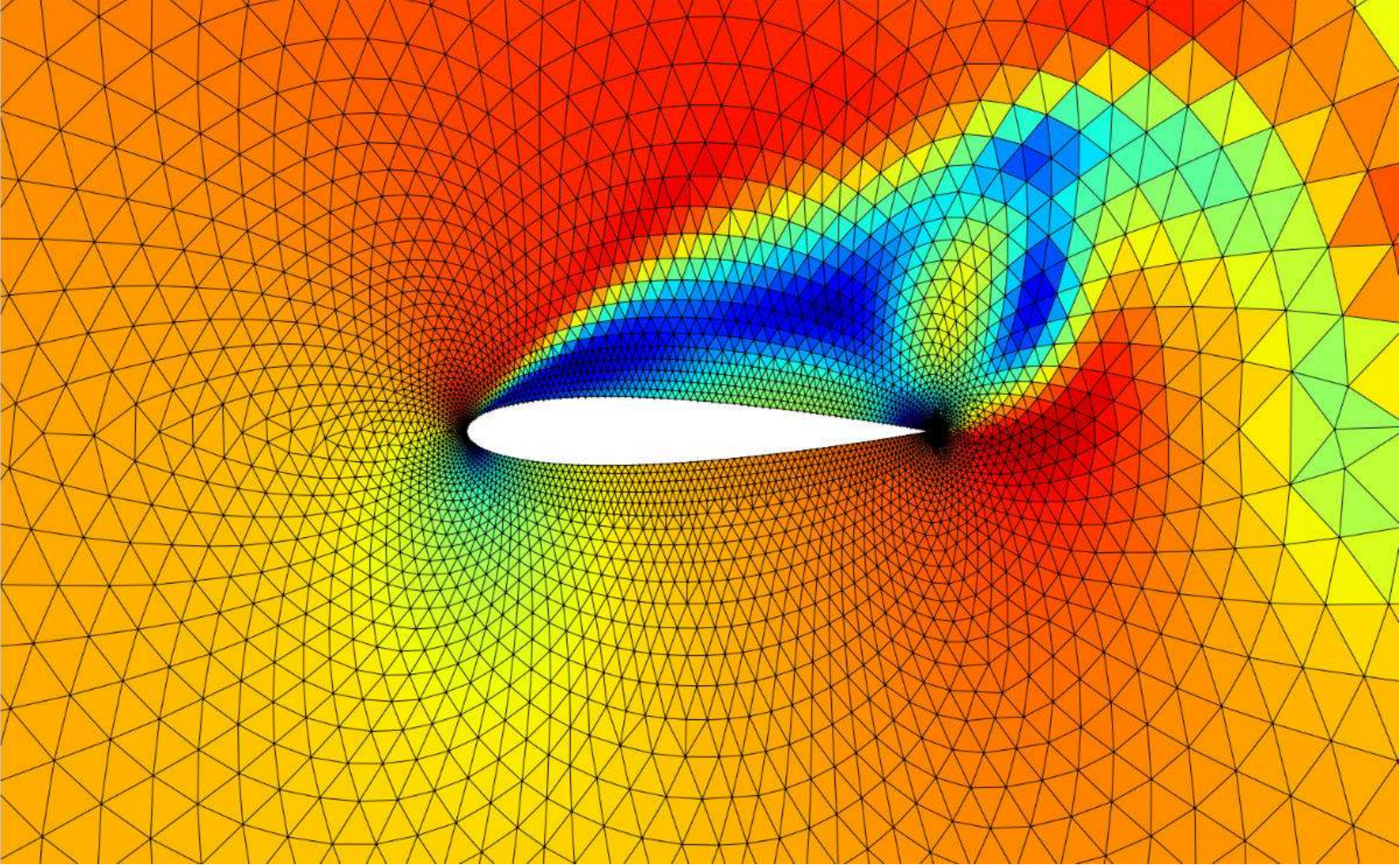}   \\
        MGN & \includegraphics[width=0.28\textwidth]{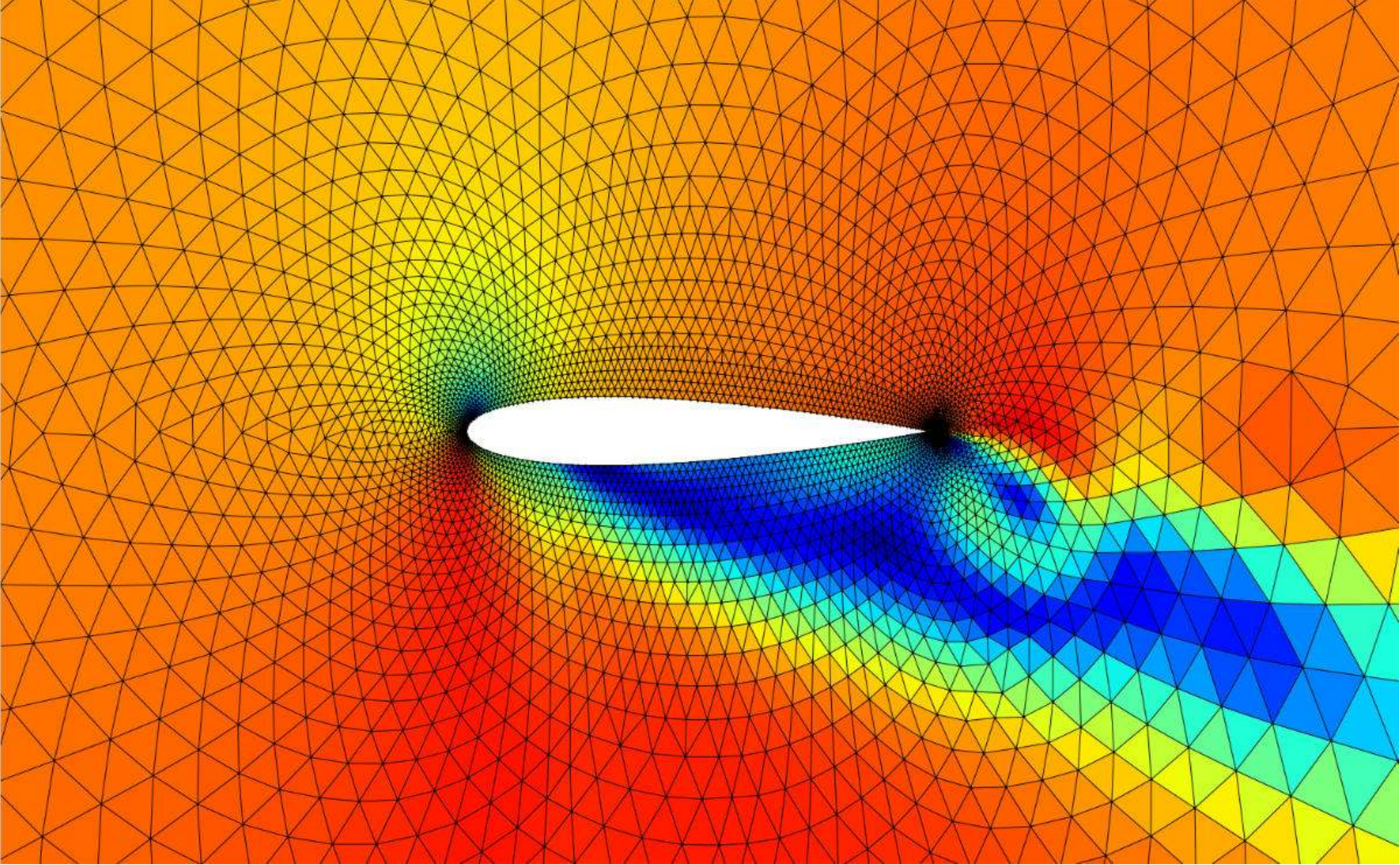} & \includegraphics[width=0.28\textwidth]{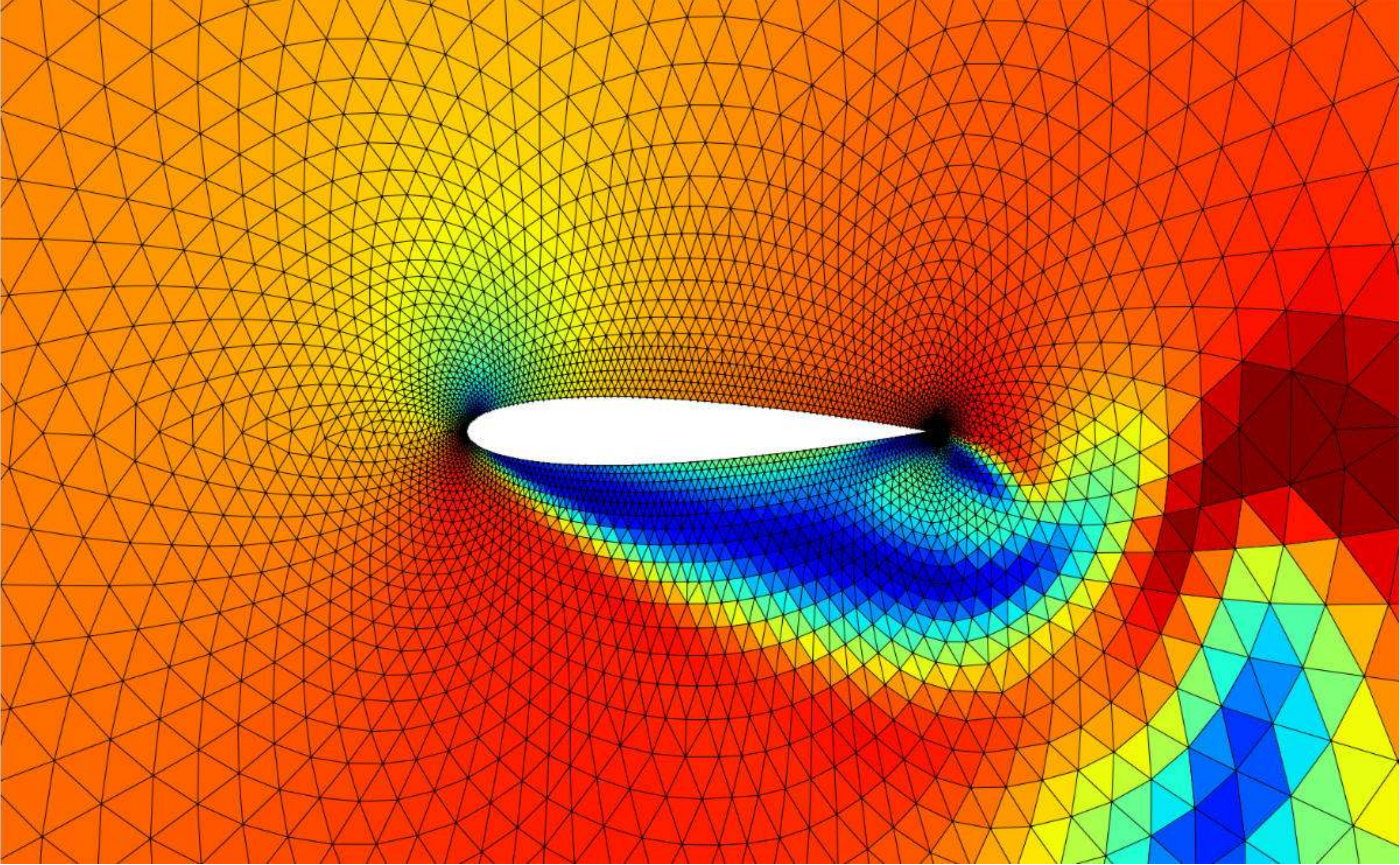} & \includegraphics[width=0.28\textwidth]{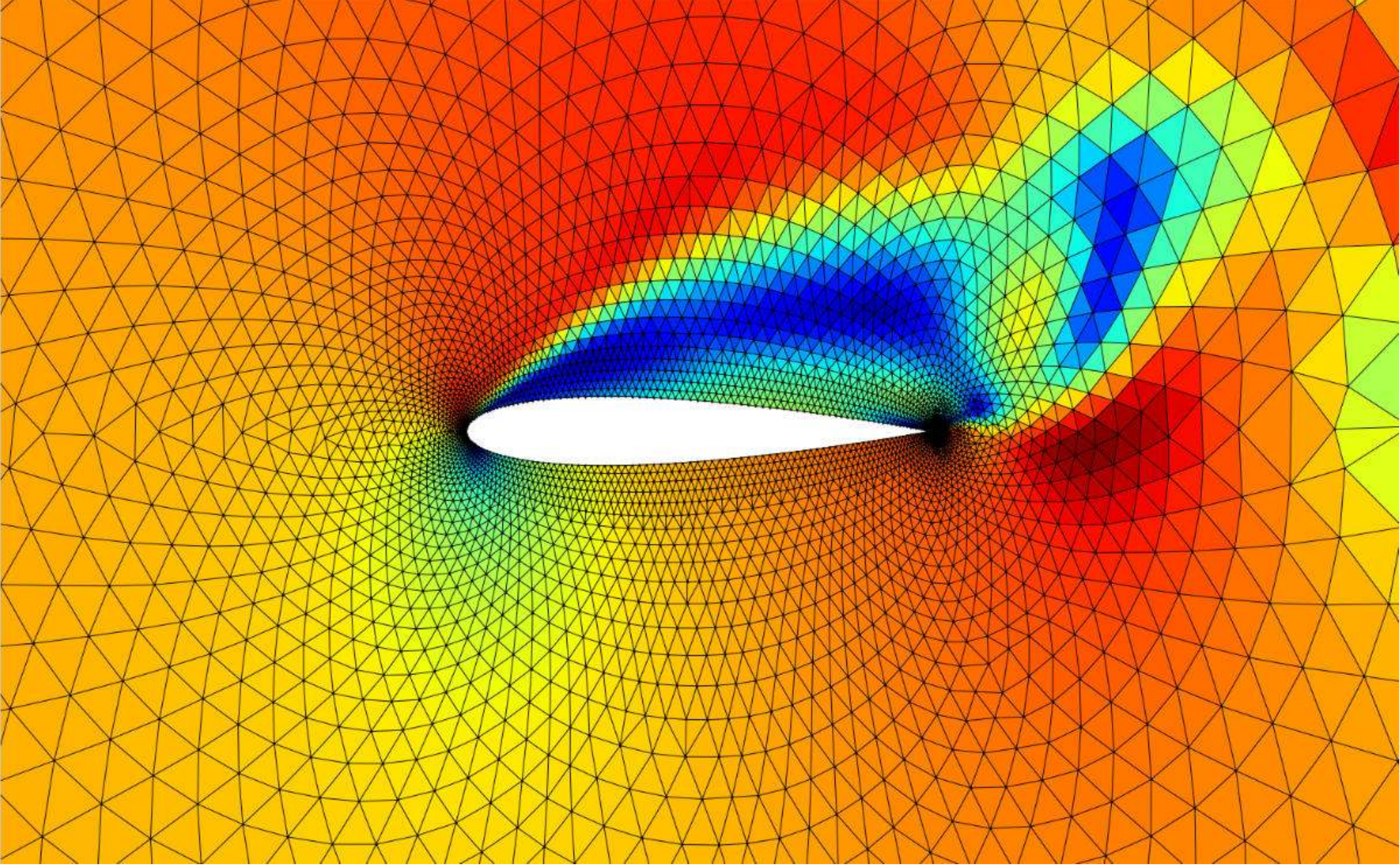} \\
        \;+ DIGL & \includegraphics[width=0.28\textwidth]{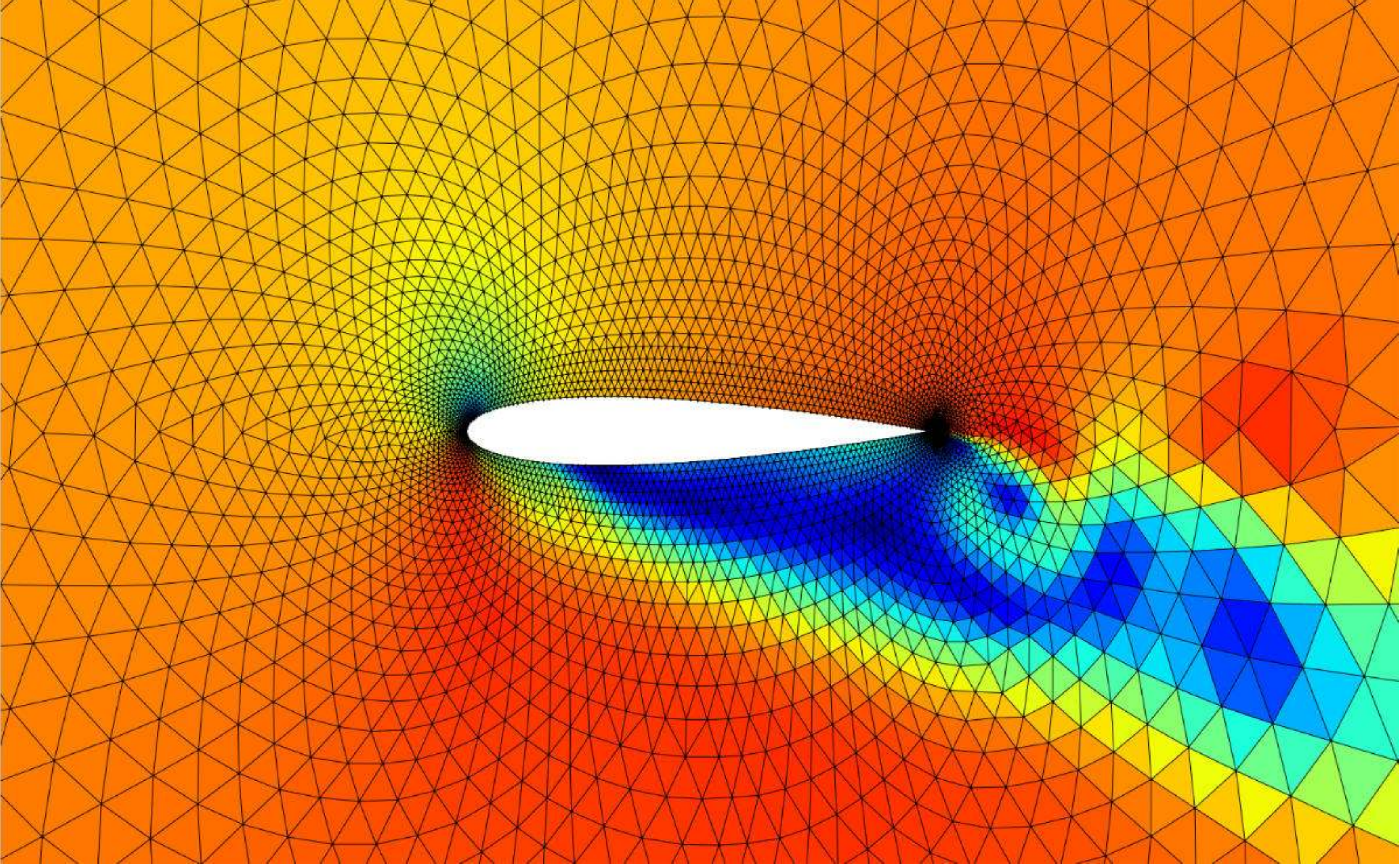} & \includegraphics[width=0.28\textwidth]{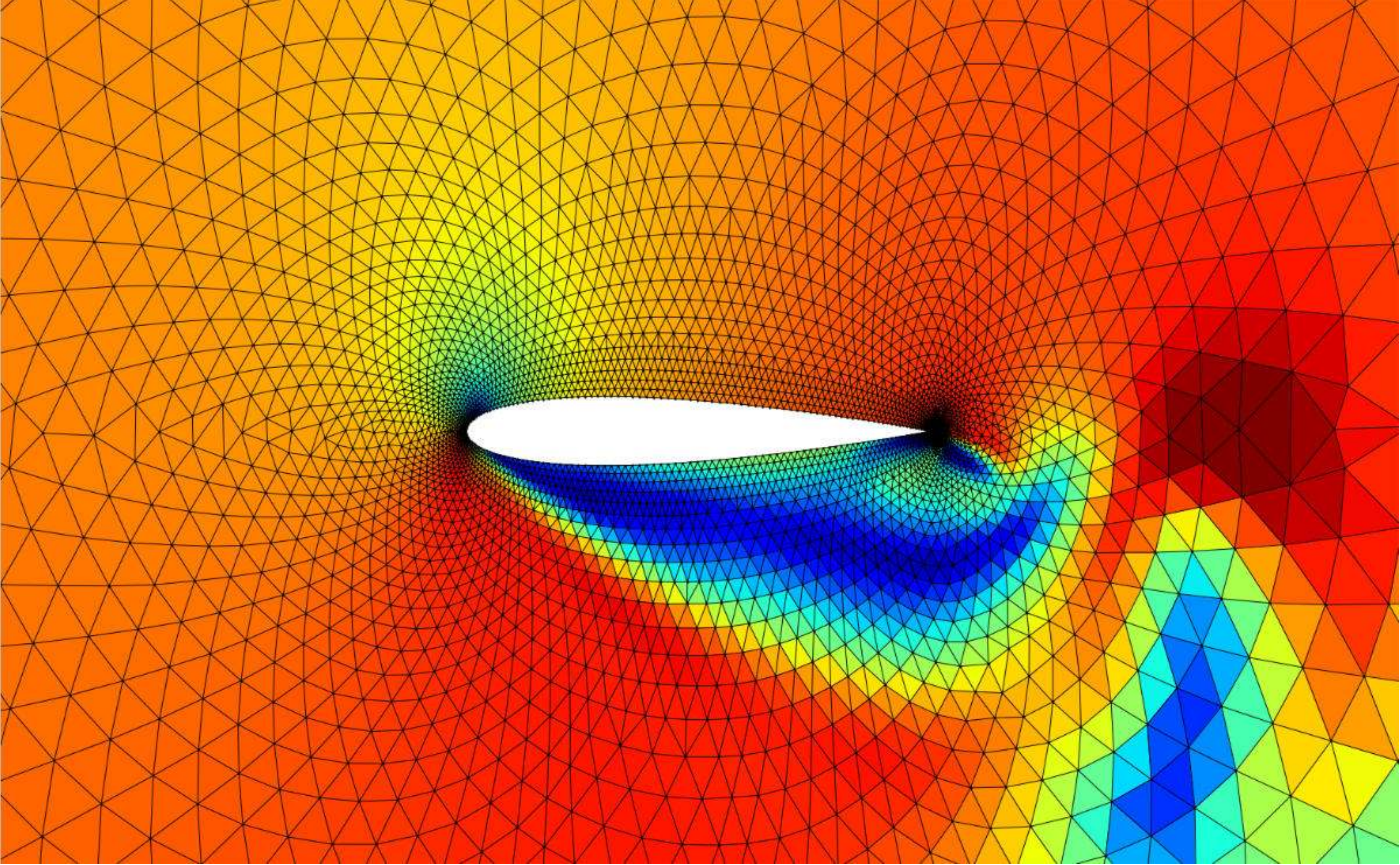} & \includegraphics[width=0.28\textwidth]{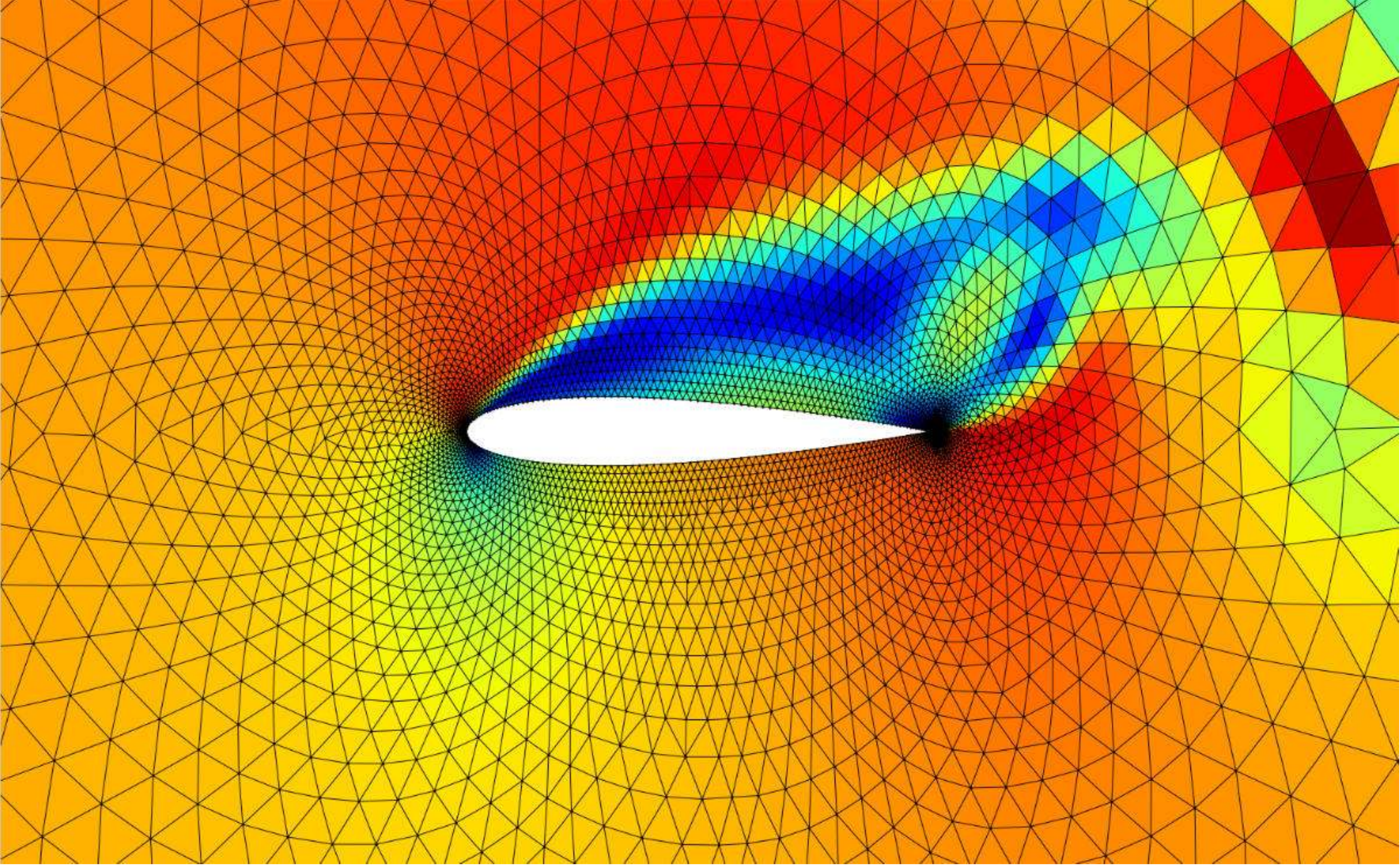} \\       
        \;+ FoSR & \includegraphics[width=0.28\textwidth]{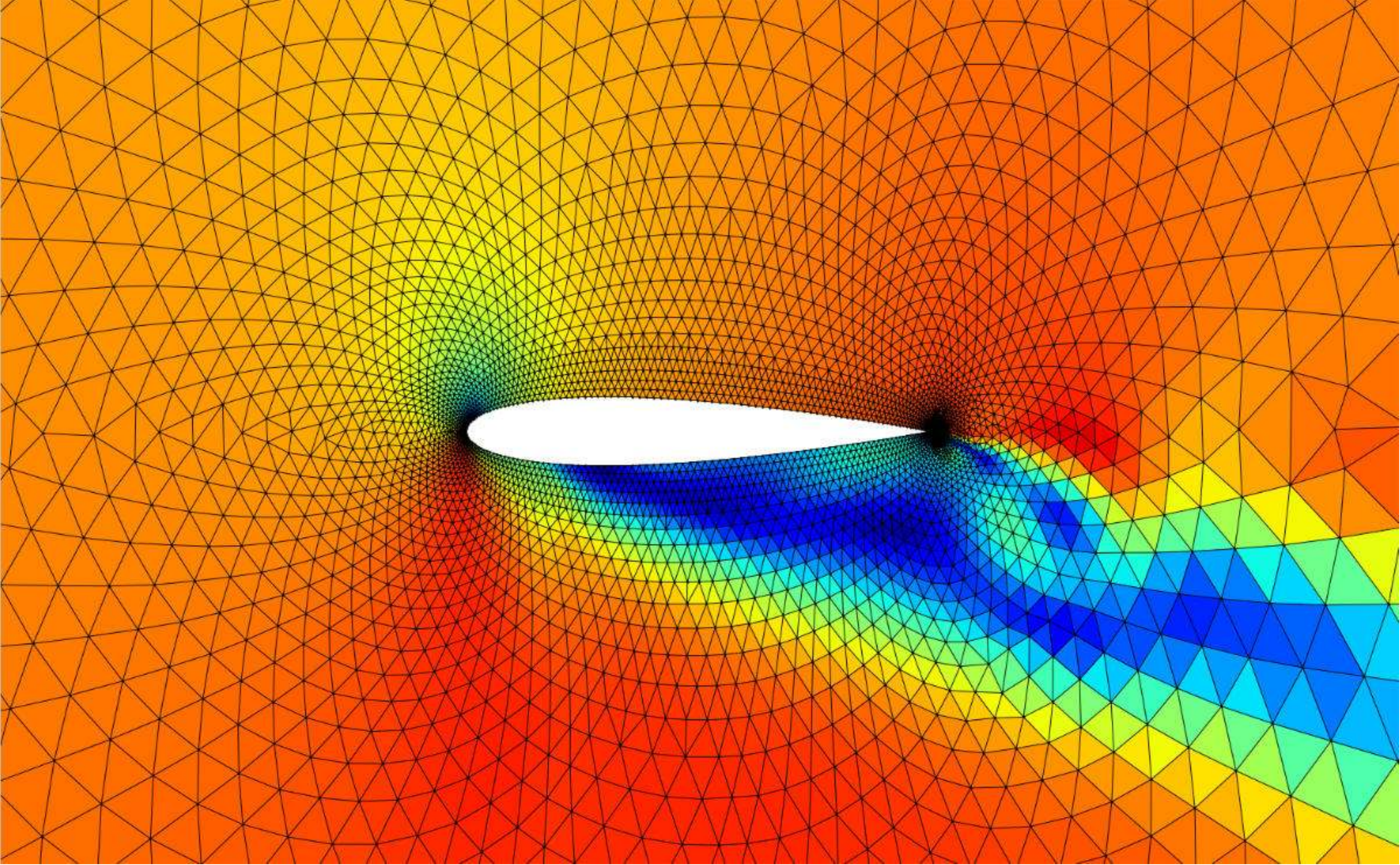} & \includegraphics[width=0.28\textwidth]{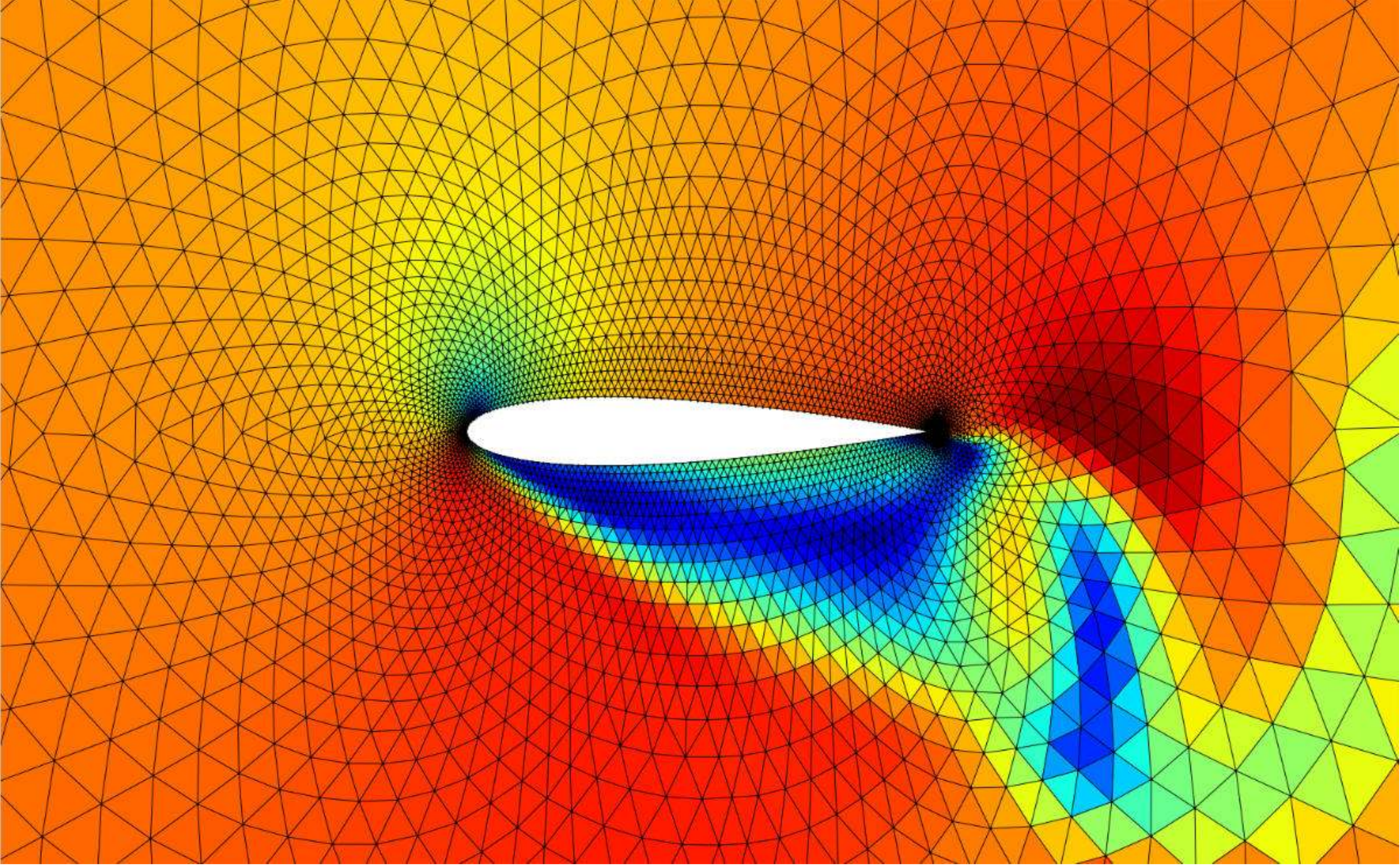} & \includegraphics[width=0.28\textwidth]{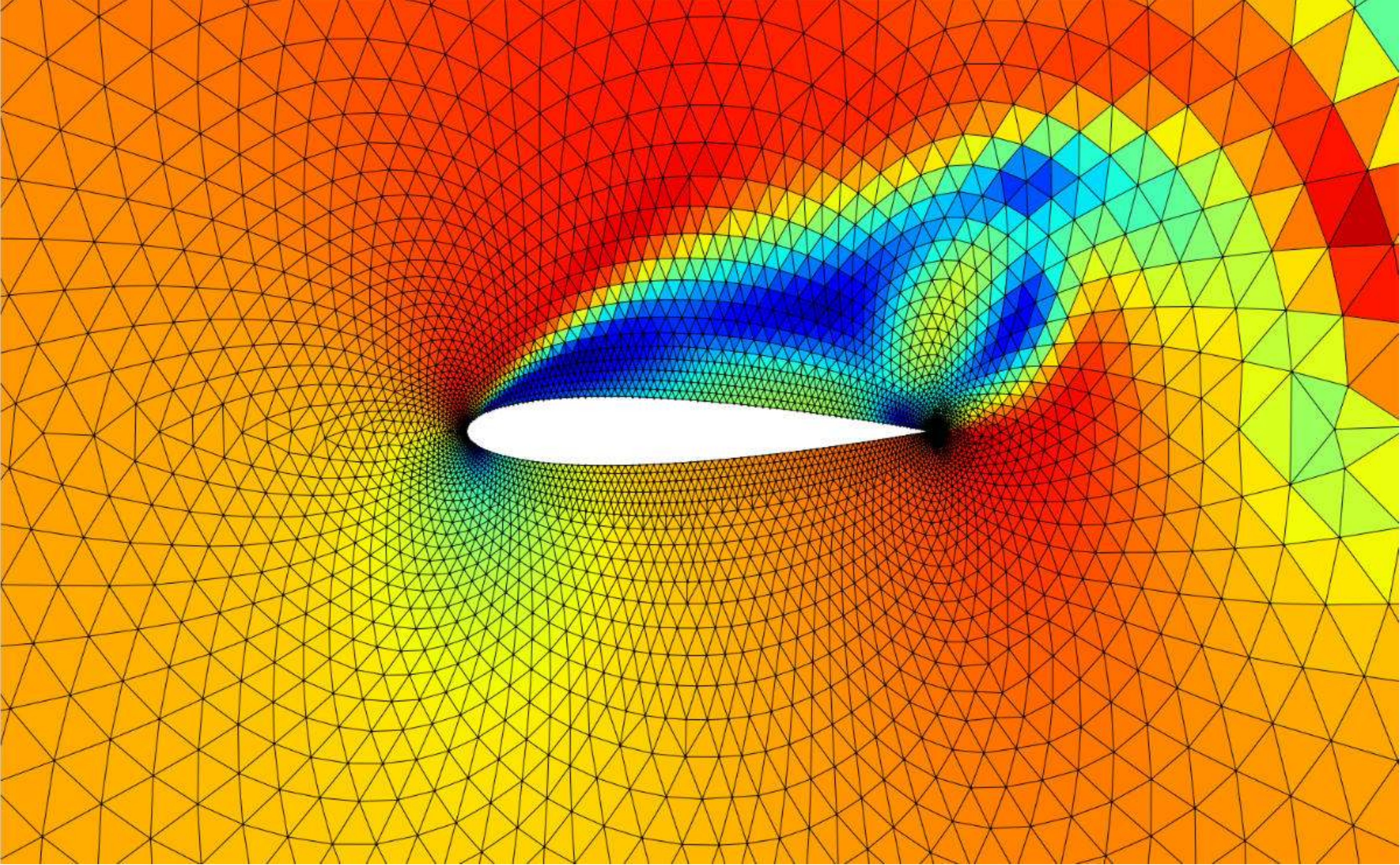} \\    
        \;+ SDRF & \includegraphics[width=0.28\textwidth]{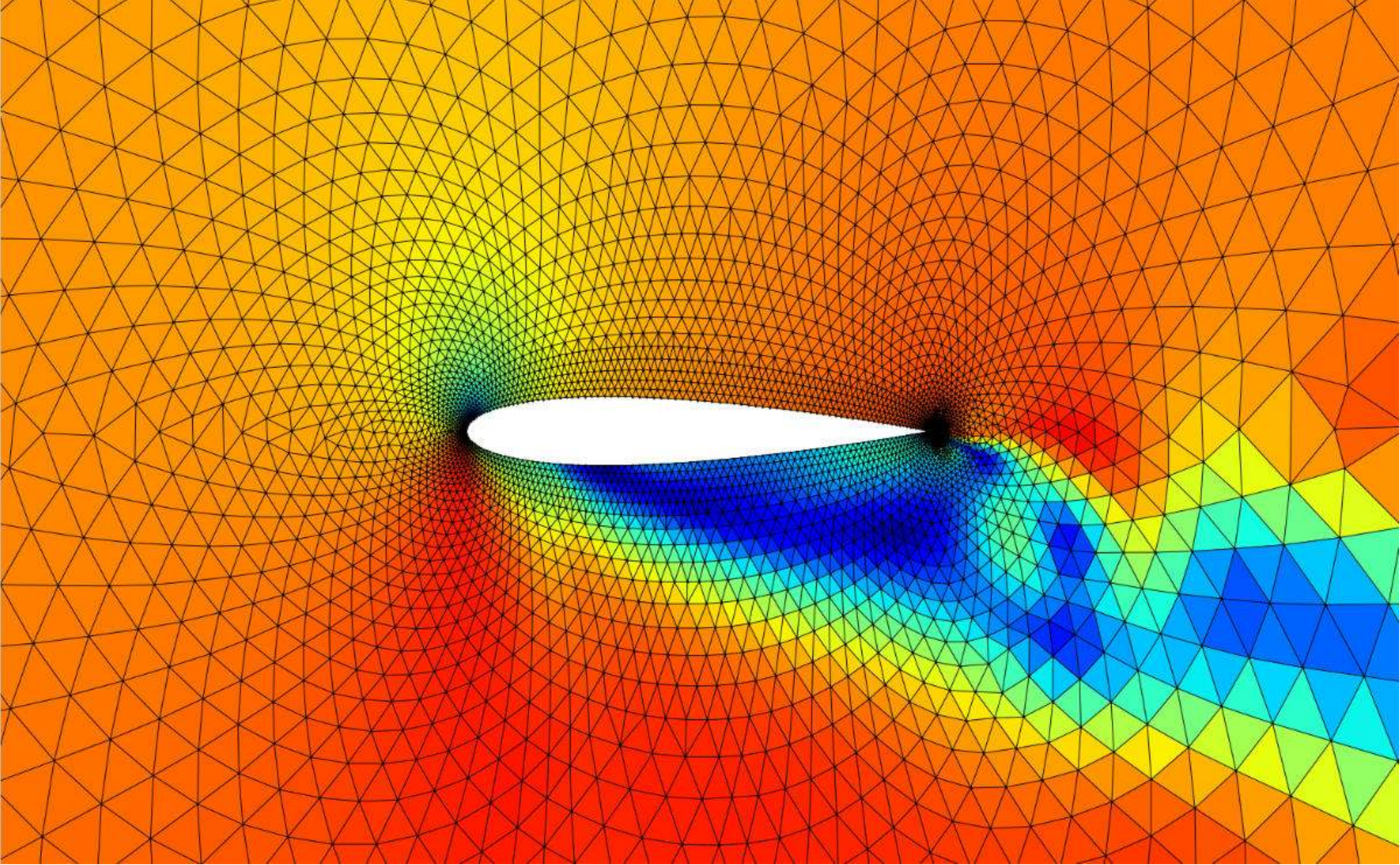} & \includegraphics[width=0.28\textwidth]{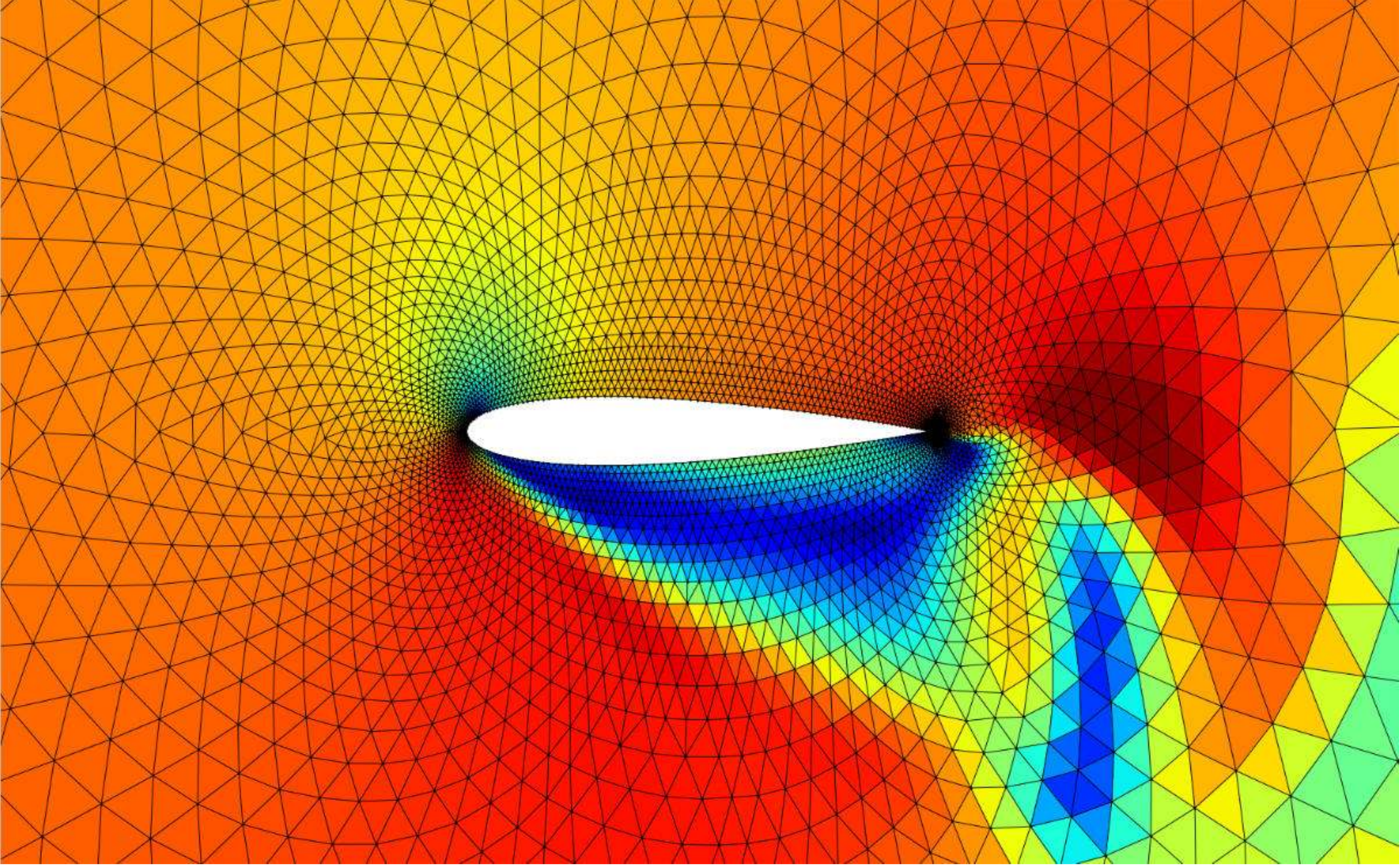} & \includegraphics[width=0.28\textwidth]{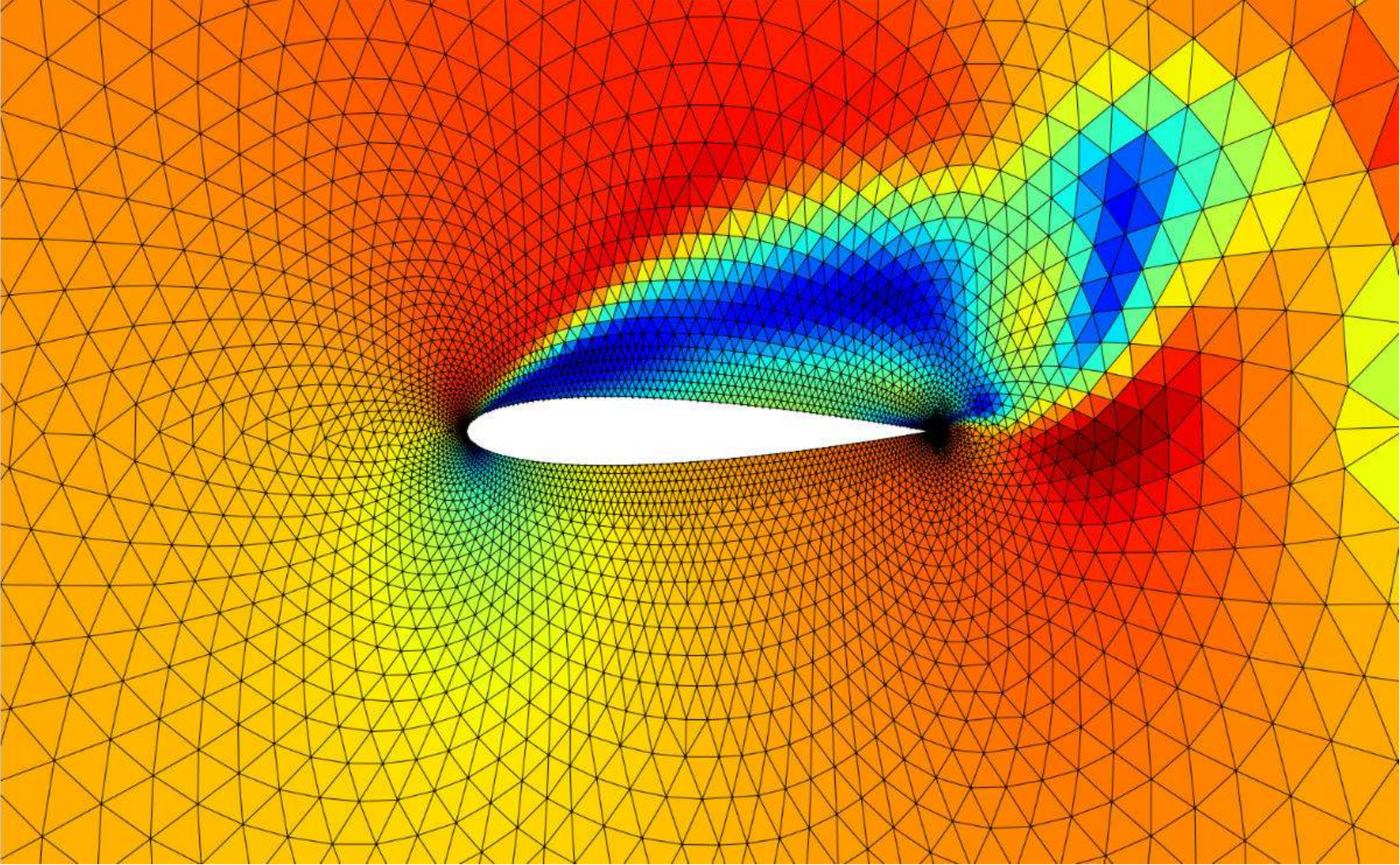}  \\   
        \;+ BORF & \includegraphics[width=0.28\textwidth]{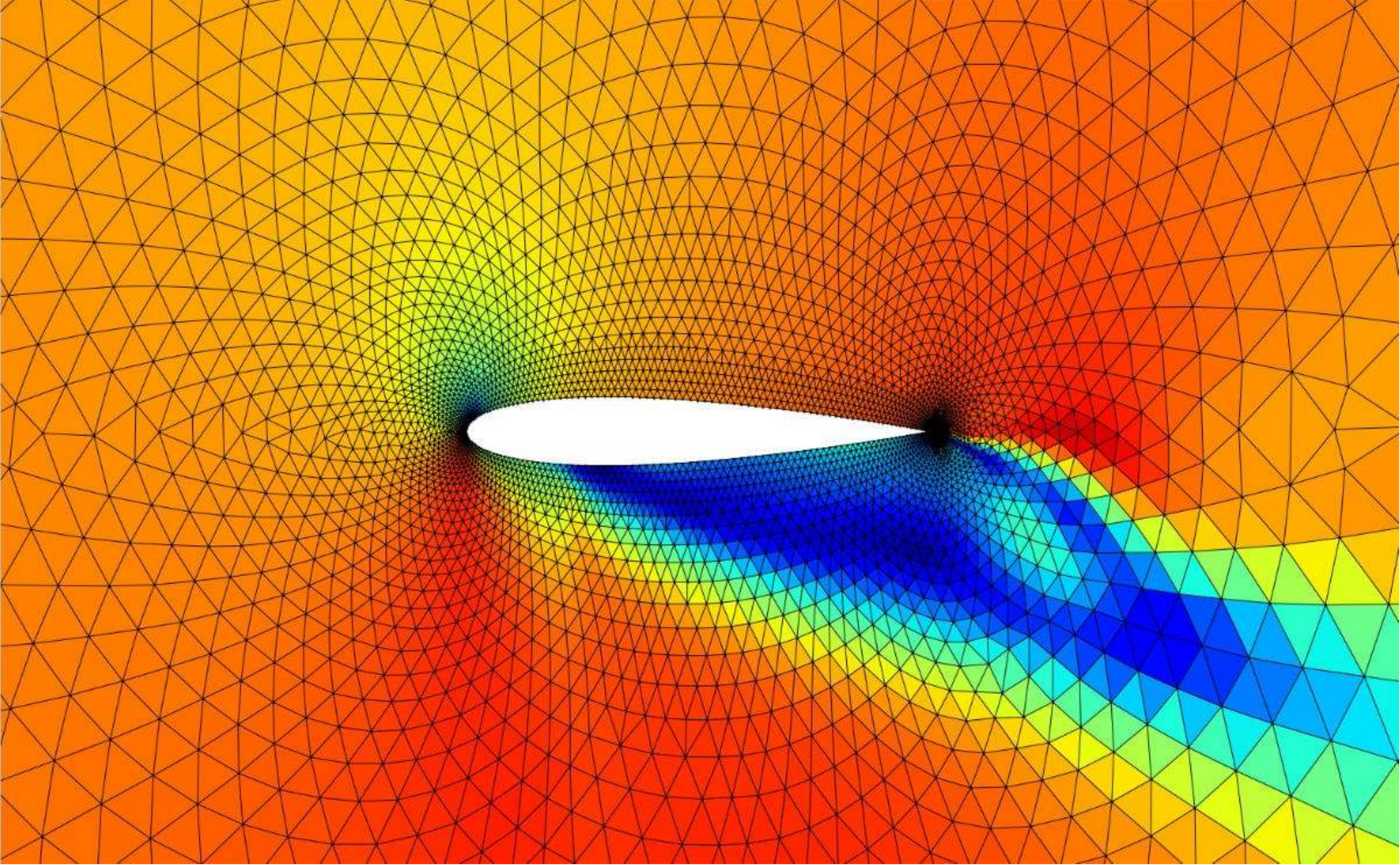} & \includegraphics[width=0.28\textwidth]{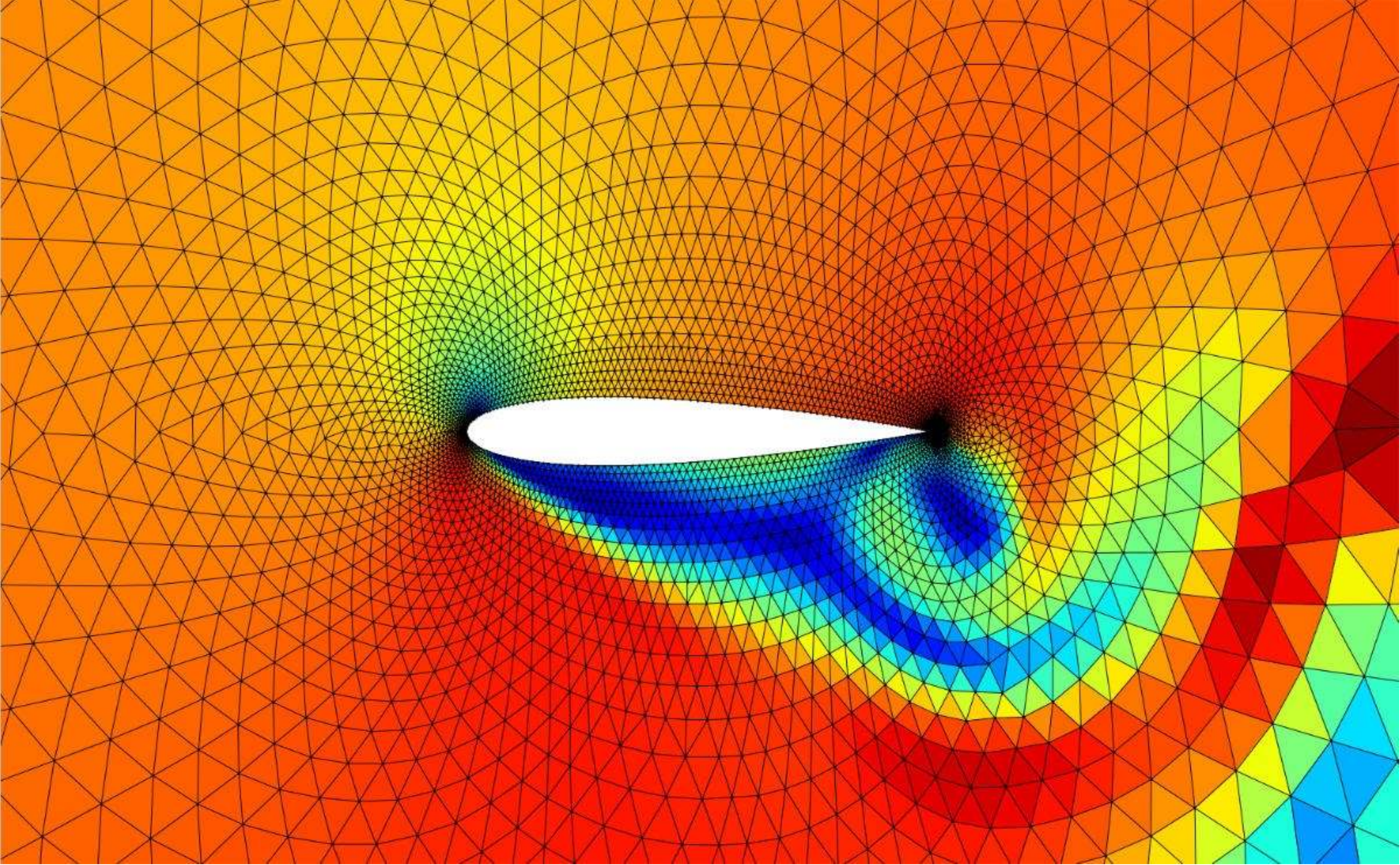} & \includegraphics[width=0.28\textwidth]{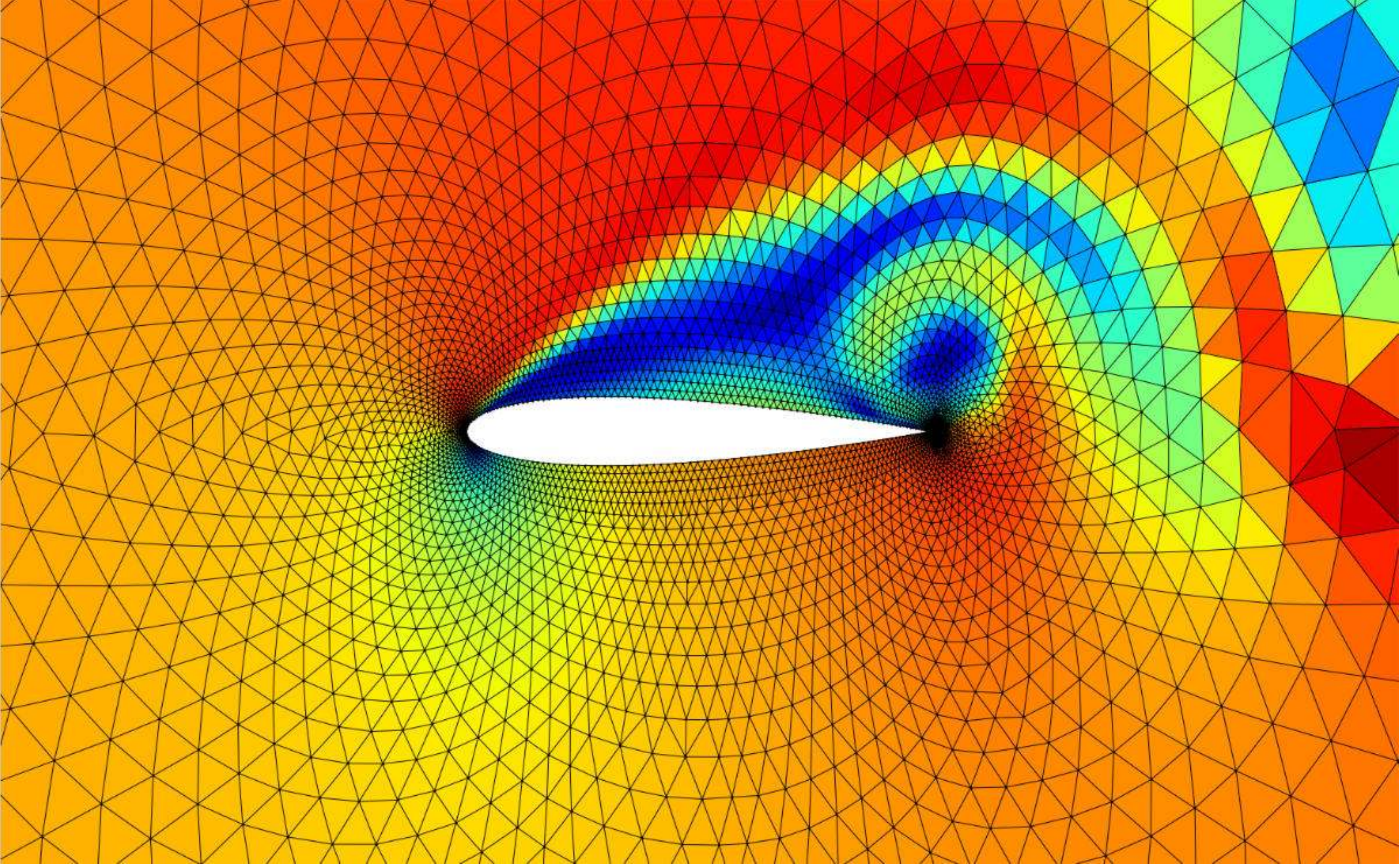} \\       
        \;+ PIORF & \includegraphics[width=0.28\textwidth]{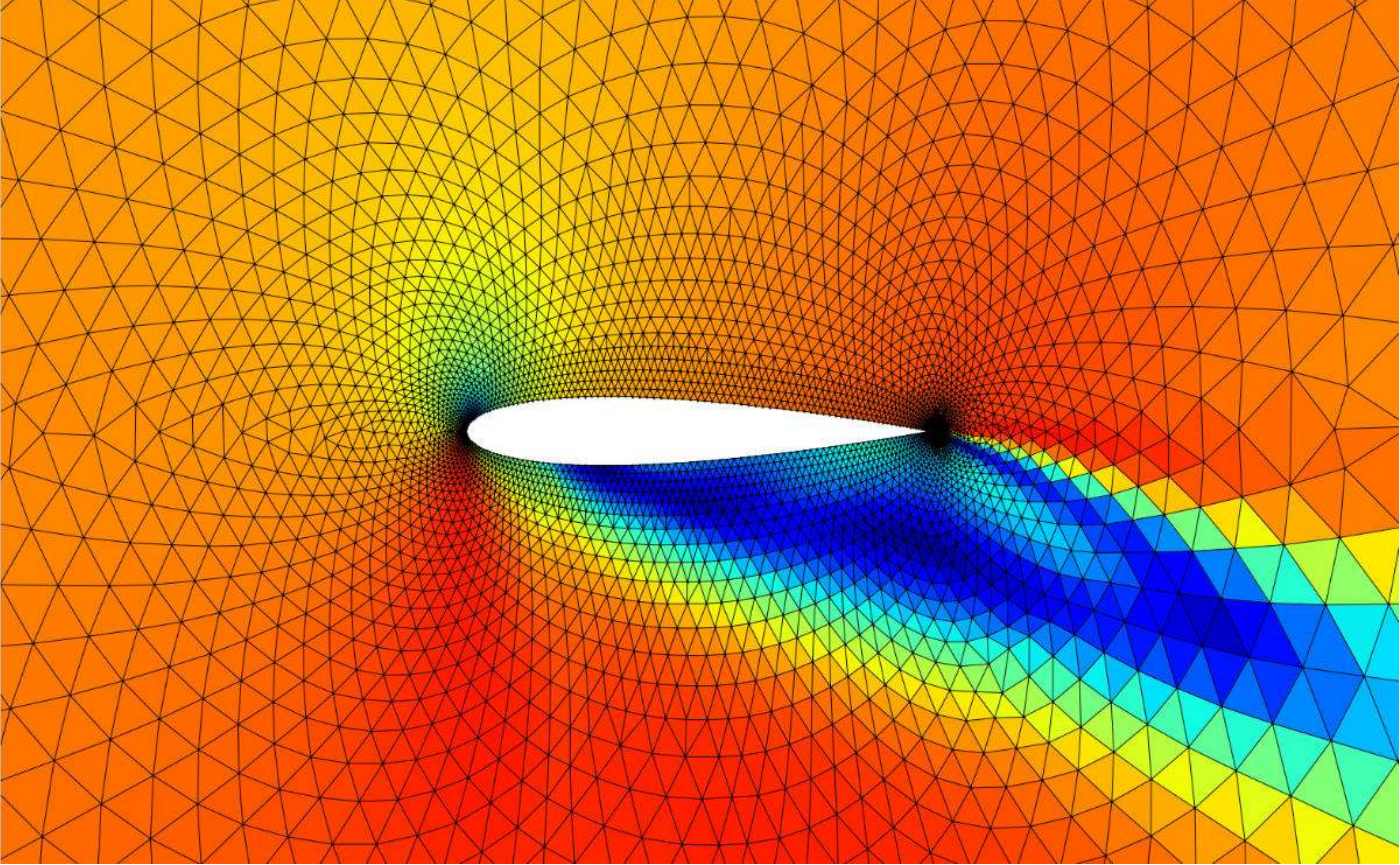} & \includegraphics[width=0.28\textwidth]{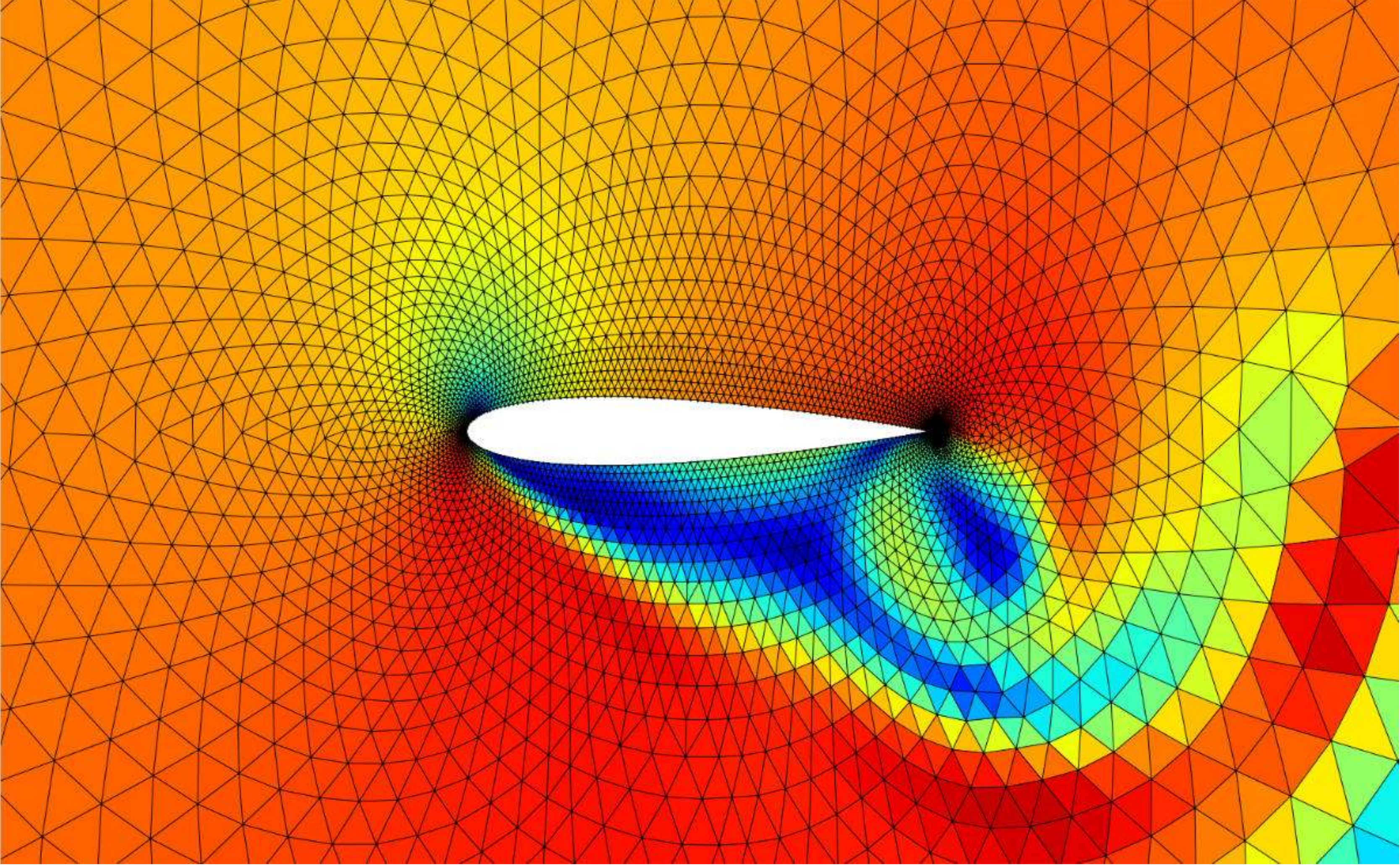} & \includegraphics[width=0.28\textwidth]{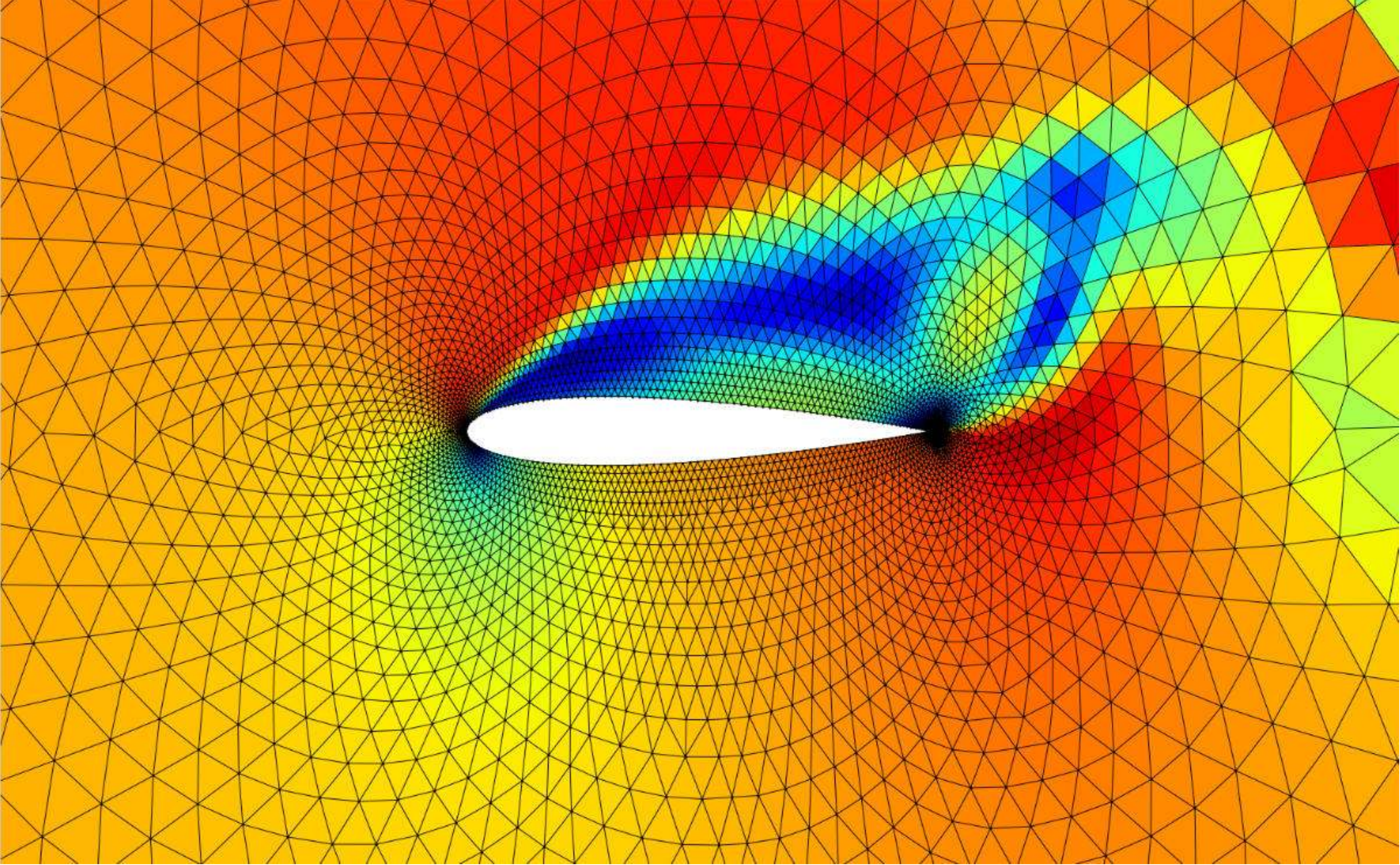} \\      
    \bottomrule
    \end{tabular}
    \includegraphics[width=0.5\textwidth]{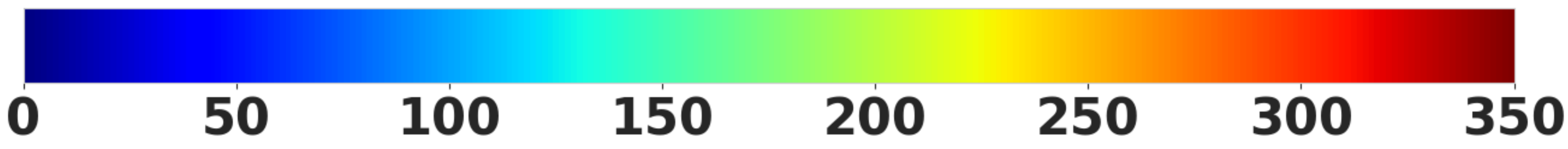}
    \caption{The velocity magnitude contours of various rewiring methods compared to the ground truth at \textsc{AirFoil}}
    \label{fig:roll_airfoil}
\end{figure}

\begin{figure}[h]
    \centering
    \setlength{\tabcolsep}{1pt}
    \begin{tabular} {l ccc}\toprule
    Models & Traj. 1 & Traj. 2 & Traj. 3 \\ \midrule
    Ground Truth & \includegraphics[width=0.28\textwidth]{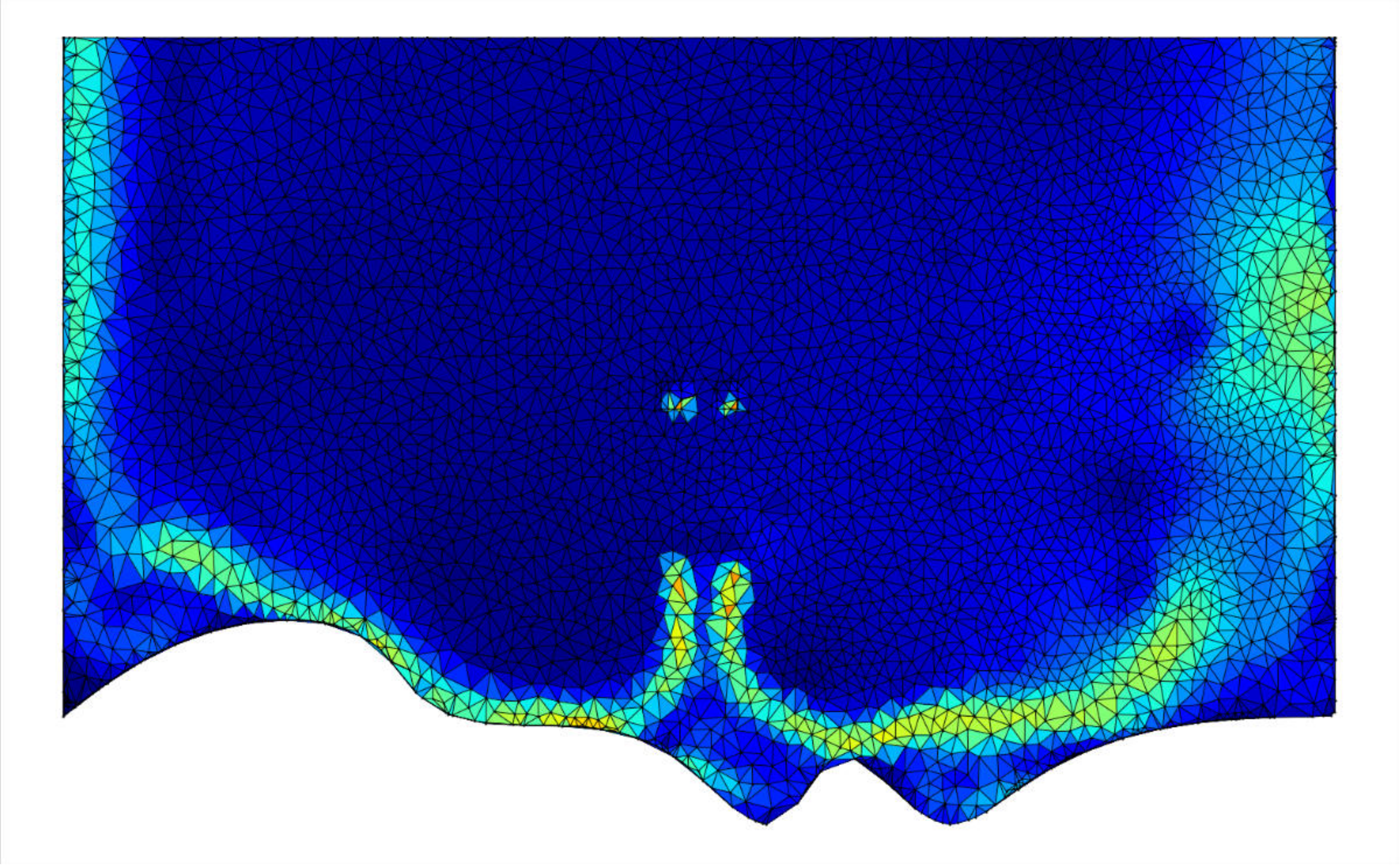} & \includegraphics[width=0.28\textwidth]{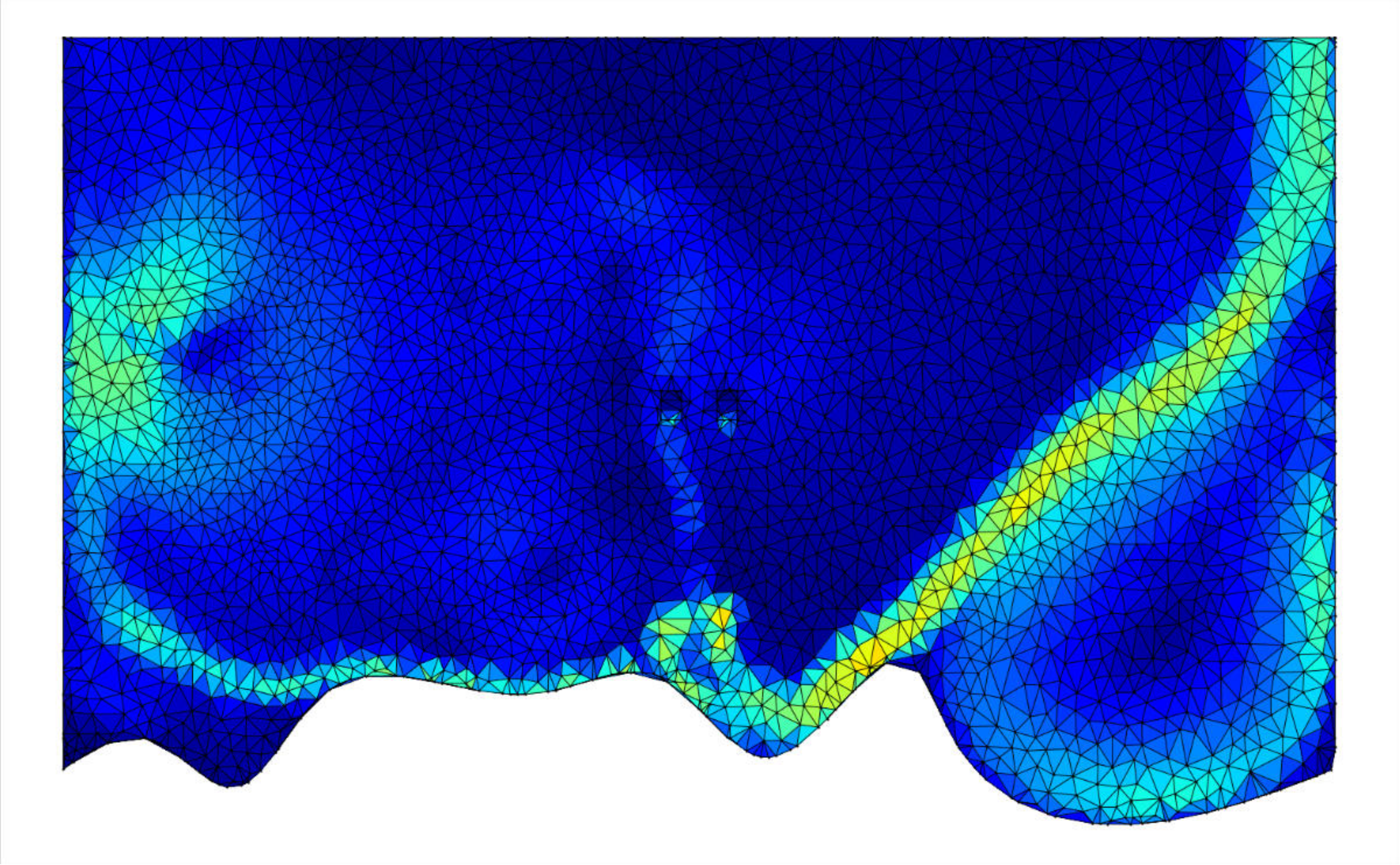} & \includegraphics[width=0.28\textwidth]{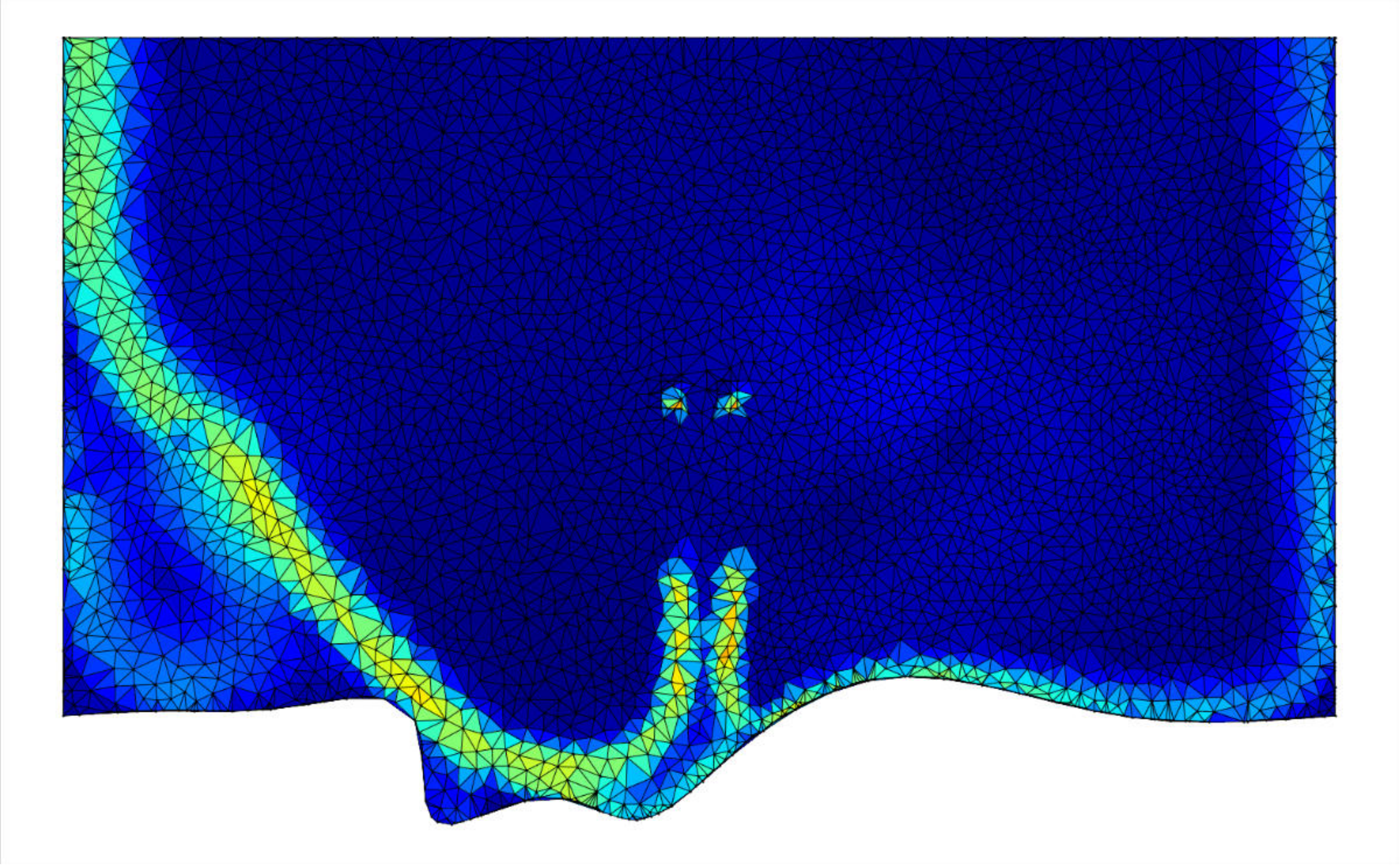}   \\
    MGN & \includegraphics[width=0.28\textwidth]{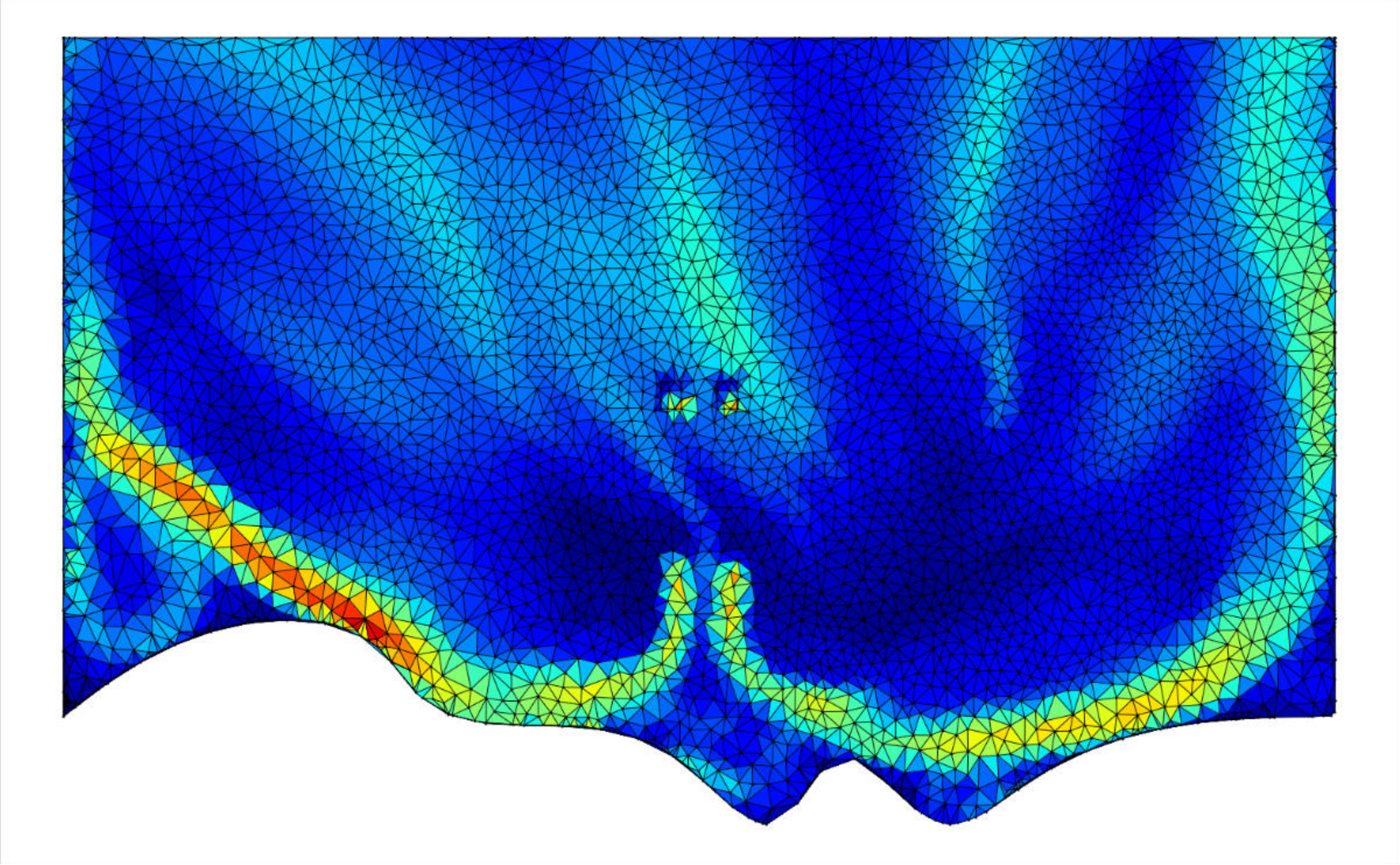} & \includegraphics[width=0.28\textwidth]{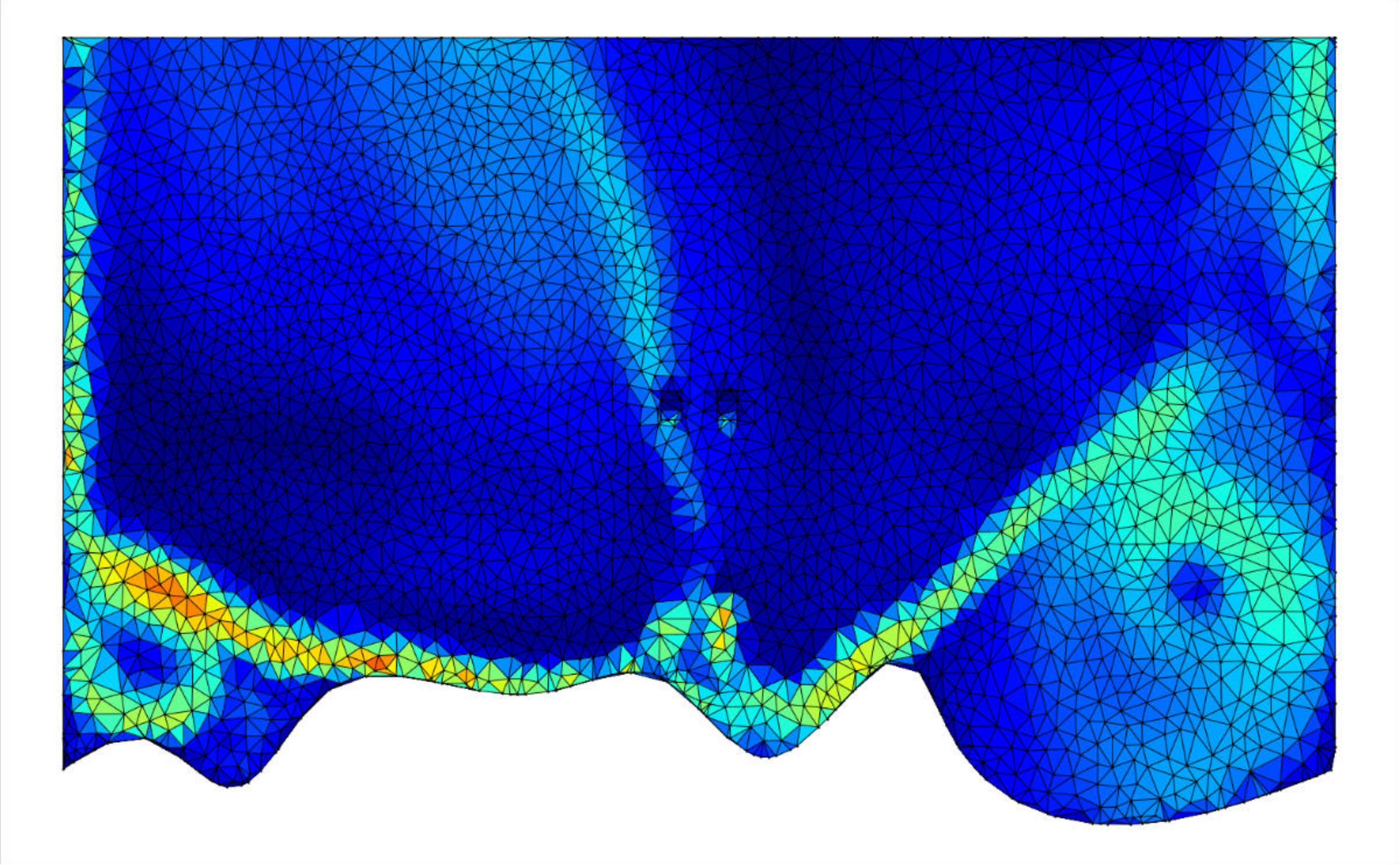} & \includegraphics[width=0.28\textwidth]{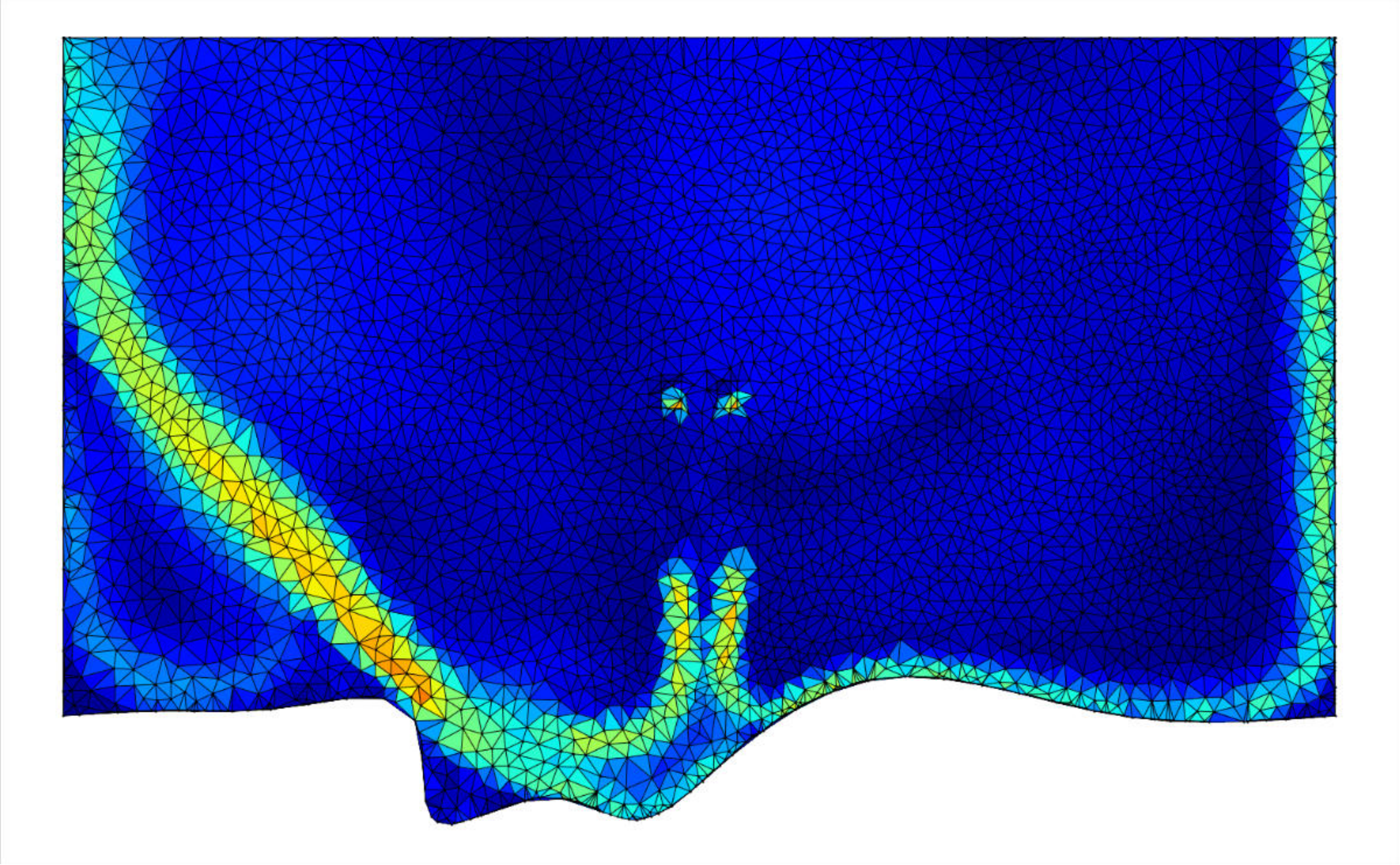} \\
    \;+ DIGL & \includegraphics[width=0.28\textwidth]{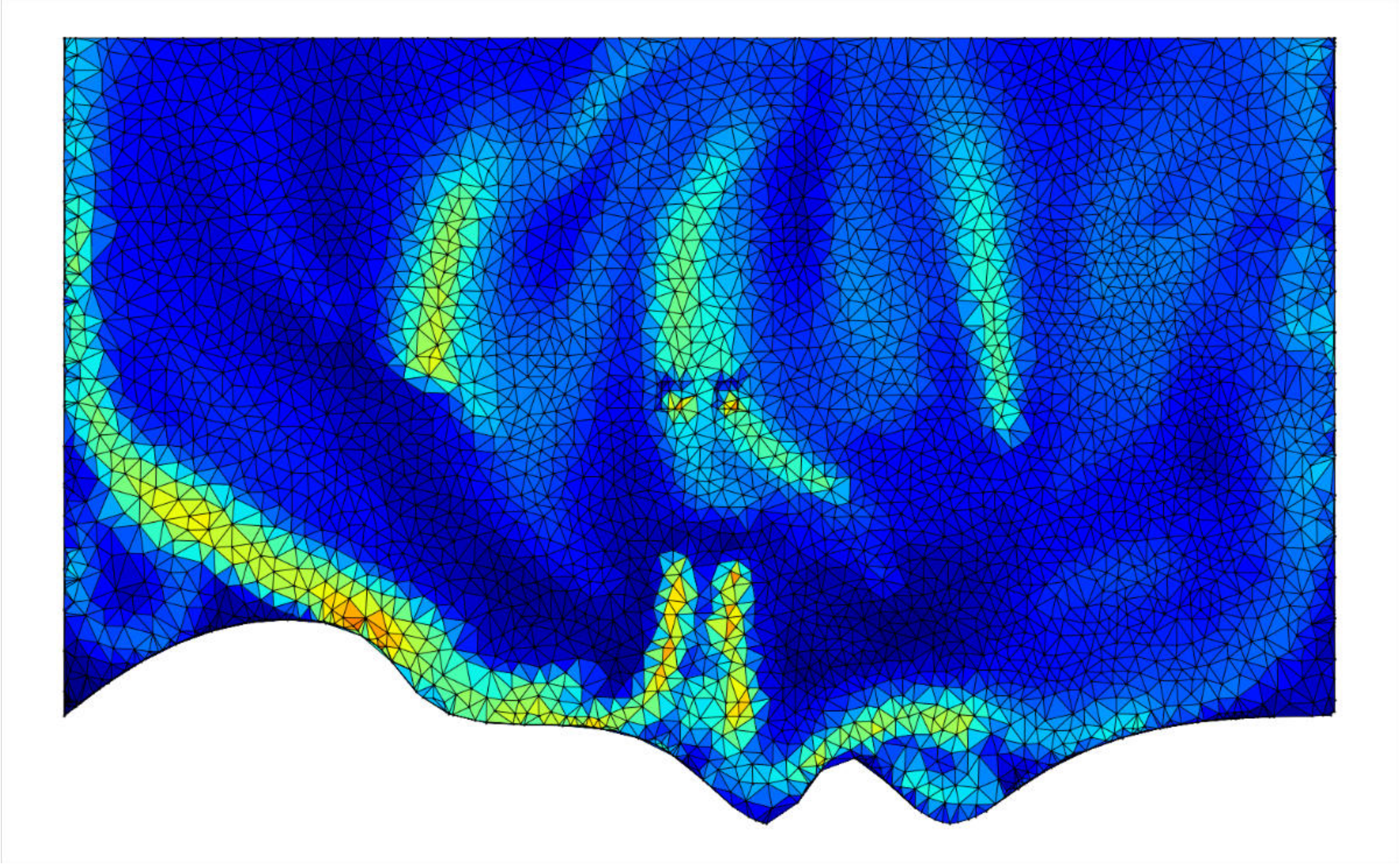} & \includegraphics[width=0.28\textwidth]{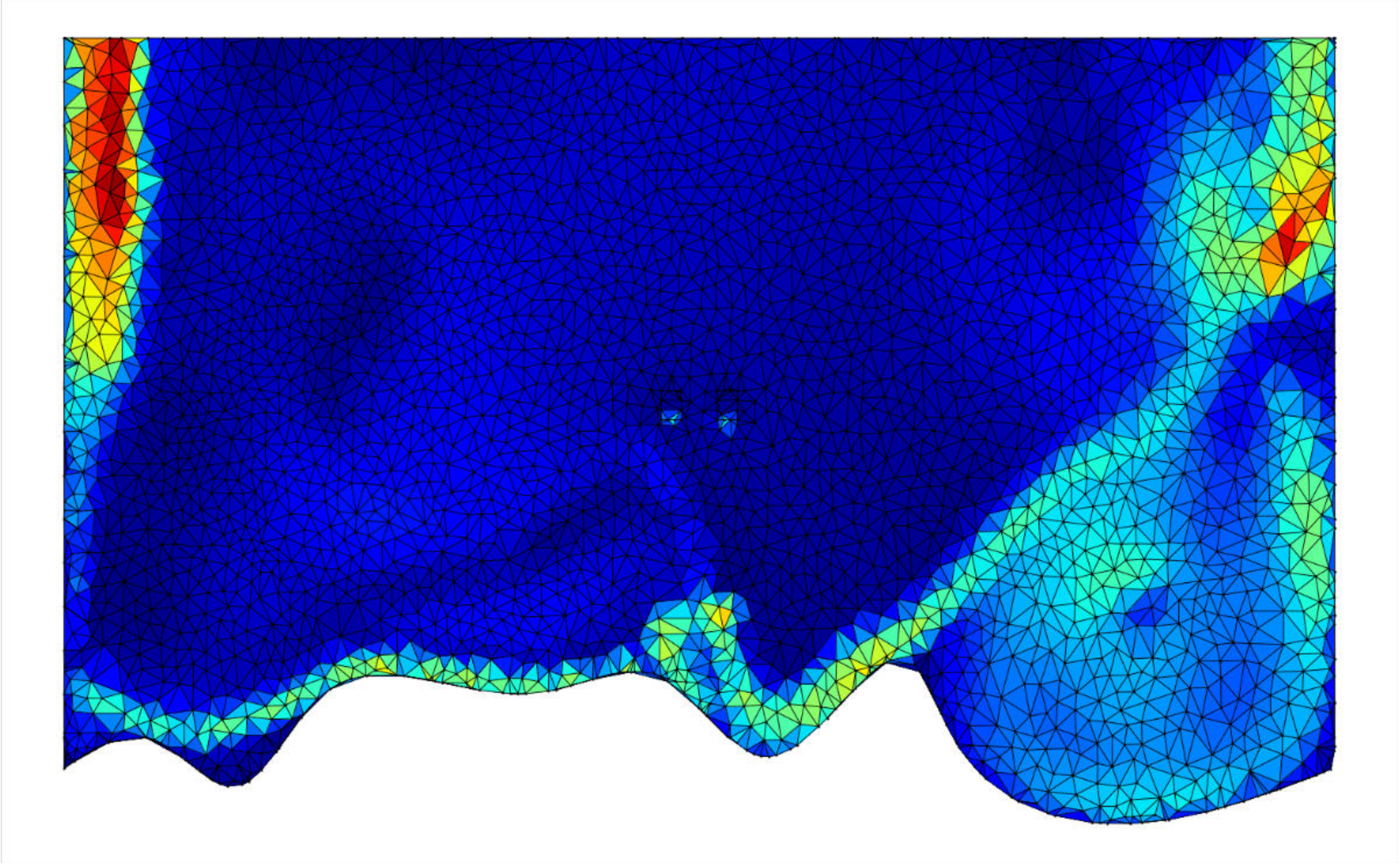} & \includegraphics[width=0.28\textwidth]{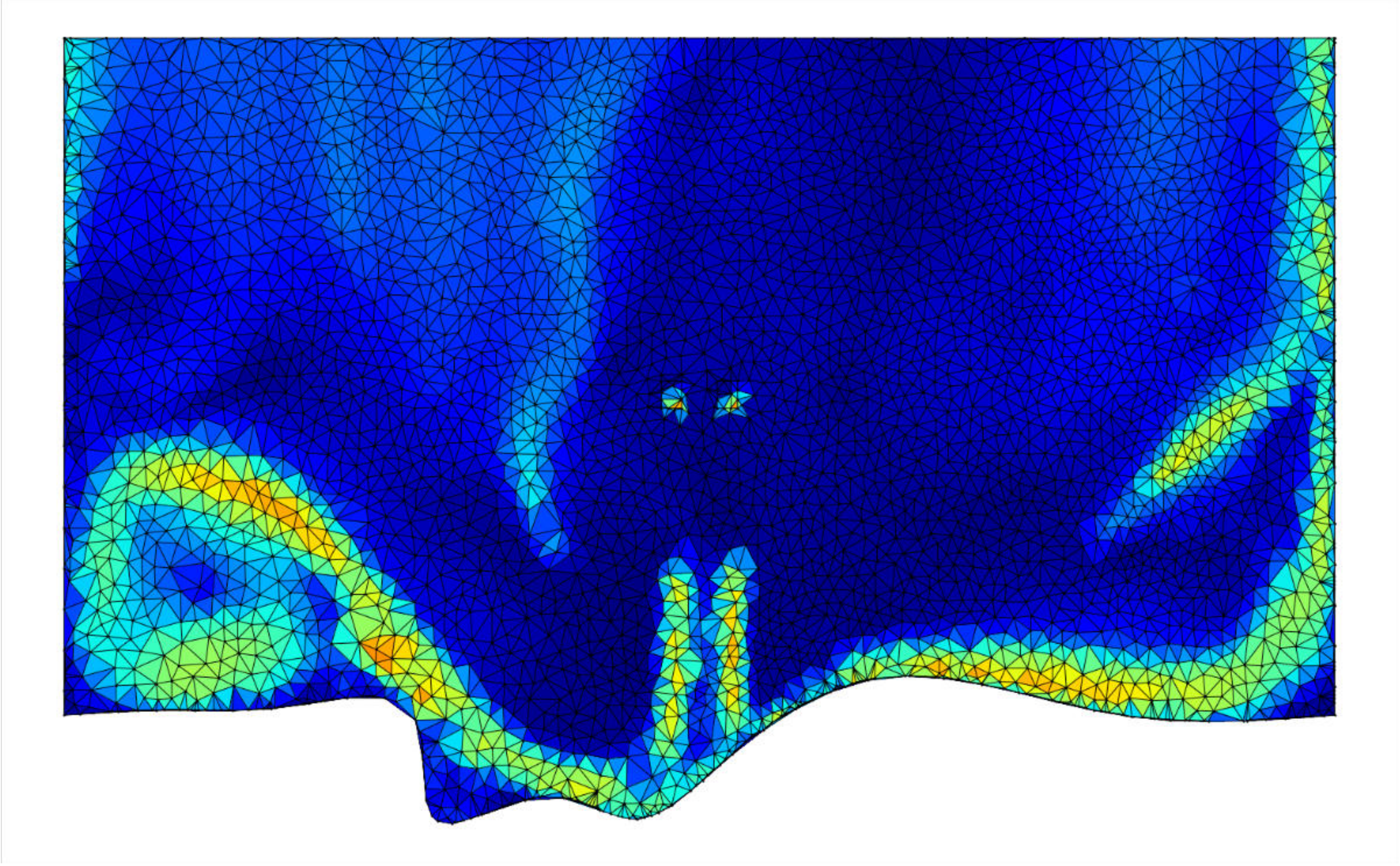}  \\    
    \;+ SDRF & \includegraphics[width=0.28\textwidth]{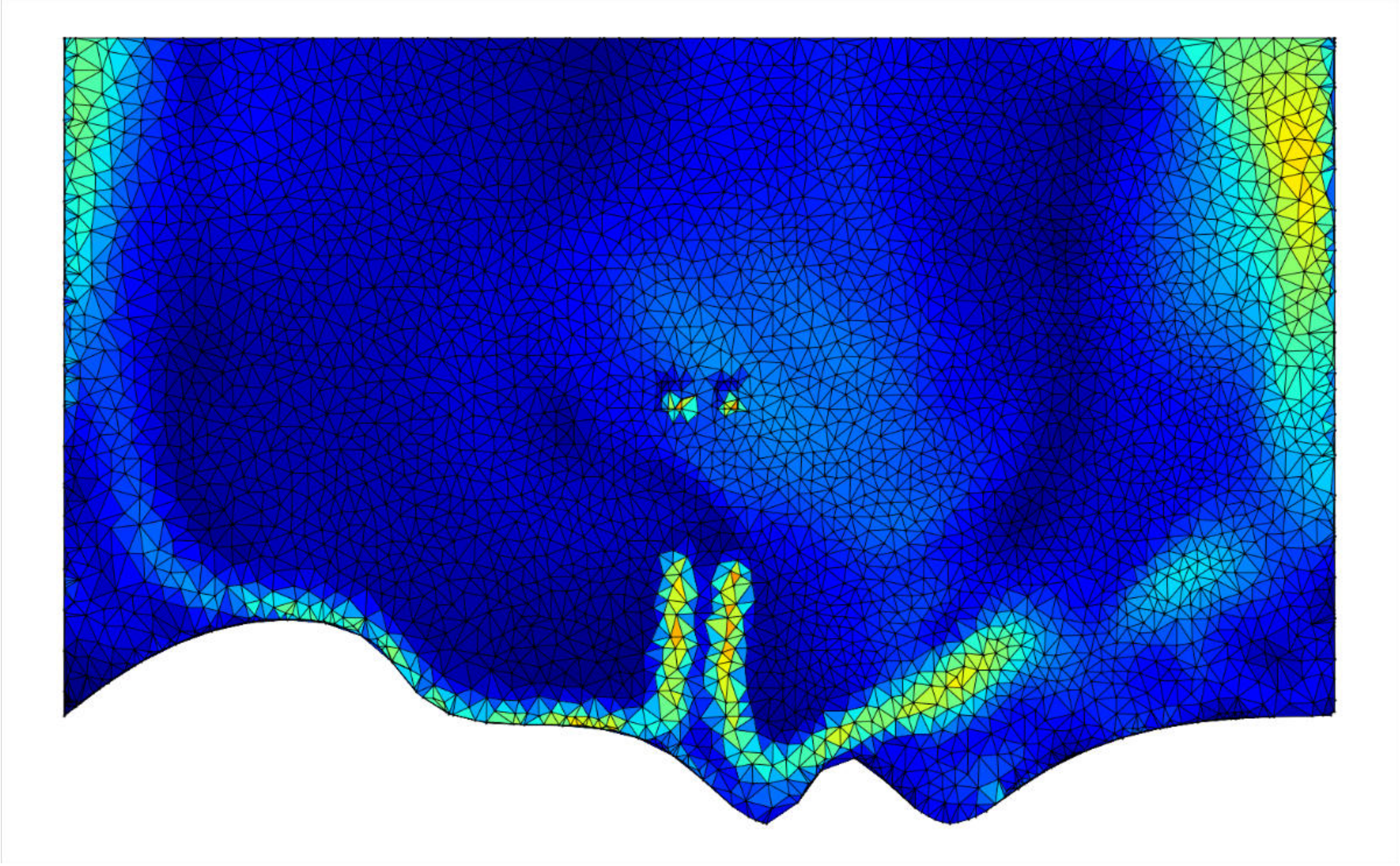} & \includegraphics[width=0.28\textwidth]{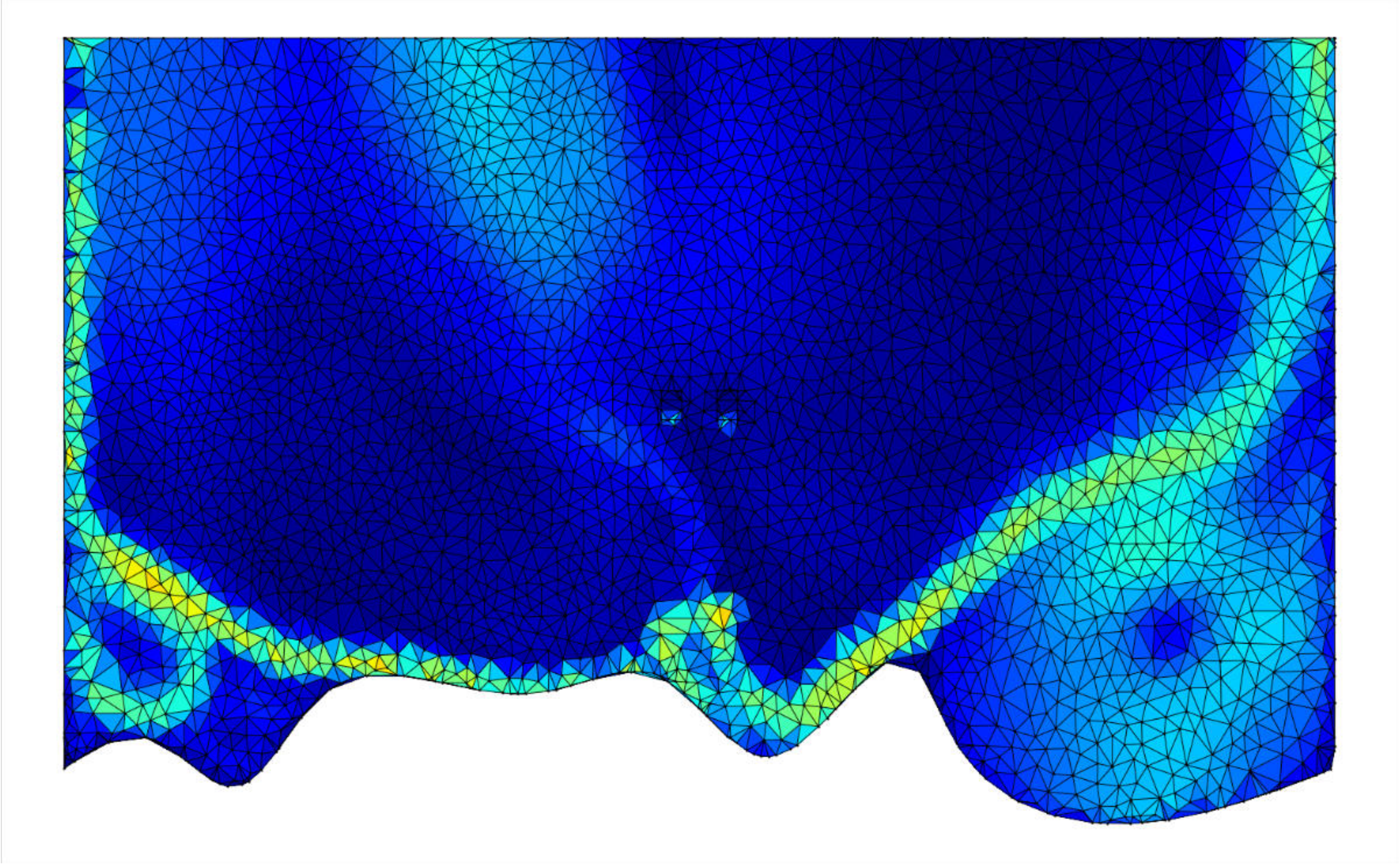} & \includegraphics[width=0.28\textwidth]{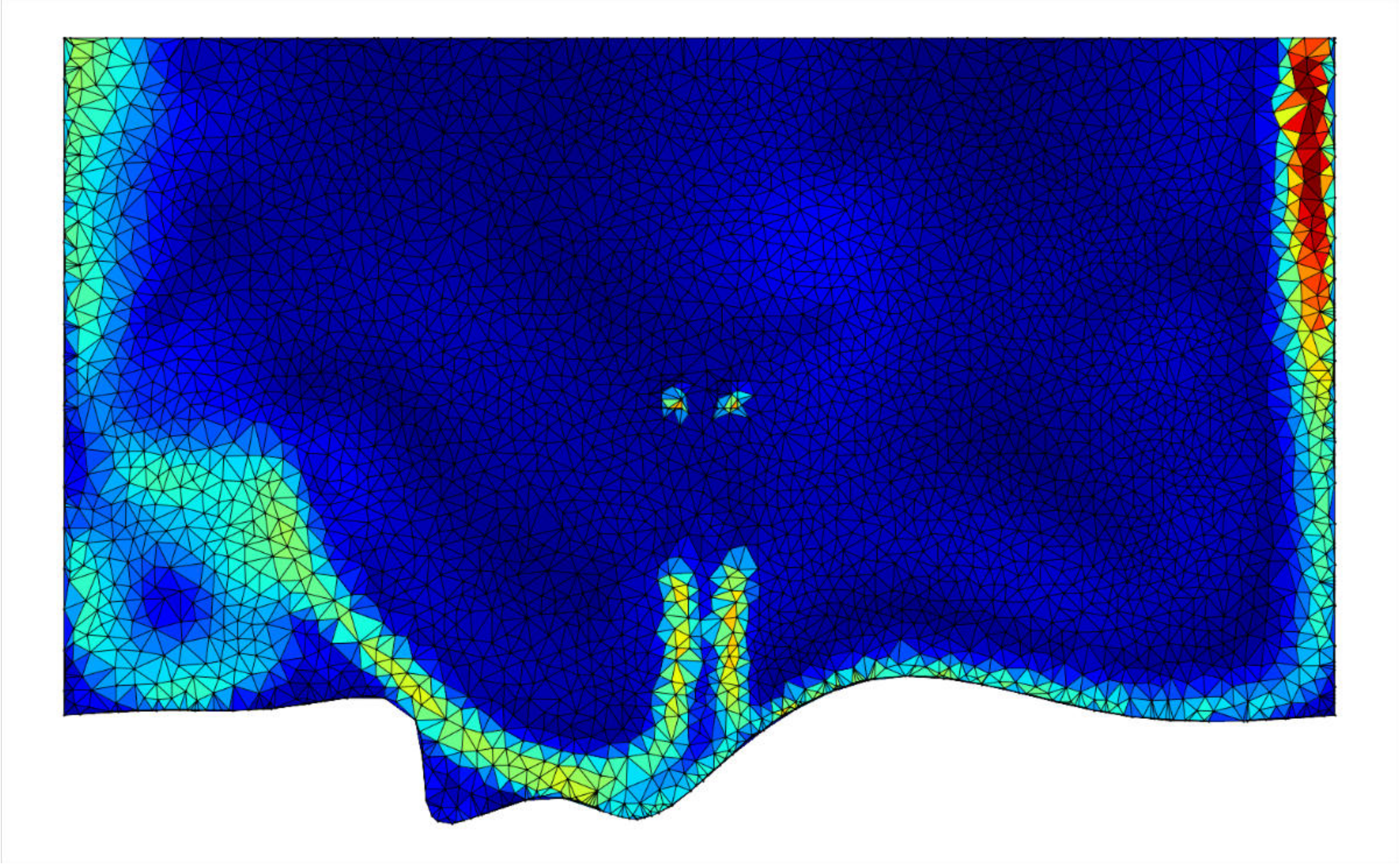}  \\    
    \;+ FoSR & \includegraphics[width=0.28\textwidth]{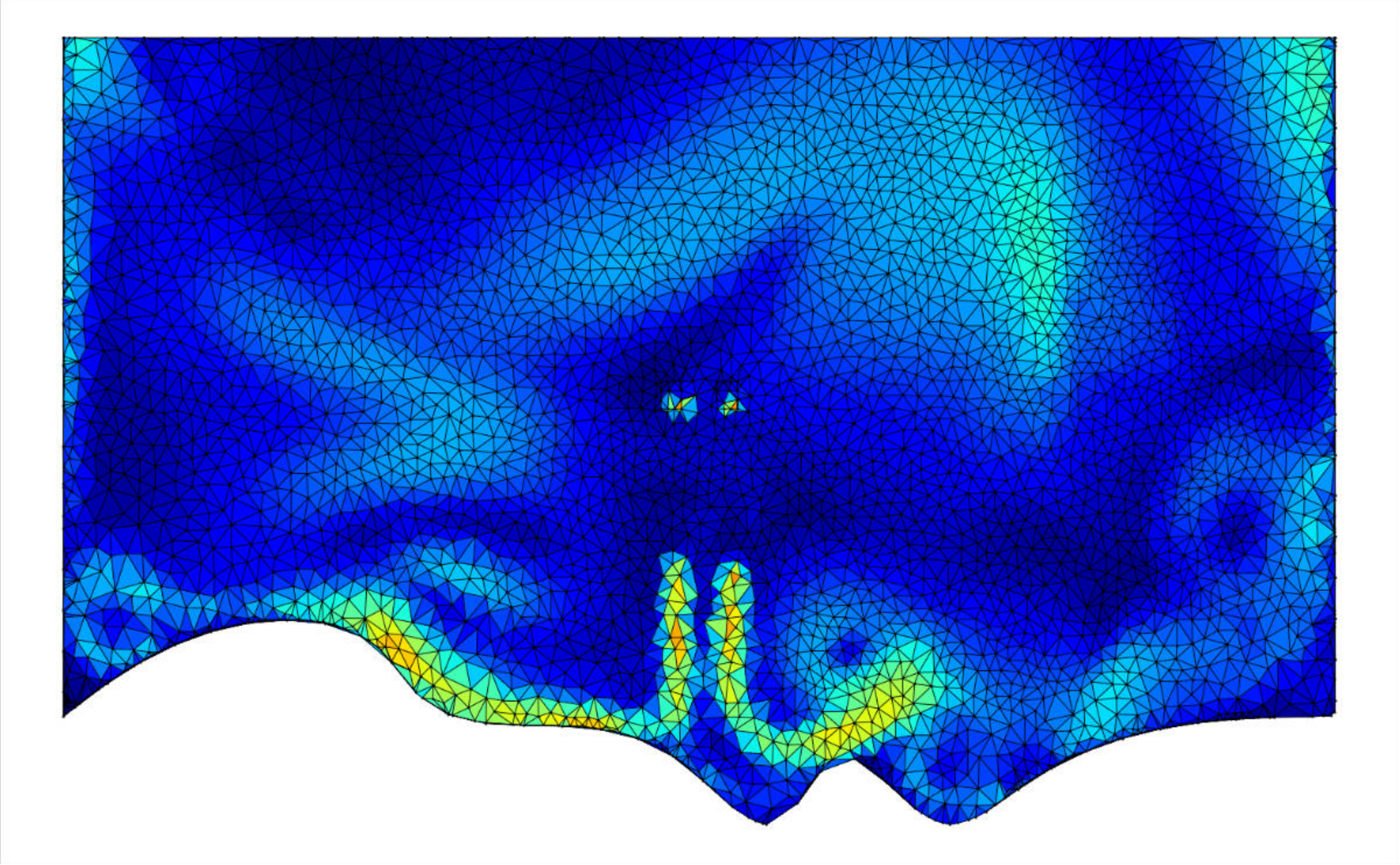} & \includegraphics[width=0.28\textwidth]{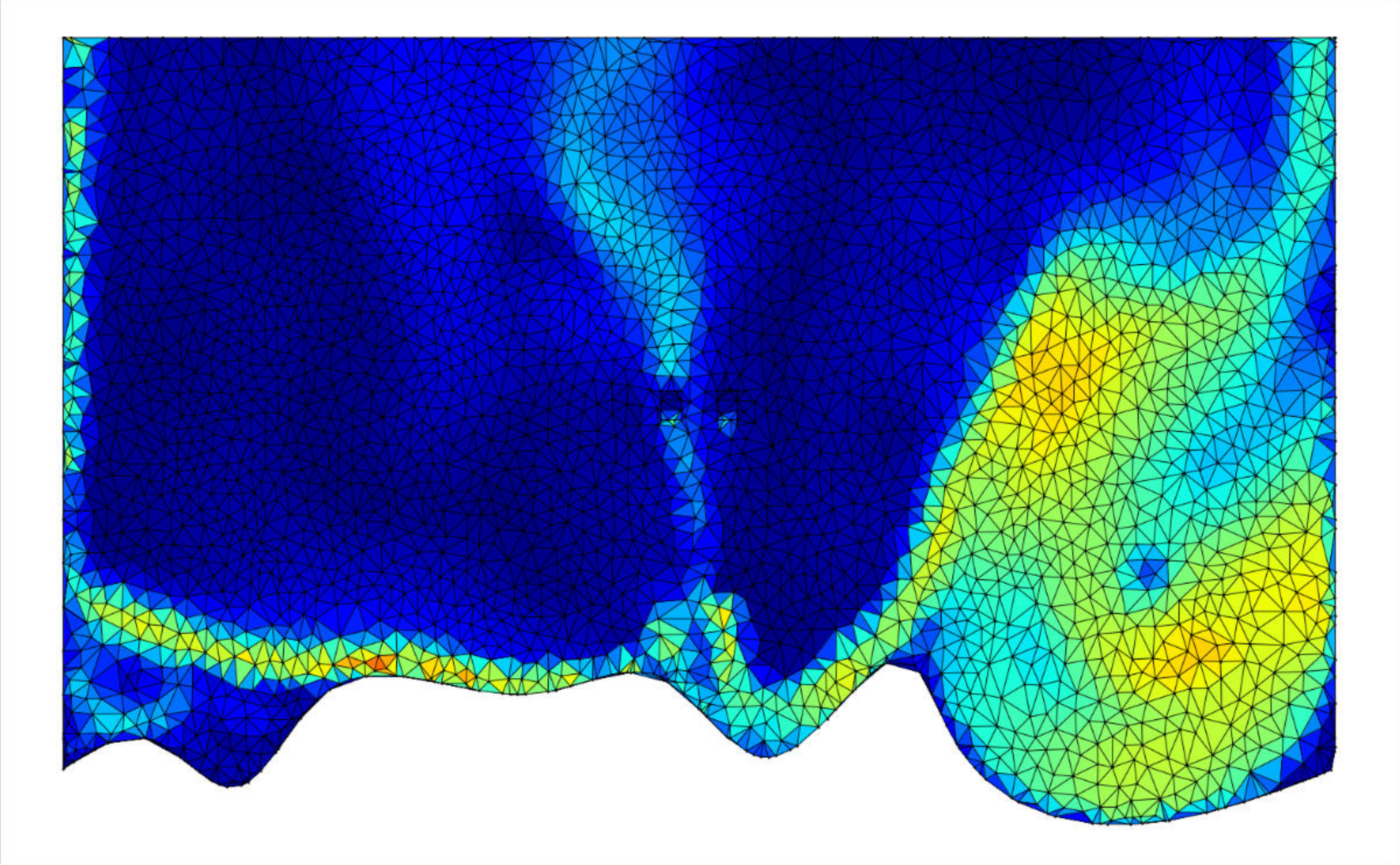} & \includegraphics[width=0.28\textwidth]{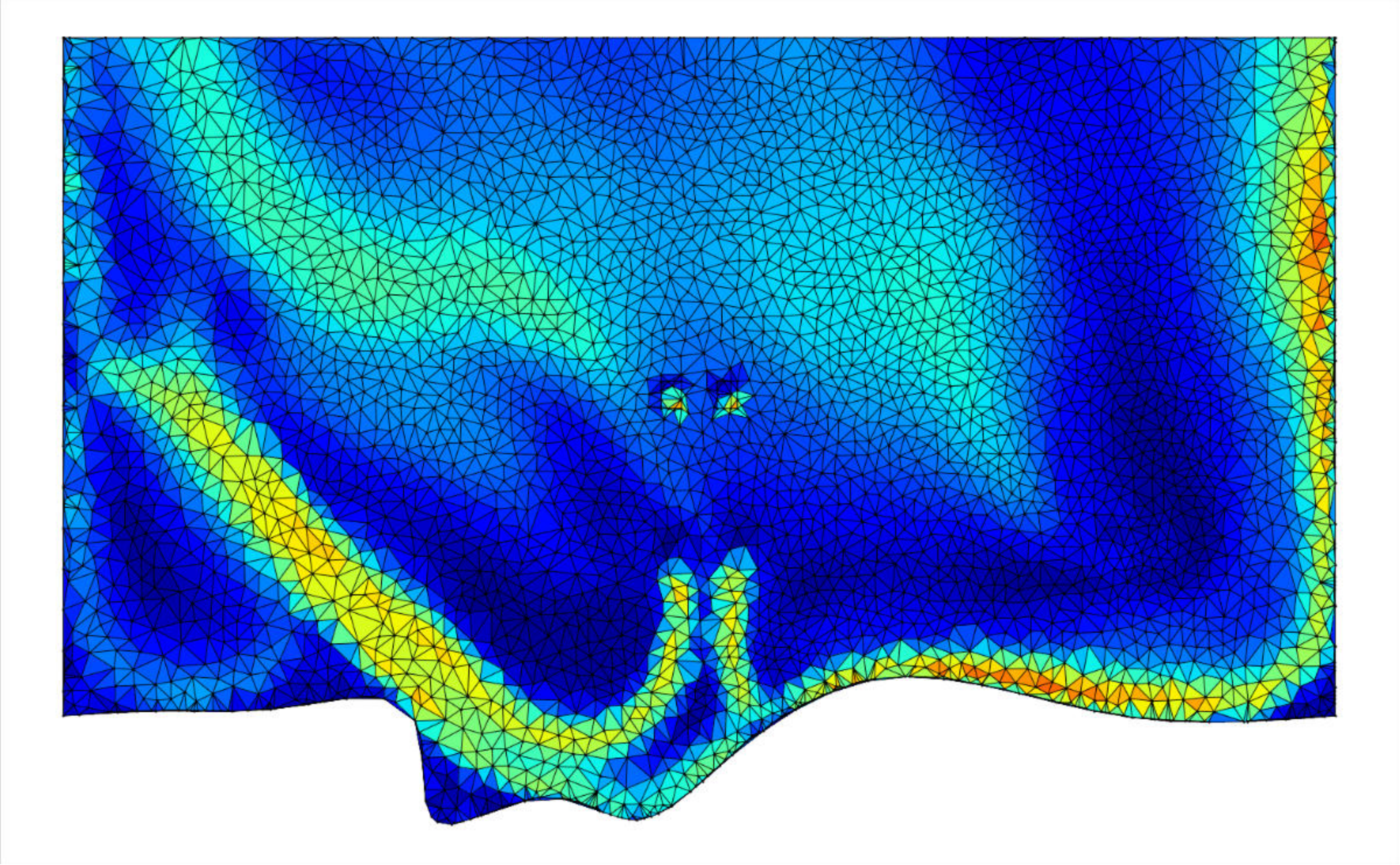}  \\    
    \;+ PIORF & \includegraphics[width=0.28\textwidth]{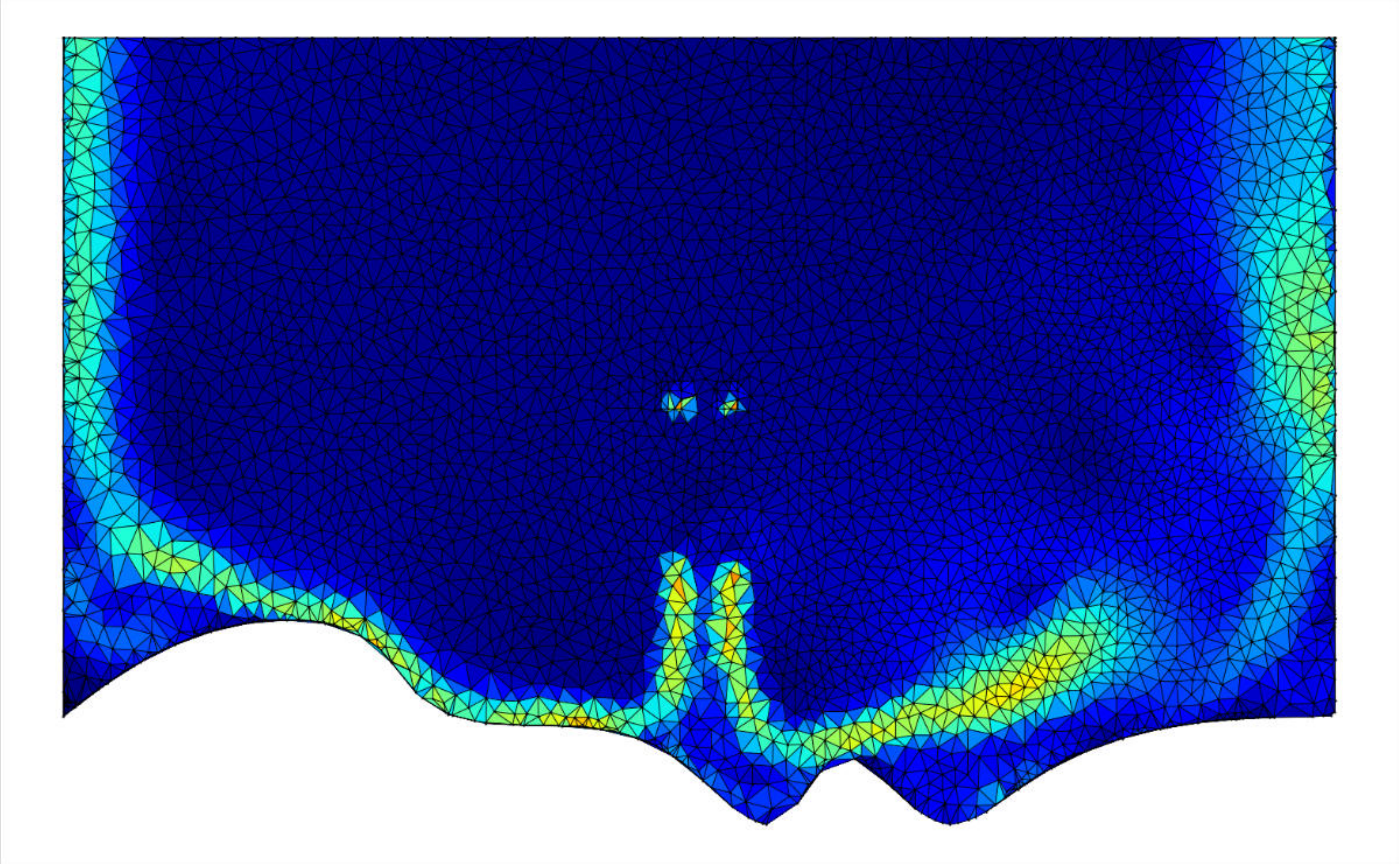} & \includegraphics[width=0.28\textwidth]{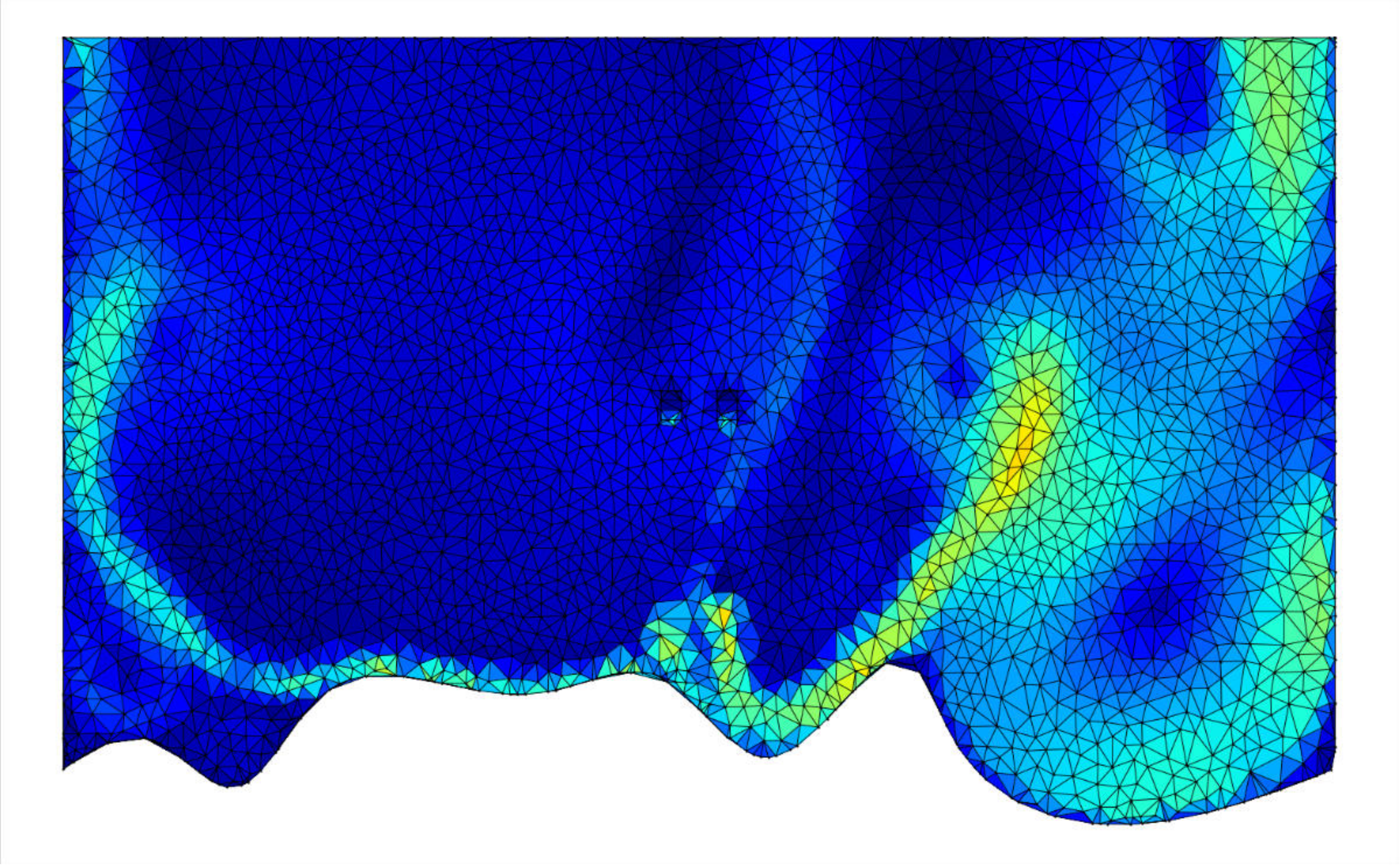} & \includegraphics[width=0.28\textwidth]{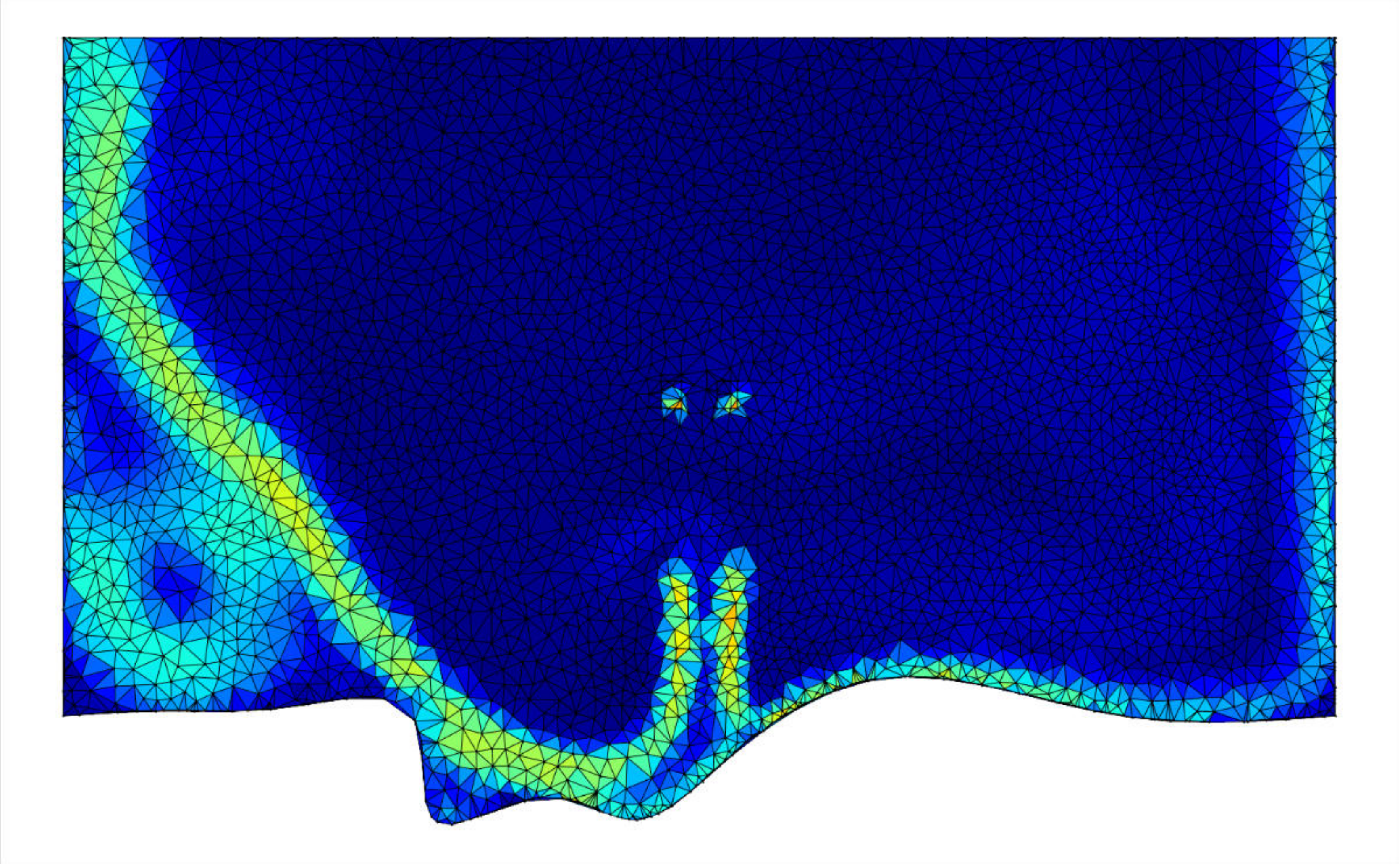}  \\      
    \bottomrule
    \end{tabular}
    \includegraphics[width=0.5\textwidth]{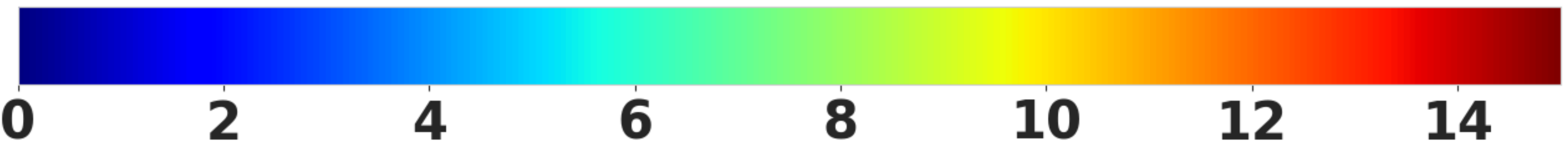}
    \caption{The velocity magnitude contours of various rewiring methods compared to the ground truth at EAGLE.}
    \label{fig:roll_eagle}
\end{figure}

\Cref{fig:roll_cylinder,fig:roll_airfoil,fig:roll_eagle} are rollout images of \textsc{CylinderFlow}, \textsc{AirFoil}, and \textsc{EAGLE}, from the last time step with the highest cumulative error.

\clearpage
\section{Notations} \label{app:notations}
\Cref{tab:notation} outlines the key notations used in the paper.

\begin{table}[h]
    \small
    \centering
    \caption{Notation summary.}
    \label{tab:notation}
    \begin{tabular}{c c}\toprule
        Name              & index     \\ \midrule
        senders                & $i$  \\ 
        receivers              & $j$  \\ 
         \midrule
        ORC(edge)                & $\kappa(i,j)$  \\
        ORC(node)                & $\gamma_i$  \\
        nodes                    & $u,v$  \\
        the shortest distance    & $d(u,v)$  \\
        distribution of 1-step random walk from node $u$ & $\mathbf{m}_u$ \\
        L1 Wassertein transport distance  & $W_1(\mu_i,\mu_j)$  \\
        sets of nodes            & $\mathcal{V}$  \\
        sets of edges            & $\mathcal{E}$  \\
        \midrule
        
        ORC Pooling Ratio & $\delta$  \\ 
        \midrule
        
        Inputs(edge features)    & $\mathbf{m}_{ij}$  \\
        Inputs(node features)    & $\mathbf{v}_i$  \\
        Outputs                  & $\mathbf{o}_i$  \\ 
        Edge Hidden Features            & $\mathbf{e}_{ij}$  \\
        Updated Edge Hidden Features    & $\mathbf{e}'_{ij}$  \\
        Node Hidden Features            & $\mathbf{v}_{ij}$  \\ 
        Updated Node Hidden Features    & $\mathbf{v}'_{ij}$  \\ 
        Node MLP                        & $\mathbf{f}^V$  \\ 
        Edge MLP                        & $\mathbf{f}^E$  \\ 
        Number of nodes                 & $|\mathcal{V}|$  \\ 
        Number of edges           & $|\mathcal{E}|$  \\ 
        \midrule
        
        Mesh Positions           & $\mathbf{x}_{i}$ \\
        Relative Mesh Positions  & $\mathbf{x}_{ij}$ \\
        Norm Relative Mesh Positions  & $|\mathbf{x}_{ij}|$ \\
        Node Type                & $\mathbf{n}_{i}$ \\
        Velocity                 & $\mathbf{w}_{i}$ \\
        Velocity Gradient        & $\dot{\mathbf{w}}_i$ \\
        Predicted Velocity Gradient & $\hat{\dot{\mathbf{w}_{i}}}$ \\
        Pressure                 & $p_i$ \\
        Predicted Pressure       & $\hat{p_i}$ \\
        Density                  & $\rho_i$ \\
        Density Gradient         & $\dot{\rho_i}$ \\
        Predicted Density Gradient         & $\hat{\dot{\rho_i}}$ \\
        \bottomrule
    \end{tabular}
\end{table}

\clearpage
\section{Additional Ablation Studies}

Our proposed rewiring method has node selection steps that depend on ingredients such as degree, ORC, and physical context. We conduct additional ablation studies to evaluate performance across different ingredient selections. The pooling ratio for all experiments is 3\%.

\Cref{tab:ablation2} shows performance based on ingredient selection. The first step is to select nodes based on curvature(``Former'', \Cref{alg:algorithm} lines 3-4), and the second is to select nodes based on physical context(``Latter'', \Cref{alg:algorithm} lines 5-6). We define the following four rewiring methods for ablation studies: i) ``Ablation 1'', where the former refers to high degree and the latter to physics, ii) ``Ablation 2'', where the former refers to random and the latter to physics, iii) ``Ablation 3'', where the former refers to random and the latter to random, and iv) ``Ablation 4'', where the former refers to ORC and the latter to random.

In CylinderFlow, ``Ablation 1'' shows results with high-degree selection, achieving improved performance compared to MGN. However, it underperforms relative to PIORF, as it exhibits varying curvature values for the same degree. ``Ablation 2'' and ``Ablation 4'' show performance based on the choice of physical context and ORC, respectively. Both outperform MGN, and Physical Context provides slightly better performance than ORC. ``Ablation 3'' is the result of randomly selecting both the former and the latter and adding edges, and is similar to the performance of MGN.

\begin{table}[h]
    \small
    % \scriptsize
    \footnotesize
    \setlength{\tabcolsep}{4pt}
    \centering
    \caption{Rollout-all RMSE ($\times10^{3}$) for PIORF and the ablations.}
    \begin{tabular}{l cc cc ccc}\toprule
        \multirow{2}{*}{Model} & \multicolumn{2}{c}{Ingredient} & \multicolumn{2}{c}{\textsc{CylinderFlow}} & \multicolumn{3}{c}{\textsc{AirFoil}} \\ \cmidrule(lr){2-3}\cmidrule(lr){4-5}\cmidrule(lr){6-8}
        & Former & Latter & Velocity & Pressure & Velocity & Pressure & Density \\\midrule
        MGN     & & & 48.8\std{5.6} & 36.7\std{2.4} & 10,261\std{832} & 3,043,186\std{282,514} & 29.4\std{2.7}\\
        PIORF       & ORC & Physics & 41.6\std{3.9} & 28.9\std{1.5} & 7,743\std{584} & 2,245,858\std{142,452} & 22.5\std{1.4}\\
        \cmidrule(lr){1-8}
        Ablation 1  & Degree & Physics & 44.9\std{5.7} & 33.3\std{1.1} & 10,379\std{607} & 3,065,807\std{276,373} & 29.6\std{2.6} \\ 
        Ablation 2  & Random & Physics & 44.6\std{0.8} & 31.1\std{1.0} & 10,150\std{505} & 2,936,397\std{177,037} & 28.4\std{1.8} \\ 
        Ablation 3  & Random & Random          & 48.4\std{1.6} & 35.3\std{0.9} & 11,220\std{538} & 3,305,957\std{176,642} & 31.9\std{1.7} \\ 
        Ablation 4  & ORC    & Random          & 43.4\std{2.4} & 32.3\std{1.2} & 10,317\std{771} & 3,115,406\std{230,796} & 30.2\std{2.2} \\
        \bottomrule
    \end{tabular}
    \label{tab:ablation2}
    \vspace{-1em}
\end{table}

\section{Additional Discussion}
\subsection{Graph topology changes.}
\begin{figure}[h]
    \centering
    \subfigure[\textsc{CylinderFlow}]
    {\includegraphics[width=0.49\textwidth]{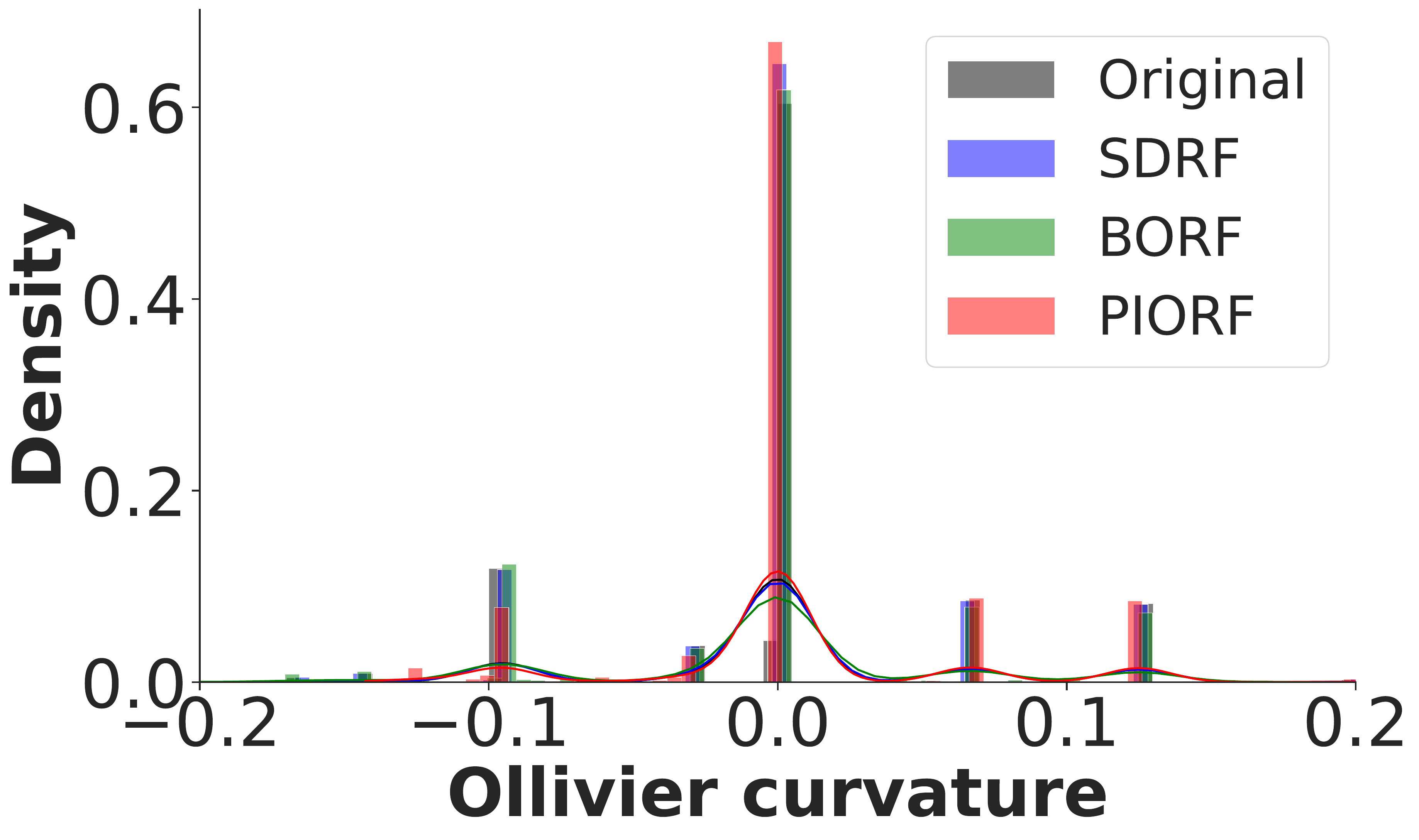}}
    \subfigure[\textsc{AirFoil}]{\includegraphics[width=0.49\textwidth]{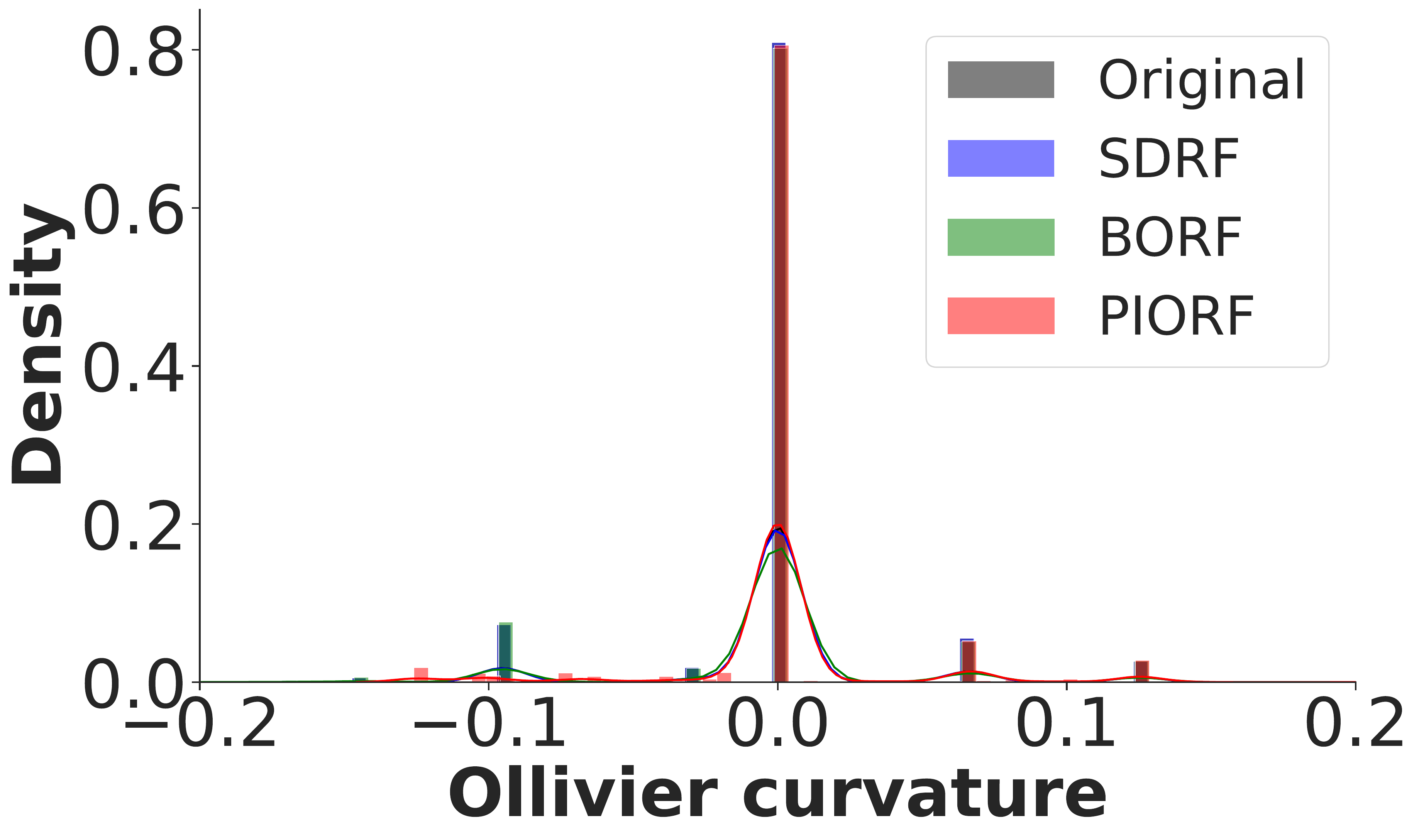}}
    \caption{Comparison of curvature distributions between the original graph and the graph using PIORF. The $x$-axis represents the Olivier curvature of the edges, and the plots show a kernel density estimate of the curvature distribution.}
    \label{fig:dist_curv}
\end{figure}

We analyze changes in graph topology in each dataset. \Cref{fig:dist_curv} shows a comparison of curvature distributions between the original graph and the graph using PIORF. The graph constructed after applying PIORF shows the removal of highly negative curvatures that cause bottlenecks~\citep{topping2021understanding}.

\subsection{Effective Resistance on Graphs.}
The effective resistance~\citep{black2023understanding} provides a metric for measuring over-squashing. We randomly pick up 10,000 sample graphs from each dataset and analyze the total resistance (the sum of the effective resistance between all pairs of nodes). \Cref{tab:resistance} shows the total effective resistance results in the original graph and the graph after applying the PIORF method in each dataset. The total effective resistance is significantly reduced, which indicates that the bottleneck is alleviated and enables long-range propagation.

\begin{table}[h]
    \centering
    \caption{Total resistance for our PIORF and baselines.}
    \begin{tabular}{ll c}\toprule
        Methods   & CylinderFlow  & Airfoil \\\midrule
        MGN       & 2,491,084     & 15,644,891 \\
        SDRF      & 2,487,198     & 15,628,620 \\
        BORF      & 2,398,661     & 15,403,149 \\
        PIORF     & 1,653,709     & 10,140,834 \\
        \bottomrule
    \end{tabular}
    \label{tab:resistance}
\end{table}

\subsection{Relationship Between Accumulated Error and Velocity Gradient.}
\begin{figure}[h]
    \centering
    \subfigure[MGN at step 200]{\includegraphics[width=0.32\textwidth]{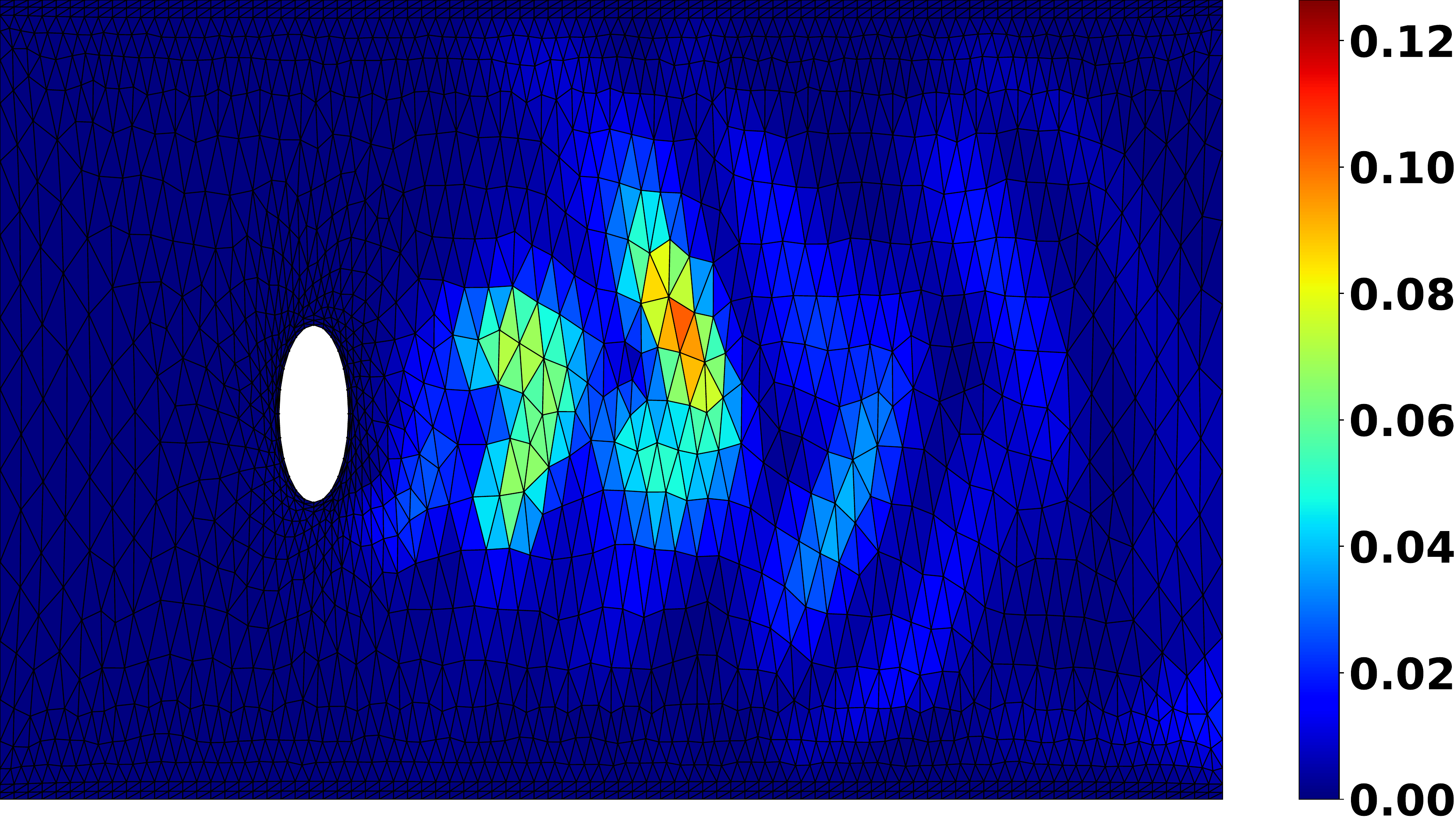}}
    \subfigure[MGN at step 300]{\includegraphics[width=0.32\textwidth]{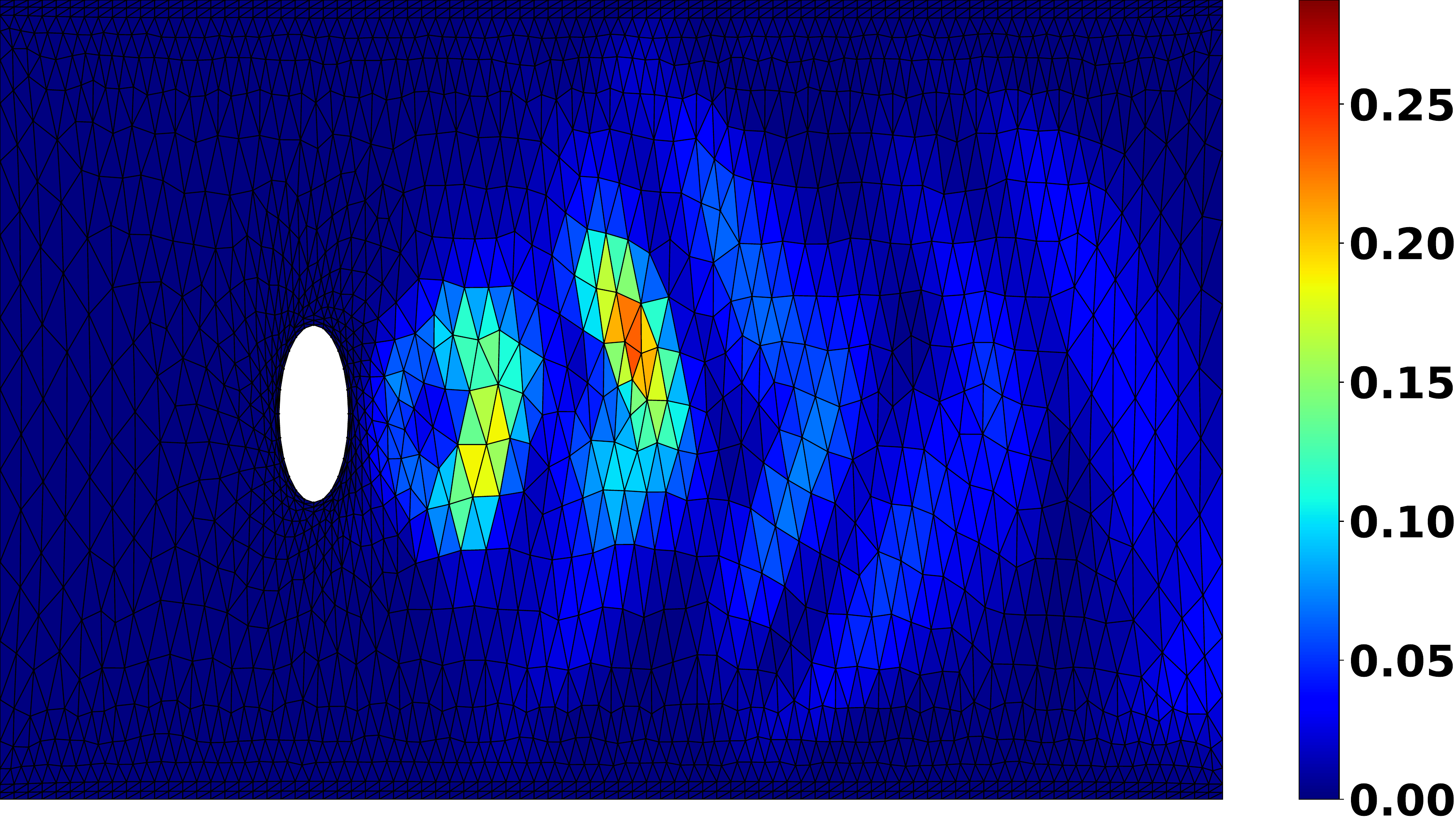}}
    \subfigure[MGN at step 400]{\includegraphics[width=0.32\textwidth]{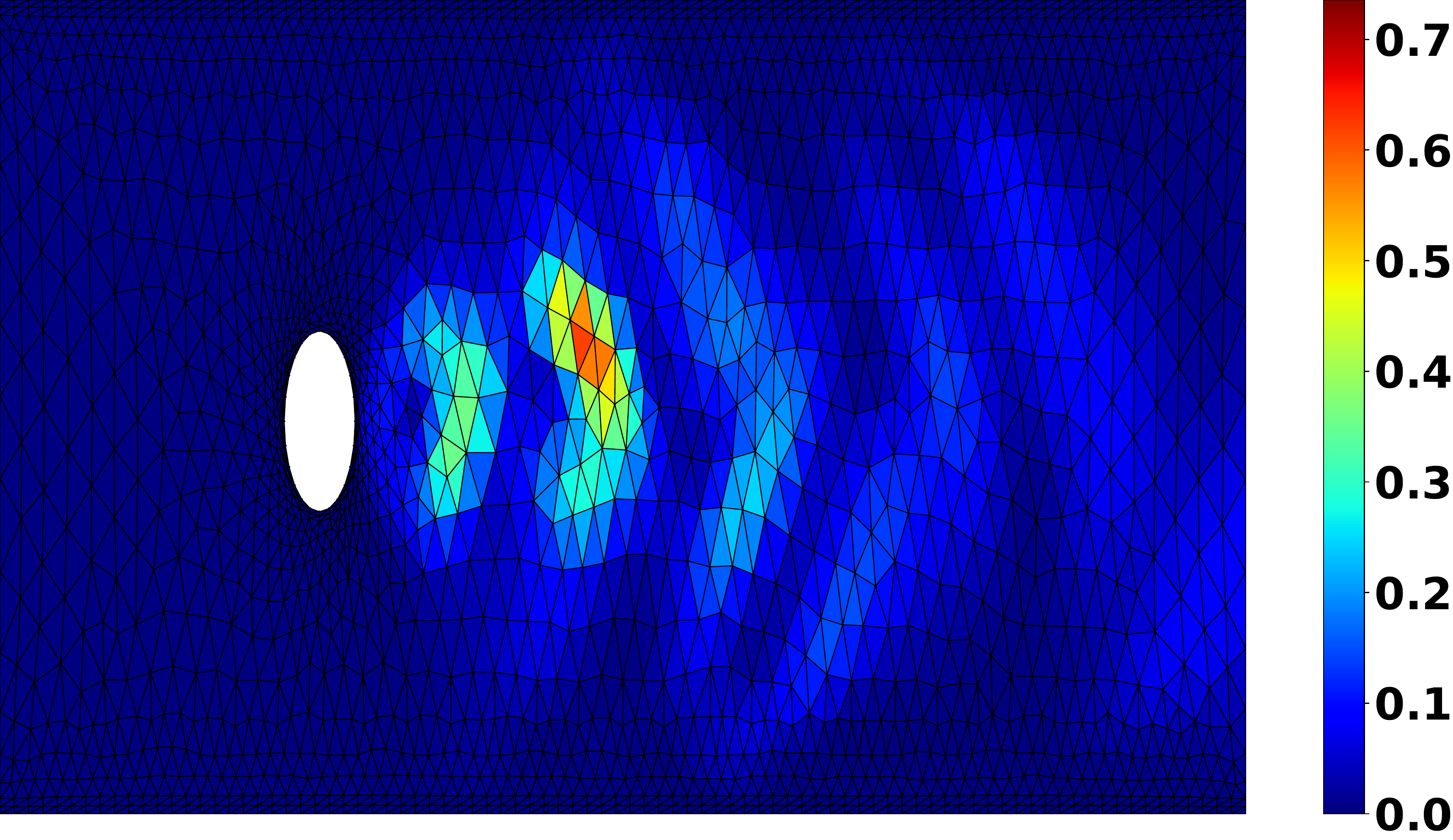}}
    \subfigure[PIORF at step 200]{\includegraphics[width=0.32\textwidth]{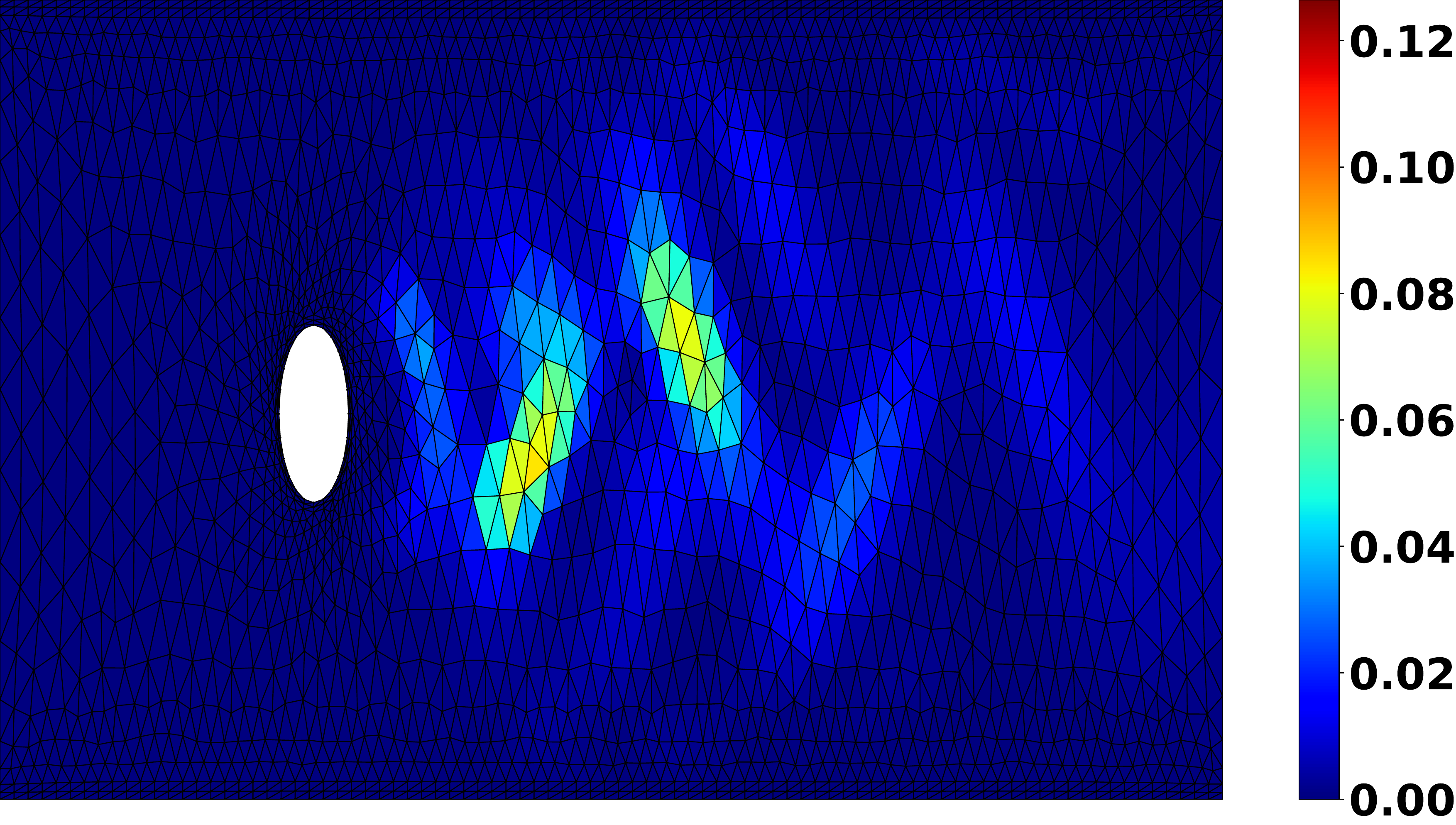}}
    \subfigure[PIORF at step 300]{\includegraphics[width=0.32\textwidth]{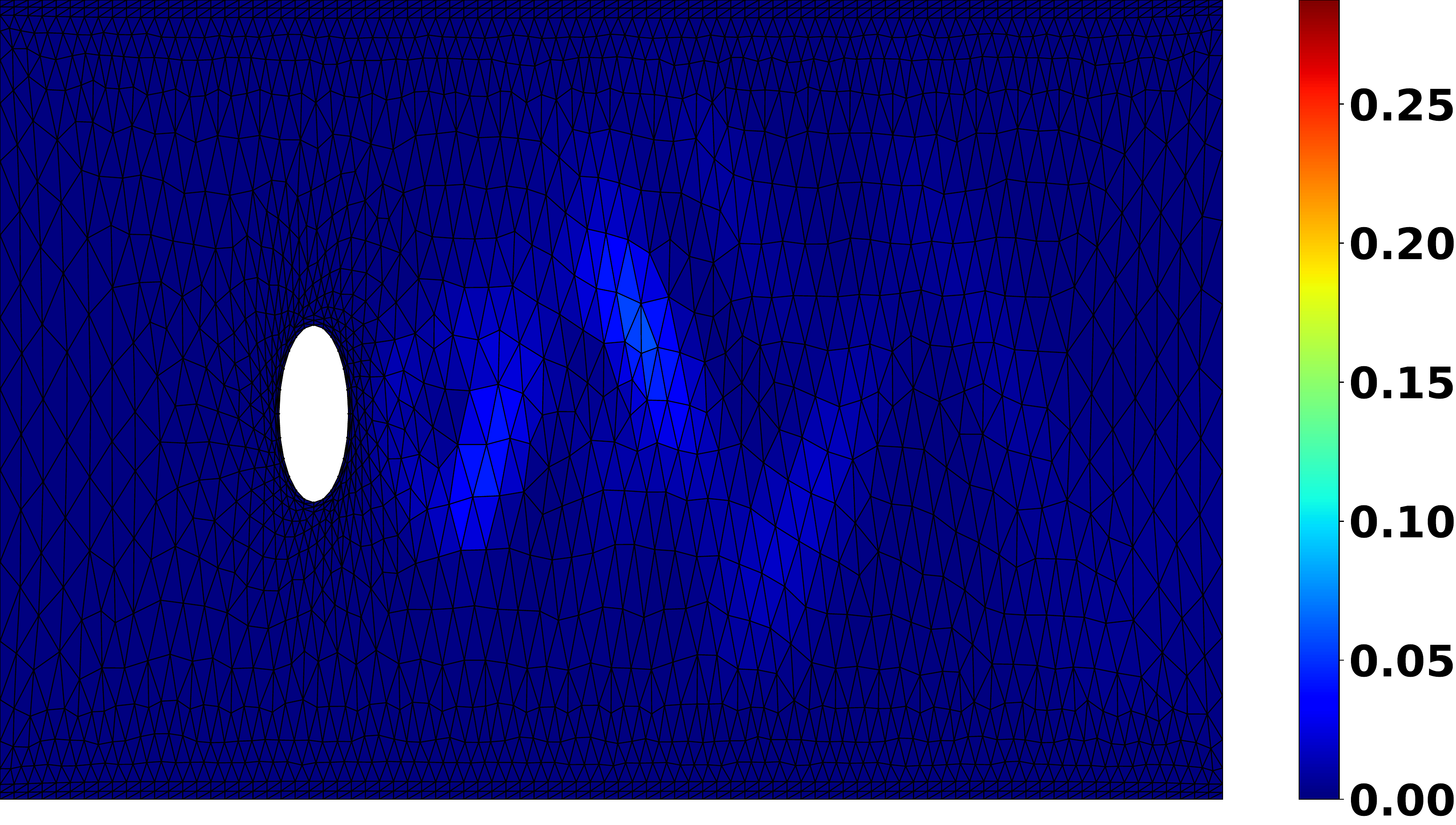}}
    \subfigure[PIORF at step 400]{\includegraphics[width=0.32\textwidth]{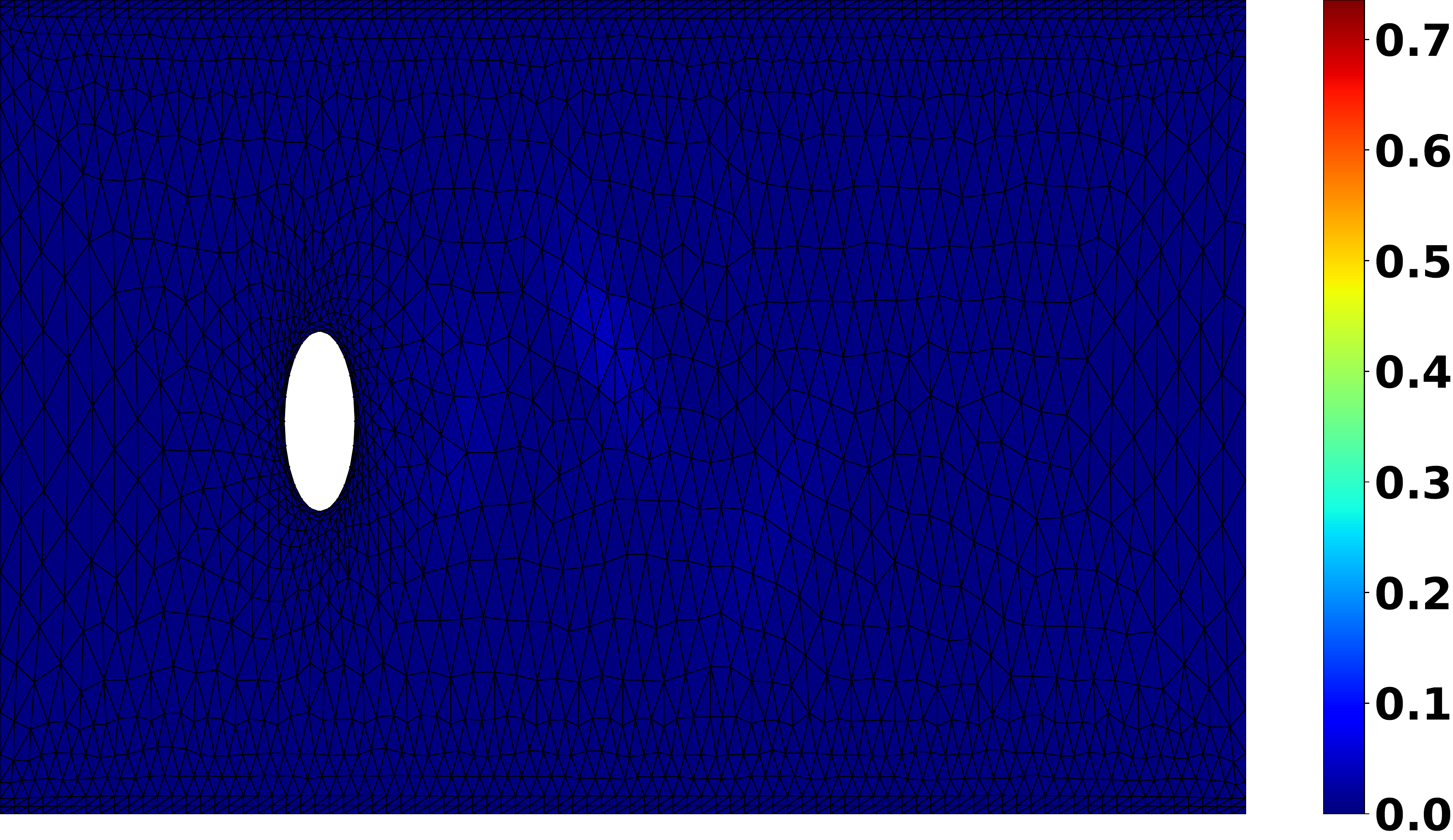}}
    \subfigure[Gradient at step 100]{\includegraphics[width=0.32\textwidth]{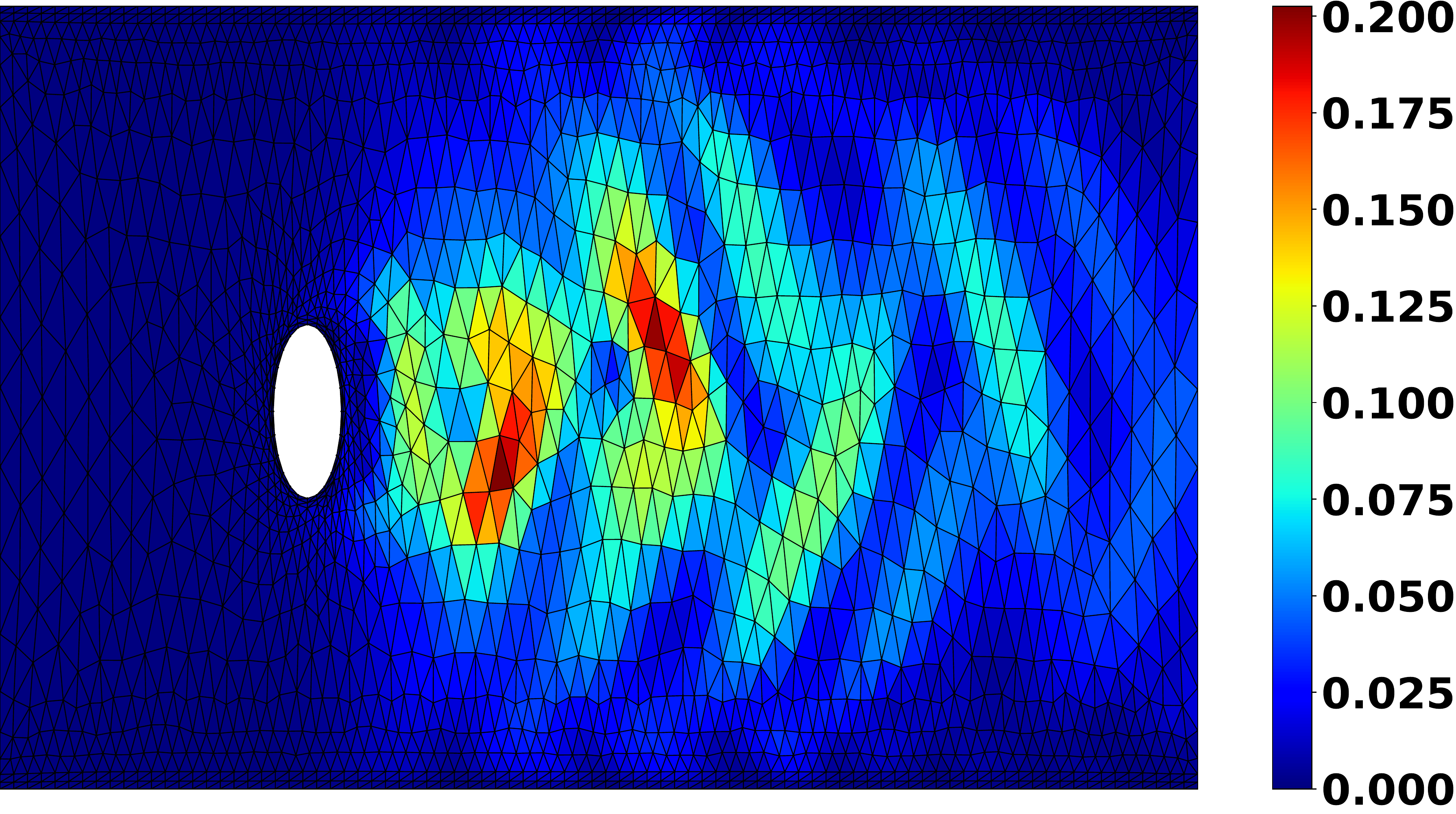}}
    \subfigure[Gradient at step 200]{\includegraphics[width=0.32\textwidth]{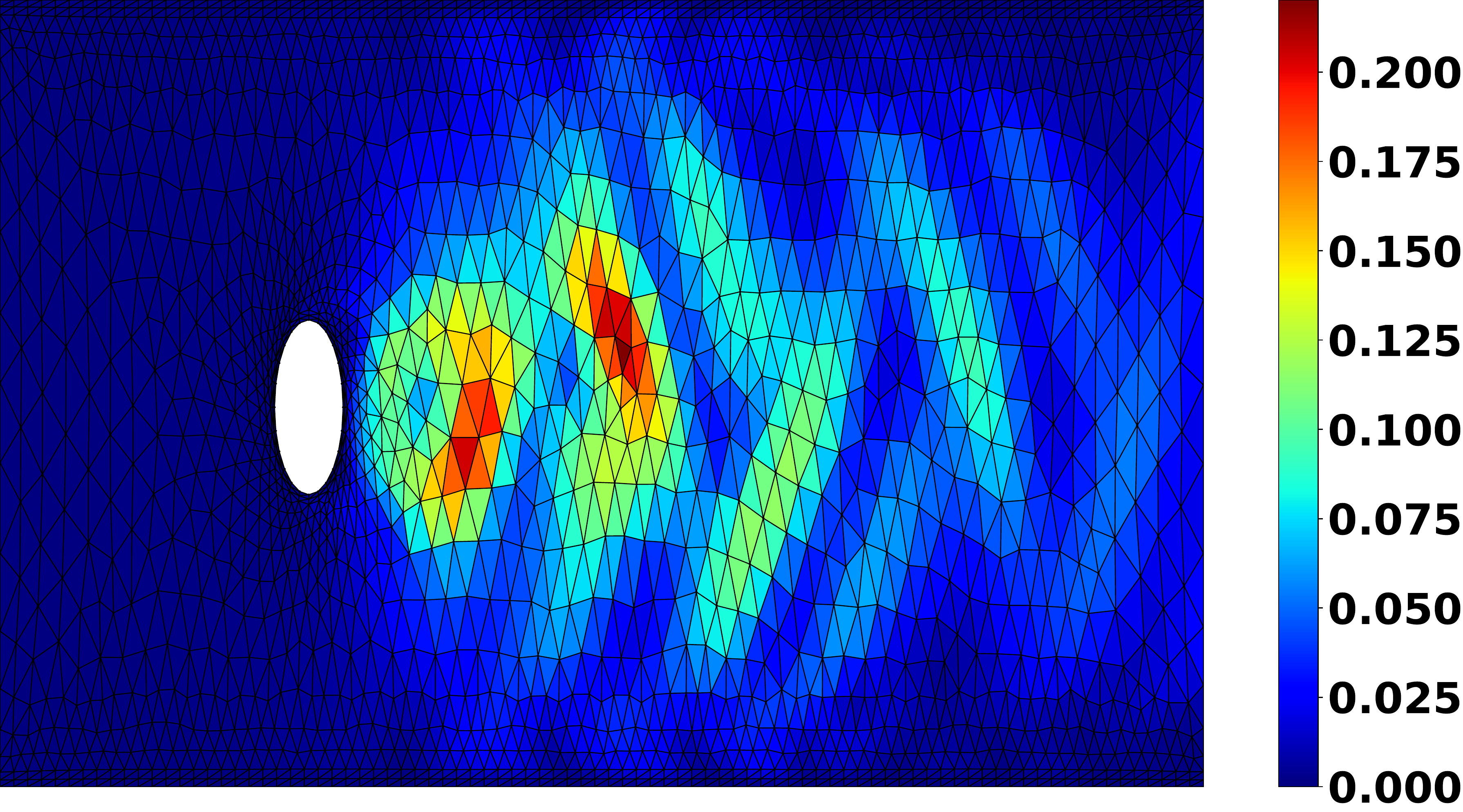}}
    \subfigure[Gradient at step 300]{\includegraphics[width=0.32\textwidth]{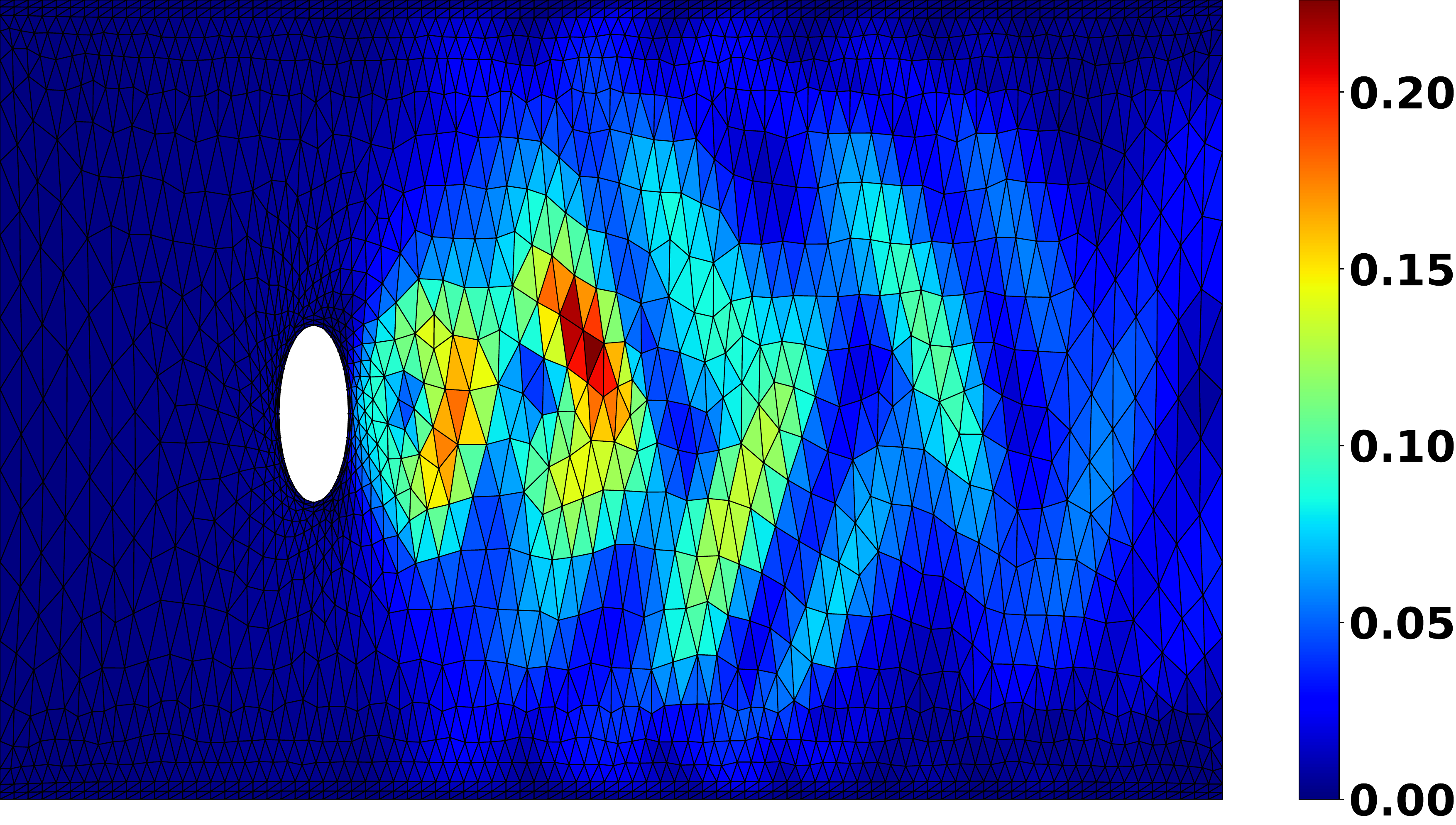}}
    \caption{Comparison of accumulated error distribution and gradient magnitude of velocity distribution according to PIORF application.}
    \label{fig:error_gradient}
\end{figure}

In the field of dynamics learning simulations, such as MGN, the model iteratively predicts the next step. The longer the simulation steps, the more accumulated error occurs during inference. \Cref{fig:error_gradient} shows the change in accumulated error and the gradient of velocity for each step after applying PIORF. Areas with significant accumulated errors depend on the velocity gradient. In PIORF, which reflects this physical quantity, the overall accumulated error is reduced compared to the original.

\subsection{Discussion on ORC in PIORF}
In studies proposing rewiring methods in the field of graph machine learning, metrics with various curvature concepts are typically used to optimize edge addition or removal by maximizing their values~\citep{topping2021understanding, nguyen2023borf}. In physics simulation, alternatives to our ORC-based approach could include methods such as SDRF~\citep{topping2021understanding} and BORF~\citep{nguyen2023borf} that use different metrics for optimization. For example, SDRF uses a balanced Forman curvature, which provides a more conservative estimate compared to ORC~\citep[Lemma 4.1]{nguyen2023borf}. While BORF similarly uses ORC for rewiring, our experiments in \Cref{tab:maintable} demonstrate why it is less suitable for physics simulation.

Metrics that can be presented as alternatives with a similar role to ORC are Forman curvature~\citep{sreejith2016forman} and betweenness centrality~\citep{barthelemy2004betweenness}. However, these metrics do not capture the area near the boundary conditions effectively. This region is where fluid flow changes and is also crucial from a domain knowledge. While Forman curvature, based on the graph Laplacian, is easier and faster to compute than Ollivier-Ricci curvature, it is less geometric~\citep{ni2019community}. We choose ORC specifically because it better captures geometric characteristic, particularly around boundary conditions where fluid flow changes dramatically. Betweenness centrality could be used for source node selection, its high complexity $\mathcal{O}(|\mathcal{V}||\mathcal{E}|)$ and need for global information make it impractical for mesh graphs with thousands of nodes and edges.

\end{document}